\documentclass[journal]{IEEEtran}
\usepackage{color}
\usepackage{times}
\usepackage{epsfig}
\usepackage{graphicx}
\usepackage{graphics}
\usepackage{amsmath}
\usepackage{amssymb}
\usepackage{multirow}
\usepackage{amsthm,amsmath,amssymb}
\usepackage{mathrsfs}
\usepackage{rotating}
\usepackage{bbm}
\usepackage{verbatim}
\usepackage{array}
\usepackage{booktabs}
\usepackage{tabularx}
\usepackage{url}
\usepackage[colorlinks,linkcolor=]{hyperref}
\usepackage[labelformat=simple]{subcaption}

\ifCLASSINFOpdf
\else
\fi

\begin{document}


%
%
%
%
\title{Consistency-Aware Map Generation \\ at Multiple Zoom Levels Using Aerial Image}
\author{
		Linwei~Chen, 
		Zheng~Fang, 
		Ying~Fu,~\IEEEmembership{Senior Member,~IEEE}
		\thanks{\color{black}
		Manuscript received December 13, 2021; revised February 17, 2022; accepted April 8, 2022. This work was supported by the National Natural Science Foundation of China under Grant 62171038, Grant 61827901, Grant 61936011, and Grant 62088101.	
		}%
		
		\thanks{L. Chen, Z. Fang and Y. Fu are with the Beijing Laboratory of Intelligent Information Technology, School of Computer Science and Technology, Beijing Institute of Technology, Beijing 100081, China.
		}%
		\thanks{Corresponding author: Ying Fu (e-mail: fuying@bit.edu.cn).}
		\thanks{Dataset is [Online]. Available: https://github.com/ying-fu/MLM.}
		\thanks{Digital Object Identifier 10.1109/JSTARS.2022.3170591}
}

%
%

\markboth{IEEE Journal of Selected Topics in Applied Earth Observations and Remote Sensing}%
{Shell \MakeLowercase{\textit{et al.}}: Bare Demo of IEEEtran.cls for Journals}
%



\newcommand{\ying}[1]{\textcolor{magenta}{{[Ying: #1]}}}
\newcommand{\linwei}[1]{\textcolor{black}{{[Linwei: #1]}}}
\def\eg{\emph{e.g.}}
\def\Eg{\emph{E.g.}}
\def\etal{\emph{et al.}}
\def\ie{\emph{i.e.}}

\maketitle

\begin{abstract}
The multi-level tiled map service is widely used and serves as a kind of digital infrastructure. 
These map tiles are usually rendered from vector data, whose update needs to walk or drive with professional equipment to check every points of interest. 
This leads to inconvenience and expensive cost in timely updating maps. 
Compared with vector data, aerial images are much easier and cheaper to obtain. 
In this paper, we propose a novel multi-level map generation framework that can automatically generate accurate and consistent maps with multiple zoom levels from aerial images. 
It consists of a level-aware map generator and a consistency-aware map generator. 
The level-aware map generator is able to generate accurate initial maps with realistic details for each zoom level.
The consistency-aware map generator regards initial maps at each zoom level as a sequence and builds the connection between them, so as to guarantee content consistency between maps at different zoom levels.  
Furthermore, we collect a large-scale high-quality dataset called Multi-Level Map (MLM) for map generation at multiple zoom levels. 
Experiments on our MLM dataset show that our method outperforms previous state-of-the-art map generation methods on both comprehensive quantitative metrics and perceptual quality. 
\end{abstract}


%
\IEEEpeerreviewmaketitle

\section{Introduction}
\IEEEPARstart{O}{nline} multi-level map service (\eg, Google map), as an important role in our lives, not only provides convenience for daily travel, but also serves as an important infrastructure for shared bike services, delivery services, logistics industry, transportation industry, and etc.

\begin{figure}[tb]
	\begin{center}
	\scalebox{0.60}{
	\begin{tabular}{cccccccccc}
		\begin{sideways}\end{sideways}
		\includegraphics[width=4.01cm, angle=0]{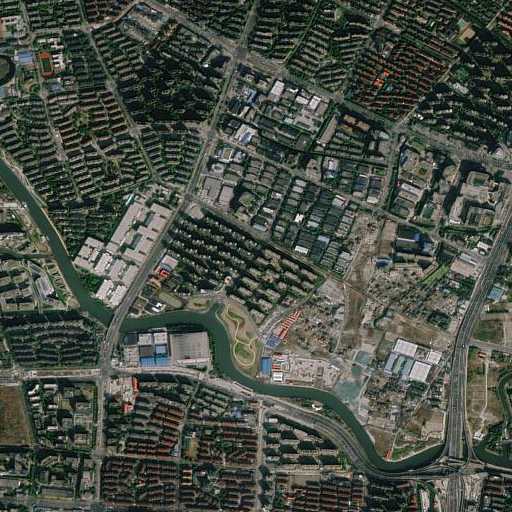}\hspace{-3mm}
		&\includegraphics[width=4.01cm, angle=0]{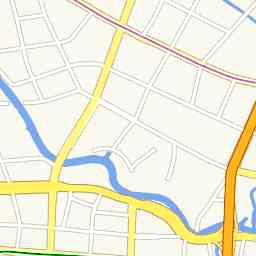}\hspace{-3mm}
		&\includegraphics[width=4.01cm, angle=0]{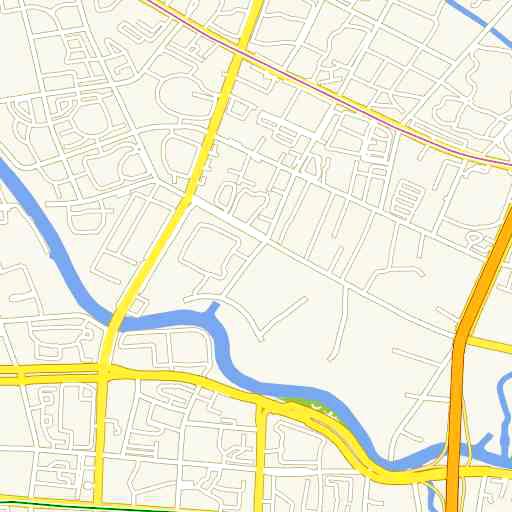}
		\begin{sideways}\end{sideways}
		\\
		(a) Aerial image & (b) map at the 15-th level & (c) map at the 16-th level\\
		   & $256\times 256$ & $512\times 512$
		\\
		\begin{sideways}\hspace{+12mm} 15-th level\end{sideways}
		\includegraphics[width=4.01cm, angle=0]{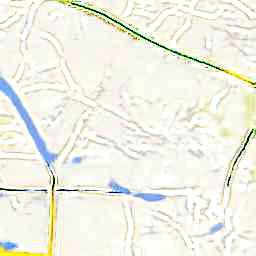}\hspace{-3mm}
		&\includegraphics[width=4.01cm, angle=0]{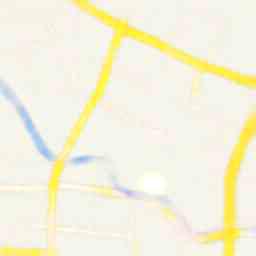}\hspace{-3mm}
		&\includegraphics[width=4.01cm, angle=0]{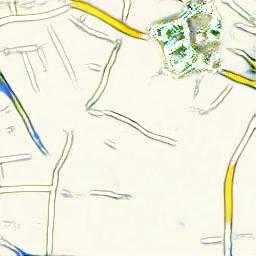}\hspace{+0mm}
		\begin{sideways}\hspace{+12mm} $256\times 256$\end{sideways}
		\\
		\begin{sideways}\hspace{+12mm} 16-th level\end{sideways}
		\includegraphics[width=4.01cm, angle=0]{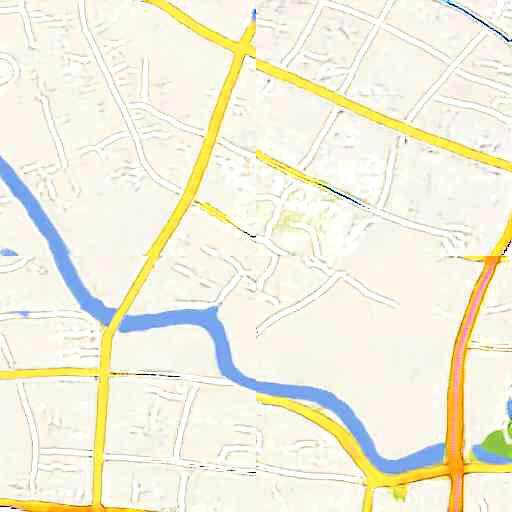}\hspace{-3mm}
		&\includegraphics[width=4.01cm, angle=0]{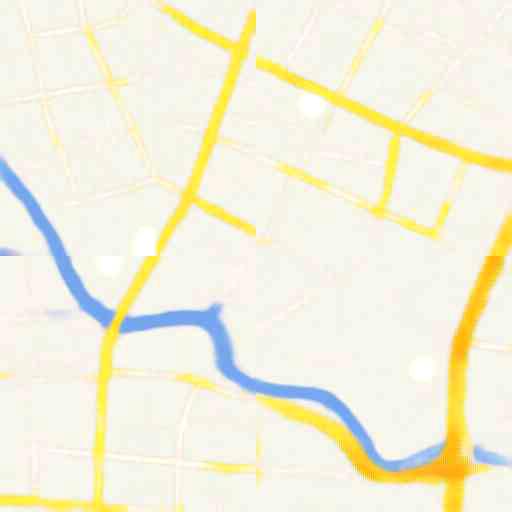}\hspace{-3mm}
		&\includegraphics[width=4.01cm, angle=0]{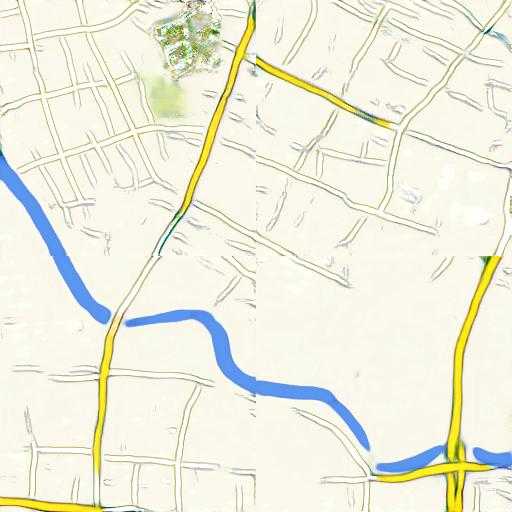}\hspace{+0mm}
		\begin{sideways}\hspace{+12mm} $512\times 512$\end{sideways}
		\\
(d) Pix2PixHD  & (e) SMAPGAN  & (f) SelectionGAN 
	\end{tabular}}
	\end{center}
	\vspace{+0mm}
	\caption{
	{\color{black}Multi-level map generation is at different zoom levels. (a) The aerial image. (b) and (c) The corresponding ground truth map at the 15-th and 16-th levels. 
	(d)-(f) show results of applying existing single-level map generation methods on multi-level generation task. 
	It can be seen that these results are inconsistent in content, blurring and have severe artifacts.
	}}
	\vspace{-6mm}
	\label{fig:intro}
\end{figure}

Since {\color{black} multi-level maps} make it much easier for users to gather cartographic information effectively by switching between different zoom levels, the tiled map service always provides maps with multiple zoom levels.  
They are usually rendered from vector data, whose update needs to walk or drive with professional equipment to check every point of interest.  
Though this ensures accuracy, it leads to expensive cost and cannot update timely in a crisis like an earthquake.
Compared with vector data, aerial images are much easier and cheaper to obtain. 
They can be collected automatically and updated timely by remote sensing instruments, \eg, airplanes, satellites and drones.
And aerial images are widely used for earth observation, they inherently contain rich information about the ground surface that can be utilized for map generation~\cite{fuying-2021-remotesensing, translation2022}. 
Therefore, multi-level maps based on aerial images can be quickly updated even in extreme conditions like earthquakes, floods, mudslides and other natural disasters and provide rescuers with important information about the ground surface in time to save people’s lives.

Generating multi-level maps from aerial images can be regarded as an image-to-image translation task~\cite{pix2pix}, \ie, extracting cartographic information from aerial image and render it as a series of RGB map images at different zoom levels, \eg, from (a) to (b) and (c) in Figure~\ref{fig:intro}.
{\color{black}Currently, there are some methods, like Pix2Pix \cite{pix2pix}, Pix2PixHD \cite{pix2pixhd}, CycleGAN \cite{cyclegan}, GeoGAN \cite{2019geogan}, and SMAPGAN \cite{2020smapgan}, that can automatically generate images in map style from aerial images.}
But they do not explicitly understand aerial images at pixel level and only consider map generation at single zoom level. 
{\color{black}As illustrated in Figure~\ref{fig:intro}(d)-(f), these methods easily result in inaccurate map generation, blur, artifacts and severe content inconsistency between different zoom levels.}
Thus, it is necessary to design a multi-level map generation method to avoid these problems.
Besides, there is still no specialized multi-level map dataset, which should provide paired samples of aerial images and corresponding map images at multiple zoom levels.
This hinders the development and evaluation of multi-level map generation methods.

\begin{figure*}[tb]
	\centering 
	\scalebox{0.8}{
	\begin{tabular}{c}
	\includegraphics[width=22cm, angle=0]{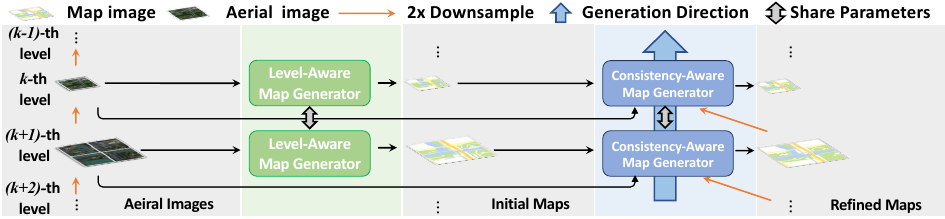}\\
	(a) The overview of our proposed method \\
	\includegraphics[width=22cm, angle=0]{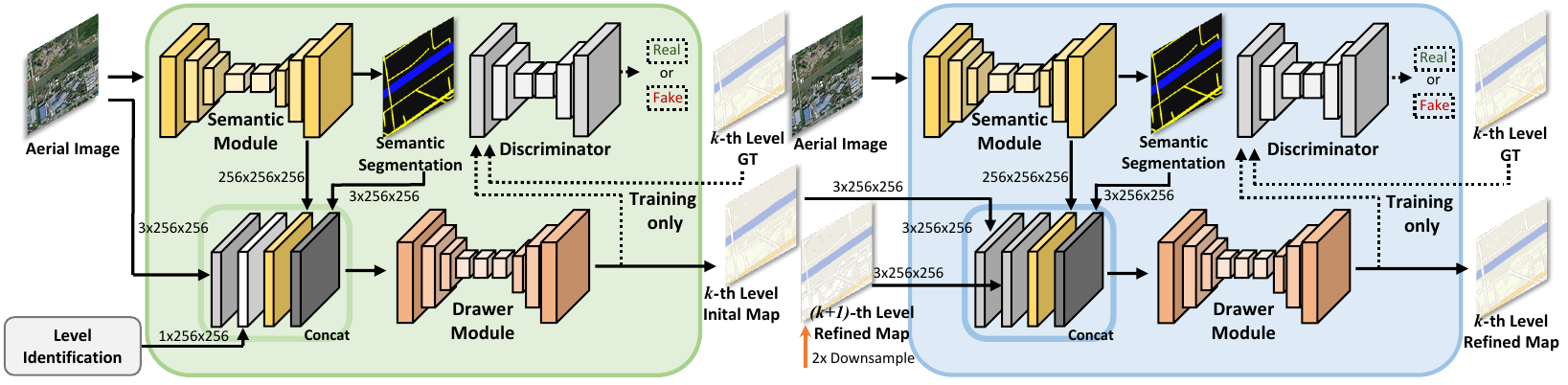}\\
	(b) Details of level-aware map generator
	\qquad\qquad\qquad\qquad\qquad\qquad\qquad
	(c) Details of consistency-aware map generator
	\end{tabular}
	}
\vspace{-0mm}
\caption{
{\color{black}The illustration of our proposed multi-level map generation framework.
(a) The overview of our pipeline. Level-aware map generator first generates accurate initial maps for each zoom level, then the consistency-aware map generator takes in the sequence of initial maps at different zoom levels and builds a connection between maps at different zoom levels, in order to keep the content consistent and refine maps at each zoom level. (b) and (c) show details of these two generators. }
}
\vspace{-0mm}
\label{fig:overview}
\end{figure*}

{\color{black}In this paper, we present a novel multi-level map generation method, which can generate accurate and consistent multilevel maps from aerial images.}
The proposed framework consists of a level-aware map generator and a consistency-aware map generator as illustrated in Figure~\ref{fig:overview}. 
To generate accurate initial maps with realistic details for each zoom level, the level-aware map generator extracts semantic information in aerial images pixel-wisely.
It learns to generate initial maps in an adversarial way \cite{2014gan} and helps to keep the topological relationship among geographical elements.
To guarantee content consistency between maps at different zoom levels, the consistency-aware map generator refines initial maps at each zoom level iteratively by building the connection between them.

{\color{black}Furthermore, we collect a large-scale high-quality Multi-Level Map dataset called MLM. }
It provides 18,700 paired samples of aerial images and corresponding maps from two representative cities (Shanghai and Rio de Janeiro) and with four zoom levels (15, 16, 17 and 18). 
Experiments on our MLM dataset show our method outperforms previous state-of-the-art map generation methods on both comprehensive quantitative metrics and perceptive
quality.

In summary, our main contributions are that
\begin{itemize} 
\vspace{-0mm}
\item We present a novel multi-level map generation method, which is able to generate accurate and consistent multi-level maps from aerial images. 

\vspace{-0mm}
\item We design a level-aware map generator and a consistency-aware map generator, of which 
the former can generate accurate initial maps for each zoom level and the latter can guarantee content consistency between maps of different zoom levels. 

\vspace{-0mm}
\item We collect a large-scale high-quality Multi-Level Map (MLM) dataset.
To our best knowledge, it is the first specialized dataset for multi-level map generation.


\end{itemize}


\begin{figure*}[tb]
	\begin{center}
		
\iffalse
	\scalebox{0.88}{
			\begin{tabular}{cccccccccc}
				\begin{sideways}\qquad 15-th\ level\end{sideways}
				\includegraphics[width=3.01cm, angle=0]{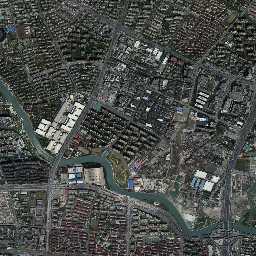}\hspace{-3mm}
				&\includegraphics[width=3.01cm, angle=0]{15_pix2pixHD_1_3_6.jpg}\hspace{-3mm}
				&\includegraphics[width=3.01cm, angle=0]{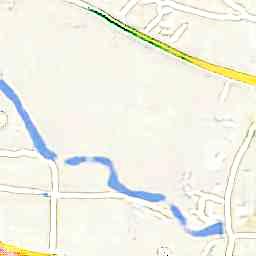}\hspace{-3mm}
				&\includegraphics[width=3.01cm, angle=0]{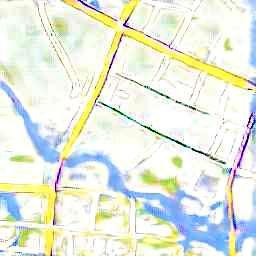}\hspace{-3mm}
				&\includegraphics[width=3.01cm, angle=0]{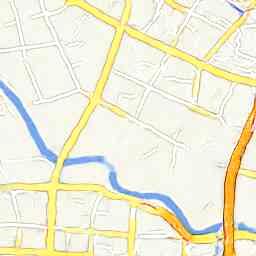}\hspace{-3mm}
				&\includegraphics[width=3.01cm, angle=0]{15_GT_1_3_6.jpg}
				\begin{sideways}\qquad $256\times 256$\end{sideways}
				\\
				\begin{sideways}\qquad 16-th\ level\end{sideways}
				\includegraphics[width=3.01cm, angle=0]{16_RS_1_3_6.jpg}\hspace{-3mm}
				&\includegraphics[width=3.01cm, angle=0]{16_pix2pixHD_1_3_6.jpg}\hspace{-3mm}
				&\includegraphics[width=3.01cm, angle=0]{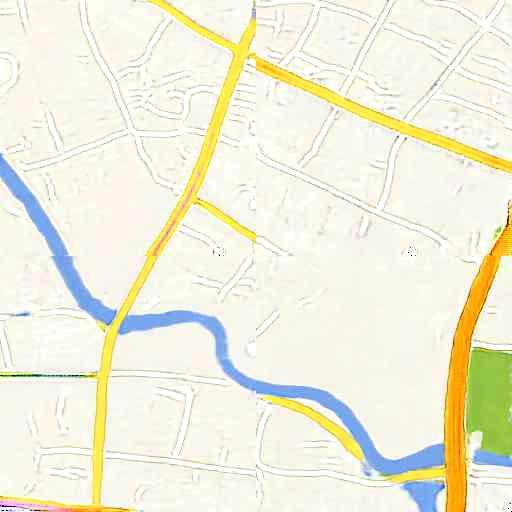}\hspace{-3mm}
				&\includegraphics[width=3.01cm, angle=0]{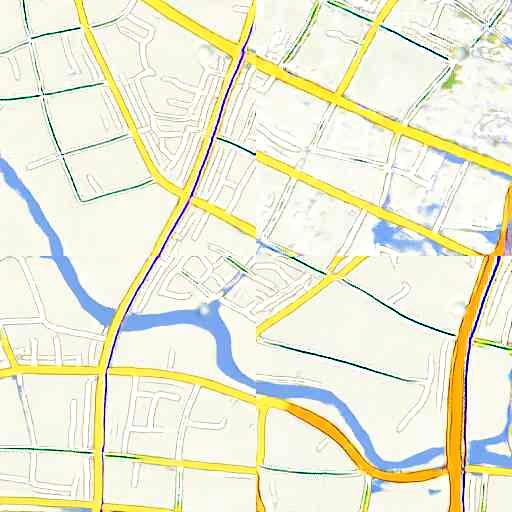}\hspace{-3mm}
				&\includegraphics[width=3.01cm, angle=0]{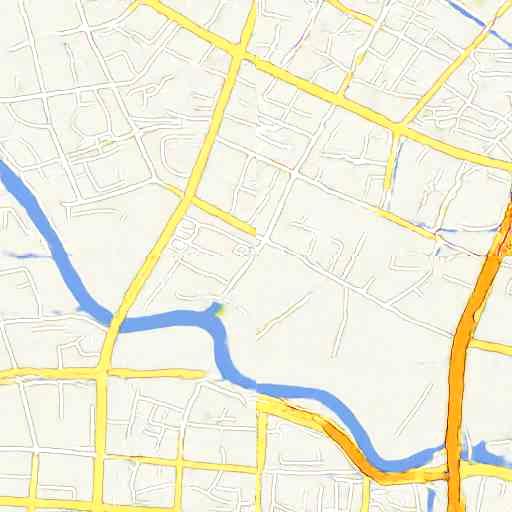}\hspace{-3mm}
				&\includegraphics[width=3.01cm, angle=0]{16_GT_1_3_6.jpg}
				\begin{sideways}\qquad $512\times 512$\end{sideways}
				\\
				\begin{sideways}\qquad 17-th\ level\end{sideways}
				\includegraphics[width=3.01cm, angle=0]{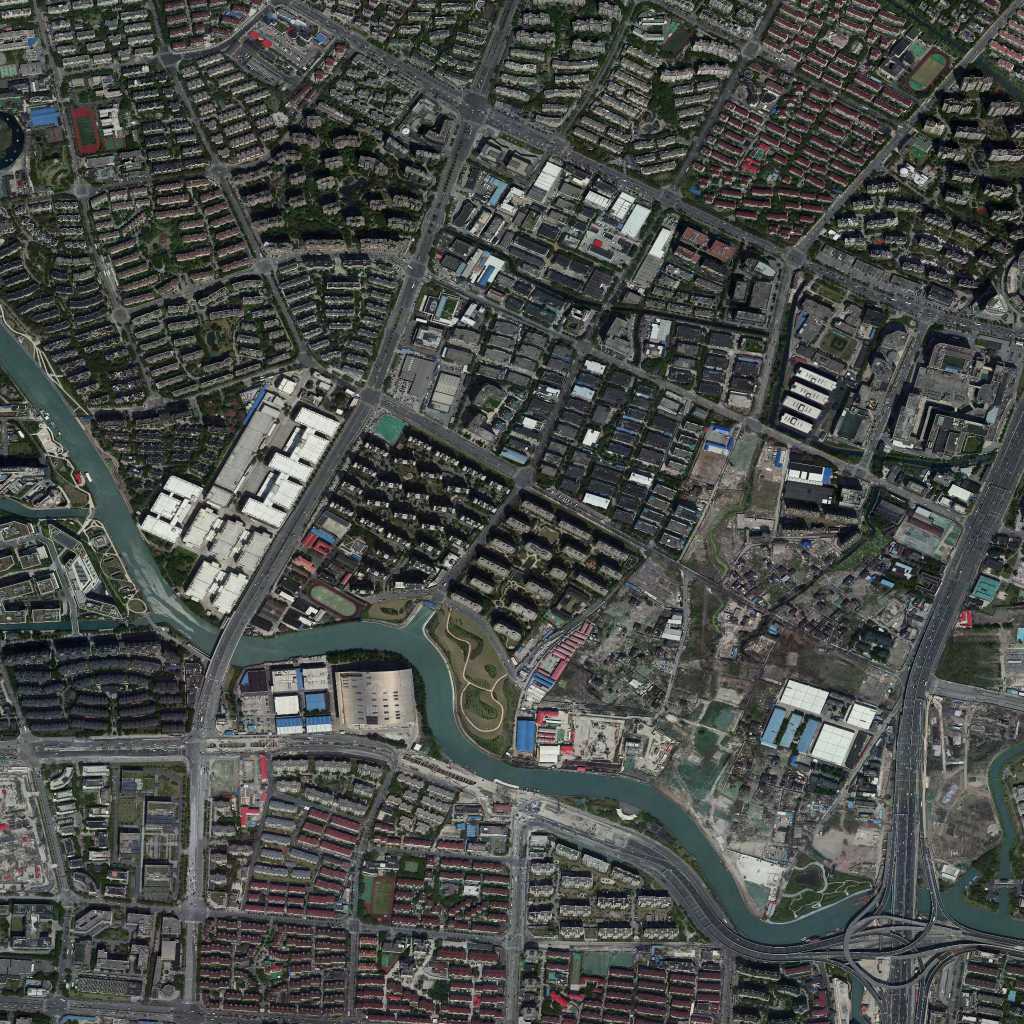}\hspace{-3mm}
				&\includegraphics[width=3.01cm, angle=0]{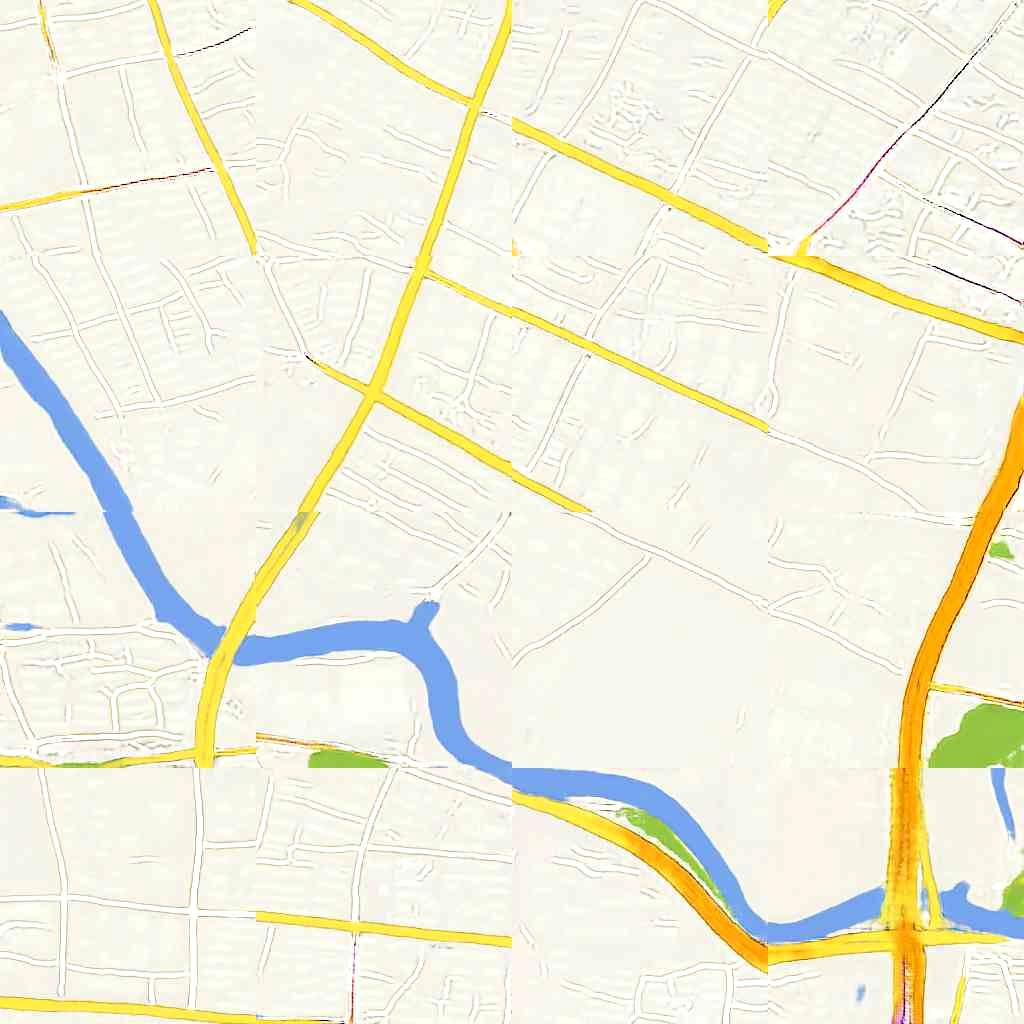}\hspace{-3mm}
				&\includegraphics[width=3.01cm, angle=0]{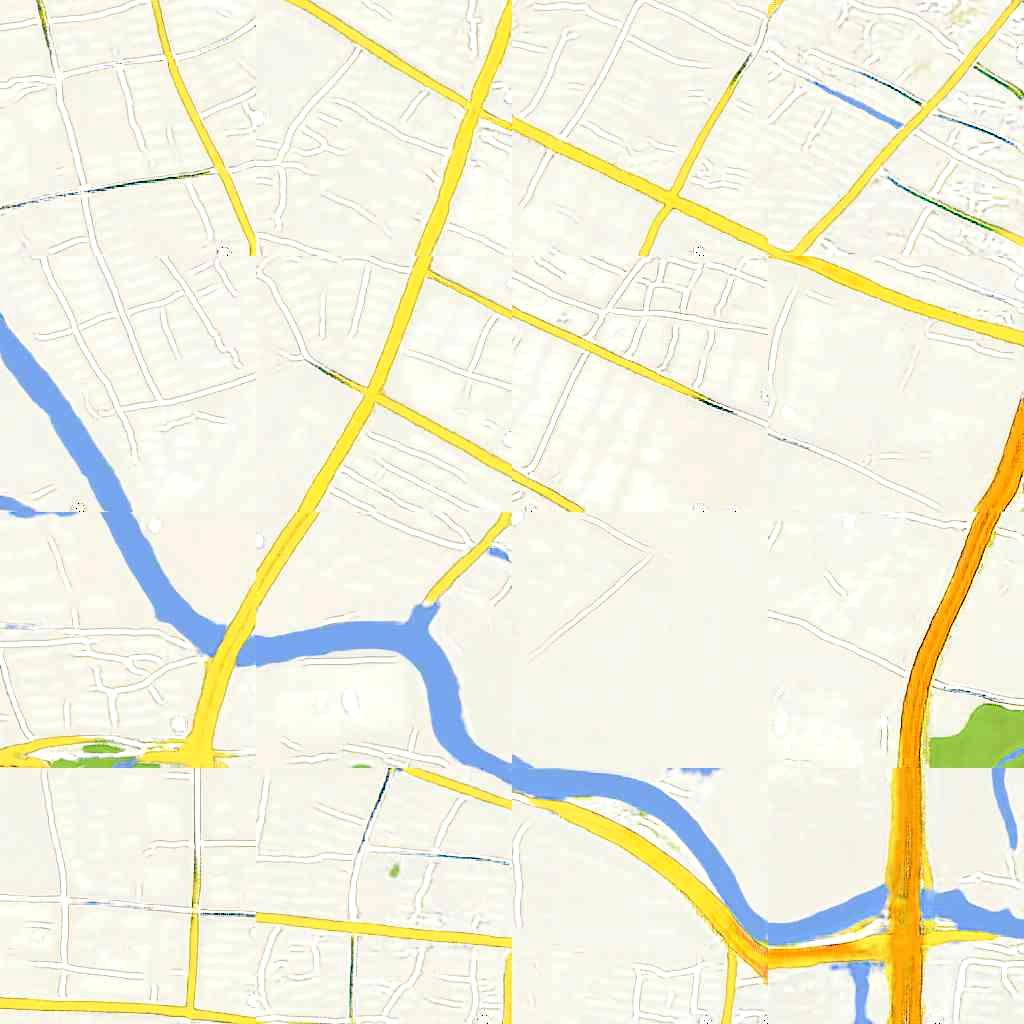}\hspace{-3mm}
				&\includegraphics[width=3.01cm, angle=0]{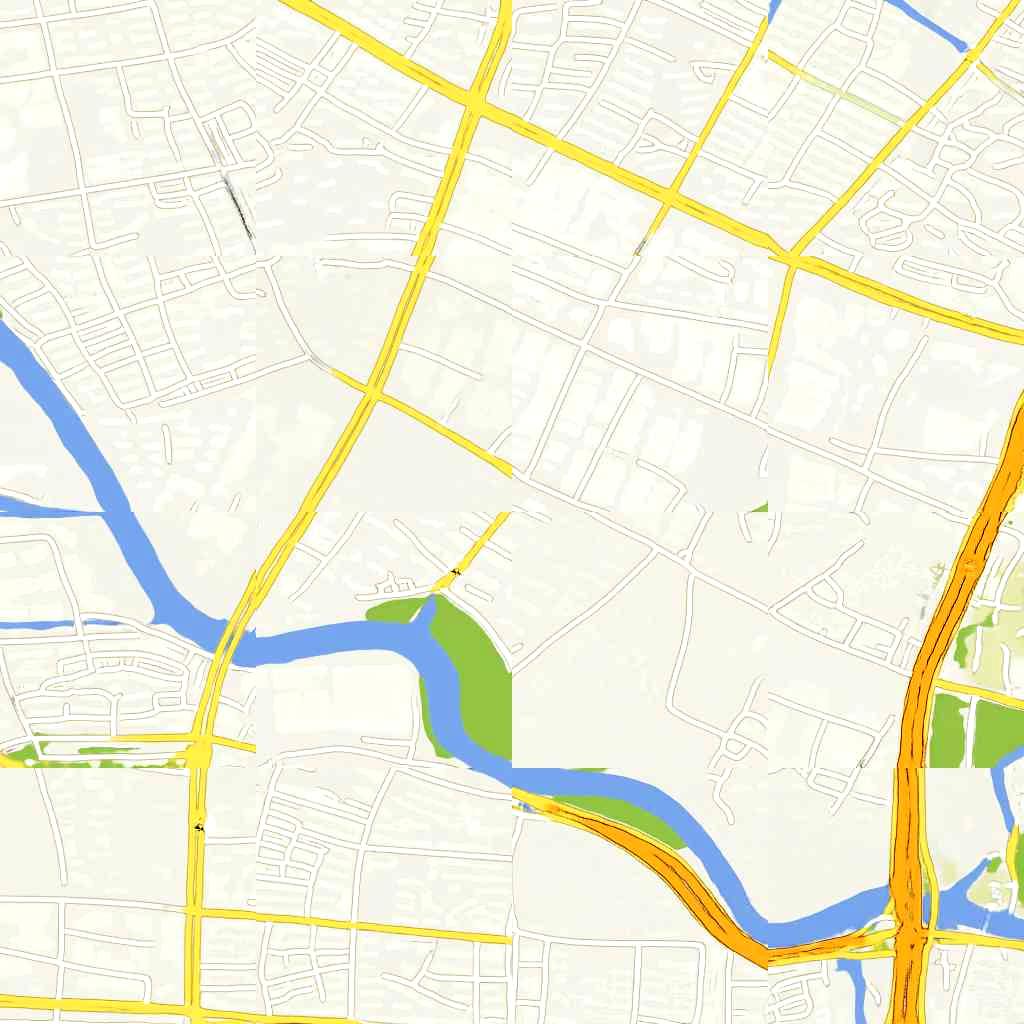}\hspace{-3mm}
				&\includegraphics[width=3.01cm, angle=0]{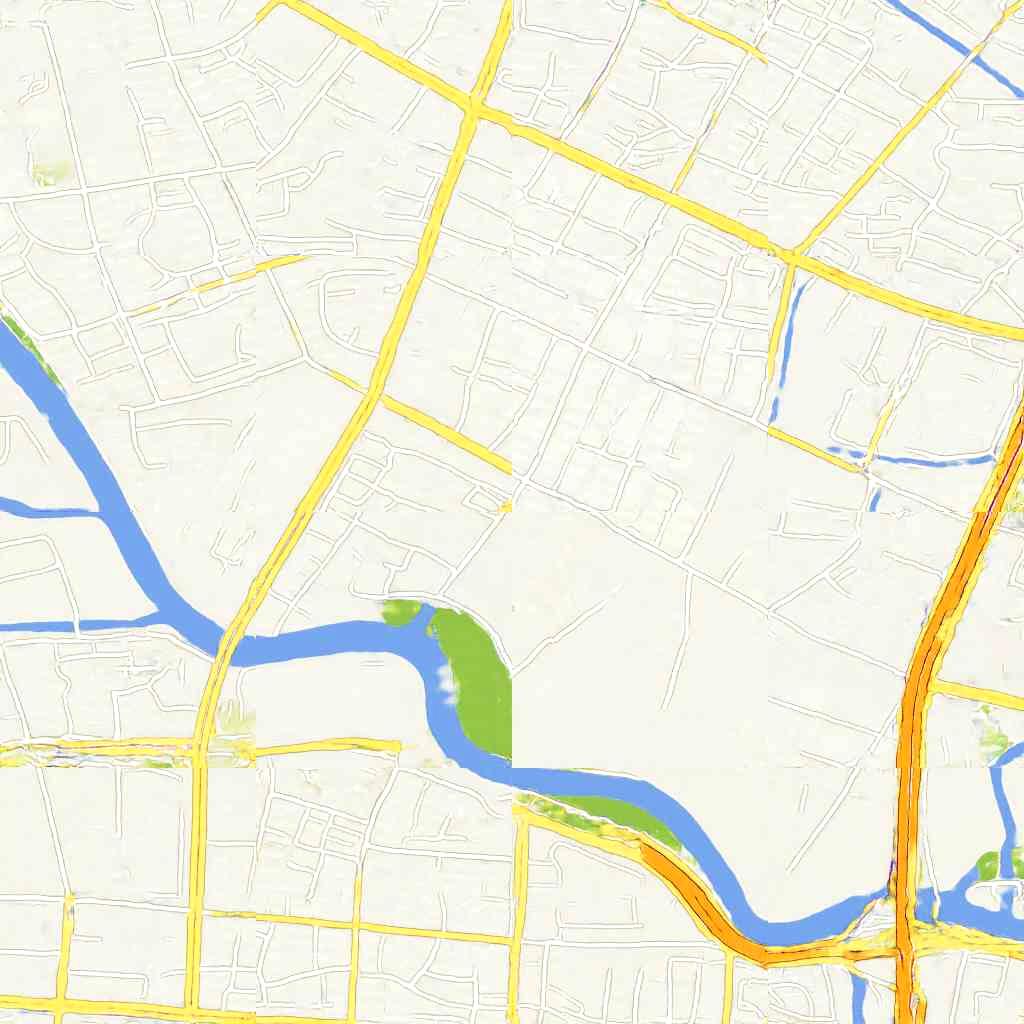}\hspace{-3mm}
				&\includegraphics[width=3.01cm, angle=0]{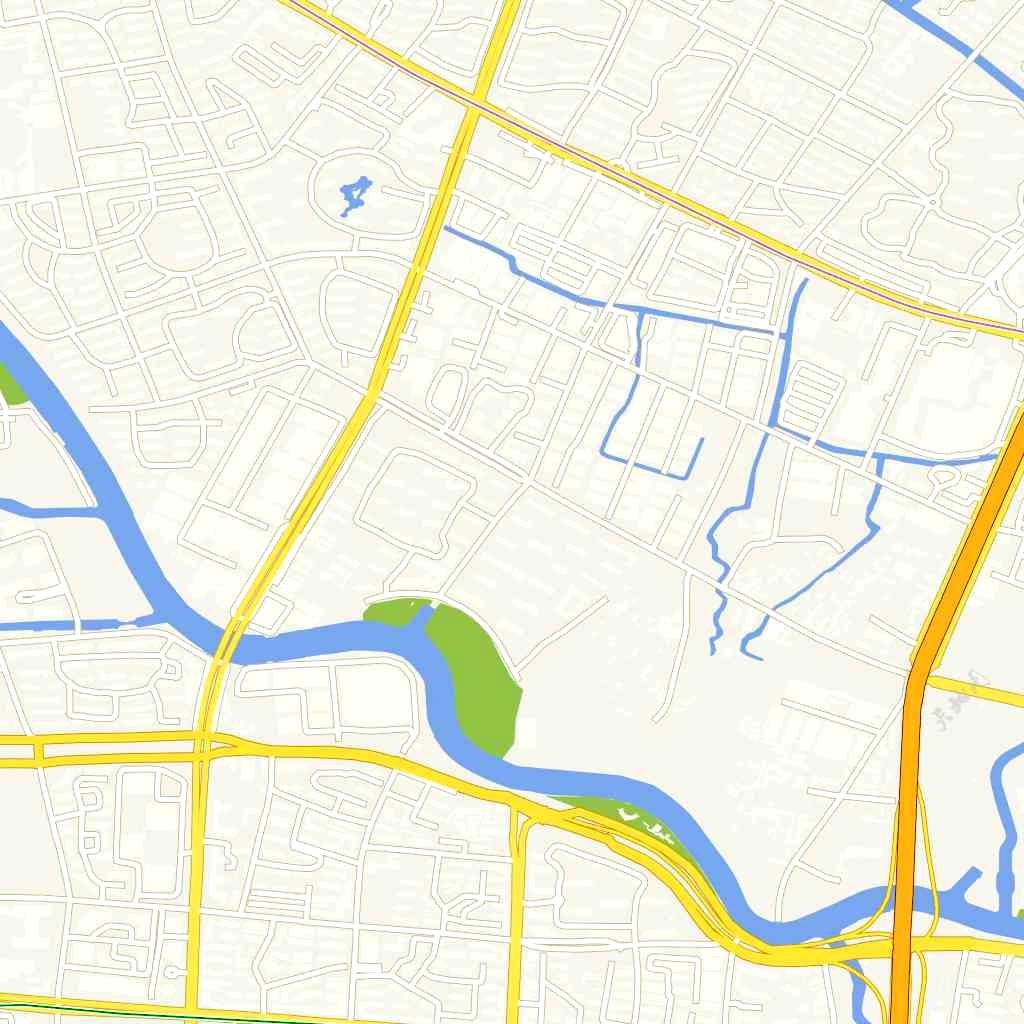}
				\begin{sideways}\qquad $1024\times 1024$\end{sideways}
				\\
				\begin{sideways}\qquad 18-th\ level\end{sideways}
				\includegraphics[width=3.01cm, angle=0]{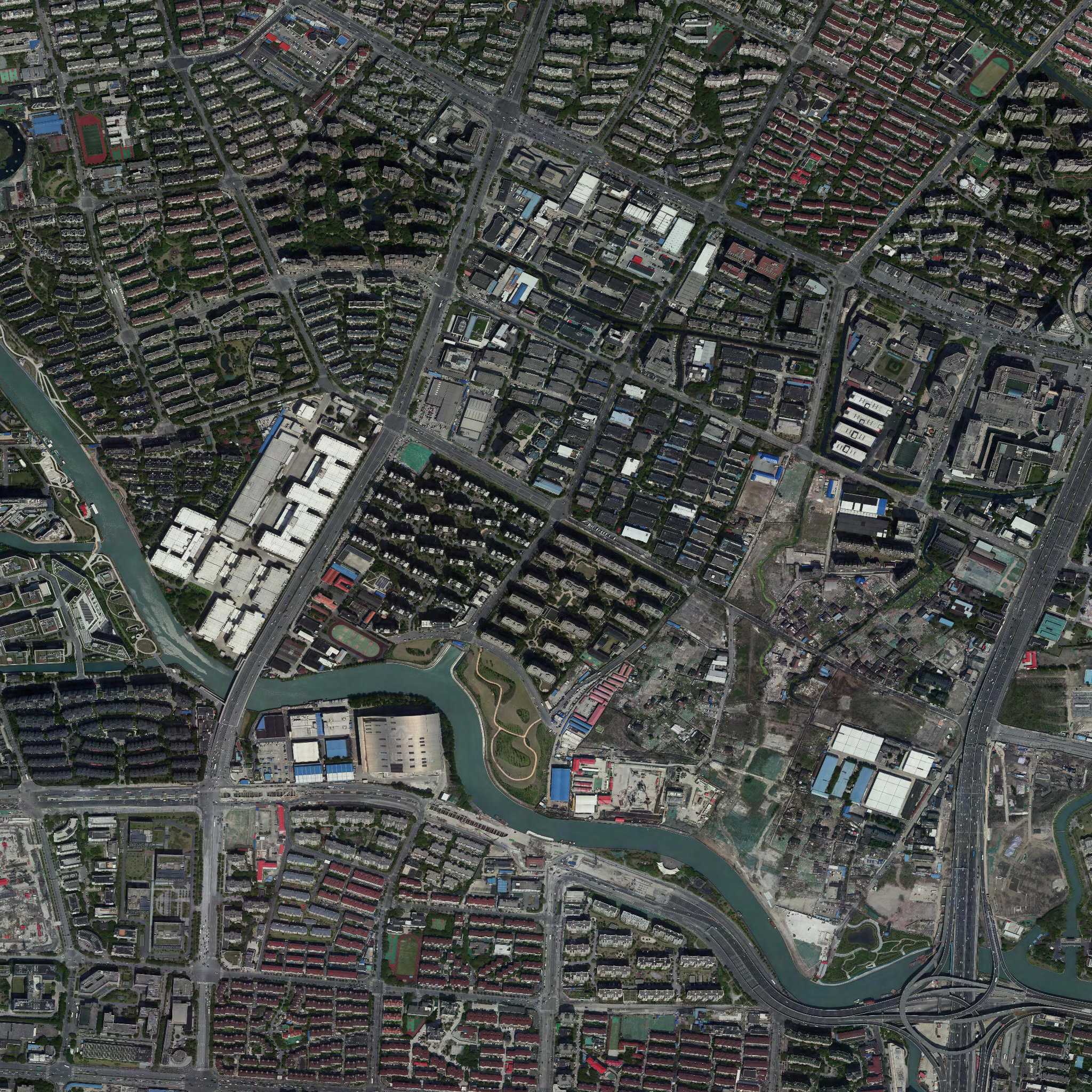}\hspace{-3mm}
				&\includegraphics[width=3.01cm, angle=0]{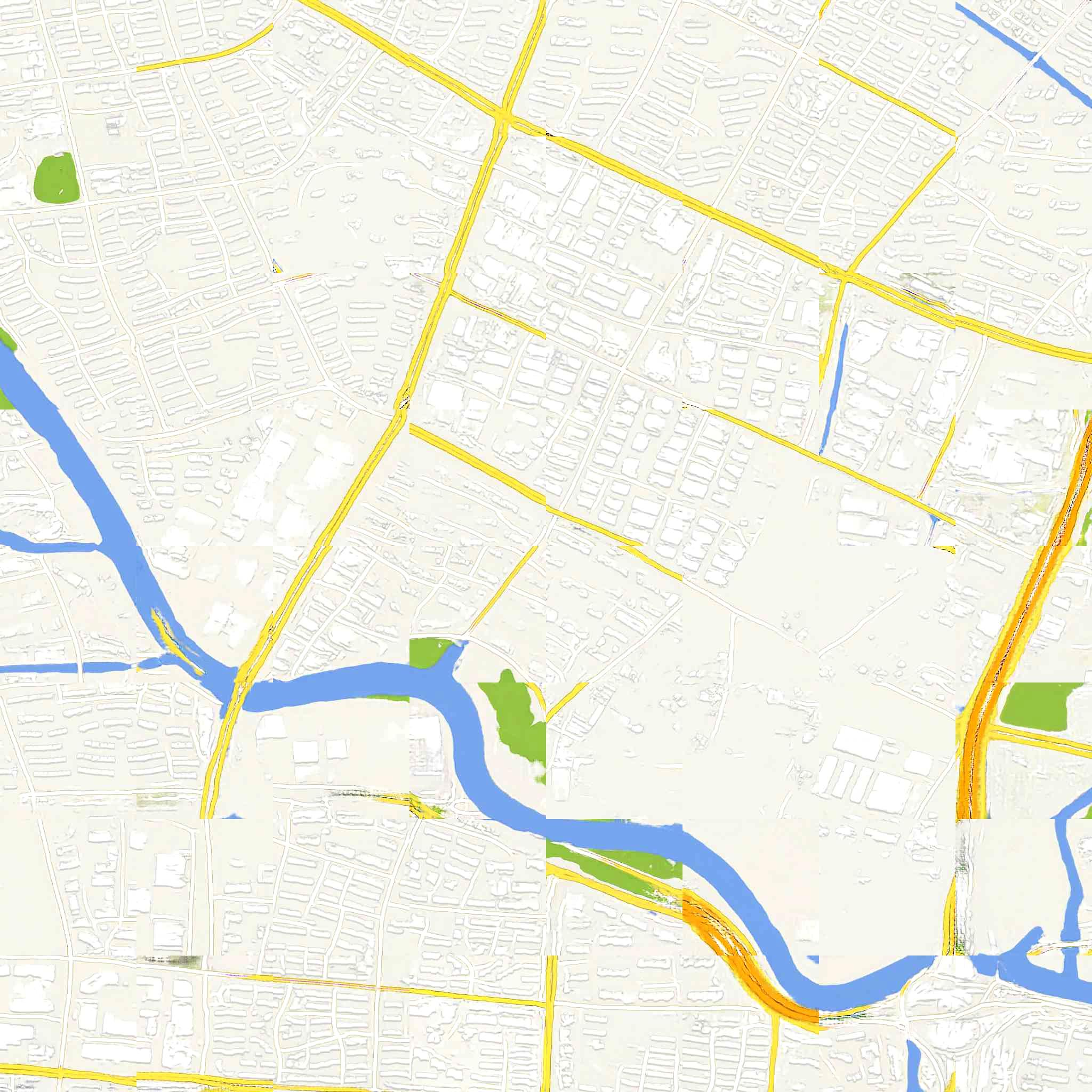}\hspace{-3mm}
				&\includegraphics[width=3.01cm, angle=0]{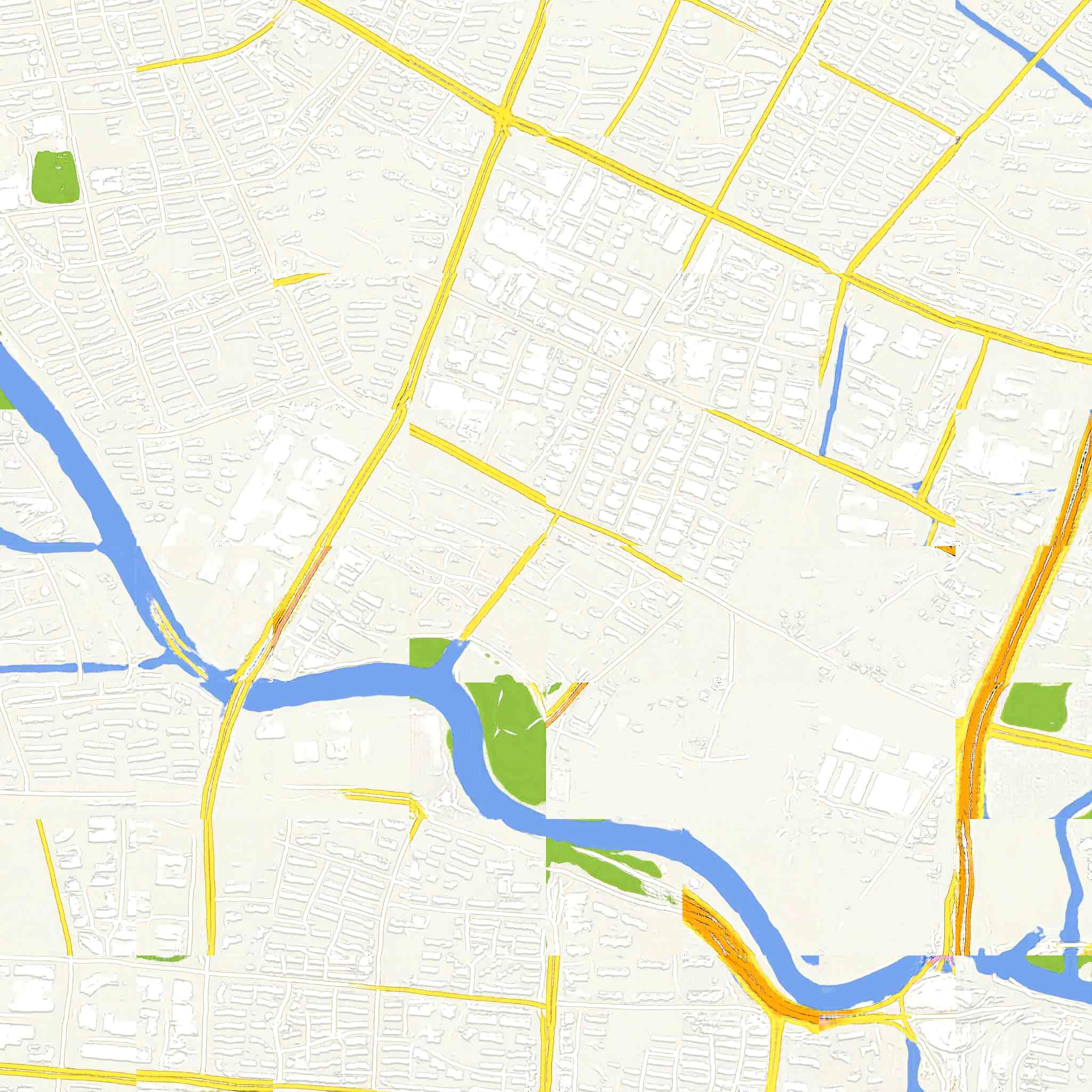}\hspace{-3mm}
				&\includegraphics[width=3.01cm, angle=0]{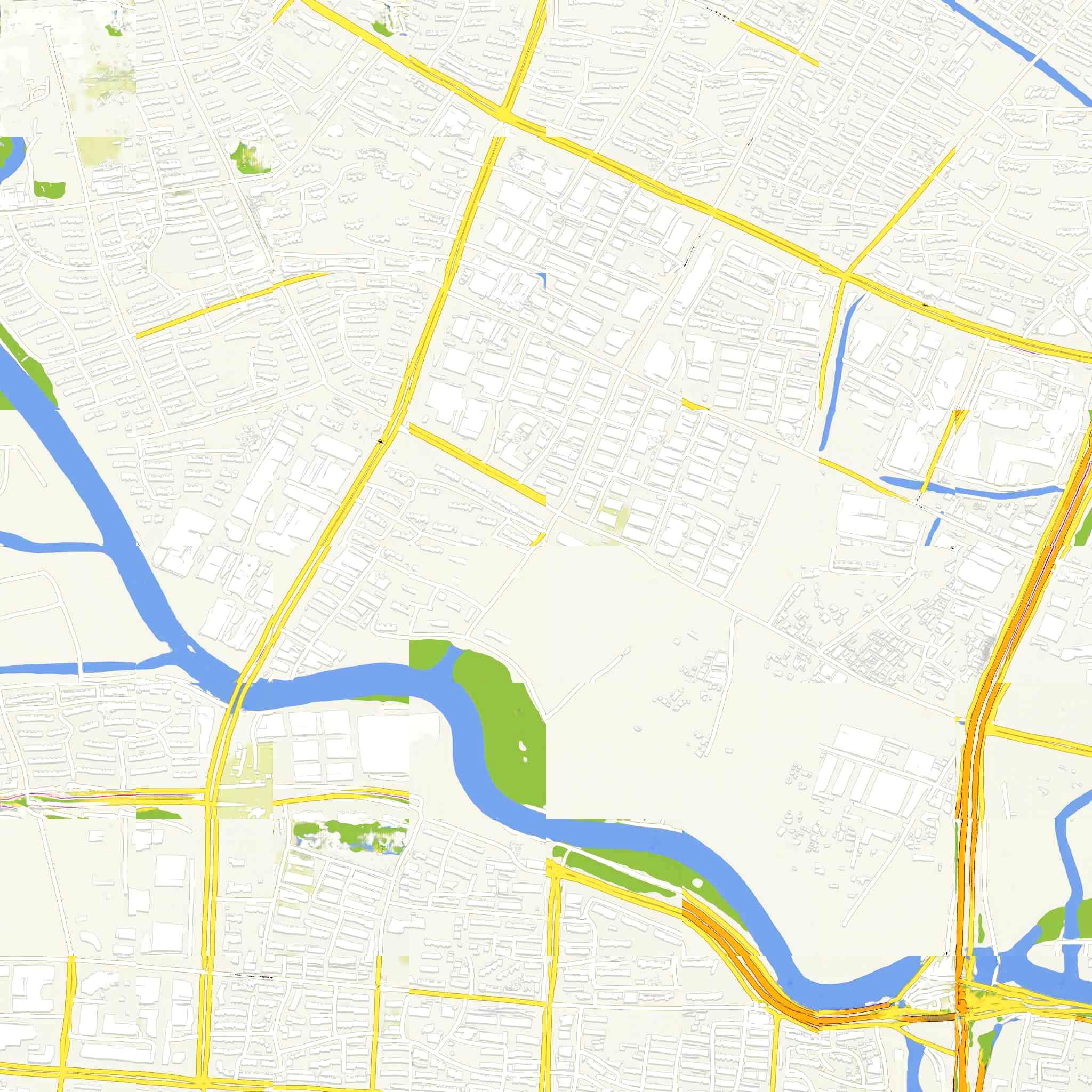}\hspace{-3mm}
				&\includegraphics[width=3.01cm, angle=0]{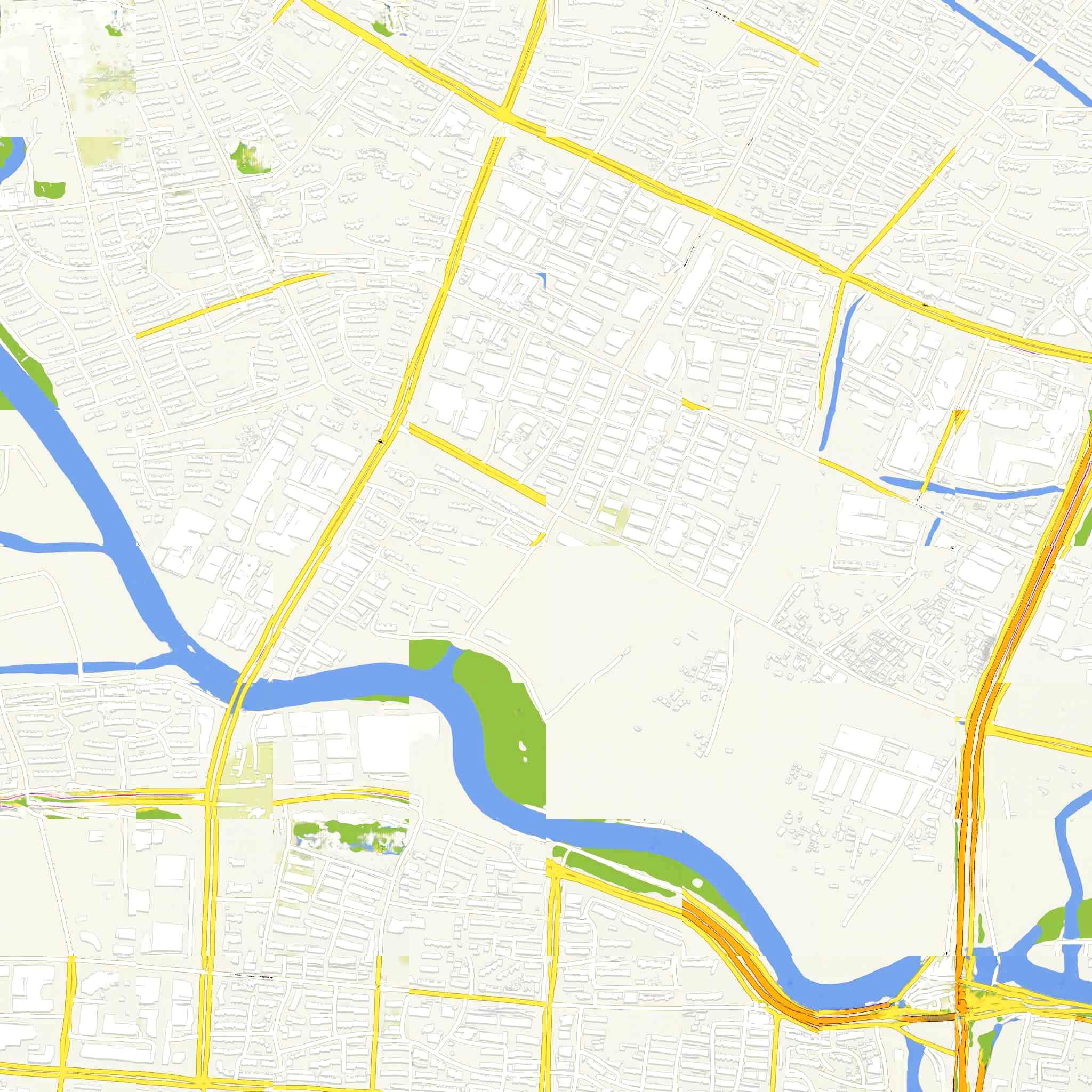}\hspace{-3mm}
				&\includegraphics[width=3.01cm, angle=0]{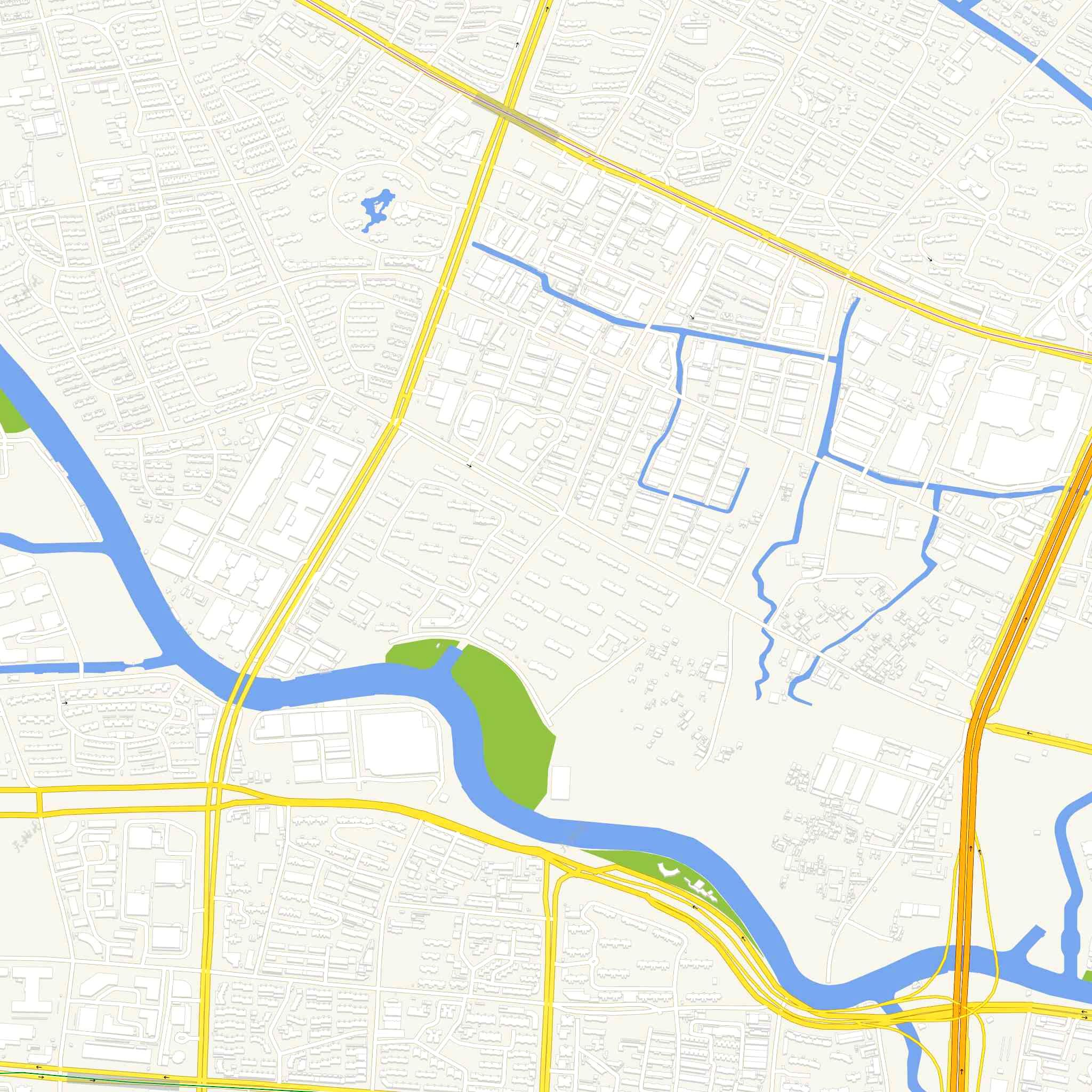}
				\begin{sideways}\qquad $2048\times 2048$\end{sideways}
				\\
				(a) Aerial image & (b) Drawer module  & (c) Drawer module
				& (d) Drawer module
				& (e) Drawer module
				& (f) Ground truth
				\\&&+Level ID&+Level ID& +Level ID &
				\\&&&+Semantic module& +Semantic module
				\\&&&(LAMG)& +CAMG
				
		\end{tabular}}
\else
\includegraphics{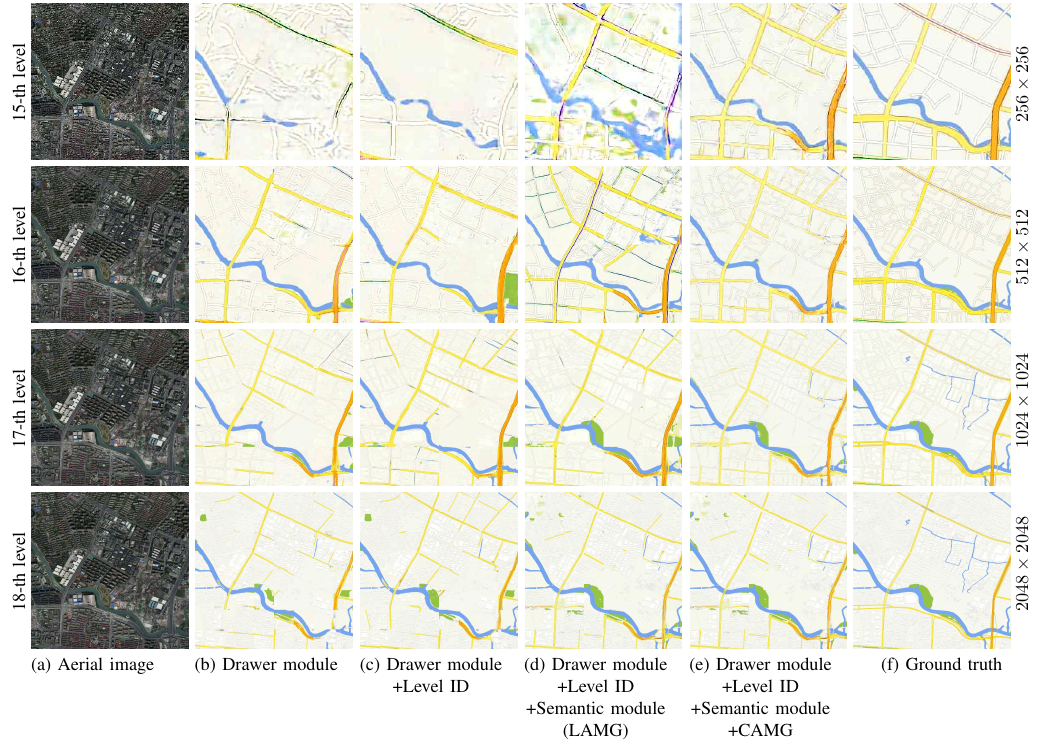}
\fi

	\end{center}
	\vspace{-4mm}
	\caption{
	{\color{black}Visualized results for each part in our pipeline. It can be seen that the accuracy and content consistency gradually improves from column (b) to column (e). LAMG indicates the Level-Aware Map Generator and CAMG indicates the Consistency-Aware Map Generator.}
	}
	\vspace{+2mm}
	\label{fig:Ablation Result}
\end{figure*}

\section{Related Work}
In this section, we review the traditional cartographic process, map generation methods and the progress of datasets for map generation. 

\vspace{+1mm}
\noindent{\bf Traditional Cartographic Process.}
As an effective tool for humans to explore and recognize the world, the map has been widely involved in our daily life for a long time. 
Online maps have immense influence for they can provide a realistic view of the world to millions of web users \cite{2013onlinemapapplication, 2017reviewmapping, 2019onlinemapservices}. 
Their production and update are done mainly in two steps, \ie, collecting vector data and visualizing the data in an appropriate way. 

The collection of vector data usually relies on surveying \cite{2009mapgps}, \eg, data collectors need to walk or drive with professional equipment to check every point of interest and collect relevant data, which is labor-intensive and time-consuming. 
This process is easily limited by weather, ground surface condition of the city, and data collectors themselves, which may lead to inaccurate or even impossible data collection. 
After the vector data is collected or updated, a team of cartographic experts is needed to process the raw data, solve the problems that cannot be done automatically \cite{haunold1993keystroke, ablameyko1993computer, 1998cartographic, 1991mapgen}, and finally convert it to the latest map tiles at multiple zoom levels with professional digital tools \cite{1995interactivemap, 2020cartography}. 
In summary, the traditional cartographic process ensures the accuracy of multi-level maps, but the high cost both in labor and time makes it hard to update timely and limits its applications. 

\vspace{+1mm}
\noindent{\bf Generating Maps from Aerial Images.}
Aerial images have found wide applications in human life~\cite{fuying-2020-remotesensing, fuying-2021-remotesensing, fu2021bidirectional, translation2022}.
They inherently contain rich information about the ground surface. Compared with vector data, aerial images are much easier and cheaper to obtain, \eg, airplanes, satellites and drones are able to collect aerial images and keep them up-to-date.

{\color{black}
Existing works like dense-global-residual network \cite{wu2020automatic}, bias U-Net \cite{chen2021reconstruction} and road-extractor model \cite{pan2021generic} focus on extracting road from aerial images.
They are based on semantic segmentation and can achieve decent performance, but they are unable to generate maps.} 
{\color{black}
By formulating map generation as an image-to-image translation task \cite{pix2pix}, there are also many methods, like Pix2Pix \cite{pix2pix}, Pix2PixHD \cite{pix2pixhd}, CycleGAN \cite{cyclegan}, GeoGAN \cite{2019geogan}, and SMAPGAN \cite{2020smapgan}, that can generate images in map style from aerial images based on generative adversarial networks~\cite{2014gan}. }
{\color{black}These methods treat the task of map generation as a simple image translation task, trying to generate maps from aerial images.}
These methods focus on generating images with diverse and realistic details but do not explicitly understand the content in aerial images, which may result in inaccurate map generation. 

{\color{black}
Besides, some researchers have presented map generation methods (\eg, GeoGAN~\cite{2019geogan} and SMAPGAN~\cite{2020smapgan}) on single zoom level based on image-to-image translation techniques~\cite{pix2pix, cyclegan}. 
}
Ganguli \etal~\cite{2019geogan} utilize conditional GAN~\cite{cgan} with reconstruction and style loss to convert aerial images to human-readable maps, while Chen \etal~\cite{2020smapgan} combine the supervised and unsupervised image-to-image translation methods \cite{pix2pix, cyclegan} to improve the map generation quality with limited training samples. 
These methods only consider generating maps on a single zoom level instead of widely used multi-level maps which are more effective in practical usage, and do not consider the connection between maps at different zoom levels. 
These easily cause content inconsistency when directly applied to multi-level map generation. 

\vspace{+1mm}
\noindent{\bf Datasets for Map Generation.}
The development of the map dataset is still in its early stage. 
Although all online tiled map services provide tiled maps with multiple zoom levels instead of single zoom level for better usage, there is no specialized dataset for multi-level map generation. 
And there seldom exist datasets \cite{2019multiscalemap, pix2pix} that are related to map generation.
The dataset in~\cite{2019multiscalemap} provides maps but has no corresponding aerial images. 
The dataset in~\cite{pix2pix} only provides paired samples of aerial images and maps at a single zoom level. 
To push the development of multi-level map generation forward, we collect and produce the first Multi-Level Map (MLM) dataset.

\section{Method}
In this section, we first formulate the problem and introduce the motivation, then we describe the level-aware map generator and consistency-aware map generator in our framework. 
Finally, learning details of our method are provided.

\subsection{Formulation and Motivation}
Multi-level map generation is a task that generates map tiles at multiple zoom levels from an aerial image pyramid, which are obtained from a single high-resolution aerial image by a series of $2\times$ downsampling that corresponds to scales of map zoom levels. 
Though map at each zoom level is different from each other (\eg, a map at higher zoom level has higher resolution and is able to show more roads in detail), they keep their content consistent because they represent the same corresponding area.

In multi-level maps, the world map at the $k$-th zoom level usually consists of $2^{k} \times 2^{k}$ map tiles, \ie, the number of map tiles at the $k$-th zoom level decreases exponentially when $k$ lowers. 
And each map tile usually has $256 \times 256$ pixels, thus the world map at the $(k+1)$-th zoom level has a double resolution of that at the $k$-th zoom level. 
In this way, the map at higher zoom level shows more details and the map at lower zoom level provides wider view. 

{\color{black}Previous map generation methods, like Pix2Pix \cite{pix2pix}, Pix2PixHD \cite{pix2pixhd}, CycleGAN \cite{cyclegan}, GeoGAN \cite{2019geogan}, and SMAPGAN \cite{2020smapgan}, mainly focus on generating photo-realistic images or images in map style. They do not consider multi-level map generation and fail to generate accurate and consistent multi-level maps.} 
To solve this, we propose a multi-level map generation framework consisting of a level-aware map generator and a consistency-aware map generator. 
The former generator can generate accurate initial maps for each zoom level and the latter one can guarantee content consistency between maps at different zoom levels. 
The overview of our proposed method is illustrated in Figure~\ref{fig:overview}{\color{black}(a)}.

\subsection{Model Sturcture}
\label{sec:model}
\begin{figure}[tb]
	\begin{center}
		\includegraphics[width=8cm,angle=0]{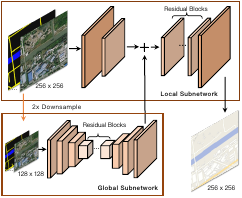}
	\end{center}
	\vspace{-0mm}
	\caption{Illustration of drawer module. 
	``+'' indicates pixel-wise addition.
	More details are provided in Table \ref{tab:model_1}, and \ref{tab:model_1_1}.}
	\label{fig1}
	\vspace{-0mm}
\end{figure}

{\color{black}As shown in Figure~\ref{fig:overview}, our framework consists of two generators, level-aware map generator (LAMG) and consistency-aware map generator (CAMG).
They share the same model structure, i.e., a semantic module and a drawer module. 

In this section, we first introduce the semantic module, and then details of the drawer module are described. 
In addition, we also introduce the structure of the discriminator that is utilized for adversarial learning.

\vspace{+1mm}
\noindent{\bf Semantic Module.}
The semantic module based on DeepLabv3+~\cite{deeplabv3+} and Xception-65~\cite{xception} serves as backbone. 
Apart from its semantic segmentation results, we also take its $4\times$ upsampled final feature maps as input of the drawer module for providing more semantic information.

\vspace{+1mm}
\noindent{\bf Drawer Module.}
We design the drawer module with a coarse-to-fine encoder-decoder structure~\cite{pix2pixhd}. 
The drawer module consists of a local subnetwork and a global subnetwork.
The global subnetwork focuses on the whole image and the local subnetwork focuses on the details of the image. 
The structure of the drawer module is shown in Figure \ref{fig1}.
We input the original image and the $2\times$ downsampled image to the local subnetwork and global subnetwork respectively, their features are fused by pixel-wise addition. 
The detail configurations of local subnetwork and global subnetwork are illustrated in Table \ref{tab:model_1} and Table \ref{tab:model_1_1}.

\vspace{+1mm}
\noindent{\bf Discriminator.}
We adopt a multi-scale discriminator~\cite{pix2pixhd} to enhance the ability to differentiate real and synthesized maps at different zoom levels. 
It consists of three discriminators, which share the same structure, as shown in Table~\ref{tab:discriminator}.
We refer to these discriminators as D1, D2 and D3, which operate at different image scales.  
D1 takes the original images as input and D2, D3 take $2\times$, $4\times$ downsampled images as input respectively. 
In this way, discriminators can have different receptive fields and focus on details at different scales that benefits the multi-level map generation. }

\subsection{Level-Aware Map Generator}
As shown in Figure~\ref{fig:overview}{\color{black}(b)}, the level-aware map generator consists of two modules, \ie, the semantic module and drawer module. 
To ensure accuracy, we employ a semantic module to explicitly extract pixel-wise semantic information from aerial images, and the drawer module is designed for drawing maps with realistic details.

To avoid the performance degradation caused by domain gap between maps at different zoom levels, we input level identification along with an aerial image to help the generator be level-aware, so as to generate high-quality maps for each zoom level. 
Thus, the level-aware map generator can be represented by
\begin{equation}
	y_{k}=g_\phi(x_{k},f_{\theta}(x_{k}),k), 
\end{equation}
where $x_{k}$, $y_{k}$ represent aerial image and initial generated map at the $k$-th zoom level, while $g_{\phi}$ and $f_{\theta}$ indicate semantic module and drawer module of level-aware map generator.

\newcommand{\tabincell}[2]{\begin{tabular}{@{}#1@{}}#2\end{tabular}}
\begin{table}[]
	\centering
	\scalebox{0.75}{
	\begin{tabular}{ccccc}
		\toprule[1.5pt]
		Layer &Layer Information&Norm&Activation& Input $\rightarrow$ Output Shape   \\ \midrule
		Conv&Conv(K7$\times$7, S1, P3)&IN&ReLU&($\frac{h}{2}$, $\frac{w}{2}$, N)$\rightarrow$($\frac{h}{2}$, $\frac{w}{2}$, 128)\\ \midrule
		Conv&Conv(K3$\times$3, S2, P1)&IN&ReLU&($\frac{h}{2}$, $\frac{w}{2}$, 128)$\rightarrow$($\frac{h}{4}$, $\frac{w}{4}$, 256)\\ \midrule
		Conv&Conv(K3$\times$3, S2, P1)&IN&ReLU&($\frac{h}{4}$, $\frac{w}{4}$, 256)$\rightarrow$($\frac{h}{8}$, $\frac{w}{8}$, 512)\\ \midrule
		Conv&Conv(K3$\times$3, S2, P1)&IN&ReLU&($\frac{h}{8}$, $\frac{w}{8}$, 512)$\rightarrow$($\frac{h}{16}$, $\frac{w}{16}$, 1024)\\ \midrule
		ResBlk&\tabincell{c}{Conv(K3$\times$3, S1, P1)\\Conv(K3$\times$3, S1, P1)}&\tabincell{c}{IN\\IN}&\tabincell{c}{ReLU\\-}&($\frac{h}{16}$, $\frac{w}{16}$, 1024)$\rightarrow$($\frac{h}{16}$, $\frac{w}{16}$, 1024)\\ \midrule
		ResBlk&\tabincell{c}{Conv(K3$\times$3, S1, P1)\\Conv(K3$\times$3, S1, P1)}&\tabincell{c}{IN\\IN}&\tabincell{c}{ReLU\\-}&($\frac{h}{16}$, $\frac{w}{16}$, 1024)$\rightarrow$($\frac{h}{16}$, $\frac{w}{16}$, 1024)\\ \midrule
		ResBlk&\tabincell{c}{Conv(K3$\times$3, S1, P1)\\Conv(K3$\times$3, S1, P1)}&\tabincell{c}{IN\\IN}&\tabincell{c}{ReLU\\-}&($\frac{h}{16}$, $\frac{w}{16}$, 1024)$\rightarrow$($\frac{h}{16}$, $\frac{w}{16}$, 1024)\\ \midrule
		ResBlk&\tabincell{c}{Conv(K3$\times$3, S1, P1)\\Conv(K3$\times$3, S1, P1)}&\tabincell{c}{IN\\IN}&\tabincell{c}{ReLU\\-}&($\frac{h}{16}$, $\frac{w}{16}$, 1024)$\rightarrow$($\frac{h}{16}$, $\frac{w}{16}$, 1024)\\ \midrule
		ResBlk&\tabincell{c}{Conv(K3$\times$3, S1, P1)\\Conv(K3$\times$3, S1, P1)}&\tabincell{c}{IN\\IN}&\tabincell{c}{ReLU\\-}&($\frac{h}{16}$, $\frac{w}{16}$, 1024)$\rightarrow$($\frac{h}{16}$, $\frac{w}{16}$, 1024)\\ \midrule
		ResBlk&\tabincell{c}{Conv(K3$\times$3, S1, P1)\\Conv(K3$\times$3, S1, P1)}&\tabincell{c}{IN\\IN}&\tabincell{c}{ReLU\\-}&($\frac{h}{16}$, $\frac{w}{16}$, 1024)$\rightarrow$($\frac{h}{16}$, $\frac{w}{16}$, 1024)\\ \midrule
		ResBlk&\tabincell{c}{Conv(K3$\times$3, S1, P1)\\Conv(K3$\times$3, S1, P1)}&\tabincell{c}{IN\\IN}&\tabincell{c}{ReLU\\-}&($\frac{h}{16}$, $\frac{w}{16}$, 1024)$\rightarrow$($\frac{h}{16}$, $\frac{w}{16}$, 1024)\\ \midrule
		ResBlk&\tabincell{c}{Conv(K3$\times$3, S1, P1)\\Conv(K3$\times$3, S1, P1)}&\tabincell{c}{IN\\IN}&\tabincell{c}{ReLU\\-}&($\frac{h}{16}$, $\frac{w}{16}$, 1024)$\rightarrow$($\frac{h}{16}$, $\frac{w}{16}$, 1024)\\ \midrule
		ResBlk&\tabincell{c}{Conv(K3$\times$3, S1, P1)\\Conv(K3$\times$3, S1, P1)}&\tabincell{c}{IN\\IN}&\tabincell{c}{ReLU\\-}&($\frac{h}{16}$, $\frac{w}{16}$, 1024)$\rightarrow$($\frac{h}{16}$, $\frac{w}{16}$, 1024)\\ \midrule
		DeConv&DeConv(K3$\times$3, S2, P1)&IN&ReLU&($\frac{h}{16}$, $\frac{w}{16}$, 1024)$\rightarrow$($\frac{h}{8}$, $\frac{w}{8}$, 512)\\ \midrule
		DeConv&DeConv(K3$\times$3, S2, P1)&IN&ReLU&($\frac{h}{8}$, $\frac{w}{8}$, 512)$\rightarrow$($\frac{h}{4}$, $\frac{w}{4}$, 256)\\ \midrule
		DeConv&DeConv(K3$\times$3, S2, P1)&IN&ReLU&($\frac{h}{4}$, $\frac{w}{4}$, 256)$\rightarrow$($\frac{h}{2}$, $\frac{w}{2}$, 128)\\ 
		\bottomrule[1.5pt]
	\end{tabular}	
	}
	\vspace{-0mm}
	\caption{Detail configuration of global sub-network in drawer module. 
	Conv: convolutional layer, DeConv: deconvolutional layer, 
	N: the number of input channels, K: kernel size, S: stride size, P: padding size, IN: instance normalization.}
	\label{tab:model_1}
\end{table}%
\begin{table}[]
	\centering
	\scalebox{0.8}{
	\begin{tabular}{ccccc}
		\toprule[1.5pt]
		Layer &Layer Information&Norm&Activation& Input $\rightarrow$ Output Shape \\ \midrule
		Conv&Conv(K7$\times$7, S1, P3)&IN&ReLU&($h$, $w$, N)$\rightarrow$($h$, $w$, 64)\\ \midrule
		Conv&Conv(K3$\times$3, S2, P1)&IN&ReLU&($h$, $w$, 64)$\rightarrow$($\frac{h}{2}$, $\frac{w}{2}$, 128)\\ \midrule
		ResBlk&\tabincell{c}{Conv(K3$\times$3, S1, P1)\\Conv(K3$\times$3, S1, P1)}&\tabincell{c}{IN\\IN}&\tabincell{c}{ReLU\\-}&($\frac{h}{2}$, $\frac{w}{2}$, 128)$\rightarrow$($\frac{h}{2}$, $\frac{w}{2}$, 128)\\ \midrule
		ResBlk&\tabincell{c}{Conv(K3$\times$3, S1, P1)\\Conv(K3$\times$3, S1, P1)}&\tabincell{c}{IN\\IN}&\tabincell{c}{ReLU\\-}&($\frac{h}{2}$, $\frac{w}{2}$, 128)$\rightarrow$($\frac{h}{2}$, $\frac{w}{2}$, 128)\\ \midrule
		ResBlk&\tabincell{c}{Conv(K3$\times$3, S1, P1)\\Conv(K3$\times$3, S1, P1)}&\tabincell{c}{IN\\IN}&\tabincell{c}{ReLU\\-}&($\frac{h}{2}$, $\frac{w}{2}$, 128)$\rightarrow$($\frac{h}{2}$, $\frac{w}{2}$, 128)\\ \midrule
		DeConv&DeConv(K3$\times$3, S2, P1)&IN&ReLU&($\frac{h}{2}$, $\frac{w}{2}$, 128)$\rightarrow$($h$, $w$, 64)\\ \midrule
		Conv&Conv(K7$\times$7, S1, P3)&-&Tanh&($h$, $w$, 64)$\rightarrow$($h$, $w$, 3)\\ 
		\bottomrule[1.5pt]
	\end{tabular}	
	}
	\vspace{-0mm}
	\caption{Detail configuration of local subnetwork in drawer module. 
	Conv: convolutional layer, DeConv: deconvolutional layer, 
	N: the number of input channels, K: kernel size, S: stride size, P: padding size, IN: instance normalization.}
	\label{tab:model_1_1}
\end{table}%
\begin{table}[]
	\centering
	\scalebox{0.69}{
	\begin{tabular}{ccccc}
		\toprule[1.5pt]
		Layer &Layer Information&Norm&Activation& Input $\rightarrow$ Output Shape \\ \midrule
		Conv&Conv(K4$\times$4, S2, P2)&-&LeakyReLU&($h$, $w$, 6)$\rightarrow$($\frac{h}{2}+2$, $\frac{w}{2}+2$, 64)\\ \midrule
		Conv&Conv(K4$\times$4, S2, P2)&IN&LeakyReLU&($\frac{h}{2}+2$, $\frac{w}{2}+2$, 64)$\rightarrow$($\frac{h}{4}+3$, $\frac{w}{4}+3$, 128)\\ \midrule
		Conv&Conv(K4$\times$4, S2, P2)&IN&LeakyReLU&($\frac{h}{4}+3$, $\frac{w}{4}+3$, 128)$\rightarrow$($\frac{h}{8}+3$, $\frac{w}{8}+3$, 256)\\ \midrule
		Conv&Conv(K4$\times$4, S1, P2)&IN&LeakyReLU&($\frac{h}{8}+3$, $\frac{w}{8}+3$, 256)$\rightarrow$($\frac{h}{8}+6$, $\frac{w}{8}+6$, 512)\\ \midrule
		Conv&Conv(K4$\times$4, S1, P2)&-&-&($\frac{h}{8}+6$, $\frac{w}{8}+6$, 512)$\rightarrow$($\frac{h}{8}+9$, $\frac{w}{8}+9$, 1)\\ 
		\bottomrule[1.5pt]
	\end{tabular}
	}
	\vspace{-0mm}
	\caption{Detail configuration of discriminator in multi-scale discriminator. 
	Conv: convolutional layer, DeConv: deconvolutional layer, 
	K: kernel size, S: stride size, P: padding size, IN: instance normalization. }
	\label{tab:discriminator}
\end{table}%

{\color{black}
\vspace{+1mm}
\noindent{\bf Semantic Module.}
To display a realistic view of the world and correct cartographic information, high-quality generated maps should not only be images in map style but also provide accurate semantic information (\eg, the locations of roads and rivers). 
Existing single-level map generation methods \cite{2019geogan, 2020smapgan} focus on generating images with realistic details and do not explicitly force the model to learn and understand the content of the input image pixel-wisely, which may lead to inferior and erroneous map generation. 
To better keep the topological relationship among geographical elements and help the drawer module to generate maps with high accuracy, we design a semantic module that can learn to assign each pixel a semantic category (\eg, road, water area and land background) and understand the aerial image pixel-wisely.
The semantic module is based on the semantic segmentation model DeepLabv3+ \cite{deeplabv3+}, we choose it for its stable performance and easy training procedure. 

The semantic module can provide pixel-wise semantic information whose role is similar to vector data in the traditional cartographic process. 
It improves the accuracy of the generated map and keeps the right topological relations of map elements as shown in Figure~\ref{fig:Ablation Result}(c) and Figure~\ref{fig:Ablation Result}(d).
}

{\color{black}
\vspace{+1mm}
\noindent{\bf Drawer Module.}
The drawer module is the basic module of the generator. 
As shown in Figure~\ref{fig:Ablation Result}(b), it can learn to generate maps with realistic details independently by adversarial learning~\cite{2014gan, pix2pix}. 
To optimize the quality of map details, it also takes aerial images along with level identification as input (see Figure~\ref{fig:Ablation Result}(c)).
To improve the generation accuracy, it takes feature maps and semantic segmentation results from the semantic module as input and translates them to map with accurate geographical information (see Figure~\ref{fig:Ablation Result}(d)).

Its role is similar to cartographic experts in the traditional cartographic process. 
On the basis of pixel-wise semantic information of aerial images from the semantic module, the drawer module can learn to draw a realistic map with accurate information by taking semantic segmentation, zoom level and aerial image into consideration.
}

{\color{black}
\vspace{+1mm}
\noindent{\bf Level Identification.}
Compared with single-level map generation, multi-level map generation is more challenging because it has to generate maps from aerial images for multiple zoom levels, and keeps their content consistent.
A naive solution is to learn a mapping model for each zoom level, but it is inconvenient and makes the training procedure complicated. 
Moreover, the amount of training samples for each zoom level decreases exponentially when the zoom level lowers, which results in insufficient samples for the training model at a low zoom level.

To solve this problem, we utilize level identification as guidance information and input it into the generator. 
The level identification is $\mathbb{R}^{1\times H\times W}$, which simply repeats the zoom level number $k$ for $H\times W$ times.
It guides the generator to generate an appropriate map for each zoom level, and all the samples from different zoom levels can be used for training the generator in a unified way, achieving better performance especially for generating maps at low zoom level, \eg, the river at the 15-th level in Figure~\ref{fig:Ablation Result}(c) is clearer than that in Figure~\ref{fig:Ablation Result}(b). 
}

{\color{black}
\subsection{Consistency-Aware Map Generator}
Level-aware map generator can generate an appropriate and accurate initial map for each zoom level. 
But due to the lack of connection between generated maps at different zoom levels, there may exist content inconsistency between maps at different zoom levels, \eg, as shown in Figure~\ref{fig:intro}{\color{black}(d)} and Figure~\ref{fig:intro}{\color{black}(e)}, generated map at the $16$-th zoom level shows river while land shows up in the same area of generated map at the $15$-th zoom level. 

Therefore, we employ the consistency-aware map generator in our framework and build the connection between generated maps at different zoom levels. 
As shown in Figure~\ref{fig:overview}{\color{black}(a)}, we view initial generated maps at $K$ different zoom levels as sequential data $\{y_1, y_2, ..., y_K\}$. 
The ``frame" in map sequence is different in shape, \ie, the height and width of the map at the $(k+1)$-th zoom level is twice the size of the map at the $k$-th zoom level.
Considering the higher resolution of map at higher zoom level is able to provide richer information for refining initial map of each level, we optimize the generation and refine multi-level map sequence from high to low zoom level in an iterative way as follows
\begin{equation}
\label{eq:CAMG}
y^{r}_{k}=g'_{\phi{'}}(f'_{\theta{'}}(x_{k}),y_{k}, y^{r}_{k+1}), 
\end{equation}
where $y^{r}_{k}$, $y^{r}_{k+1}$ are refined map at the $k$-th and the $(k+1)$-th zoom level, and
$g'_{\phi'}$, $f'_{\theta'}$ indicate semantic module and drawer module of the consistency-aware map generator.
Notice that we regard the initial map at highest zoom level $y_{K}$ as refined map $y_{K}^r$ for the refinement of $y_{K-1}$.
The refinement defined in Equation~\eqref{eq:CAMG} builds connection between adjacent zoom levels.
And by applying it iteratively, we can build connection across different zoom levels to ensure their consistency, \eg, the refined map $y_{k-1}^r$ can be obtained from $y_{k}$ and $y_{k+1}^r$ as follows
\begin{equation}
\label{eq:CAMG2}
y^{r}_{k-1}=g'_{\phi{'}}(f'_{\theta{'}}(x_{k-1}),y_{k-1}, g'_{\phi{'}}(f'_{\theta{'}}(x_{k}),y_{k}, y^{r}_{k+1})).
\end{equation}
It largely improves the accuracy of the generated map and keeps the content consistency between different levels as shown in Figure~\ref{fig:Ablation Result}(e).

The structure of the consistency-aware map generator is the same as the level-aware map generator, which also consists of a semantic module and a drawer module. 
The main difference is their input, \ie, the level-aware map generator takes an aerial image and level identification as input, while the consistency-aware map generator takes an aerial image, initial maps from the $k$-th and the $(k+1)$-th zoom level as input, as illustrated in Figure~\ref{fig:overview}{\color{black}(c)}. 
}

\begin{figure}[tb]
	\begin{center}
	
		\scalebox{0.55}{
			\begin{tabular}{cccccccccc}
				\includegraphics[width=3cm, angle=0]{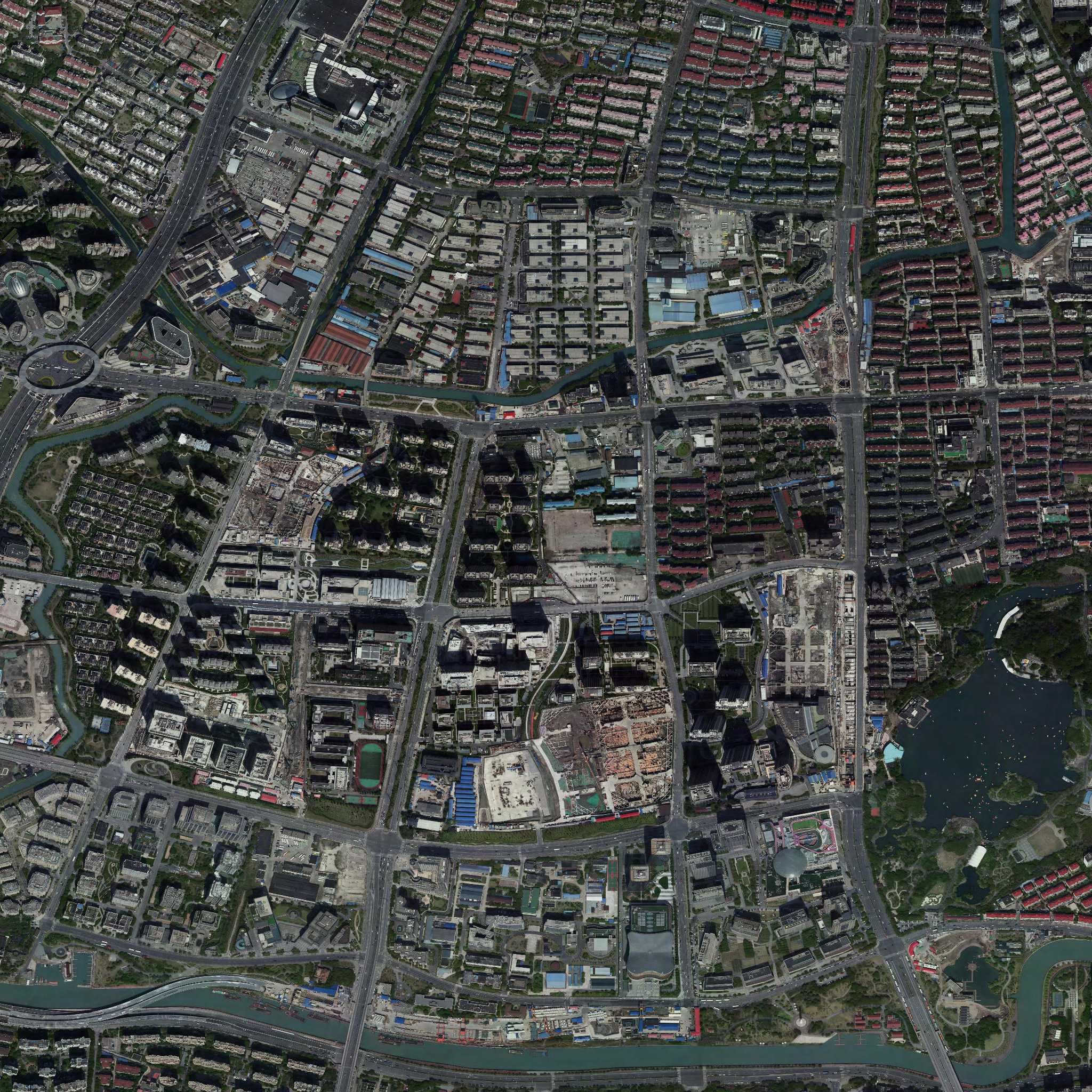}\hspace{-3mm}
				&\includegraphics[width=3cm, angle=0]{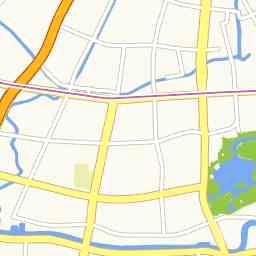}\hspace{-3mm}
				&\includegraphics[width=3cm, angle=0]{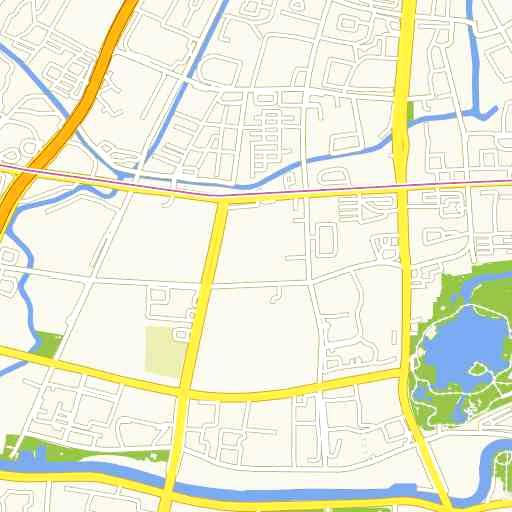}\hspace{-3mm}
				&\includegraphics[width=3cm, angle=0]{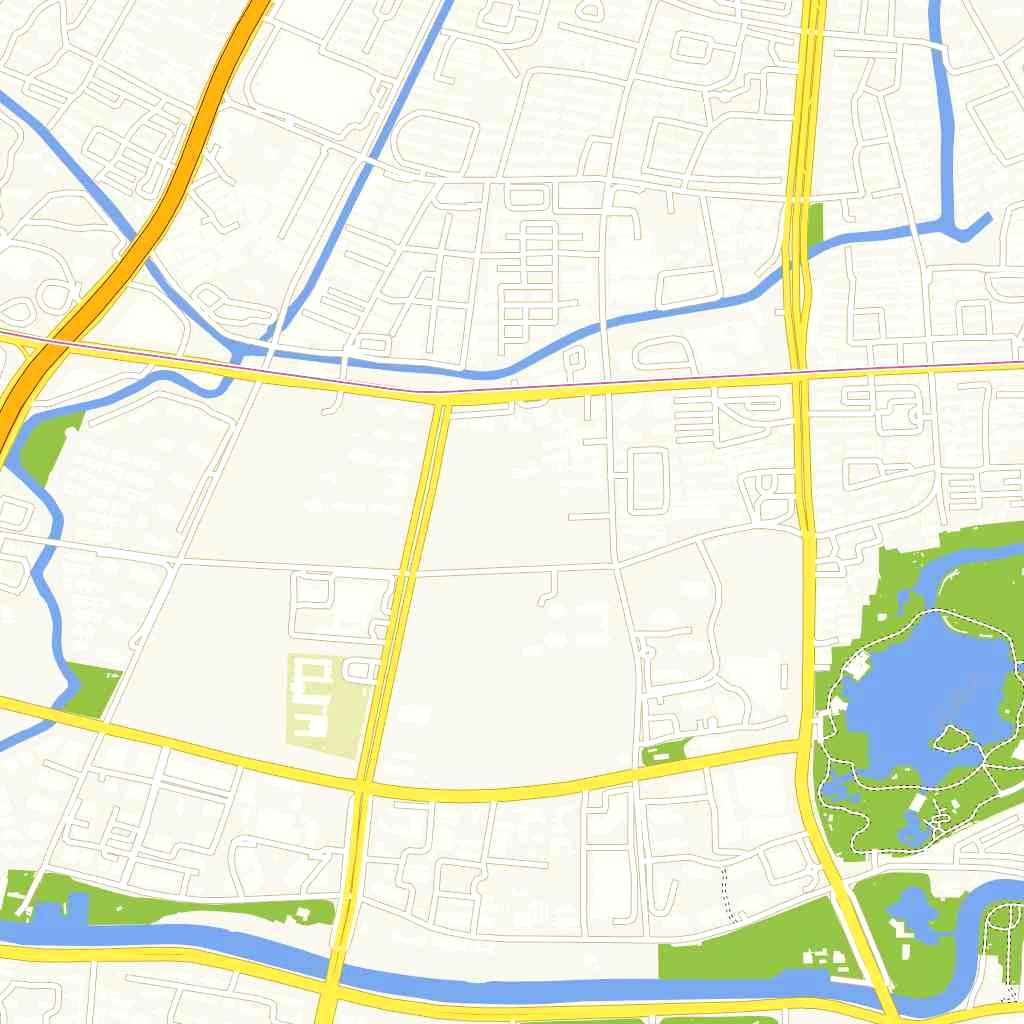}\hspace{-3mm}
				&\includegraphics[width=3cm, angle=0]{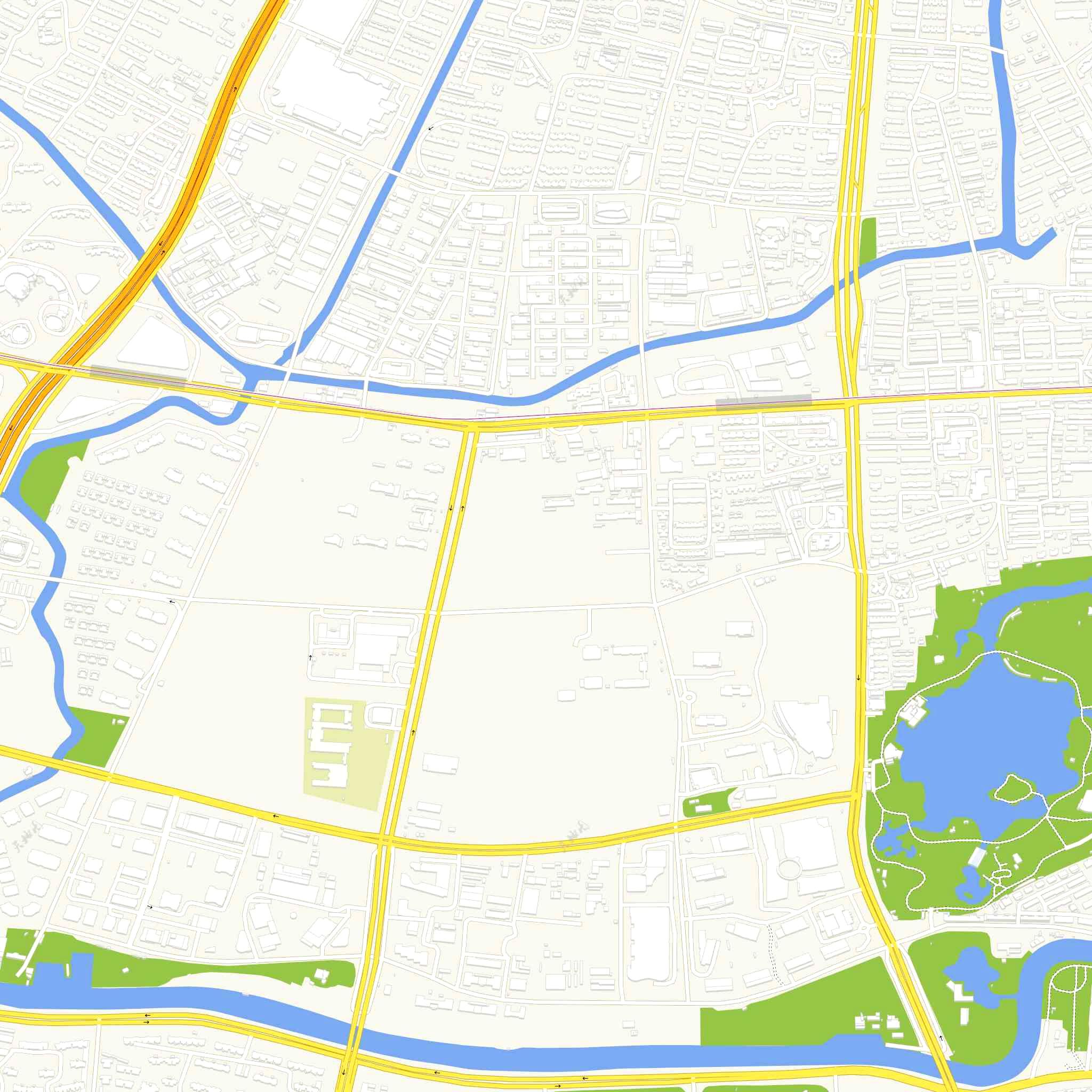}
				\\
				\includegraphics[width=3cm, angle=0]{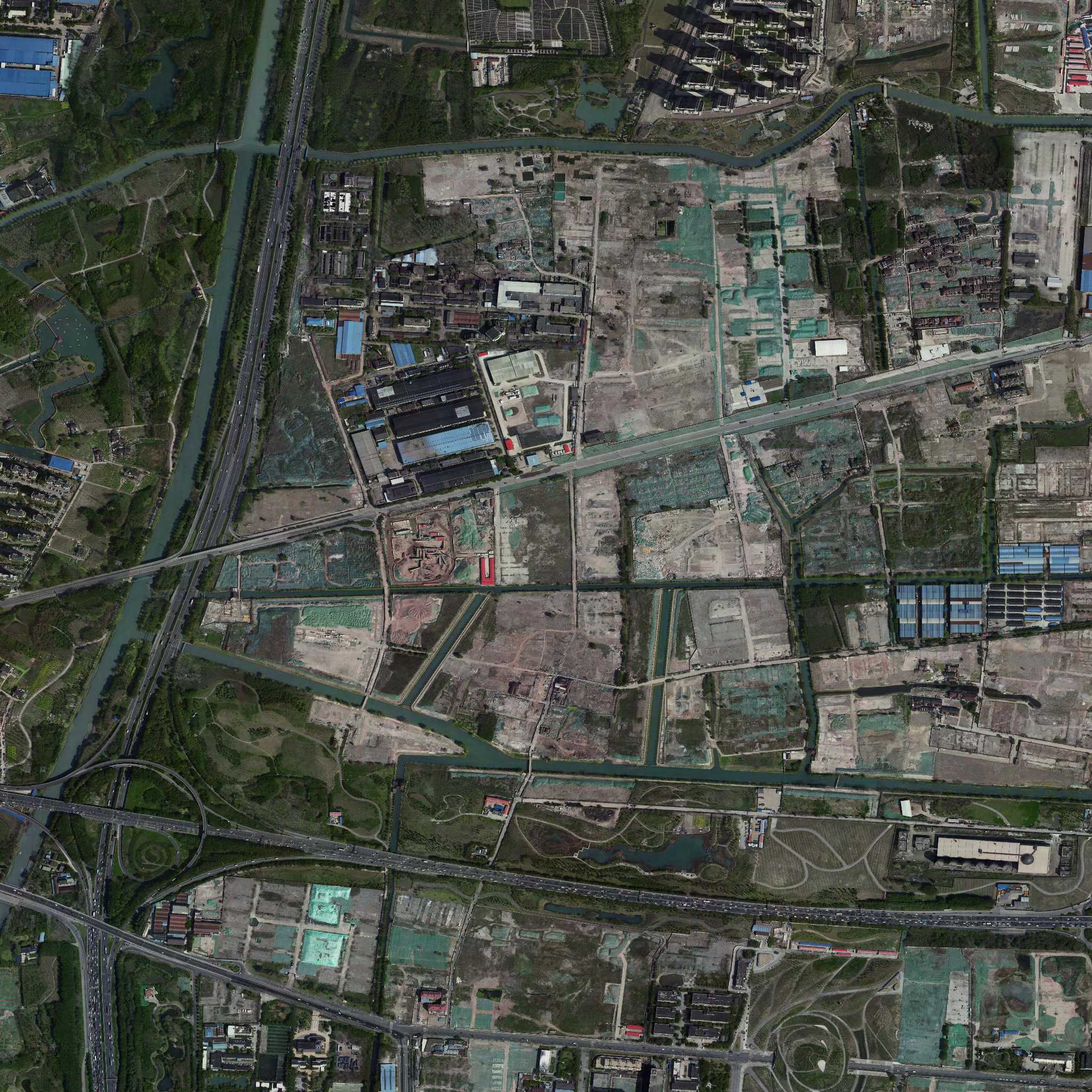}\hspace{-3mm}
				&\includegraphics[width=3cm, angle=0]{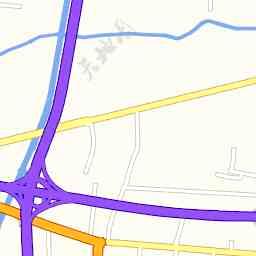}\hspace{-3mm}
				&\includegraphics[width=3cm, angle=0]{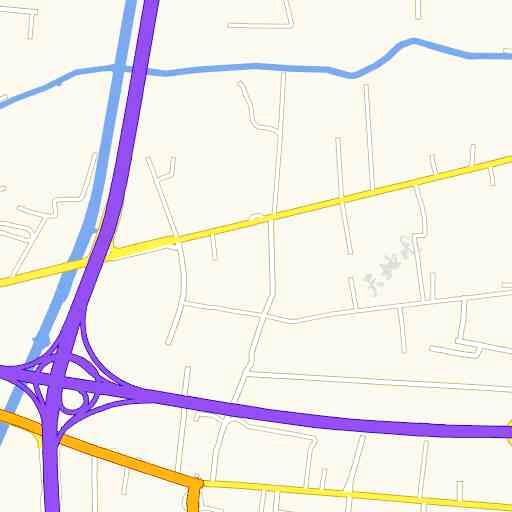}\hspace{-3mm}
				&\includegraphics[width=3cm, angle=0]{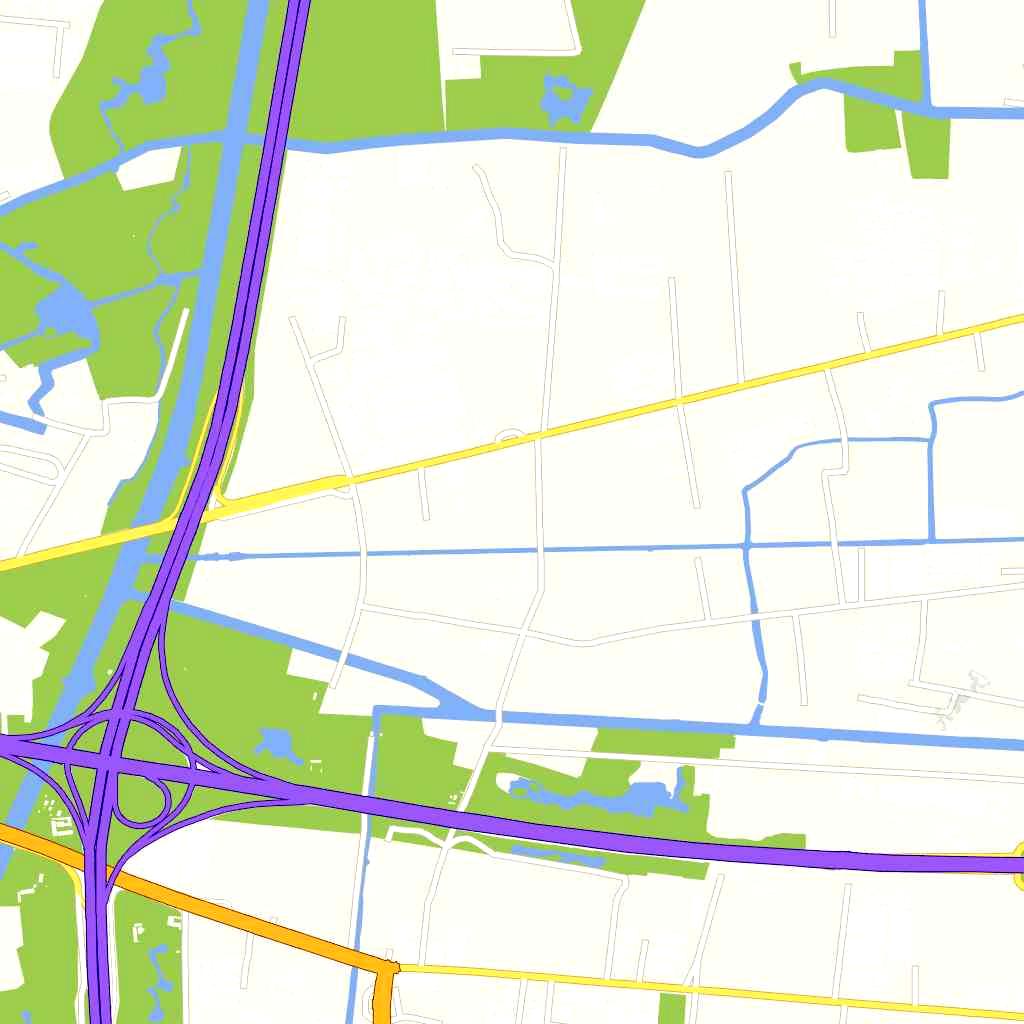}\hspace{-3mm}
				&\includegraphics[width=3cm, angle=0]{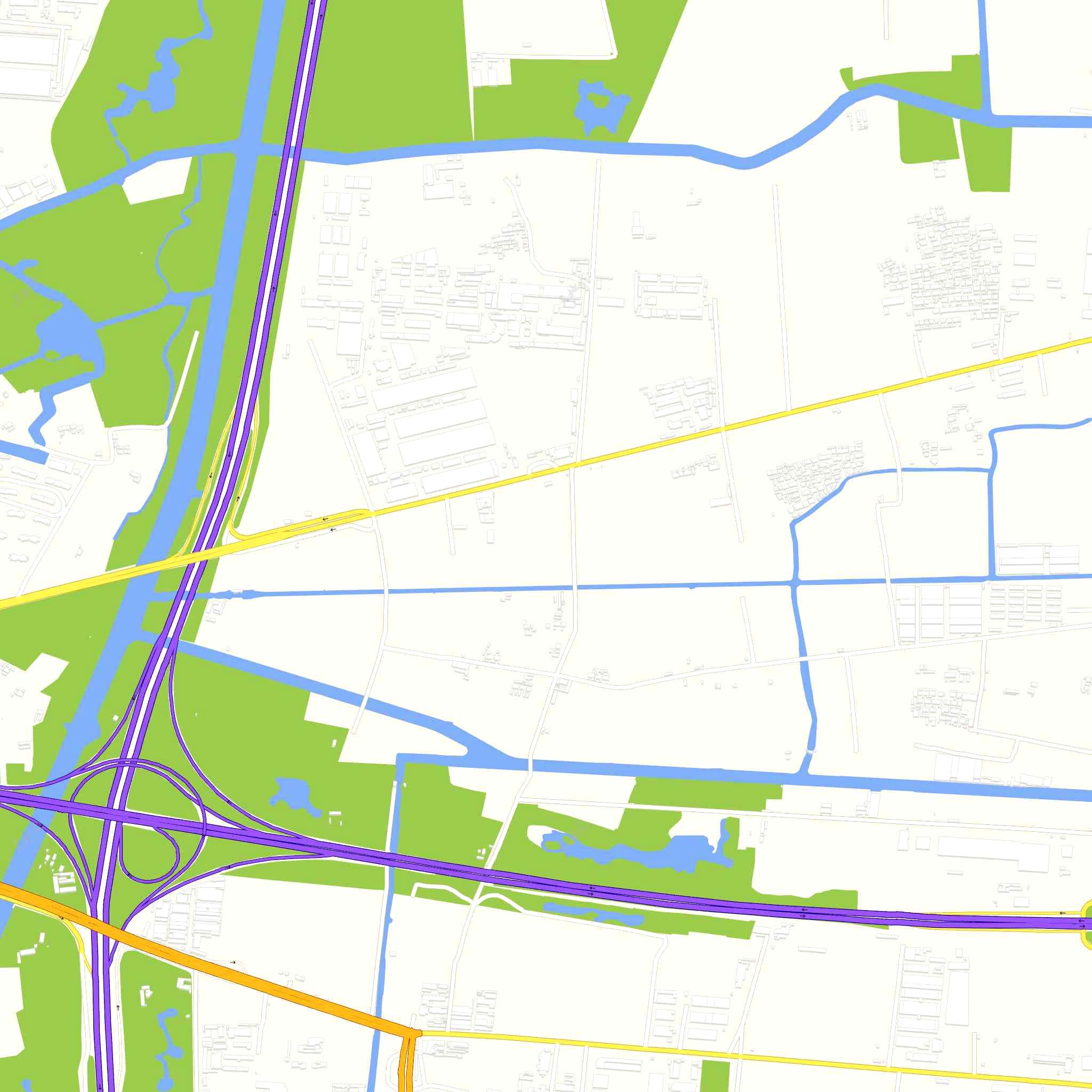}
				\\
				\includegraphics[width=3cm, angle=0]{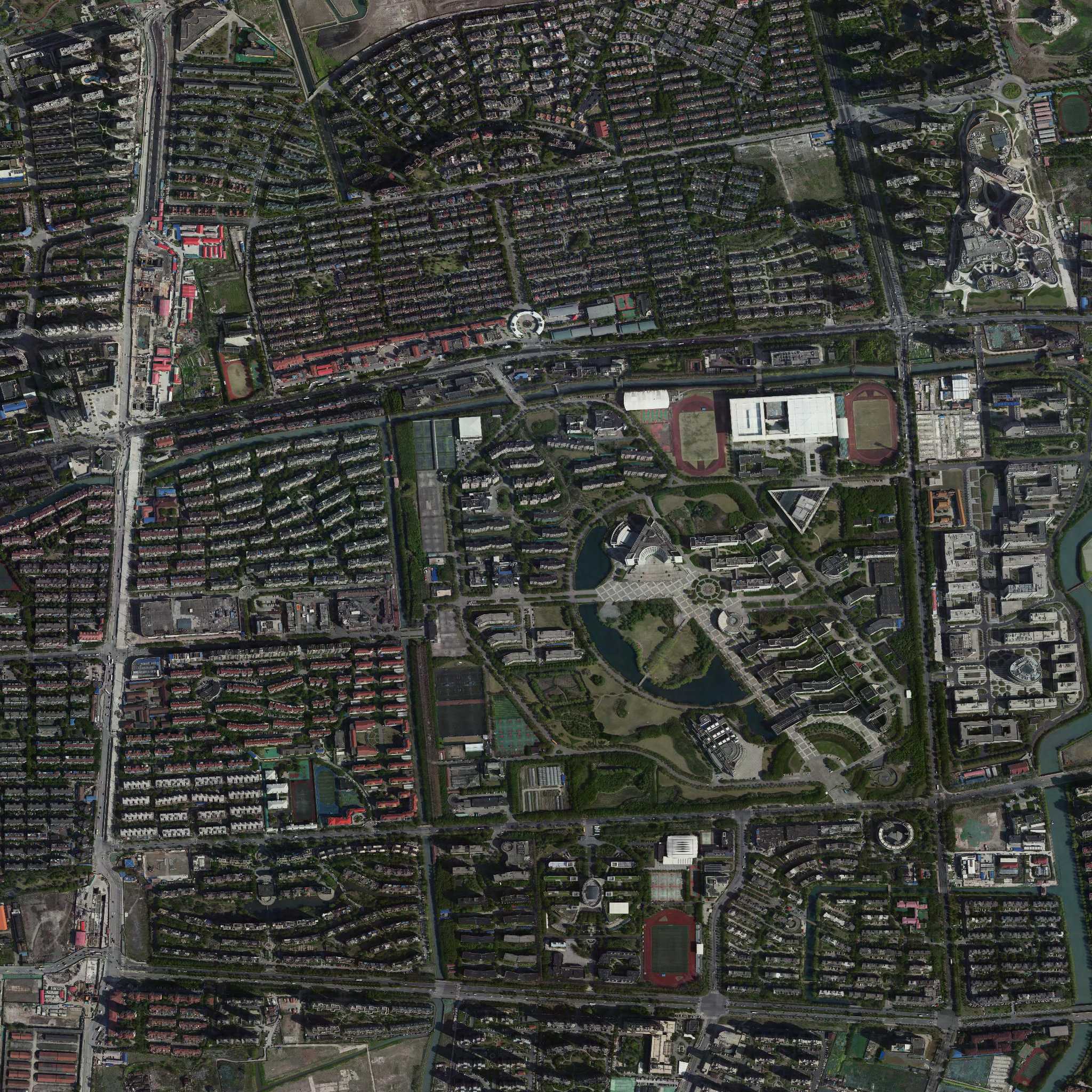}\hspace{-3mm}
				&\includegraphics[width=3cm, angle=0]{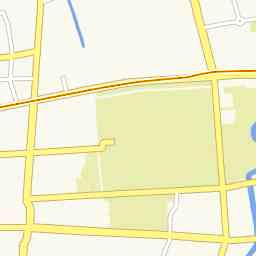}\hspace{-3mm}
				&\includegraphics[width=3cm, angle=0]{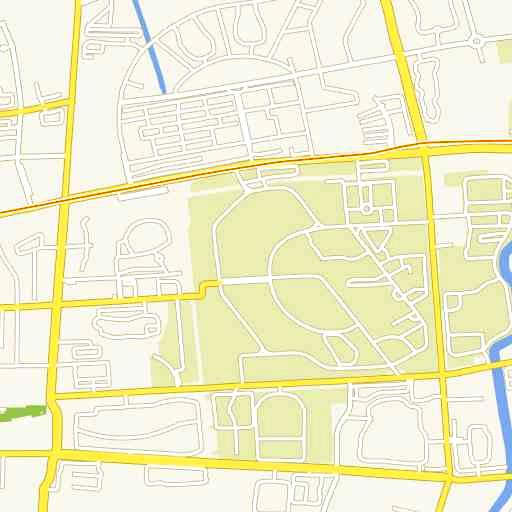}\hspace{-3mm}
				&\includegraphics[width=3cm, angle=0]{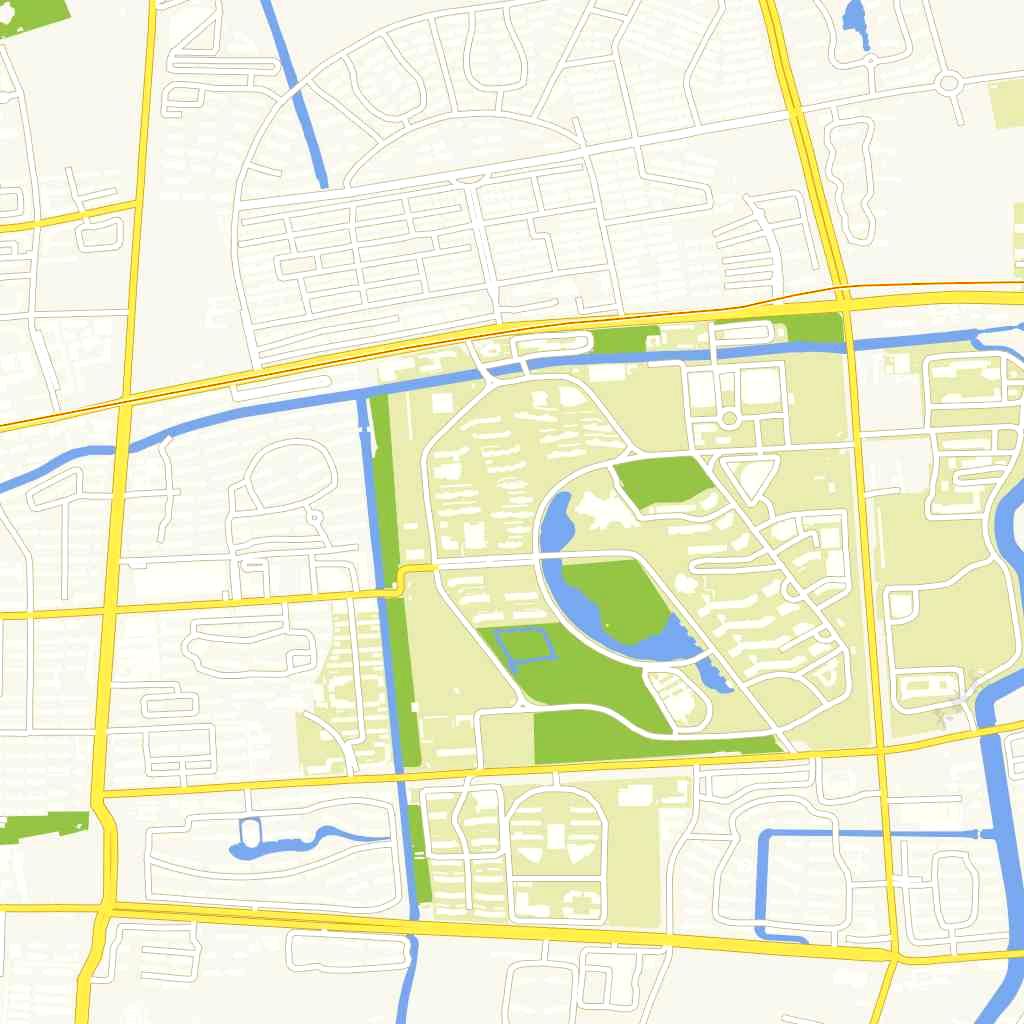}\hspace{-3mm}
				&\includegraphics[width=3cm, angle=0]{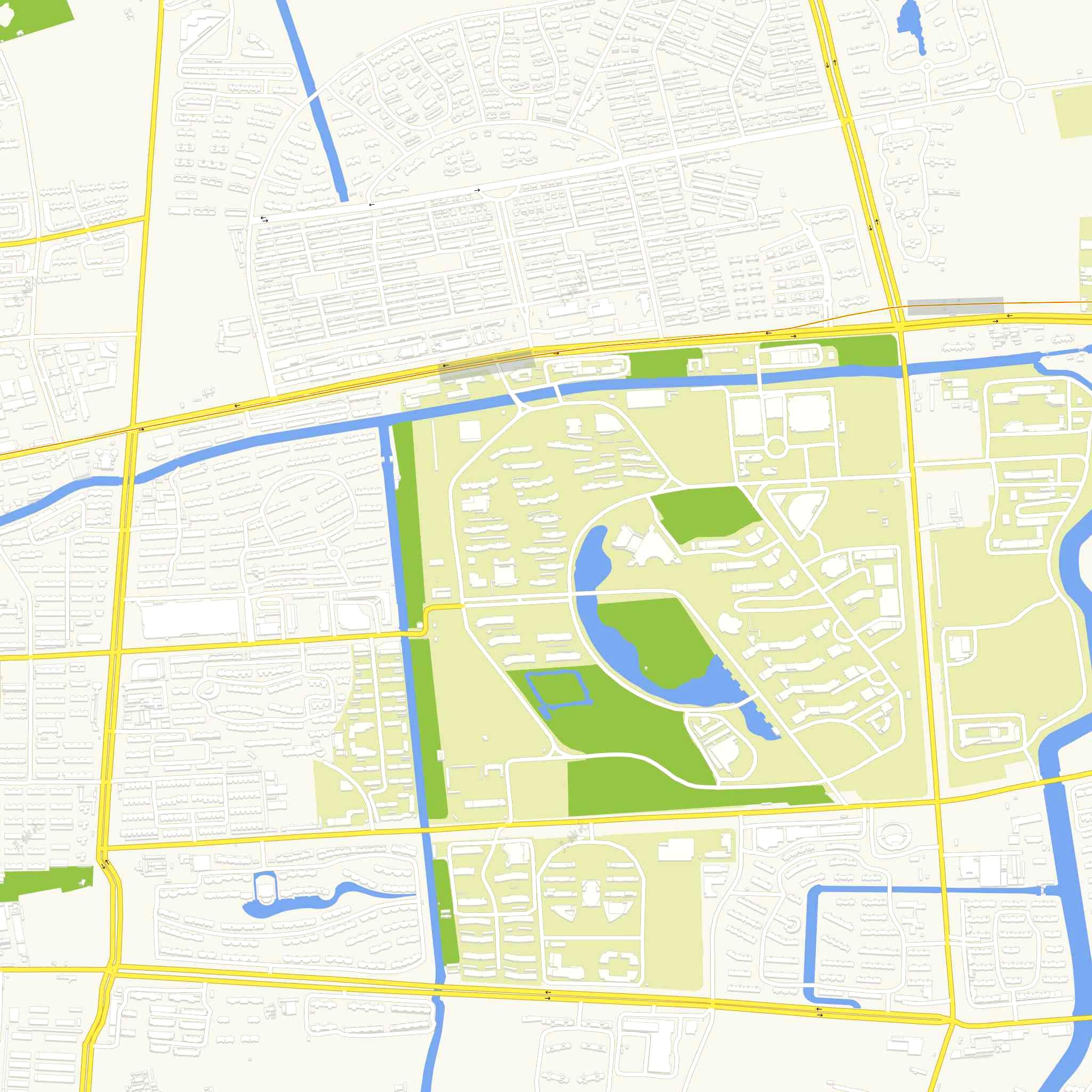}
				\\
				\includegraphics[width=3cm, angle=0]{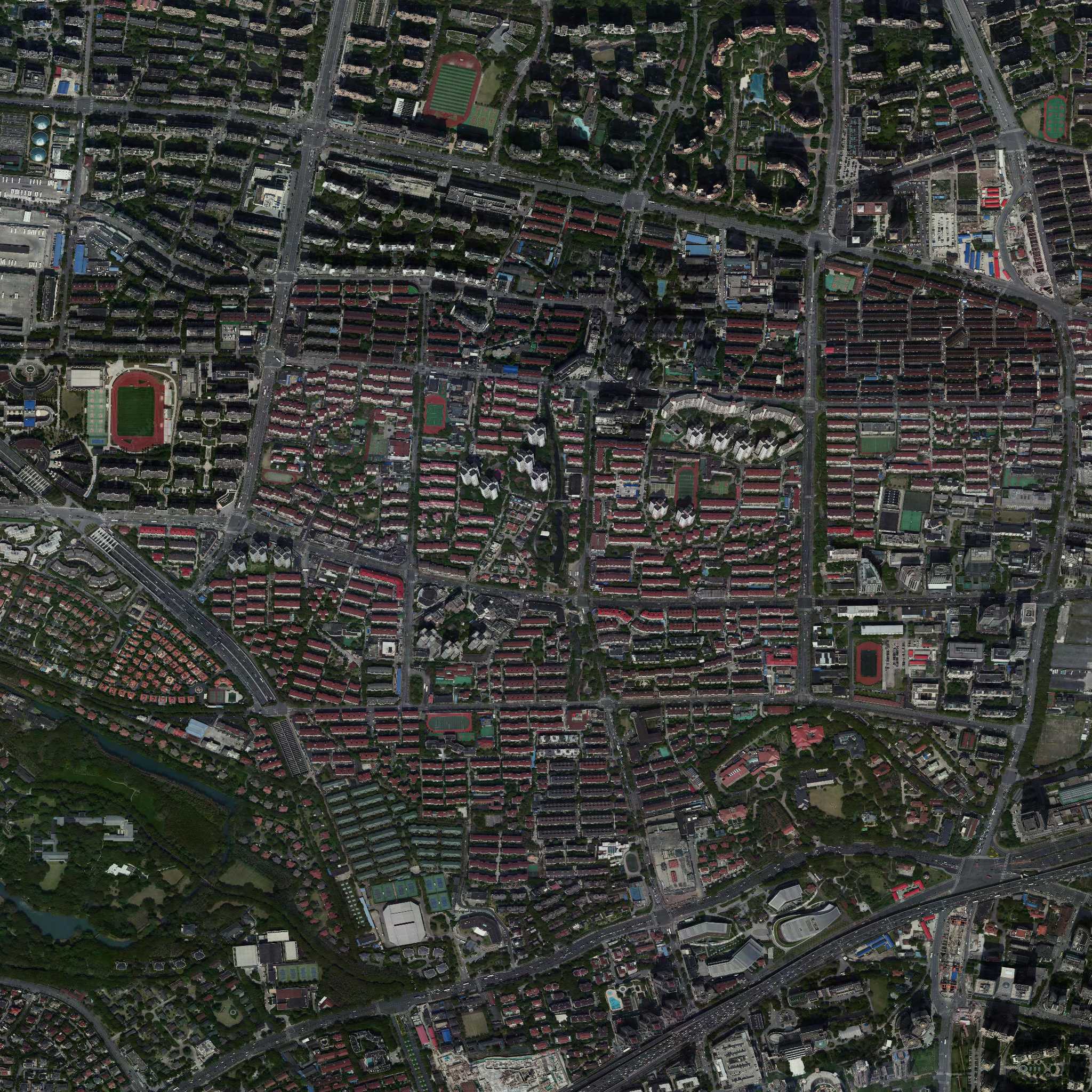}\hspace{-3mm}
				&\includegraphics[width=3cm, angle=0]{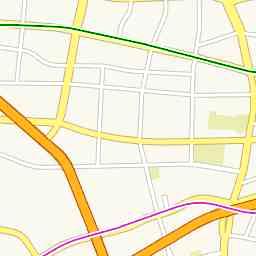}\hspace{-3mm}
				&\includegraphics[width=3cm, angle=0]{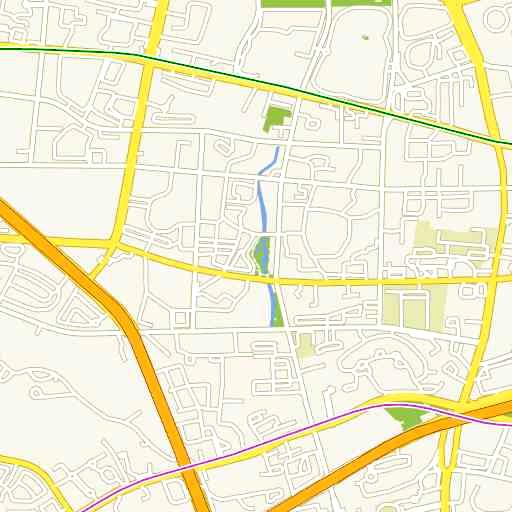}\hspace{-3mm}
				&\includegraphics[width=3cm, angle=0]{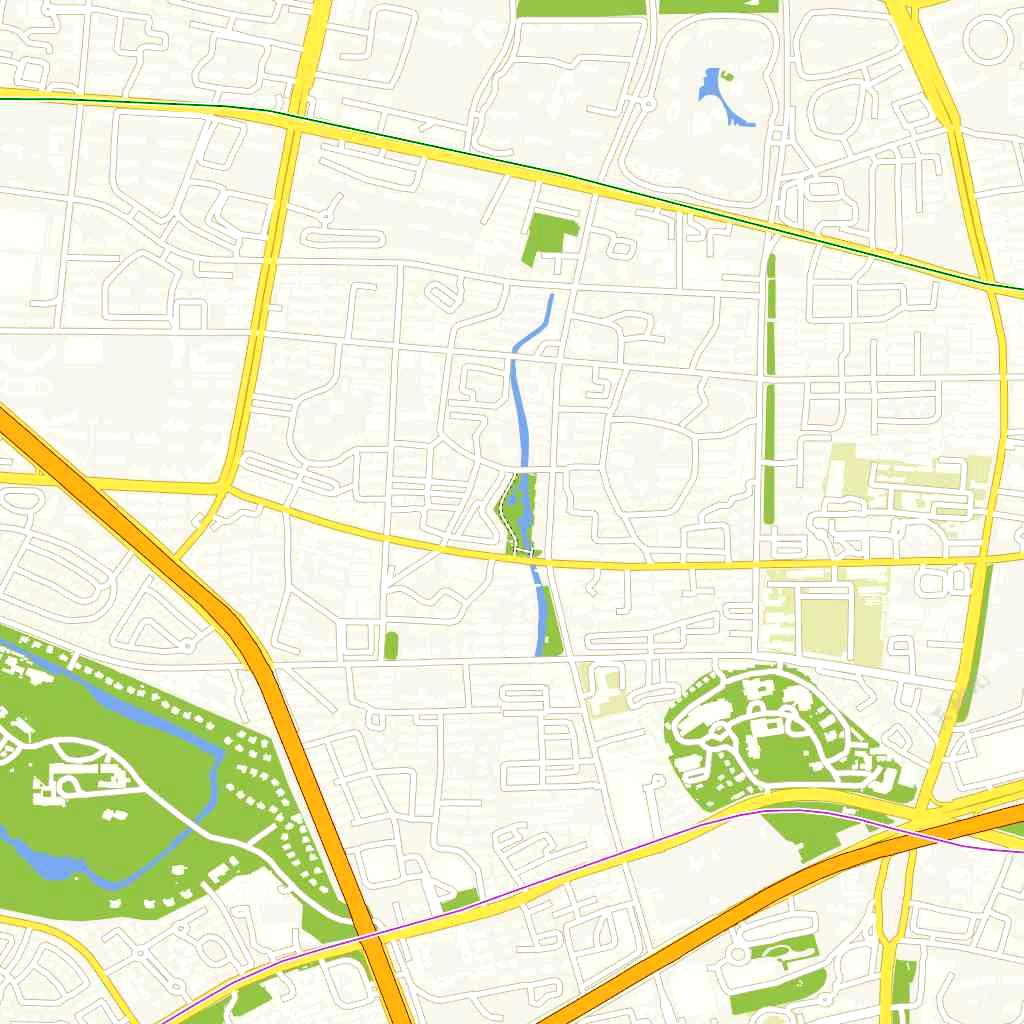}\hspace{-3mm}
				&\includegraphics[width=3cm, angle=0]{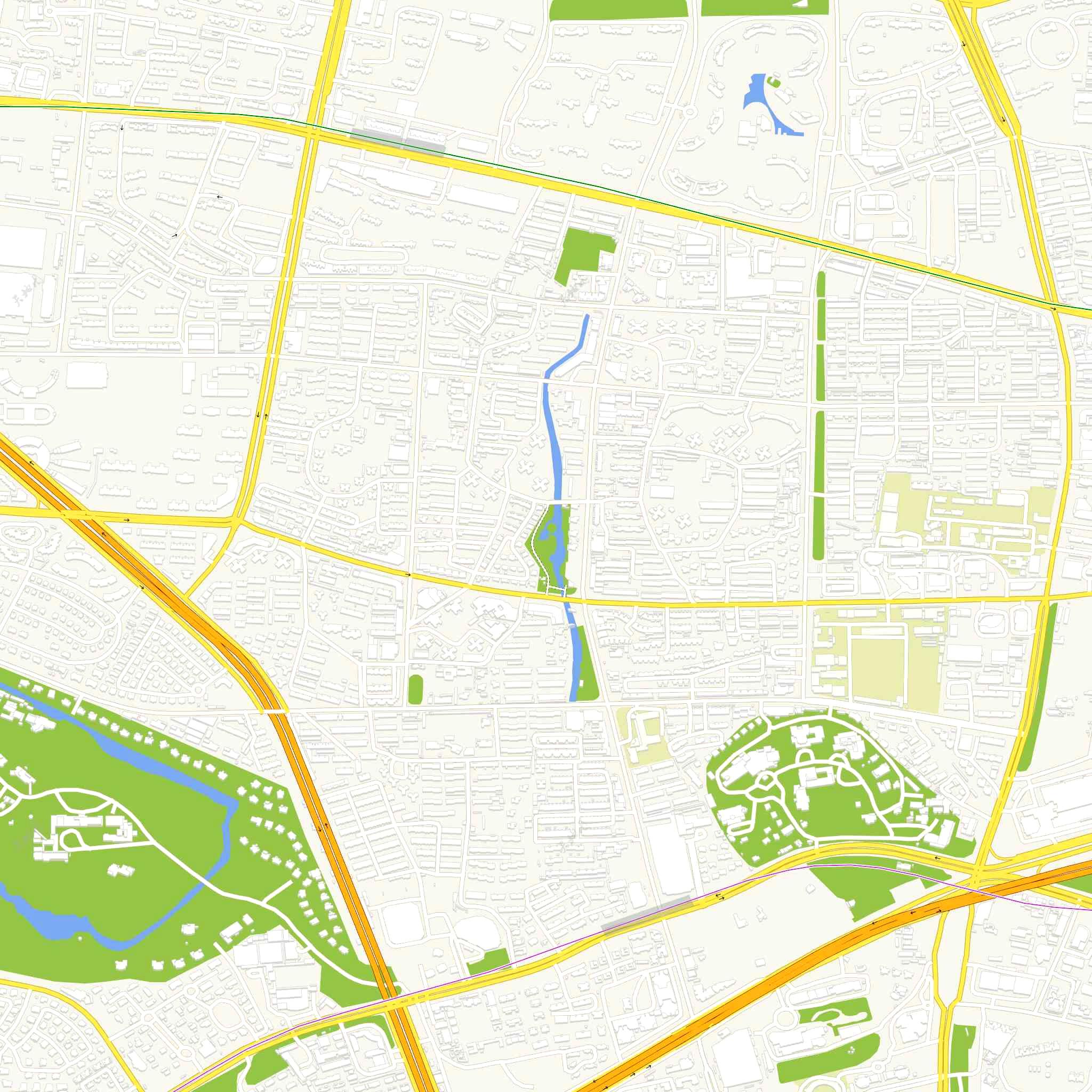}
				\\
				(a) Aerial image & (b) 15-th level & (c) 16-th level & (d) 17-th level & (e) 18-th level
				\\
				$2048\times 2048$& $256\times 256$ & $512\times 512$ & $1024\times 1024$ & $2048\times 2048$
		\end{tabular}}

	\end{center}
	\vspace{-4mm}
	\caption{Examples of our Multi-Level Map dataset from Shanghai. 
	}
	\vspace{+2mm}
	\label{fig:SH}
\end{figure}
\begin{figure}[tb]
	\begin{center}
		\scalebox{0.55}{
			\begin{tabular}{cccccccccc}
				\includegraphics[width=3cm, angle=0]{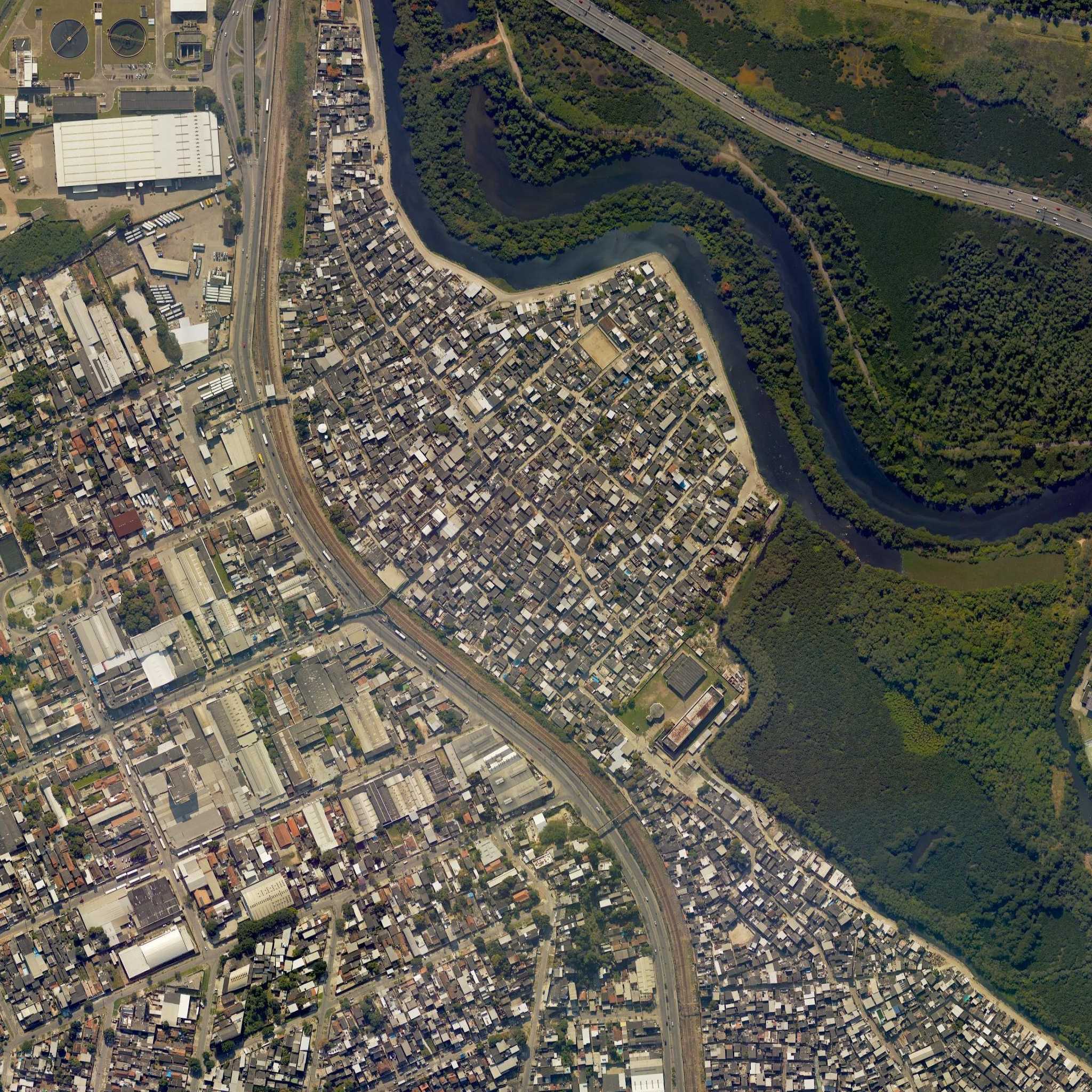}\hspace{-3mm}
				&\includegraphics[width=3cm, angle=0]{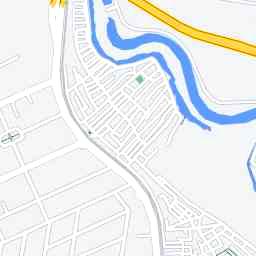}\hspace{-3mm}
				&\includegraphics[width=3cm, angle=0]{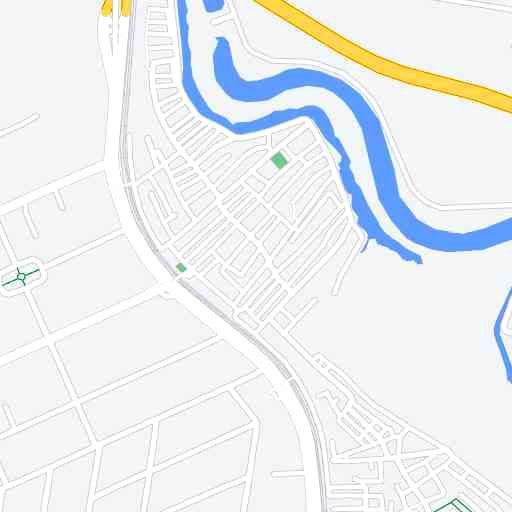}\hspace{-3mm}
				&\includegraphics[width=3cm, angle=0]{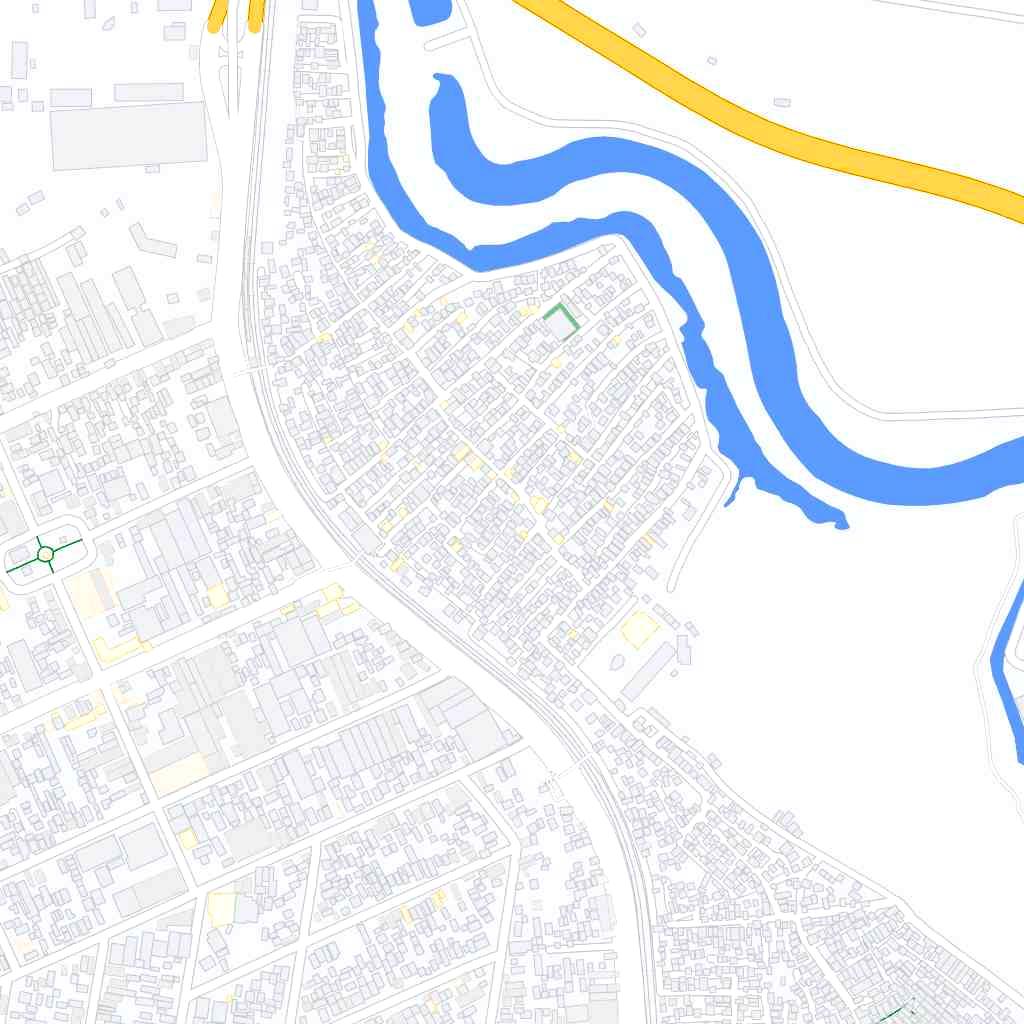}\hspace{-3mm}
				&\includegraphics[width=3cm, angle=0]{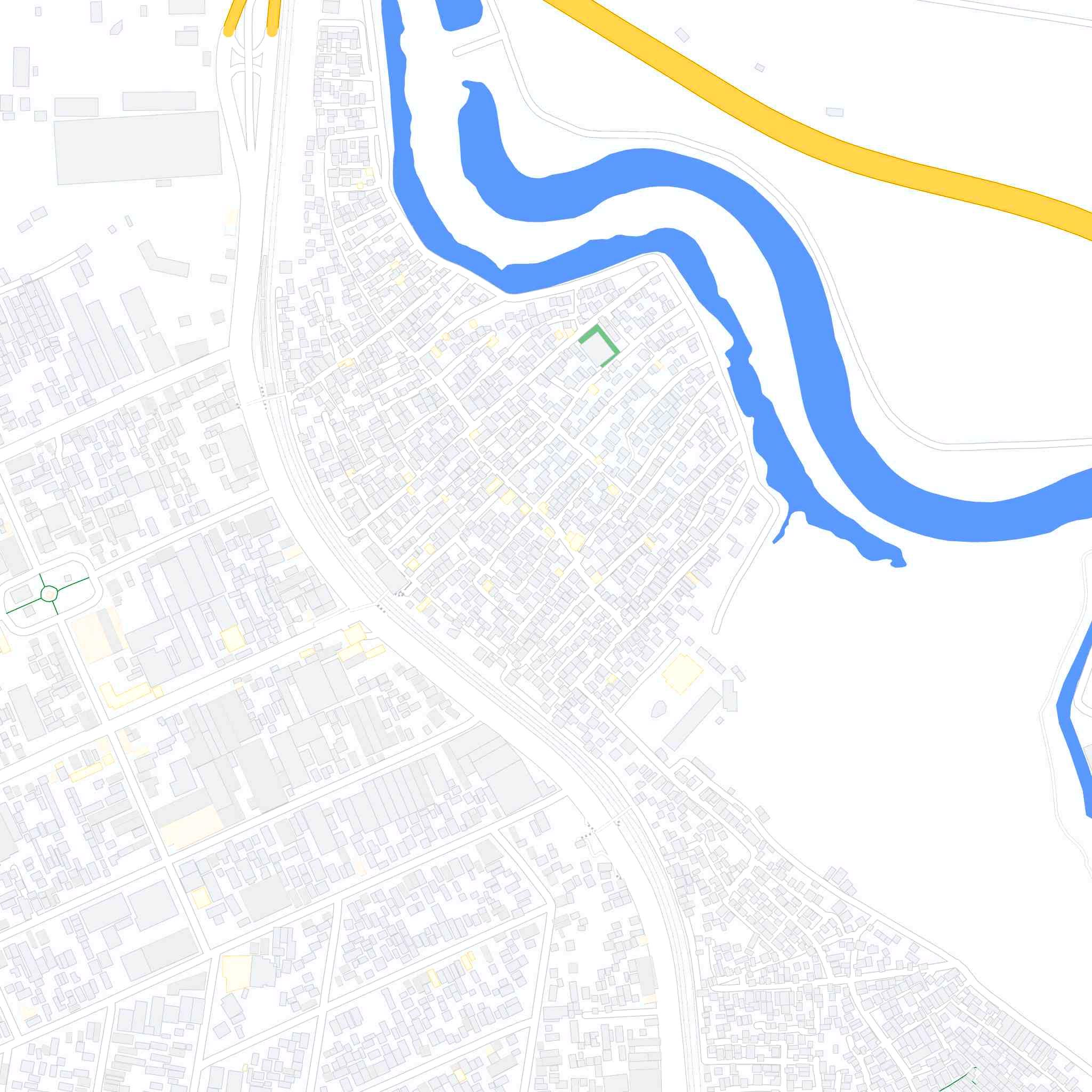}
				\\
				\includegraphics[width=3cm, angle=0]{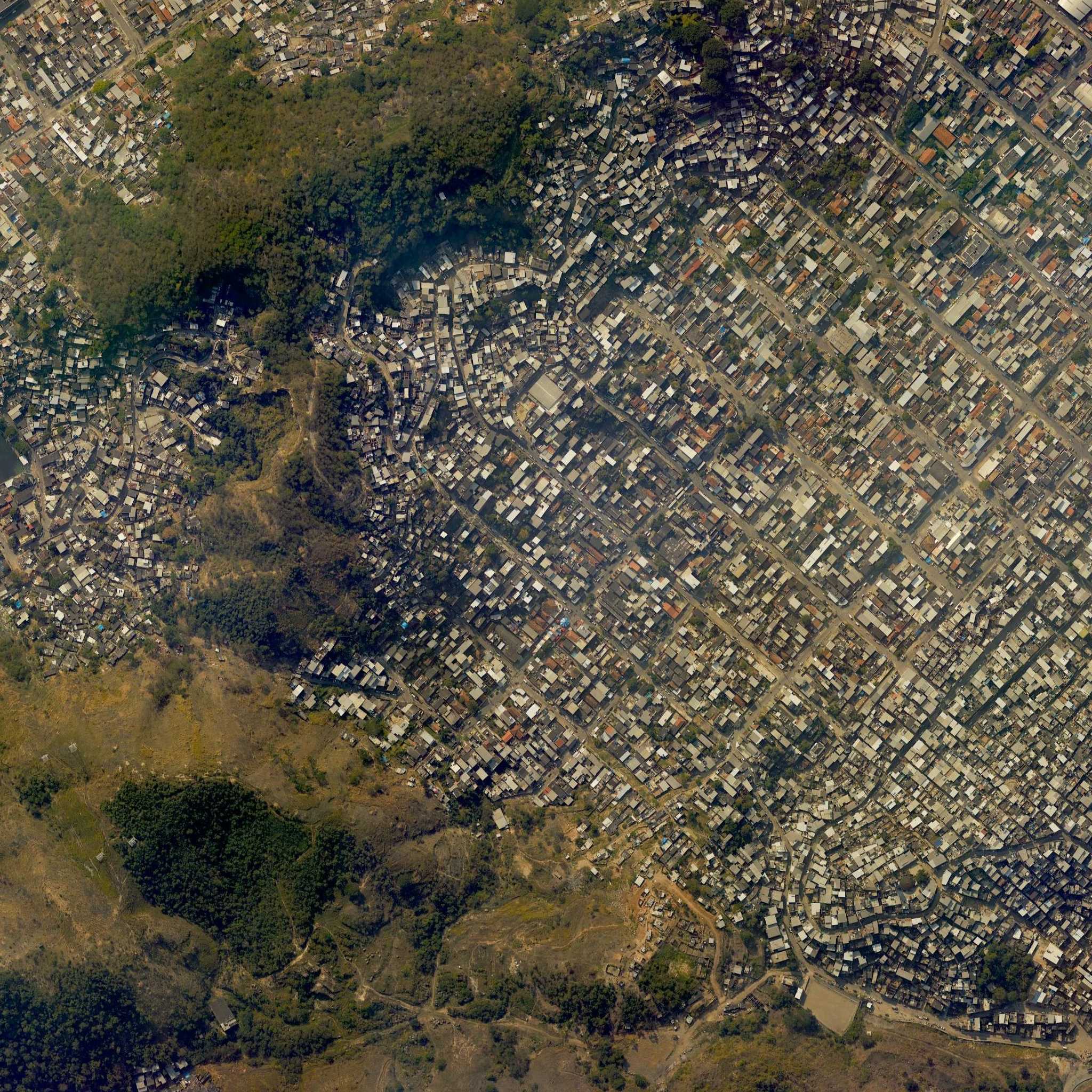}\hspace{-3mm}
				&\includegraphics[width=3cm, angle=0]{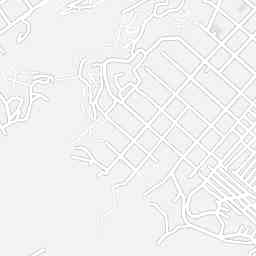}\hspace{-3mm}
				&\includegraphics[width=3cm, angle=0]{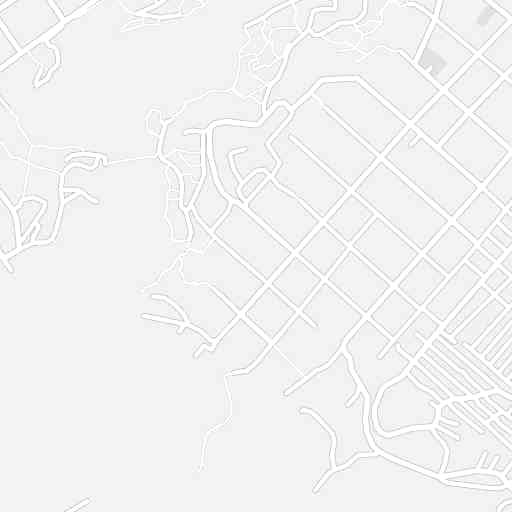}\hspace{-3mm}
				&\includegraphics[width=3cm, angle=0]{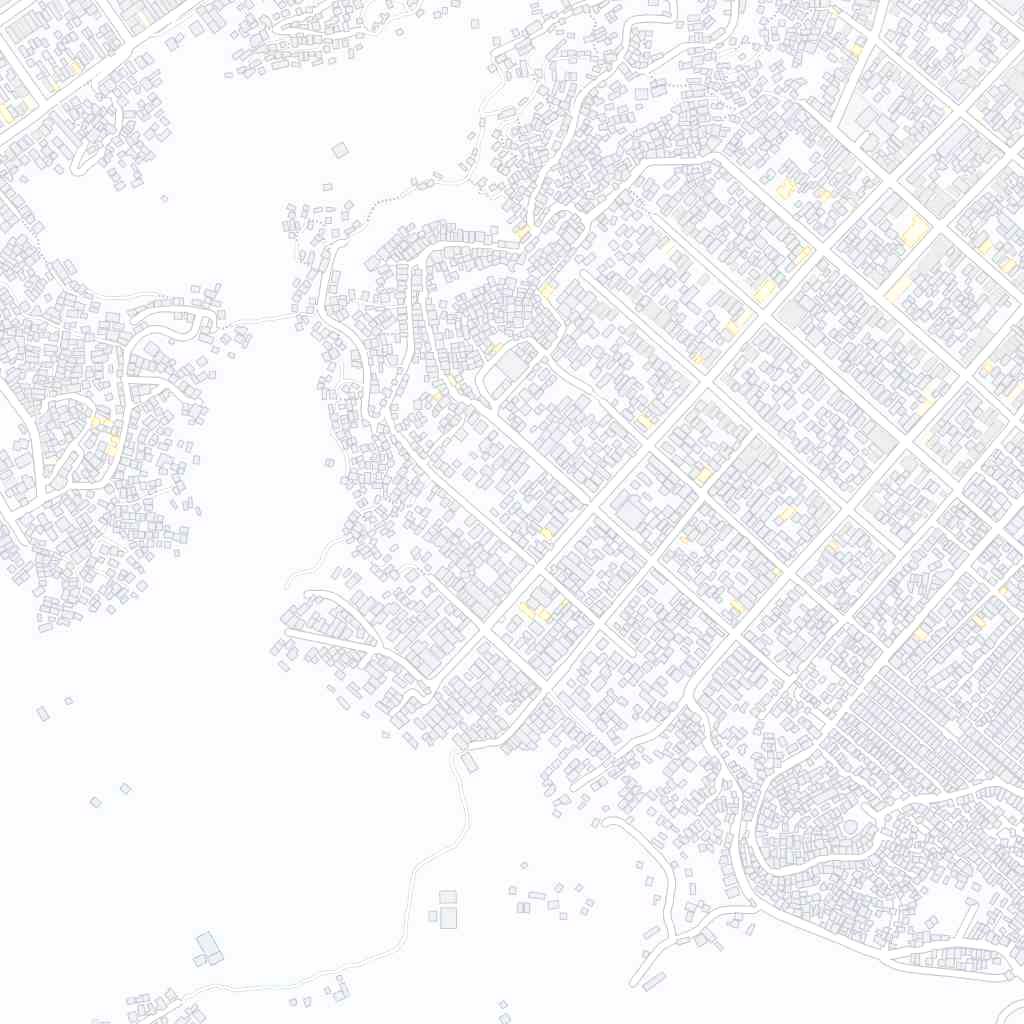}\hspace{-3mm}
				&\includegraphics[width=3cm, angle=0]{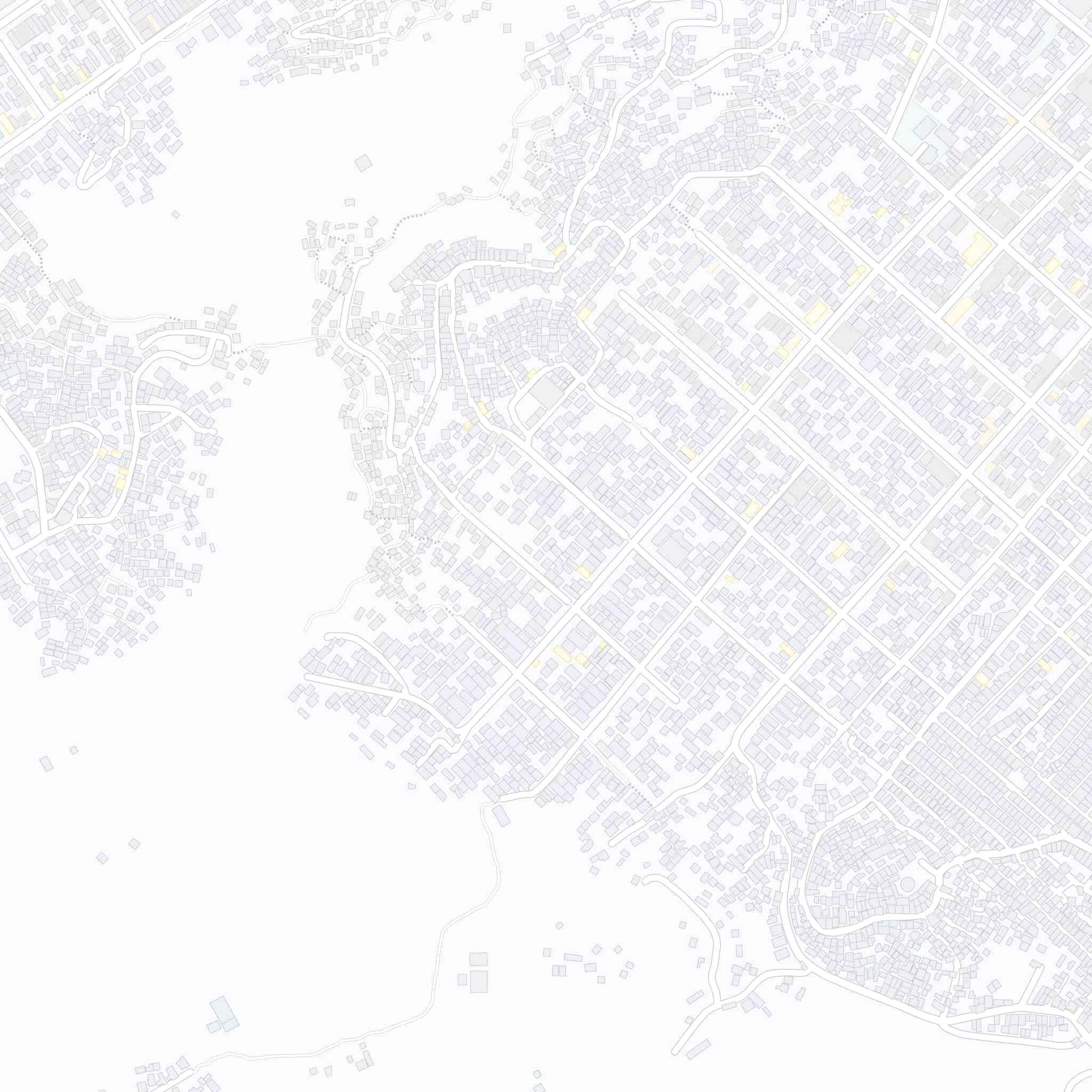}
				\\
				\includegraphics[width=3cm, angle=0]{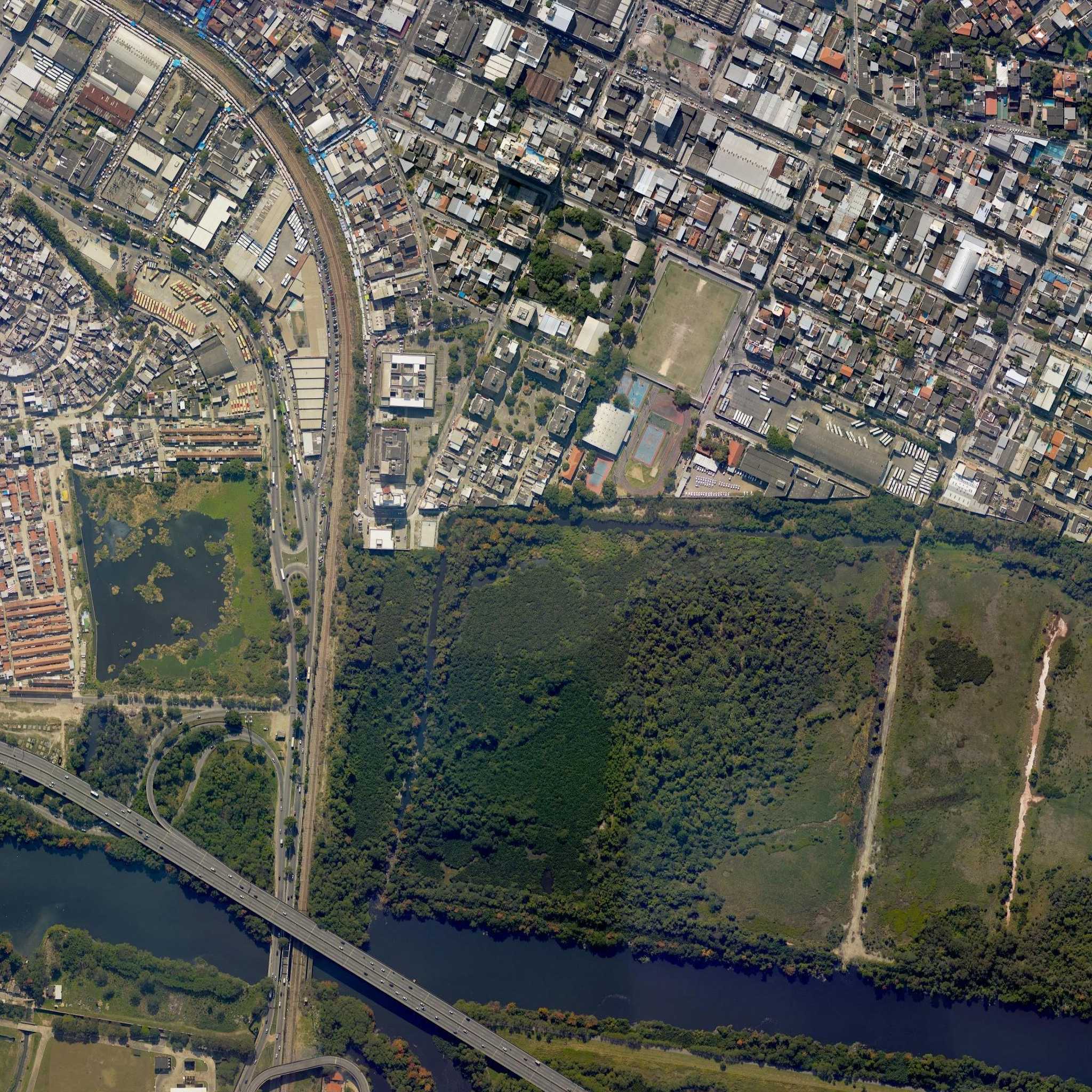}\hspace{-3mm}
				&\includegraphics[width=3cm, angle=0]{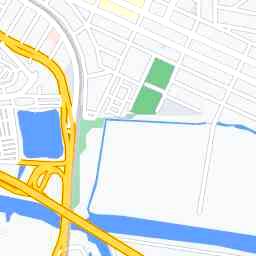}\hspace{-3mm}
				&\includegraphics[width=3cm, angle=0]{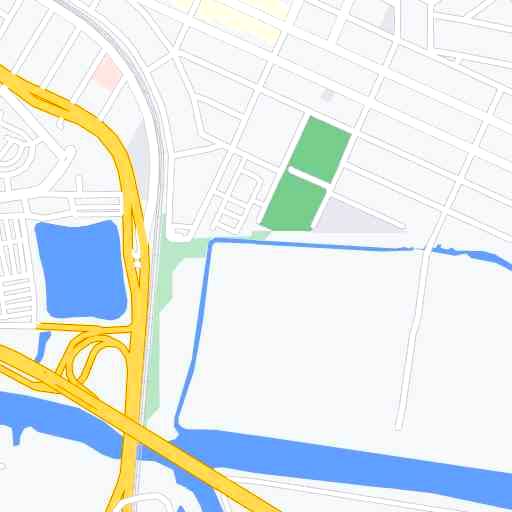}\hspace{-3mm}
				&\includegraphics[width=3cm, angle=0]{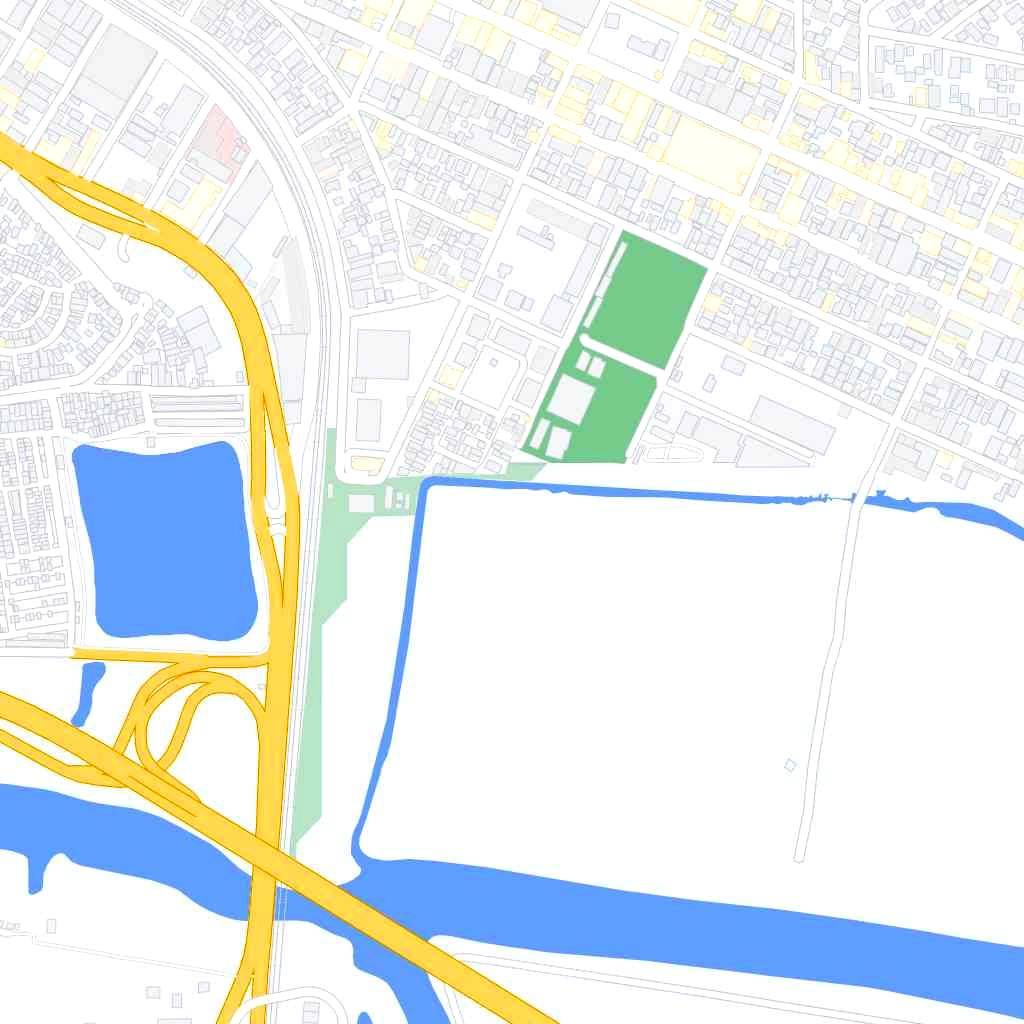}\hspace{-3mm}
				&\includegraphics[width=3cm, angle=0]{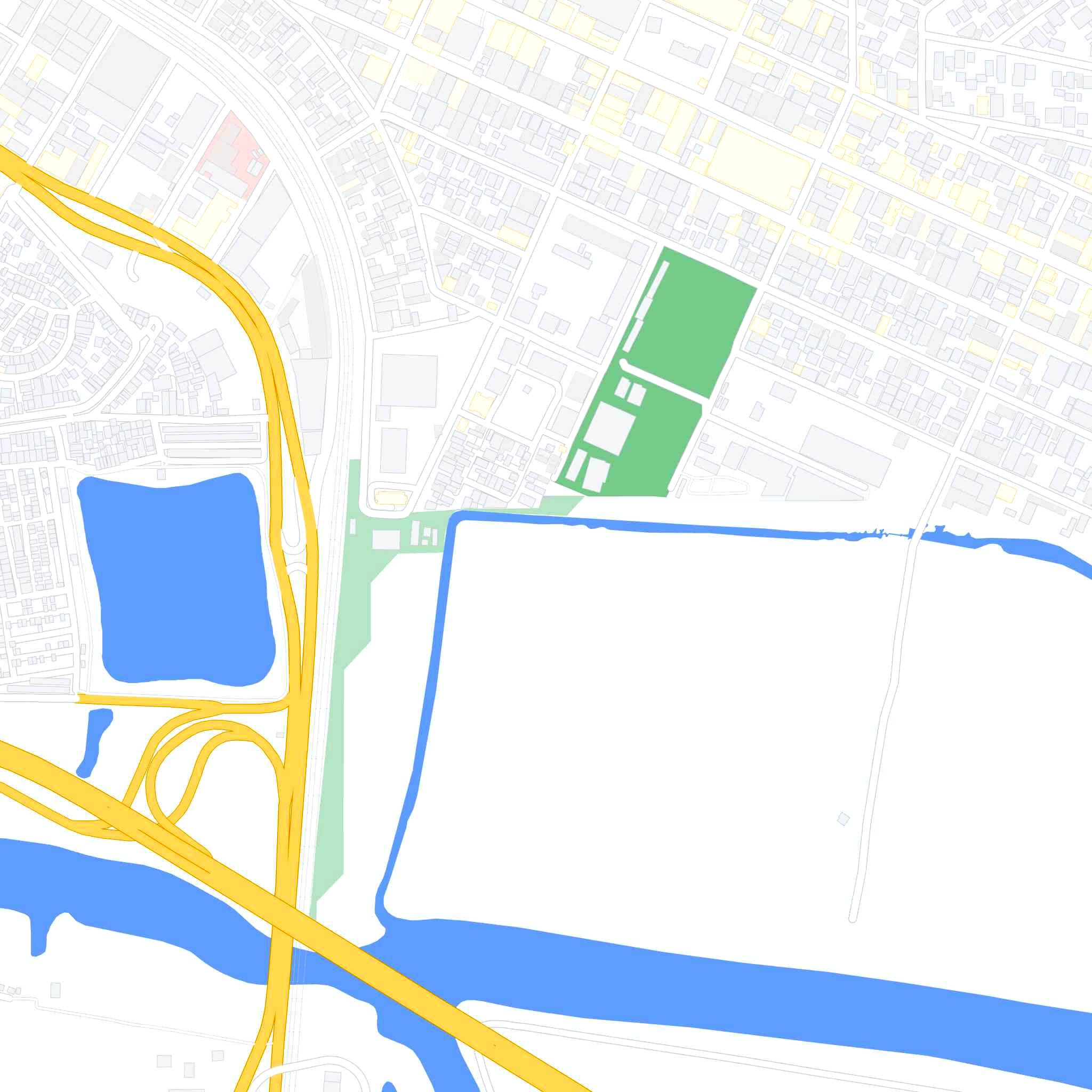}
				\\
				\includegraphics[width=3cm, angle=0]{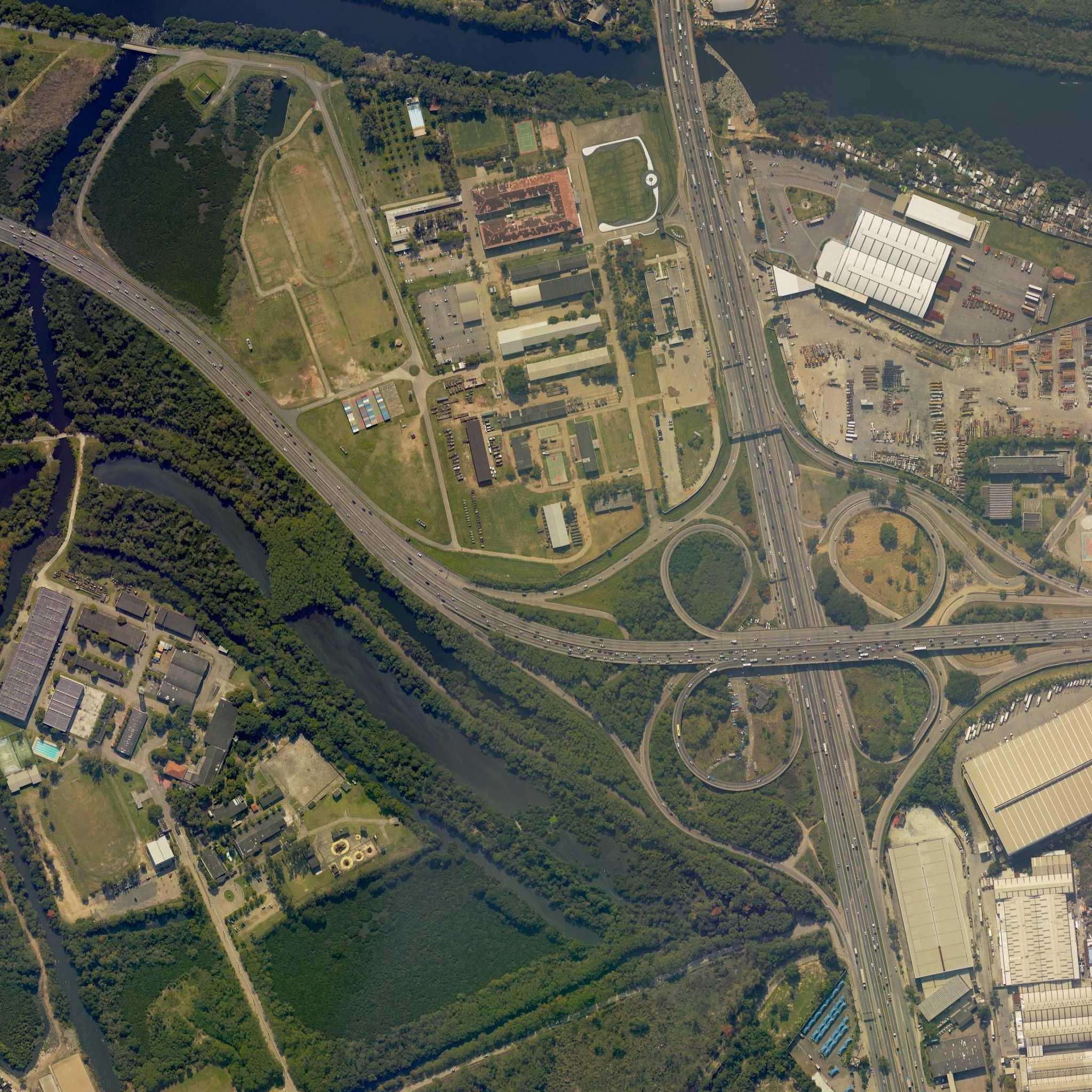}\hspace{-3mm}
				&\includegraphics[width=3cm, angle=0]{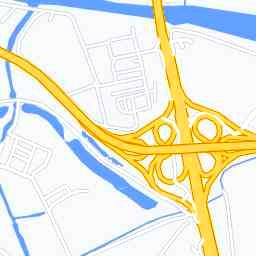}\hspace{-3mm}
				&\includegraphics[width=3cm, angle=0]{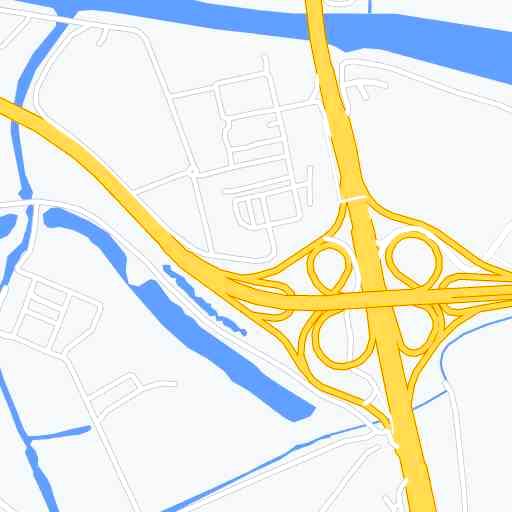}\hspace{-3mm}
				&\includegraphics[width=3cm, angle=0]{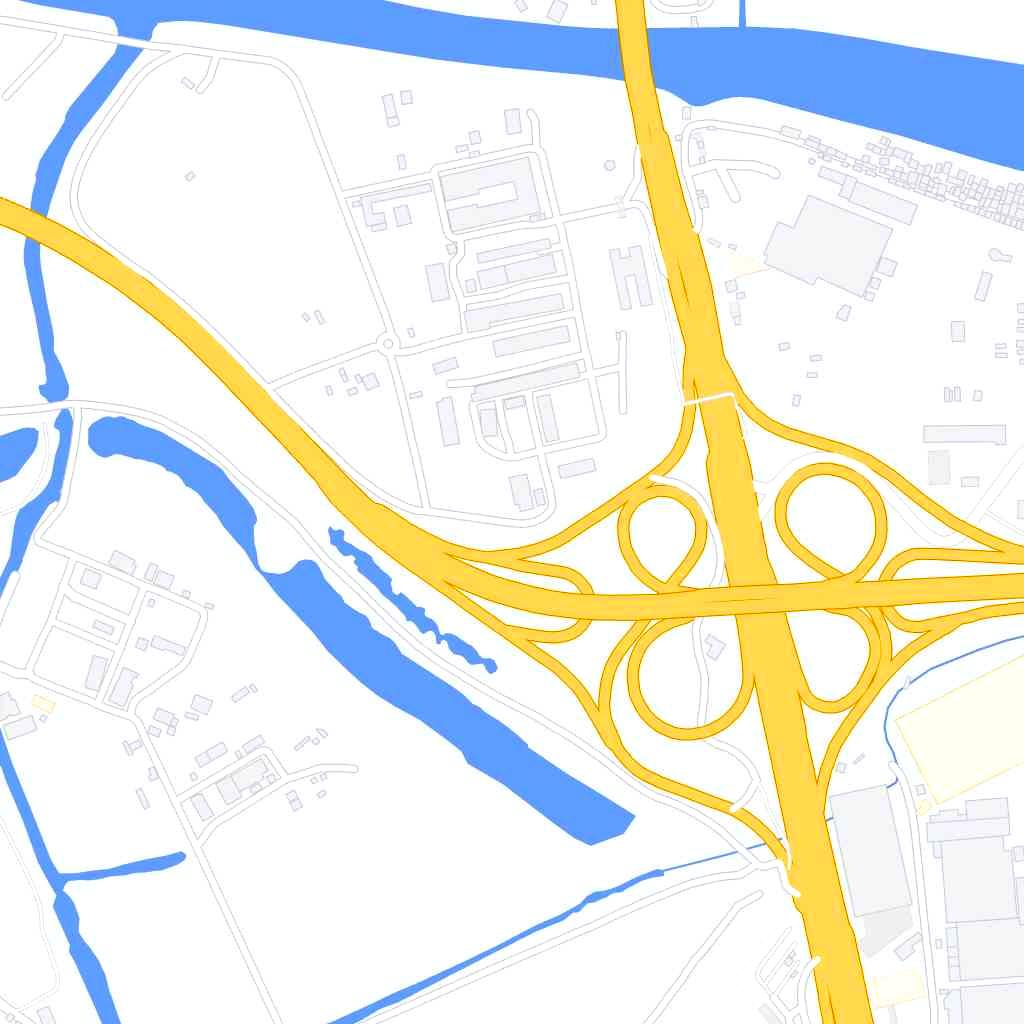}\hspace{-3mm}
				&\includegraphics[width=3cm, angle=0]{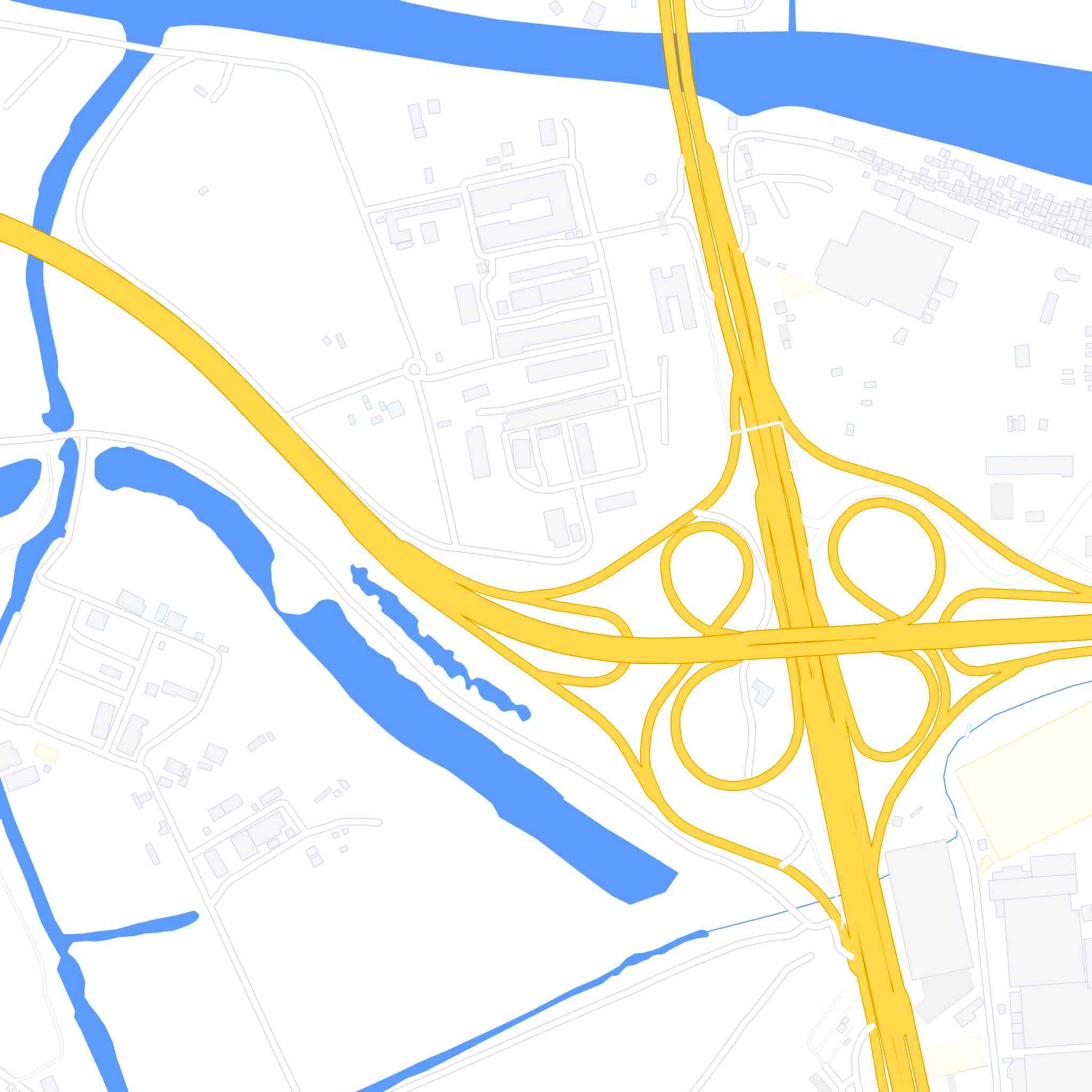}
				\\
				
				(a) Aerial image & (b) 15-th level & (c) 16-th level & (d) 17-th level & (e) 18-th level
				\\
				$2048\times 2048$& $256\times 256$ & $512\times 512$ & $1024\times 1024$ & $2048\times 2048$
				
		\end{tabular}}
	\end{center}
	\vspace{-4mm}
	\caption{Examples of our Multi-Level Map dataset from Rio de Janeiro. 
	}
	\vspace{+2mm}
	\label{fig:RIO}
\end{figure}

\begin{table}[tb]
	\begin{center}
	\scalebox{0.99}{
		\begin{tabular}{cccccc}
			\toprule[1.0pt]
			Dataset & Resolution & Samples & Aerial & Map & Levels \\
			\midrule
			Kang \etal~\cite{2019multiscalemap}& $256\times 256$ 
			&10,500 & $\times$& $\surd$ & 2\\ 
			Isola \etal~\cite{pix2pix} & $256\times 256$
			&2,194  & $\surd$ & $\surd$ & 1\\ 
			MLM (ours) & $256\times 256$ 
			&18,700  & $\surd$& $\surd$ & 4 \\
			\bottomrule[1.pt]
		\end{tabular}}
	\end{center}
	\vspace{-0mm}
	\caption{
	Comparison with existing map datasets.
	MLM provides larger number of samples and in more zoom levels.
	}
	\vspace{-0mm}
	\label{tab:datasets}
\end{table}

\begin{table}[tb]
	\begin{center}
	\scalebox{0.90}{
		\begin{tabular}{ccccc}
			\toprule[1.pt]
			Zoom & Train Samples & Test Samples &Train + Test & Spatial  \\
			Level & SH $|$ RJ & SH $|$ RJ & SH $|$ RJ & Resolution \\
			\midrule
			15 & 96 $|$ 80 & 24 $|$ 20& 120 $|$ 100& 4.8 m/pixel \\ 
			16 & 384 $|$ 320 & 96 $|$ 80&480 $|$ 400&  2.4 m/pixel \\ 
			17 & 1,536 $|$ 1,280 & 384 $|$ 320&1,920 $|$ 1,600& 1.2 m/pixel \\
			18 & 6,144 $|$ 5,120& 1,536 $|$ 1,280&7,680 $|$ 6,400& 0.6 m/pixel  \\
			\midrule
			Total & 8,160 $|$ 6,800 & 2,040 $|$ 1,700 &10,200 $|$ 8,500\ \ \ & {\bf All 18,700}  \\
			\bottomrule[1.pt]
		\end{tabular}}
	\end{center}
	\vspace{-0mm}
	\caption{Details of our Multi-Level Map dataset. 
	It consists of samples from Shanghai (SH) region and Rio de Janeiro (RJ) region.  }
	\vspace{-0mm}
	\label{tab:MLM}
\end{table}

\begin{table*}[!htbp]
	\centering
	\begin{tabularx}{\linewidth}{lXX}
		\toprule[1.5pt]
		& \textbf{Introduction } 
		& \textbf{Details }
		\\
		\midrule
		FID$\downarrow$
		& FID~\cite{FID} is robust to noise and consistent with human perception. 
		It uses the Fréchet distance between two multivariate Gaussians
		\begin{equation*}
			\begin{aligned}
				\mathrm{FID}(\mathbb{P}_r,\mathbb{P}_g)=\|\mu_r-\mu_g\|^2
				+\mathtt{Tr}(\mathtt{C}_r+\mathtt{C}_g-2(\mathtt{C}_r\mathtt{C}_g)^{\frac{1}{2}}). 
			\end{aligned} 
		\end{equation*}
		
		& 
		$\mathbb{P}_r$, $\mathbb{P}_g$ indicate the distribution of real samples and generated samples respectively. 
		$\mathtt{C}$ is the covariance matrix of feature vector. $\mathtt{Tr}$ is the trace of matrix. We follow \cite{FID} to use pre-trained InceptionV3~\cite{InceptionV3} network to extract features.
		\\
		\midrule
		KID$\downarrow$
		&
		KID~\cite{KID} can be seen as an improved version of FID~\cite{FID}, its estimates are unbiased and asymptotically normal
		\begin{equation*}
			\begin{aligned}
				\mathrm{KID}(\mathbb{P}_r,\mathbb{P}_g)=& \mathbb{E}_{\substack{\mathtt{x}_r,\mathtt{x}^{\prime}_r\sim \mathbb{P}_r,\\ \mathtt{x}_g,\mathtt{x}^{\prime}_g\sim \mathbb{P}_g}}[k(I(\mathtt{x}_r),I(\mathtt{x}^{\prime}_r))
				\\ &-2k(I(\mathtt{x}_r),I(\mathtt{x}_g))
				+k(I(\mathtt{x}_g),I(\mathtt{x}^{\prime}_g))].
			\end{aligned} 
		\end{equation*}
		
		&
		$I(\mathtt{x}_r)$, $I(\mathtt{x}_g)$ are features obtained from real samples and generated samples respectively. 
		$k$ indicates the polynomial kernel function, $k(a, b) = (\frac{1}{d} a^{\mathrm{T}}b+1)^{3}$, where $d$ is the representation dimension.
		We follow \cite{KID} to use pre-trained InceptionV3~\cite{InceptionV3} network to extract features.
		
		\\
		\midrule
		KMMD$\downarrow$
		& 
		KMMD~\cite{KMMD} uses fixed kernel function to measure the dissimilarity between real samples and generated samples
		\begin{equation*}
			\begin{aligned}
				\mathrm{MMD^{2}}(\mathbb{P}_r,\mathbb{P}_g)=& \mathbb{E}_{\substack{\mathtt{x}_r,\mathtt{x}^{\prime}_r\sim \mathbb{P}_r,\\ \mathtt{x}_g,\mathtt{x}^{\prime}_g\sim \mathbb{P}_g}}[k(\mathtt{x}_r,\mathtt{x}^{\prime}_r)
				\\&-2k(\mathtt{x}_r,\mathtt{x}_g)+k(\mathtt{x}_g,\mathtt{x}^{\prime}_g)].
			\end{aligned} 
		\end{equation*}
		
		&
		$\mathtt{x}_r$ and $\mathtt{x}_g$ is the sample feature from true distribution and generated distribution. $k$ is the gaussian kernel function. 
		Results in \cite{GAN_metric} show KMMD appears to be a good metric in terms of discriminability, robustness and efficiency with pre-trained ResNet~\cite{Resnet}. 
		Thus, we use pre-trained ResNet~\cite{Resnet} for feature extraction.
		\\
		\midrule
		IS$\uparrow$
		&IS~\cite{IS} has a reasonable correlation with human judgment of image quality and is able to show the diversity of generated images
		\begin{equation*}
			\begin{aligned}
				\mathrm{IS}(\mathbb{P}_g)=& \exp(\mathbb{E}_{\mathtt{x}\sim \mathbb{P}_g} [KL(p_{\mathcal{M}}(y|\mathtt{x})||p_{\mathcal{M}}(y))]).
			\end{aligned} 
		\end{equation*}
		%
		
		&$p_{\mathcal{M}}(y|\mathtt{x})$ is the label distribution of $\mathtt{x}$ as predicted by model $\mathcal{M}$, and $p_{\mathcal{M}}(y)=\int_\mathtt{x} p_{\mathcal{M}}(y|\mathtt{x})\mathrm{d} \mathbb{P}_g$. 
		$KL$ indicates Kullback-Leibler divergence. 
		We follow \cite{IS} to use InceptionV3~\cite{InceptionV3} network to extract the probability distribution of each category.
		\\
		\midrule
		PSNR$\uparrow$
		&
		PSNR is commonly used to measure the quality of image by comparing samples with original images pixel-wisely
		\begin{equation*}
			\begin{aligned}
				\mathrm{PSNR}=& 10\times\log_{10}(\frac{MAX^2_I}{MSE}).
			\end{aligned} 
		\end{equation*}
		
		&$MAX_I$ is the maximum possible pixel value of the image. 
		MSE is the mean-square error. 
		\\
		\bottomrule[1.5pt]%
	\end{tabularx}%
	
	\caption{Details of evaluation metrics. 
	Up arrow means higher is better, and down arrow means lower is better.}
	\label{tab:metric}%
\end{table*}%

\begin{table}[tb]
	\begin{center}
		\scalebox{0.86}{
			\begin{tabular}{c|c|ccccc}
				\specialrule{0em}{1.5pt}{1.5pt}
				\toprule[1.5pt]
				Level & Method & FID$\downarrow$ & KID$\downarrow$ & KMMD$\downarrow$ & IS$\uparrow$ &PSNR$\uparrow$ \\
				\midrule
				\multirow{10}{*}{15}
				& Pix2Pix \cite{pix2pix}       & 380.3 & 0.4179 & 0.6770 & 2.057 & 19.98  \\ 
				& Pix2PixHD \cite{pix2pixhd}  & 371.7 & 0.2914 & 0.6324 & 2.565 & 19.50    \\ 
				& CycleGAN \cite{cyclegan}   & 345.5 & 0.3437 & 0.6641 & 2.178 & 19.65  \\
				& GeoGAN \cite{2019geogan}  & 534.6 & 0.6836 & 1.0008 & 1.374 & 20.30 \\ 
				& SMAPGAN \cite{2020smapgan}  & 391.9 & 0.4281 & 0.7027 & 1.938 & 20.96  \\ 
				& SPADE \cite{2019spade} & 516.7  & 0.5265  & 0.9095 & 1.496  & 19.91  \\ 
				& SelectionGAN \cite{2019selectiongan} & 328.4 & 0.2672 & 0.5678 & 2.139 & 19.39 \\ 
				& TSIT \cite{2020tsit} & 323.3  & 0.3488 & 0.6833 & 1.633 & 19.47  \\ 
				& LPTN \cite{2021LPTN} & 411.2 & 0.4380 & 0.8249 & 1.977 & 17.64  \\ 
				& Ours  & \bf{239.2} & \bf{0.1268} & \bf{0.3823} & \bf{3.019} & \bf{21.75} \\ 
				\midrule
				\multirow{10}{*}{16}
				& Pix2Pix \cite{pix2pix}       & 291.0 & 0.3181 & 0.6035 & 2.250 & 20.24  \\ 
				& Pix2PixHD \cite{pix2pixhd}  & 296.7 & 0.1839 & 0.5031 & 3.019 & 20.17    \\ 
				& CycleGAN \cite{cyclegan}   & 232.7 & 0.2333 & 0.5325 & 2.784 & 19.91  \\
				& GeoGAN \cite{2019geogan}  & 448.1 & 0.5827 & 0.9164 & 1.528 & 20.61 \\ 
				& SMAPGAN \cite{2020smapgan}  & 392.7 & 0.4228 & 0.6709 & 2.155 & 21.42  \\ 
				& SPADE \cite{2019spade} & 375.0 & 0.4369 & 0.7665 & 1.717 & 20.16  \\ 
				& SelectionGAN \cite{2019selectiongan} & 262.7 & 0.2332 & 0.4724 & 2.705 & 19.91  \\ 
				& TSIT \cite{2020tsit} & 177.8 & 0.1775 & 0.4864 & 2.056 & 19.73  \\ 
				& LPTN \cite{2021LPTN} & 322.4 & 0.3700 & 0.7045 & 2.299 & 17.12  \\ 
				& Ours  &\bf 144.3 & \bf{0.1032} & \bf{0.3726} & \bf{3.251} & \bf{21.55} \\
				\midrule
				\multirow{10}{*}{17}
				& Pix2Pix \cite{pix2pix}       & 254.9 & 0.2160 & 0.5068 & 2.627 & 20.33  \\ 
				& Pix2PixHD \cite{pix2pixhd}  & 246.7 & 0.1138 & 0.3943 & 3.170 & 20.64    \\ 
				& CycleGAN \cite{cyclegan}   & 206.0 & 0.1461 & 0.4657 & 3.029 & 19.99  \\
				& GeoGAN \cite{2019geogan}  & 384.9 & 0.3543 & 0.7941 & 1.735 & 21.22 \\ 
				& SMAPGAN \cite{2020smapgan}  & 308.1 & 0.2754 & 0.5967 & 2.484 & 21.96  \\ 
				& SPADE \cite{2019spade} & 321.2 & 0.2829 & 0.6625 & 2.031 & 20.13  \\ 
				& SelectionGAN \cite{2019selectiongan}  & 236.8 & 0.1557 & 0.4183 & 3.019 & 20.64  \\ 
				& TSIT \cite{2020tsit} & 143.5 & 0.1111 & 0.3863 & 2.352 & 19.81  \\ 
				& LPTN \cite{2021LPTN} & 276.6 & 0.2067 & 0.6000 & 2.574 & 16.73  \\ 
				& Ours  & \bf 116.0 & \bf{0.0684} & \bf{0.3008} & \bf{3.322} & \bf{22.05} \\ 
				\midrule
				\multirow{10}{*}{18}
				& Pix2Pix \cite{pix2pix}       & 92.3 & 0.0667 & 0.2770 & 3.590 & 21.71  \\ 
				& Pix2PixHD \cite{pix2pixhd}  & 140.8 & 0.0850 & 0.3015 & 3.055 & 22.30    \\ 
				& CycleGAN \cite{cyclegan}   & 68.4 & 0.0477 & 0.2455 & 2.787 & 21.14  \\
				& GeoGAN \cite{2019geogan}  & 291.8 & 0.2541 & 0.6772 & 1.866 & 22.49 \\ 
				& SMAPGAN \cite{2020smapgan}  & 213.8 & 0.2118 & 0.4767 & 2.307 & \bf{23.79}  \\ 
				& SPADE \cite{2019spade} & 156.9 & 0.1317 & 0.4232 & 2.285 & 21.69  \\ 
				& SelectionGAN \cite{2019selectiongan}  & 112.3 & 0.0802 & 0.2662 & 3.101 & 22.04  \\ 
				& TSIT \cite{2020tsit} & 80.1 & 0.0596 & 0.2419 & 2.688 & 21.45  \\ 
				& LPTN \cite{2021LPTN} & 219.5 & 0.1503 & 0.5142 & 3.065 & 17.55  \\ 
				& Ours  & \bf{51.1} & \bf{0.0237} & \bf{0.1273} & \bf{3.656} & 22.71 \\ 
				\midrule
				\multirow{10}{*}{Avg.}
				& Pix2Pix \cite{pix2pix}        & 254.6 & 0.2547 & 0.5161 & 2.631 & 20.57  \\ 
				& Pix2PixHD \cite{pix2pixhd}  & 264.0 & 0.1685 & 0.4578 & 2.952 & 20.65   \\ 
				& CycleGAN \cite{cyclegan}   & 213.2 & 0.1927 & 0.4770 & 2.695 & 20.17 \\
				& GeoGAN \cite{2019geogan}  & 414.9 & 0.4687 & 0.8471 & 1.626 & 21.16 \\ 
				& SMAPGAN \cite{2020smapgan}  & 326.6 & 0.3345 & 0.6117 & 2.221 & \bf{22.03} \\ 
				& SPADE \cite{2019spade}  & 342.5 & 0.3445 & 0.6905 & 1.882 & 20.47  \\ 
				& SelectionGAN \cite{2019selectiongan} & 235.0 & 0.1841 & 0.4312 & 2.741 & 20.50  \\ 
				& TSIT \cite{2020tsit} & 181.2 & 0.1743 & 0.4495 & 2.182 & 20.11  \\ 
				& LPTN \cite{2021LPTN} & 307.4 & 0.2913 & 0.6609 & 2.479 & 17.26  \\ 
				& Ours  & \bf{137.7} & \bf{0.0805} & \bf{0.2957} & \bf{3.312} & 22.01 \\ 
				\bottomrule[1.5pt]
		\end{tabular}}
	\end{center}
	\vspace{-0mm}
	\caption{Quantitative results on Multi-Level Map Shanghai (MLM-SH) dataset.
		``Avg.'' indicates results that average over 15-18 zoom levels and also shows consistency results.
	}
	\vspace{-0mm}
	\label{tab:mainresults1}
\end{table}

\begin{table}[tb]
	\begin{center}
		\scalebox{0.86}{
			\begin{tabular}{c|c|ccccc}
				\specialrule{0em}{1.5pt}{1.5pt}
				\toprule[1.5pt]
				Level & Method & FID$\downarrow$ & KID$\downarrow$ & KMMD$\downarrow$ & IS$\uparrow$ &PSNR$\uparrow$ \\
				\midrule
				\multirow{10}{*}{15}
				& Pix2Pix \cite{pix2pix}       & 280.6 & 0.2388 & 0.5640 & 1.961 & 22.90  \\ 
				& Pix2PixHD \cite{pix2pixhd}  & 249.5 & 0.1790 & 0.5244 & 1.989 & 24.48    \\ 
				& CycleGAN \cite{cyclegan}   & 319.7 & 0.2945 & 0.6151 & 1.686 & 23.39  \\
				& GeoGAN \cite{2019geogan}  & 548.7 & 0.4809 & 0.9097 & 1.170 & 25.51 \\ 
				& SMAPGAN \cite{2020smapgan}  & 218.4 & 0.1031 & 0.3486 & 1.979 & \bf{25.81}  \\ 
				& SPADE \cite{2019spade} & 358.3 & 0.3637 & 0.6665 & 2.068 & 24.22  \\ 
				& SelectionGAN \cite{2019selectiongan}  & 282.3 & 0.2328 & 0.5261 & 1.797 & 24.83  \\ 
				& TSIT \cite{2020tsit} & 286.7 & 0.2486 & 0.5245 & 1.628 & 24.25  \\ 
				& LPTN \cite{2021LPTN} & 379.7 & 0.3364 & 0.7346 & 1.753 & 21.48  \\ 
				& Ours  & \bf{172.2} & \bf{0.0644} & \bf{0.2548} & \bf{2.212} & 25.73 \\ 
				
				\midrule
				\multirow{10}{*}{16}
				& Pix2Pix \cite{pix2pix}       & 243.5 & 0.1677 & 0.5587 & 1.896 & 24.27  \\ 
				& Pix2PixHD \cite{pix2pixhd}  & 174.0 & 0.0608 & 0.2714 & 2.824 & 26.14    \\ 
				& CycleGAN \cite{cyclegan}   & 277.1 & 0.2354 & 0.5900 & 1.881 & 24.73  \\
				& GeoGAN \cite{2019geogan}  & 491.4 & 0.3908 & 0.8700 & 1.307 & 27.16 \\ 
				& SMAPGAN \cite{2020smapgan}  & 181.5 & 0.0670 & 0.2398 & 2.774 & \bf{27.88}  \\ 
				& SPADE \cite{2019spade} & 323.5 & 0.2766 & 0.6547 & 2.228 & 25.61  \\ 
				& SelectionGAN \cite{2019selectiongan}  & 217.5 & 0.1142 & 0.3756 & 2.759 & 26.02  \\ 
				& TSIT \cite{2020tsit} & 247.0 & 0.1713 & 0.5434 & 2.003 & 25.63  \\ 
				& LPTN \cite{2021LPTN} & 356.1 & 0.2440 & 0.7114 & 1.796 & 25.40  \\ 
				& Ours  & \bf{167.7} & \bf{0.0652} & \bf{0.2371} & \bf{3.333} & 27.52 \\ 
				
				\midrule
				\multirow{10}{*}{17}
				& Pix2Pix \cite{pix2pix}       & 104.6 & 0.0670 & 0.3668 & 2.059 & 24.93  \\ 
				& Pix2PixHD \cite{pix2pixhd}  & 54.9 & 0.0177 & 0.1516 & 2.557 & 24.81    \\ 
				& CycleGAN \cite{cyclegan}   & 53.4 & 0.0203 & 0.2244 & 2.372 & 25.09  \\
				& GeoGAN \cite{2019geogan}  & 380.1 & 0.3367 & 0.8179 & 1.374 & 26.49 \\ 
				& SMAPGAN \cite{2020smapgan}  & 303.5 & 0.2495 & 0.5861 & 2.674 & \bf{26.72}  \\ 
				& SPADE \cite{2019spade} & 145.1 & 0.1070 & 0.4700 & 2.372 & 25.03  \\ 
				& SelectionGAN \cite{2019selectiongan}  & 84.6 & 0.0450 & 0.2362 & \bf 2.728 & 24.75  \\ 
				& TSIT \cite{2020tsit} & 97.5 & 0.0734 & 0.3369 & 2.196 & 25.10  \\ 
				& LPTN \cite{2021LPTN} & 241.7 & 0.1519 & 0.5968 & 1.879 & 25.26  \\ 
				& Ours  & \bf 52.1  & \bf 0.0153 & \bf 0.1177 & 2.527 & 25.14 \\
				
				\midrule
				\multirow{10}{*}{18}
				& Pix2Pix \cite{pix2pix}       & 52.0 & 0.0204 & 0.1912 & 1.985 & 26.91  \\ 
				& Pix2PixHD \cite{pix2pixhd}  & 42.1 & 0.0126 & 0.1245 & 2.528 & 27.02    \\ 
				& CycleGAN \cite{cyclegan}   & 44.0 & 0.0139 & 0.1395 & 2.181 & 26.75  \\
				& GeoGAN \cite{2019geogan}  & 407.6 & 0.2494 & 0.8428 & 1.405 & 28.33 \\ 
				& SMAPGAN \cite{2020smapgan}  & 317.1 & 0.1600 & 0.6077 & 2.109 & \bf{28.88}  \\ 
				& SPADE \cite{2019spade} & 116.3 & 0.0705 & 0.3956 & 2.373 & 26.69  \\ 
				& SelectionGAN \cite{2019selectiongan}  & 78.3 & 0.0451 & 0.2046 & 2.440 & 26.64  \\ 
				& TSIT \cite{2020tsit} & 95.3 & 0.0517 & 0.3493 & 2.312 & 26.79  \\ 
				& LPTN \cite{2021LPTN} & 200.4 & 0.0916 & 0.5360 & 2.043 & 26.08  \\ 
				& Ours  & \bf{32.7} & \bf{0.0061} & \bf{0.0762} & \bf{2.719} & 26.92 \\ 
				
				\midrule
				\multirow{10}{*}{Avg.}
				& Pix2Pix \cite{pix2pix}       & 170.2 & 0.1235 & 0.4202 & 1.975 & 24.75  \\ 
				& Pix2PixHD \cite{pix2pixhd}  & 130.1 & 0.0675 & 0.2680 & 2.475 & 25.61    \\ 
				& CycleGAN \cite{cyclegan}   & 173.6 & 0.1410 & 0.3922 & 2.030 & 24.99  \\
				& GeoGAN \cite{2019geogan}  & 457.0 & 0.3644 & 0.8601 & 1.314 & 26.87 \\ 
				& SMAPGAN \cite{2020smapgan}  & 255.1 & 0.1449 & 0.4456 & 2.384 & \bf{27.32}  \\ 
				& SPADE \cite{2019spade} & 235.8 & 0.2044 & 0.5467 & 2.260 & 25.39  \\ 
				& SelectionGAN \cite{2019selectiongan}  & 165.7 & 0.1093 & 0.3356 & 2.431 & 25.56  \\ 
				& TSIT \cite{2020tsit} & 181.6 & 0.1363 & 0.4385 & 2.035 & 25.44  \\ 
				& LPTN \cite{2021LPTN} & 294.5 & 0.2060 & 0.6447 & 1.868 & 24.56  \\ 
				& Ours  & \bf{106.2} & \bf{0.0378} & \bf{0.1715} & \bf{2.698} & 26.33 \\ 
				\bottomrule[1.5pt]
		\end{tabular}}
	\end{center}
	\vspace{-0mm}
	\caption{Quantitative results on Multi-Level Map Rio de Janeiro (MLM-RJ) dataset.
		 ``Avg.'' indicates results that average over 15-18 zoom levels and also shows consistency results.
	}
	\vspace{-0mm}
	\label{tab:mainresults2}
\end{table}

\begin{table}[tb]
	\begin{center}
		\scalebox{0.73}{
			{\color{black}
			\begin{tabular}{c|c|c|ccccc}
				\specialrule{0em}{1.5pt}{1.5pt}
				\toprule[1.5pt]
				Level & Method& Train on & FID$\downarrow$ & KID$\downarrow$ & KMMD$\downarrow$ & IS$\uparrow$ &PSNR$\uparrow$ \\
				\midrule
				\multirow{6}{*}{15}
				& \multirow{2}{*}{TSIT \cite{2020tsit}}   &Pix2pix    &362.3 & 0.3058 & 0.6371 & \bf 2.079 & 21.53 \\ &&MLM-SH    & \bf 282.8 & \bf 0.2877 & \bf 0.5879 & 1.791 & \bf 23.81 \\
				\cmidrule{2-8}
				& \multirow{2}{*}{SelectionGAN \cite{2019selectiongan}}   &Pix2pix    & 454.2 & 0.4004 & 0.7490 & 1.585 & 19.57 \\ &&MLM-SH    & \bf 433.8 & \bf 0.3495 & \bf 0.6784 & \bf 2.262 & \bf 24.94 \\
				\cmidrule{2-8}
				& \multirow{2}{*}{LAMG}   &Pix2pix    & 435.0 & 0.3877 & 0.7856 & 1.445 & 16.02 \\ &&MLM-SH    & \bf 409.2 & \bf 0.3062 & \bf 0.5618 & \bf 2.883 & \bf 20.83 \\
				
				\midrule
				\multirow{6}{*}{16}
				& \multirow{2}{*}{TSIT \cite{2020tsit}}   &Pix2pix    &323.3 & 0.2608 & \bf 0.5685 & \bf 2.301 & 21.87 \\ &&MLM-SH    & \bf 263.6 & \bf 0.2168 & 0.5771 & 1.874 & \bf 24.56 \\
				\cmidrule{2-8}
				& \multirow{2}{*}{SelectionGAN \cite{2019selectiongan}}   &Pix2pix    & 414.9 & 0.3236 & 0.6914 & 2.136 & 20.34 \\ &&MLM-SH    & \bf 395.8 & \bf 0.2765 & \bf 0.6094 & \bf 2.200 & \bf 26.01 \\
				\cmidrule{2-8}
				& \multirow{2}{*}{LAMG}   &Pix2pix    & 416.1 & 0.3255 & 0.7443 & 1.531 & 16.06 \\ &&MLM-SH    & \bf 296.0 & \bf 0.1562 & \bf 0.4338 & \bf 2.781 & \bf 24.60 \\
				
				\midrule
				\multirow{6}{*}{17}
				& \multirow{2}{*}{TSIT \cite{2020tsit}}   &Pix2pix    &290.1 & 0.3177 & 0.5628 & \bf 2.346 & 21.34 \\ &&MLM-SH    & \bf 194.6 & \bf 0.1464 & \bf 0.4613 & 2.113 & \bf 24.21 \\
				\cmidrule{2-8}
				& \multirow{2}{*}{SelectionGAN \cite{2019selectiongan}}   &Pix2pix    & 341.0 & 0.2905 & 0.6476 & 2.258 & 19.93 \\ &&MLM-SH    & \bf 338.4 & \bf 0.2809 & \bf 0.6395 & \bf 2.367 & \bf 24.79 \\
				\cmidrule{2-8}
				& \multirow{2}{*}{LAMG}   &Pix2pix    & 336.4 & 0.3042 & 0.7351 & 1.576 & 15.99 \\ &&MLM-SH    & \bf 267.5 & \bf 0.1940 & \bf 0.5526 & \bf 2.792 & \bf 24.14 \\
				
				\midrule
				\multirow{6}{*}{18}
				& \multirow{2}{*}{TSIT \cite{2020tsit}}   &Pix2pix    &264.8 & 0.2454 & 0.6394 & \bf 2.747 & 20.73 \\ &&MLM-SH    & \bf 175.7 & \bf 0.1182 & \bf 0.5180 & 2.164 & \bf 24.75 \\
				\cmidrule{2-8}
				& \multirow{2}{*}{SelectionGAN \cite{2019selectiongan}}   &Pix2pix    & 373.5 & 0.2422 & 0.6720 & \bf 2.537 & 19.08 \\ &&MLM-SH    & \bf 339.5 & \bf 0.1891 & \bf 0.6027 & 2.448 & \bf 25.38 \\
				\cmidrule{2-8}
				& \multirow{2}{*}{LAMG}   &Pix2pix    & 362.6 & 0.2443 & 0.7301 & 1.625 & 16.15 \\ &&MLM-SH    & \bf 225.1 & \bf 0.0999 & \bf 0.4521 & \bf 2.859 & \bf 25.24 \\
				
				\midrule
				\multirow{6}{*}{Avg.}
				& \multirow{2}{*}{TSIT \cite{2020tsit}}   &Pix2pix    &310.1 & 0.2824 & 0.6019 & \bf 2.368 & 21.37 \\ &&MLM-SH    & \bf 229.2 & \bf 0.1923 & \bf 0.5361 & 1.986 & \bf 24.33 \\
				\cmidrule{2-8}
				& \multirow{2}{*}{SelectionGAN \cite{2019selectiongan}}   &Pix2pix    & 395.9 & 0.3142 & 0.6900 & 2.129 & 19.73 \\ &&MLM-SH    & \bf 376.9 & \bf 0.2740 & \bf 0.6325 & \bf 2.319 & \bf 25.28 \\
				\cmidrule{2-8}
				& \multirow{2}{*}{LAMG}   &Pix2pix    & 387.5 & 0.3154 & 0.7488 & 1.544 & 16.06 \\ &&MLM-SH    & \bf 299.4 & \bf 0.1891 & \bf 0.5001 & \bf 2.829 & \bf 23.70 \\ 
				\bottomrule[1.5pt]
		\end{tabular}}}
	\end{center}
	\vspace{-0mm}
	\caption{{\color{black}Quantitative results of different methods training on Pix2pix dataset and MLM-SH dataset. 
		``Avg.'' indicates results that average over 15-18 zoom levels and also shows consistency results.}
	}
	\vspace{-0mm}
	\label{tab:DatasetExperiment}
\end{table}

\subsection{Learning Details}
In this section, we first introduce the loss function and then describe the implementation details. 
Notice that the level-aware map generator and consistency-aware map generator utilize the same loss function.

\vspace{+1mm}
\noindent{\bf Losses for Semantic Module.} 
The semantic module is trained to learn to understand aerial images pixel-wisely, we follow \cite{fcn, deeplabv3+, denseaspp, danet} to use pixel-wise cross entropy loss for training. 
And to better adapt to hard samples, we adopt focal loss \cite{2017retinanet} version of it:
\begin{equation}
\begin{aligned}
    	\mathcal{L}_{SM} &\!=\! \frac{1}{H\times W} \sum_{i=1}^{H\times W} FL(p_{i},  y^{seg}_{i}),  \\
	FL(p_{i}, y^{seg}_{i}) &\!=\!
	\begin{cases}
  		-\alpha_t(1-p_{it})^\gamma\log(p_{it})&\!\!\!\!\! \text{if $ y^{seg}_{i}=t$}\\
    		-\alpha_t\ p_{it}^\gamma\log(1-p_{it})&\!\!\!\!\! \text{otherwise}
    	\end{cases}\!,
\end{aligned}
\end{equation}
where $H$, $W$ indicate the height and width of the image, $y^{seg}$ is ground truth semantic segmentation label and $p_{it}\in[0,1]$ is predicted confidence value of pixel $i$ for category $t$.
$\alpha_t$, $\gamma$ are hyper-parameters of focal loss \cite{2017retinanet}, whose values are 1.0, 2.0 respectively. 
The ground truth semantic segmentation label is obtained from the real map by mapping different colors of the map to semantic categories, \eg, blue indicates rivers, yellow indicates roads, etc. 
We mainly focus on segmenting three semantic categories, \ie, roads, water area and land.

\vspace{+1mm}
\noindent{\bf Losses for Drawer Module.}
To generate high-quality maps with realistic details, we adopt content loss, adversarial loss and perceptual loss for the training drawer module.

The content loss \cite{pix2pix, pix2pixhd} improves the similarity between generated map and real map:
\begin{equation}
	\mathcal{L}_{con}=\frac{1}{H\times W} \sum_{i=1}^{H\times W} \lVert G(x) -y^{map} \rVert_1, 
\end{equation}
where $y^{map}$ is ground truth map and $G$ is generator. 

{\color{blue} The adversarial loss \cite{2014gan, cgan} is able to make the details of image more realistic: 
\begin{equation}
    \begin{aligned}
    \mathcal{L}_{adv}=& \mathop{min}_{G}\mathop{max}_{D}V(D,G)
	\\=& \mathbb{E}_{x, y^{map}}[\log D(x,y^{map})]
	\\&+\mathbb{E}_{y^{map}}[\log(1-D(x,G(x)))],
    \end{aligned} 
\end{equation}
where $D$ is discriminator, $G$ is generator, and $x, y^{map}$ indicate aerial image and map image.
To enhance the discriminator's ability to differentiate real and synthesized maps at different zoom level, we adopt multi-scale discriminator~\cite{pix2pixhd}.
}

The perceptual loss \cite{2016perceptual, pix2pixhd} is utilized to improve the perceptual quality of generated map:
\begin{equation}
\mathcal{L}_{per}\!=\! 
\sum_{i=1}^{L}
\frac{1}{N_i}
 \lVert F^{(i)}(y^{map})-F^{(i)}(G(x)) \rVert_1, 
\end{equation}
where $F$ is feature extraction network, $F^{(i)}$ indicates the $i$-th layer of feature extraction network, $N_i$ is the number of pixels of corresponding feature map, and $L$ is the number of layers. 
We use a pre-trained VGG model \cite{vgg} as feature extraction network $F$.

To better stabilize the training procedure \cite{pix2pixhd}, we adopt feature-matching loss for training. 
Specifically, we extract features from multiple layers of the discriminator, and learn to match these intermediate representations from the real map and generated map:
\begin{equation}
\mathcal{L}_{fm}=
\sum\limits_{i=1}^{L}
\frac{1}{N_i}
\lVert D^{(i)}(x,y^{map})-D^{(i)}(x,G(x)) \rVert_1, 
\end{equation}
where $D^{(i)}$ means the first $i$ layers of discriminator $D$, $L$ means the total number of layers, and $N_i$ denotes the number of elements in each layer.

Thus the total loss function for drawer module is 
\begin{equation}
	\mathcal{L}_{DM}=\lambda_{con} \mathcal{L}_{con} + \lambda_{adv} \mathcal{L}_{adv}
	+\lambda_{per} \mathcal{L}_{per} + \lambda_{fm} \mathcal{L}_{fm}, 
\end{equation}
where we let $\lambda_{con} = \lambda_{per} = \lambda_{fm} = 10$ and $\lambda_{adv} = 1$ for loss balancing.

\vspace{+1mm}
\noindent{\bf Implementation Details.}
We first train the level-aware map generator, then fix it to help to optimize the consistency-aware map generator. 
The training procedures of the two generators are the same. 
The input size of paired map tile is $256\times256$. 
We train the semantic module and drawer module jointly with Adam optimizer \cite{2014adam}. 
For balancing the learning of each part, we set the initial learning rate of the semantic module, the backbone of semantic module and the drawer module to $7\times 10^{-4}$, $7\times 10^{-5}$ and $2\times 10^{-4}$ respectively.
And the batch size is 4. 
Both generators are trained for 100 epochs on a GeForce RTX 3090.
The semantic module adopts a polynomial learning rate decay schedule \cite{deeplabv3+}
while the learning rate of the drawer module is fixed for the first 50 epochs, and linearly decays to 0 for the rest 50 epochs.

\section{Multi-Level Map Dataset}
\label{sec:mlm}
To develop and evaluate learning-based multi-level map generation methods, a dataset consisting of paired samples of aerial images and maps with multiple zoom levels is necessarily needed.
Thus, we collect and produce a large-scale high-quality Multi-Level Map (MLM) dataset for map generation at multiple zoom levels.  
The data comes from Google Maps and Tianditu.
It has the following distinctive characteristics. 
\begin{itemize} 
\item {\bf Region Diversity.} 
As shown in Figure~\ref{fig:SH} and Figure~\ref{fig:RIO}, our MLM dataset provides data for two regions with different geographical features, \ie, Shanghai and Rio de Janeiro, which are representative cities from the Northern Hemisphere and the Southern Hemisphere. 
For brevity, we indicate the subset of samples in Shanghai as {\bf MLM-SH} and the subset of samples in Rio de Janeiro as {\bf MLM-RJ}. 
\item {\bf Multi-Level.} 
As shown in Table~\ref{tab:MLM}, the MLM dataset provides paired samples in four different zoom levels, \ie, 15, 16, 17 and 18 that cover the central region of two cities. 
The samples at zoom levels lower than 15 are too few due to area size and high distance per pixel ratio, while those at zoom levels higher than 19 are hard to access due to usage limitations. 
All of the samples form a complete multi-level map as illustrated in Figure~\ref{fig:SH} and Figure~\ref{fig:RIO}. 
\item {\bf Large Scale.}
MLM dataset has 18,700 high-quality paired samples, \ie, 10,200 samples of Shanghai and 8,500 samples of Rio de Janeiro. 
It is much larger than existing map dataset in~\cite{2019multiscalemap} and dataset in~\cite{pix2pix} as shown in Table~\ref{tab:datasets}. 

{\color{black}
\item {\bf High-quality.}
During data collection, we ensure the image pairs at different levels show the same corresponding geographic area.
And each map image is semantically aligned with the corresponding aerial image.
}
\end{itemize}


\section{Experiments}
The proposed method is systematically evaluated on our MLM-SH and MLM-RJ datasets. 
We use their train set for training and test set for evaluation. 
In this section, evaluation metrics are first introduced.
Then, we quantitatively and qualitatively evaluate our method. 
Finally, the ablation studies are performed to analyze our proposed method.

\begin{figure*}[tb!]
	\begin{center}
	
\iffalse
	\scalebox{0.66}{
	\begin{tabular}{cccccccc}
		\begin{sideways}\qquad \quad 15-th\ level\end{sideways}
		\includegraphics[width=4.01cm, angle=0]{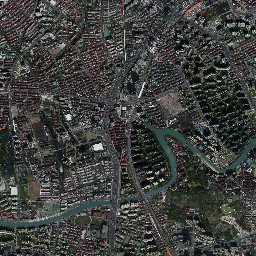}\hspace{-3mm}
		&\includegraphics[width=4.01cm, angle=0]{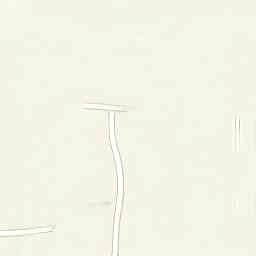}\hspace{-3mm}
		&\includegraphics[width=4.01cm, angle=0]{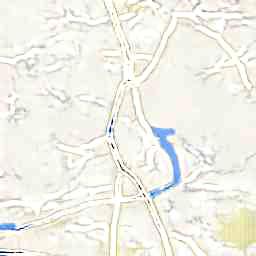}\hspace{-3mm}
		&\includegraphics[width=4.01cm, angle=0]{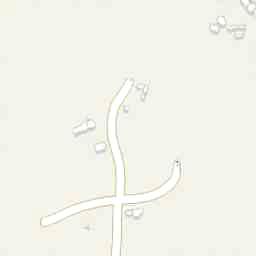}\hspace{-3mm}
		&\includegraphics[width=4.01cm, angle=0]{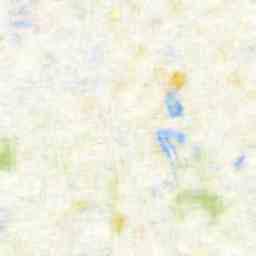}\hspace{-3mm}
		&\includegraphics[width=4.01cm, angle=0]{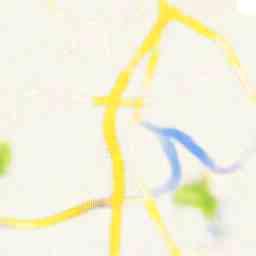}\hspace{-3mm}\quad
		\begin{sideways}\qquad \quad $256\times 256$\end{sideways}
		\\
		\begin{sideways}\qquad \quad 16-th\ level\end{sideways}
		\includegraphics[width=4.01cm, angle=0]{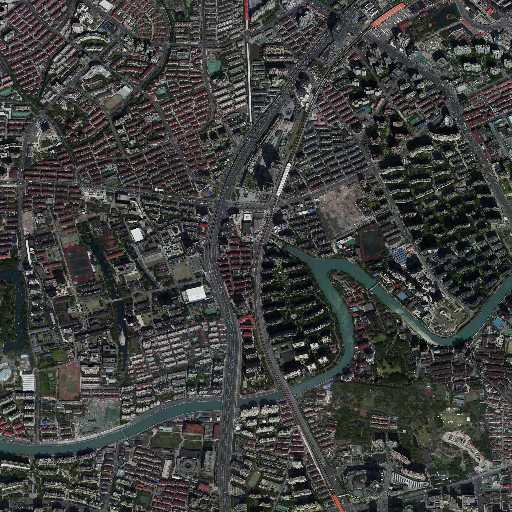}\hspace{-3mm}
		&\includegraphics[width=4.01cm, angle=0]{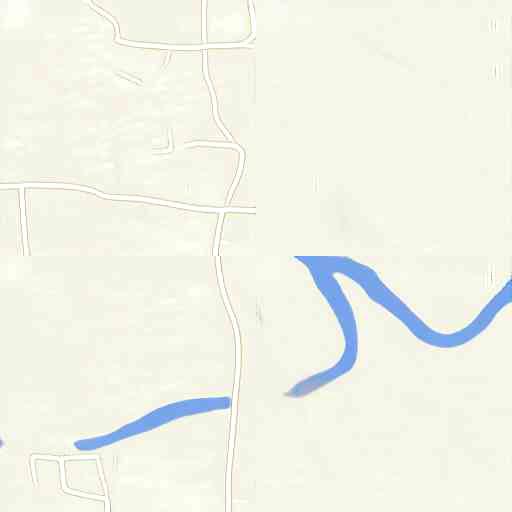}\hspace{-3mm}
		&\includegraphics[width=4.01cm, angle=0]{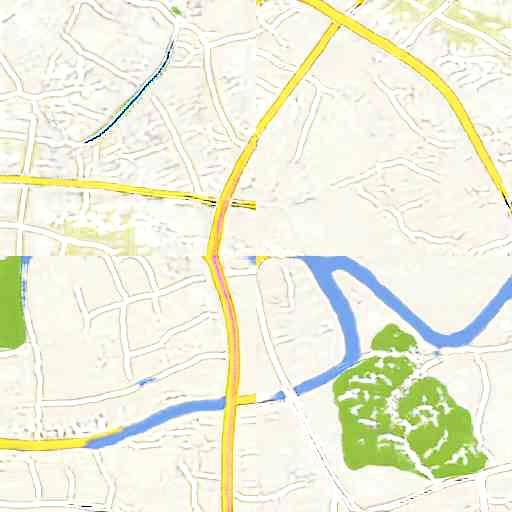}\hspace{-3mm}
		&\includegraphics[width=4.01cm, angle=0]{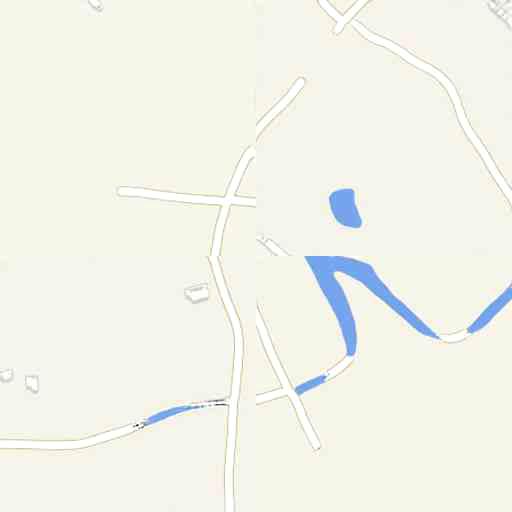}\hspace{-3mm}
		&\includegraphics[width=4.01cm, angle=0]{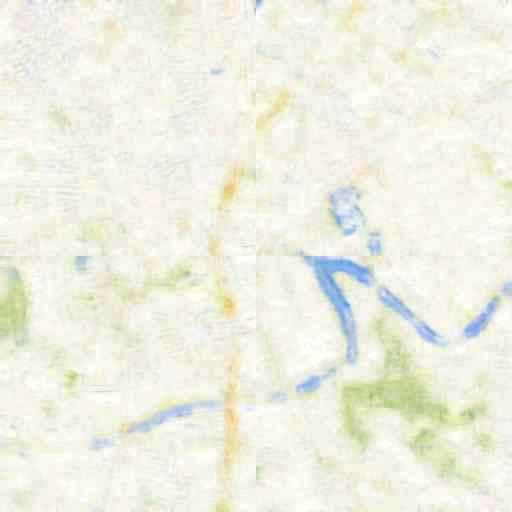}\hspace{-3mm}
		&\includegraphics[width=4.01cm, angle=0]{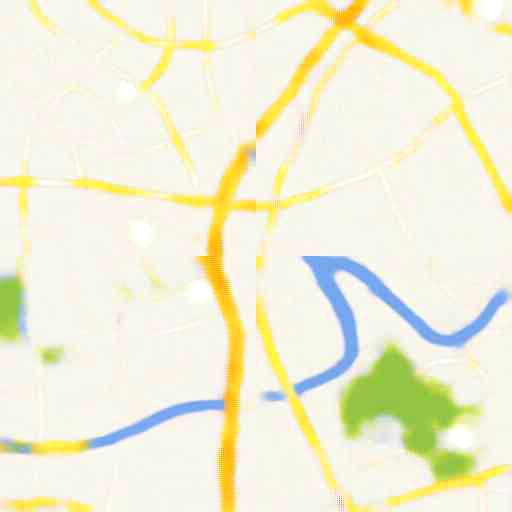}\hspace{-3mm}\quad
		\begin{sideways}\qquad \quad $512\times 512$\end{sideways}
		\\
		\begin{sideways}\qquad \quad 17-th\ level\end{sideways}
		\includegraphics[width=4.01cm, angle=0]{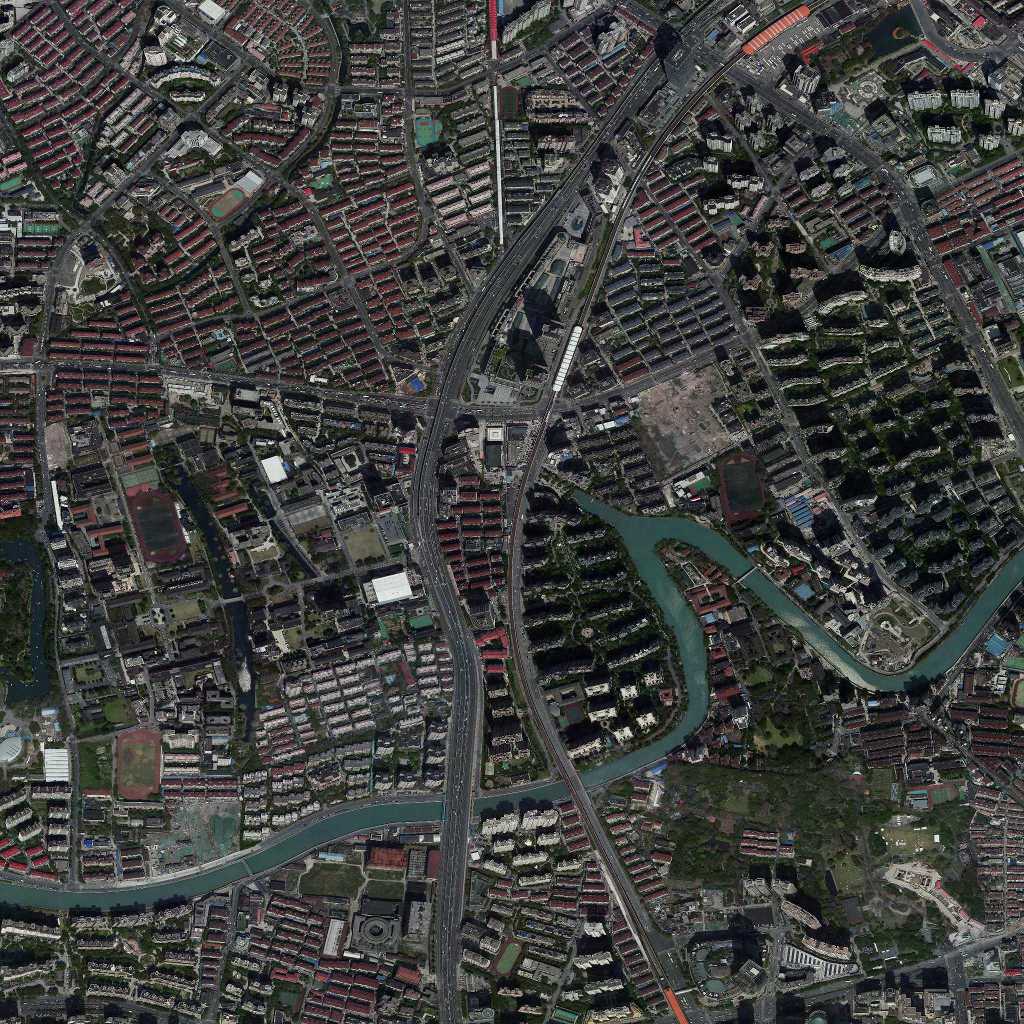}\hspace{-3mm}
		&\includegraphics[width=4.01cm, angle=0]{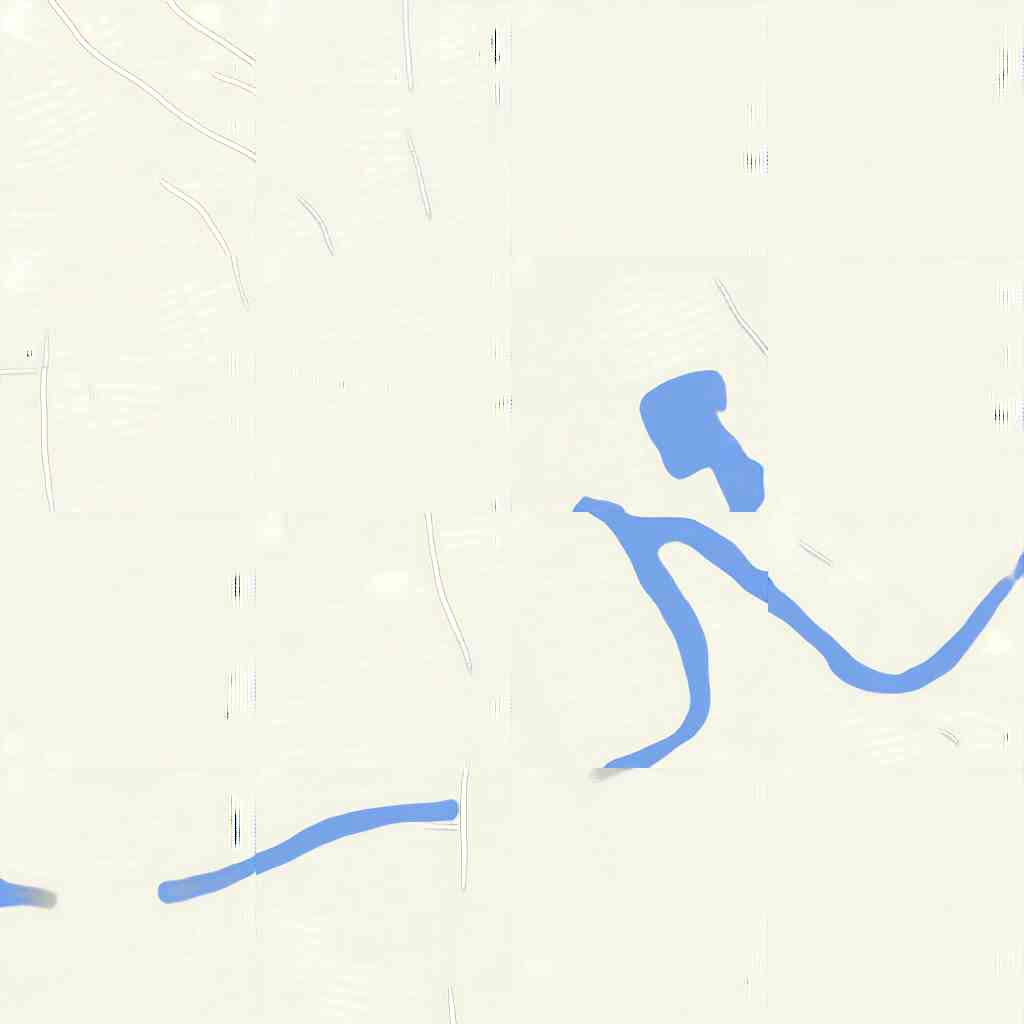}\hspace{-3mm}
		&\includegraphics[width=4.01cm, angle=0]{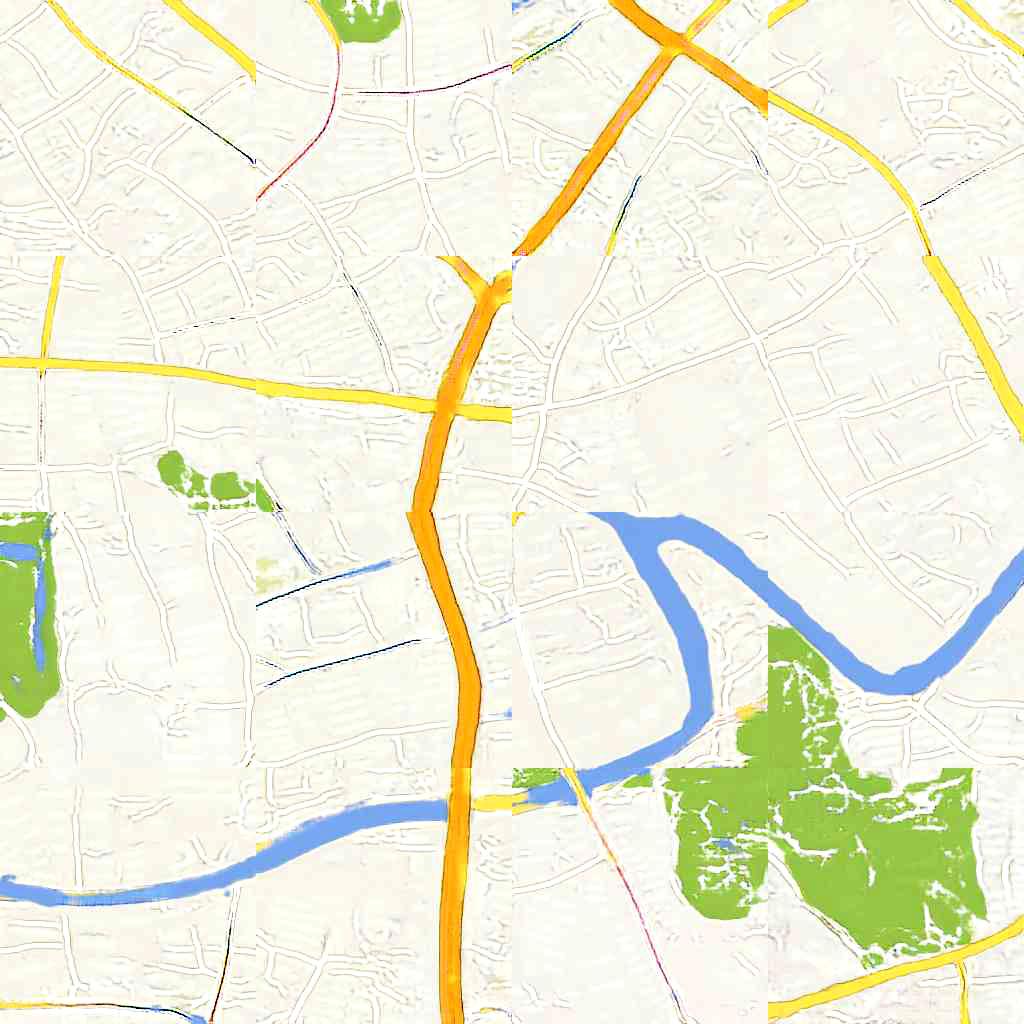}\hspace{-3mm}
		&\includegraphics[width=4.01cm, angle=0]{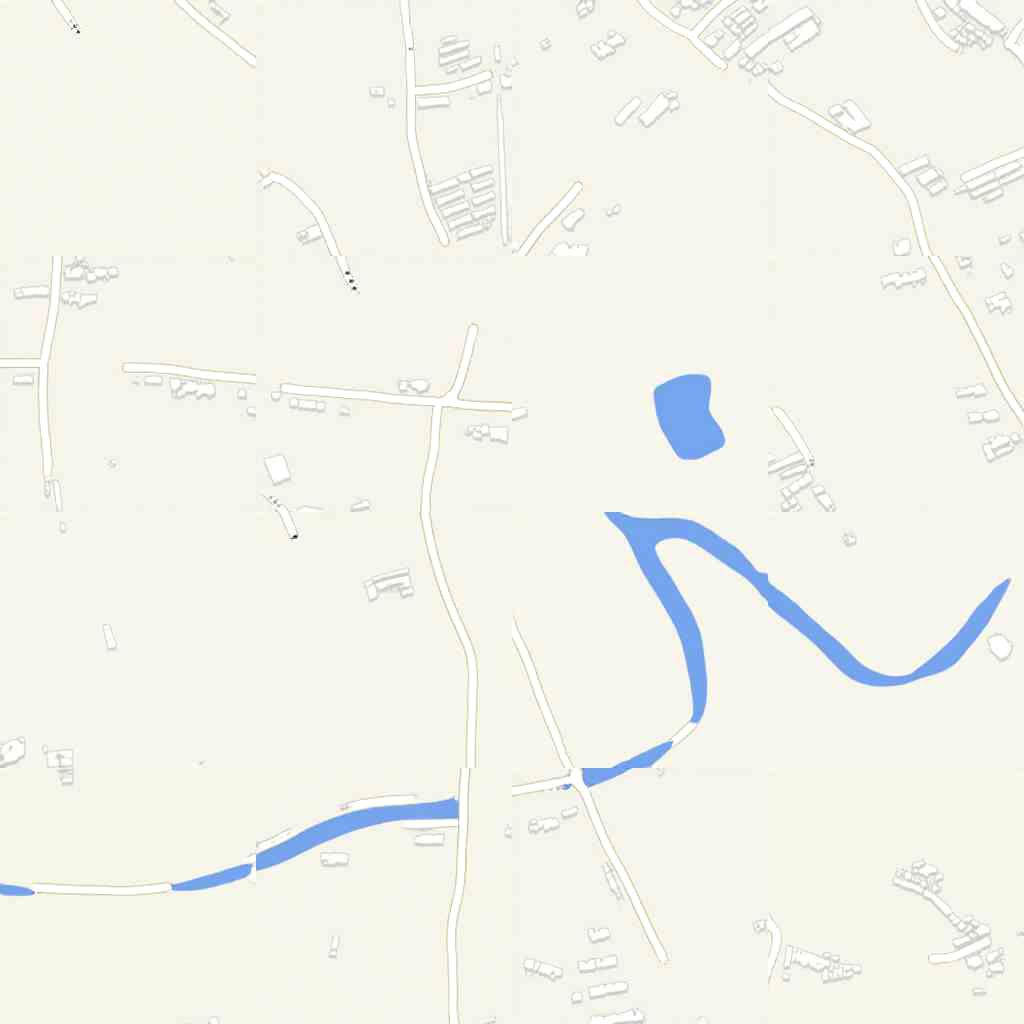}\hspace{-3mm}
		&\includegraphics[width=4.01cm, angle=0]{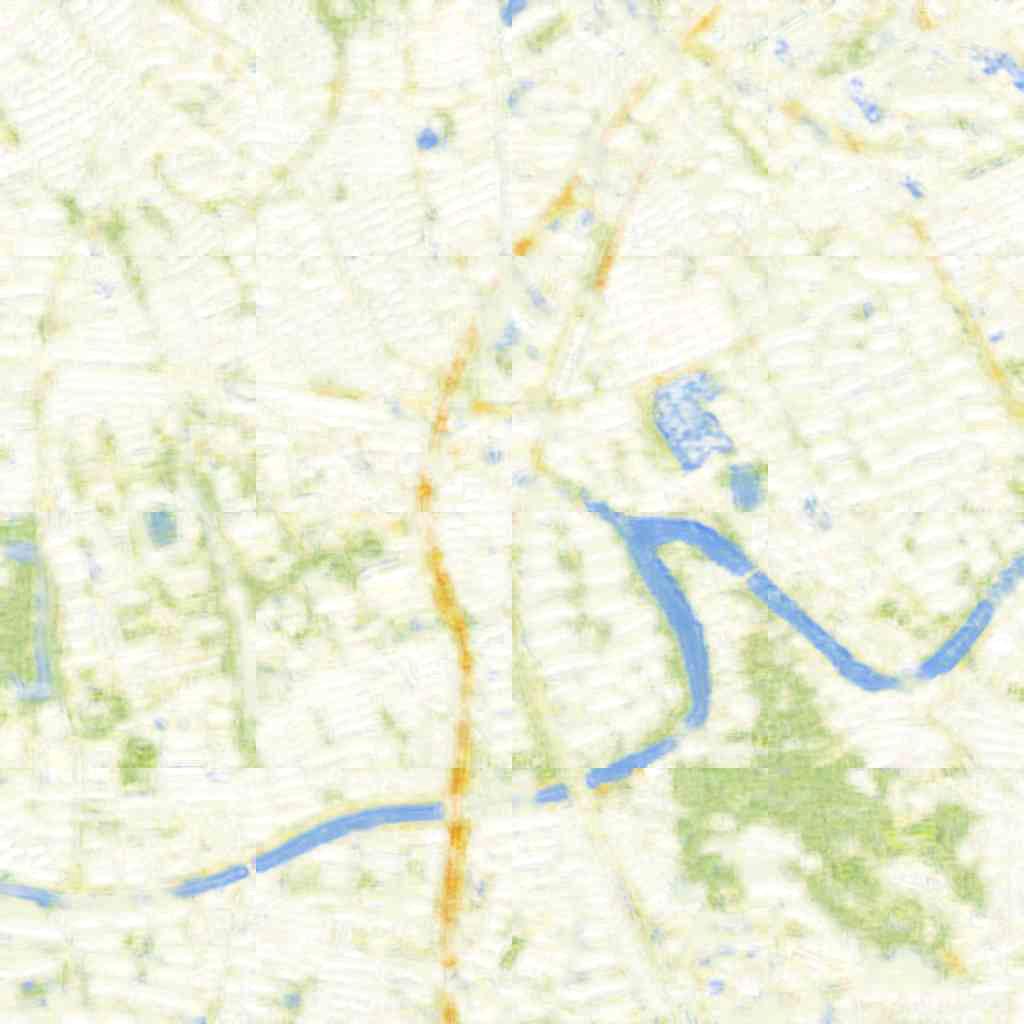}\hspace{-3mm}
		&\includegraphics[width=4.01cm, angle=0]{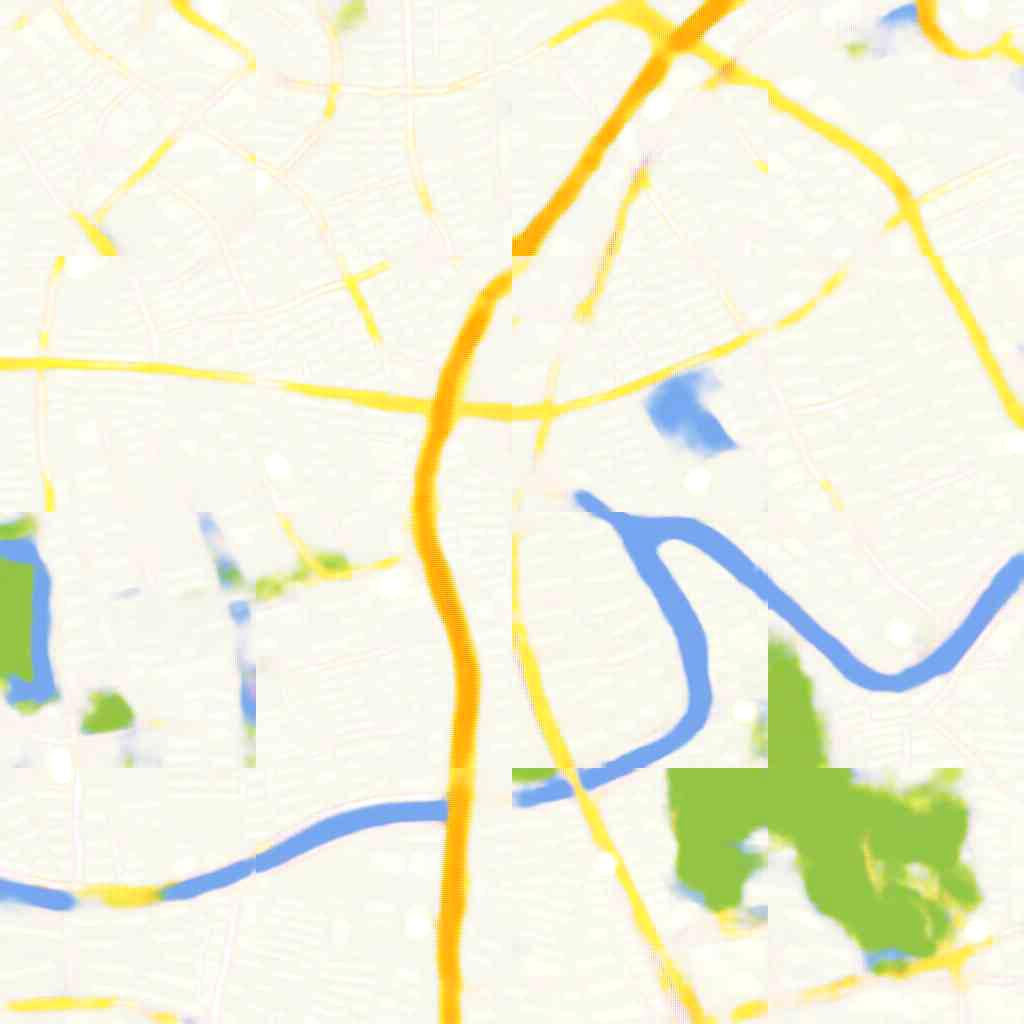}\hspace{-3mm}\quad
		\begin{sideways}\qquad \quad $1024\times 1024$\end{sideways}
		\\
		
		\begin{sideways}\qquad \quad 18-th\ level\end{sideways}
		\includegraphics[width=4.01cm, angle=0]{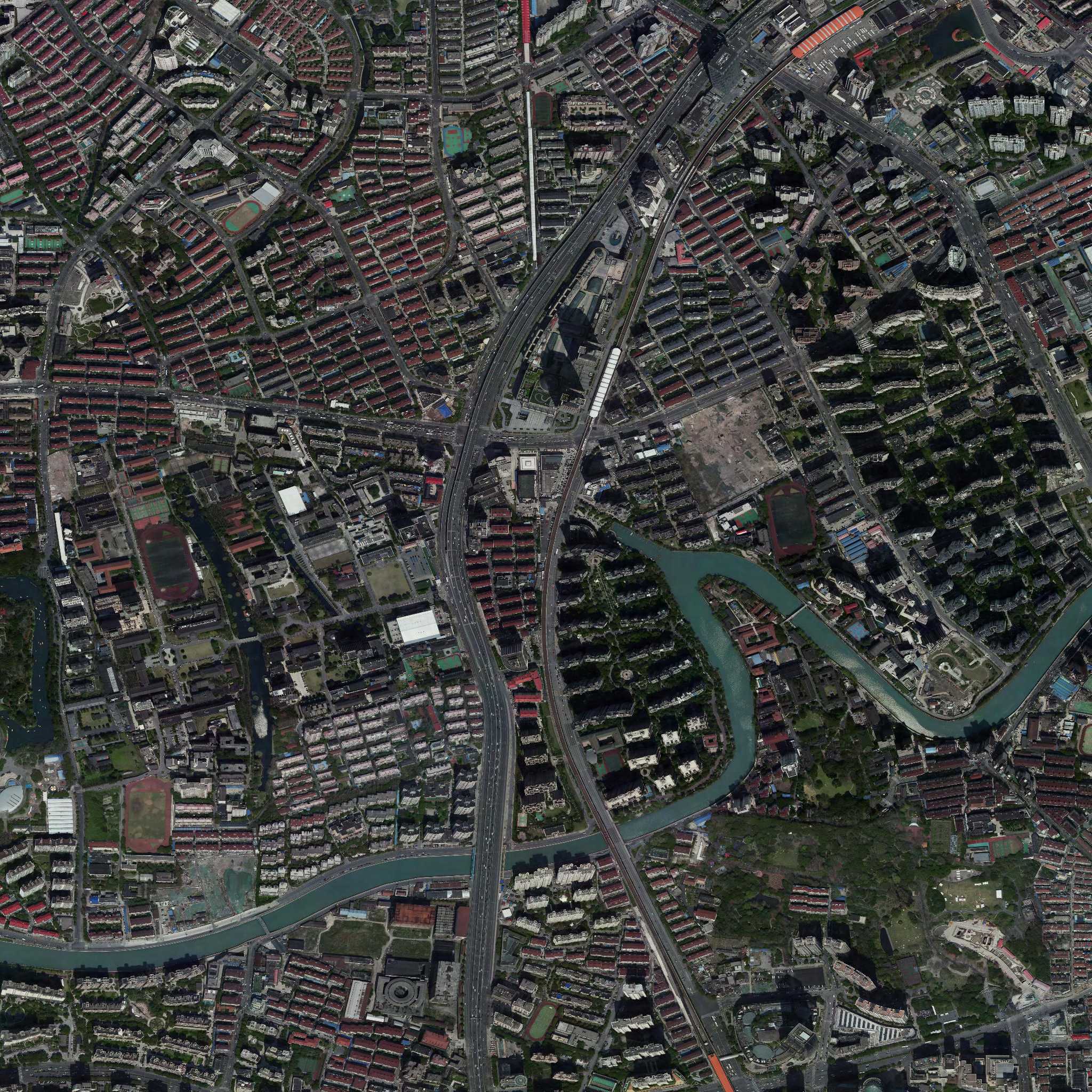}\hspace{-3mm}
		&\includegraphics[width=4.01cm, angle=0]{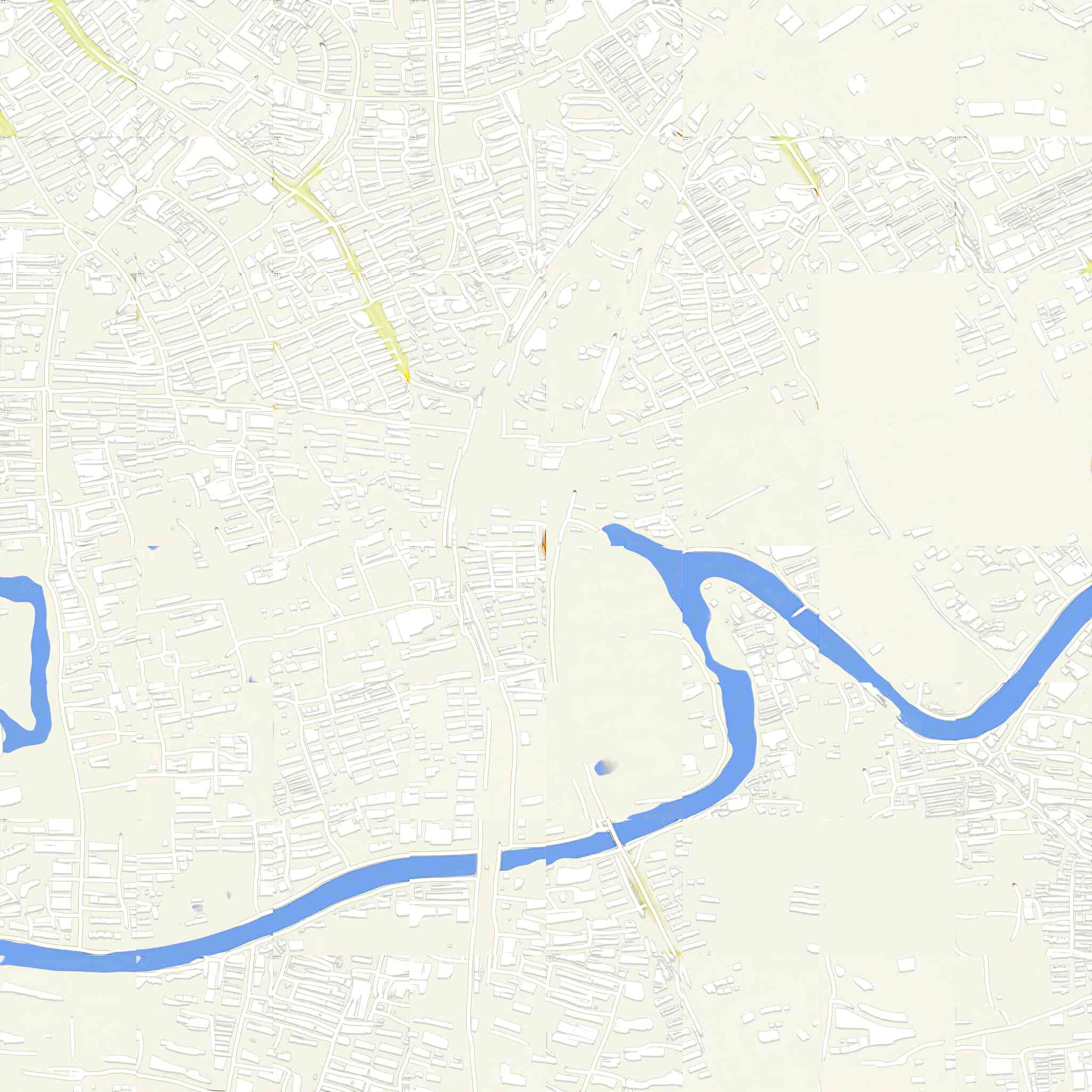}\hspace{-3mm}
		&\includegraphics[width=4.01cm, angle=0]{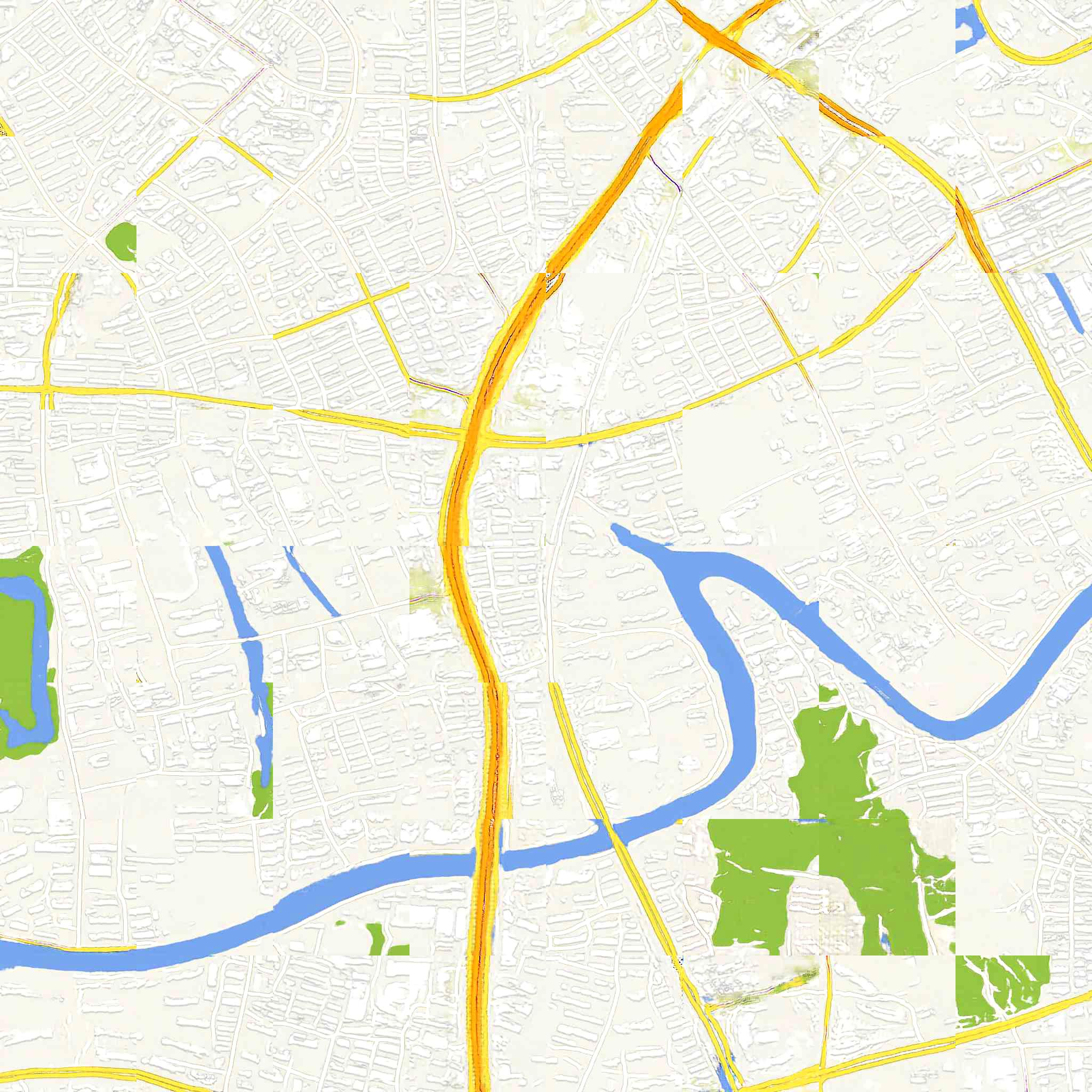}\hspace{-3mm}
		&\includegraphics[width=4.01cm, angle=0]{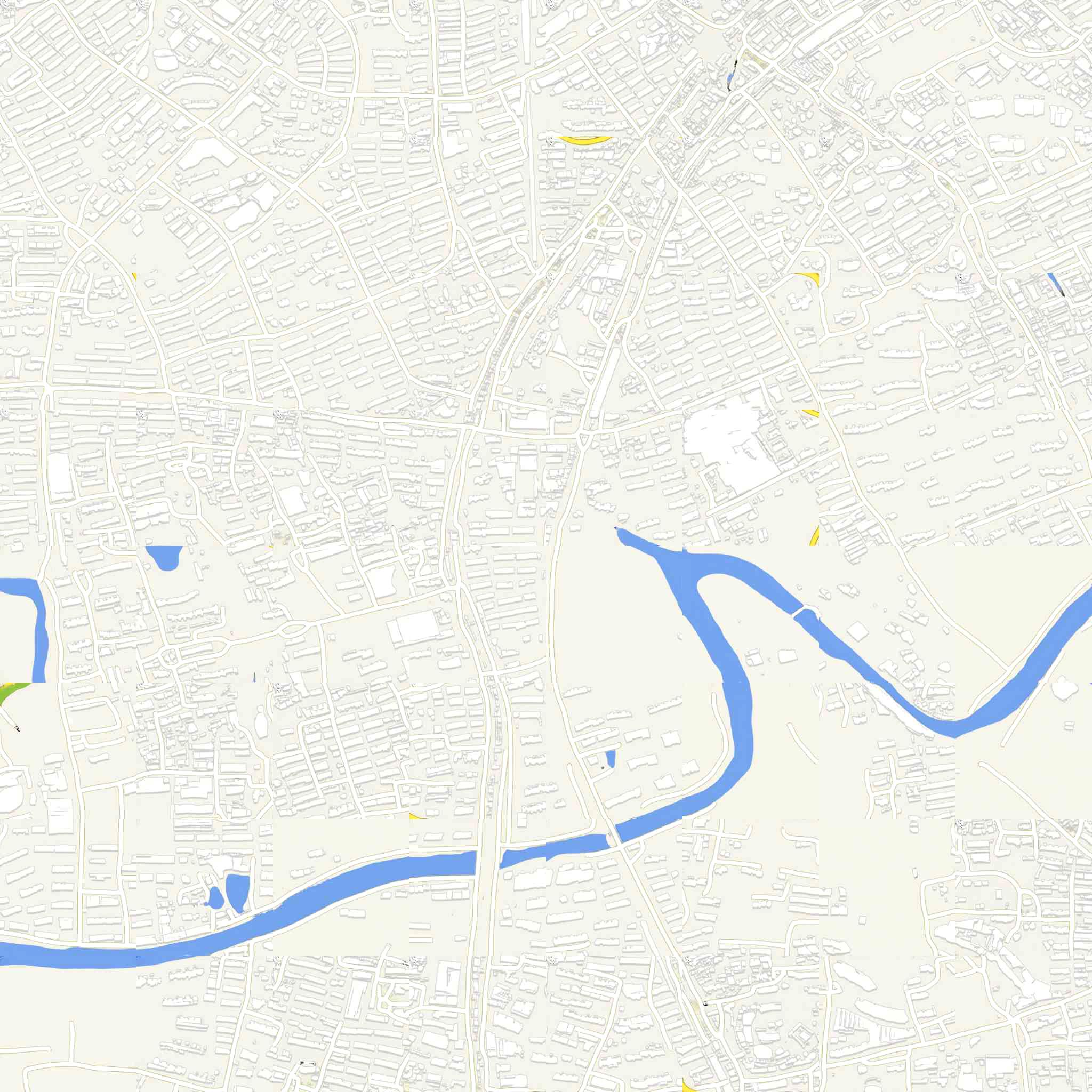}\hspace{-3mm}
		&\includegraphics[width=4.01cm, angle=0]{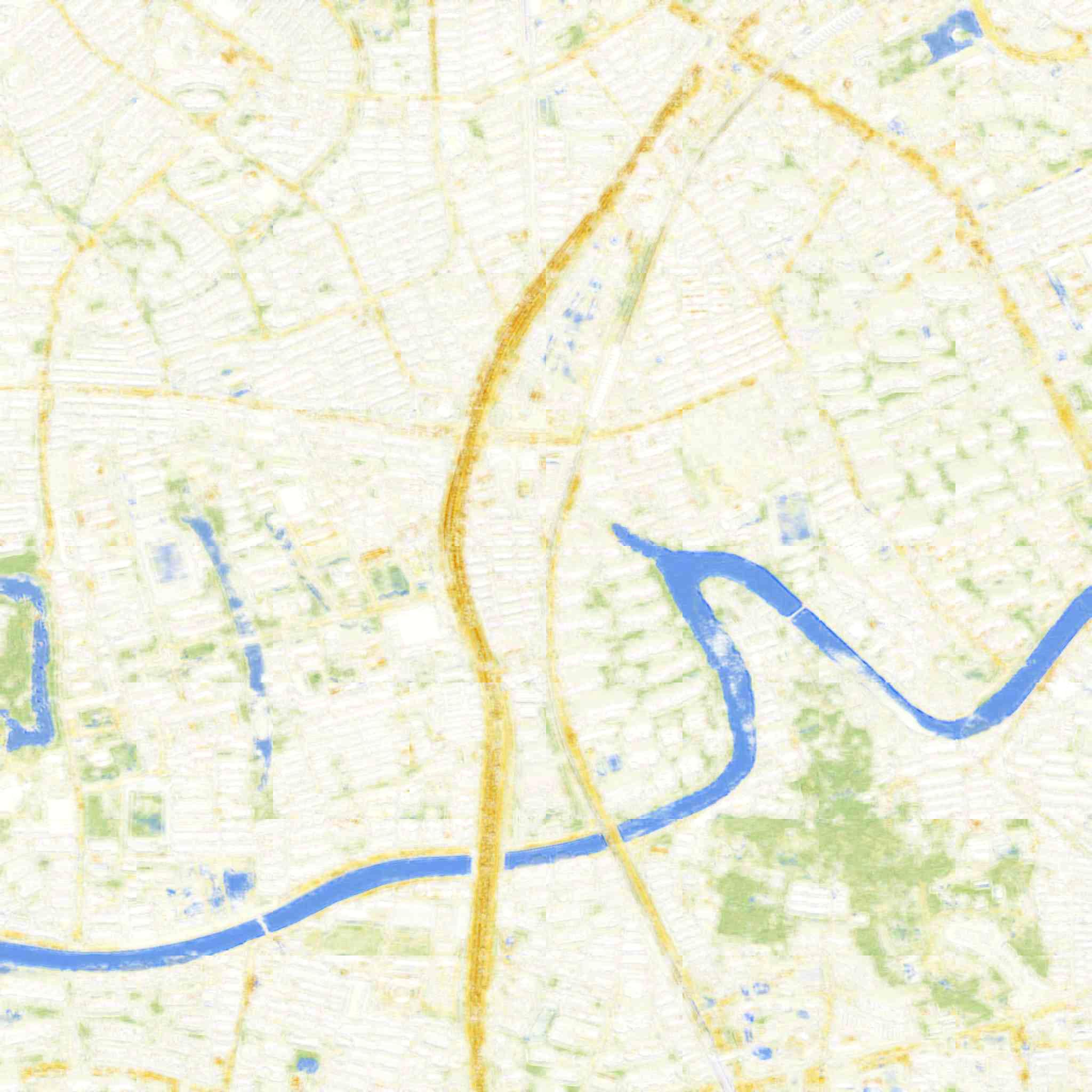}\hspace{-3mm}
		&\includegraphics[width=4.01cm, angle=0]{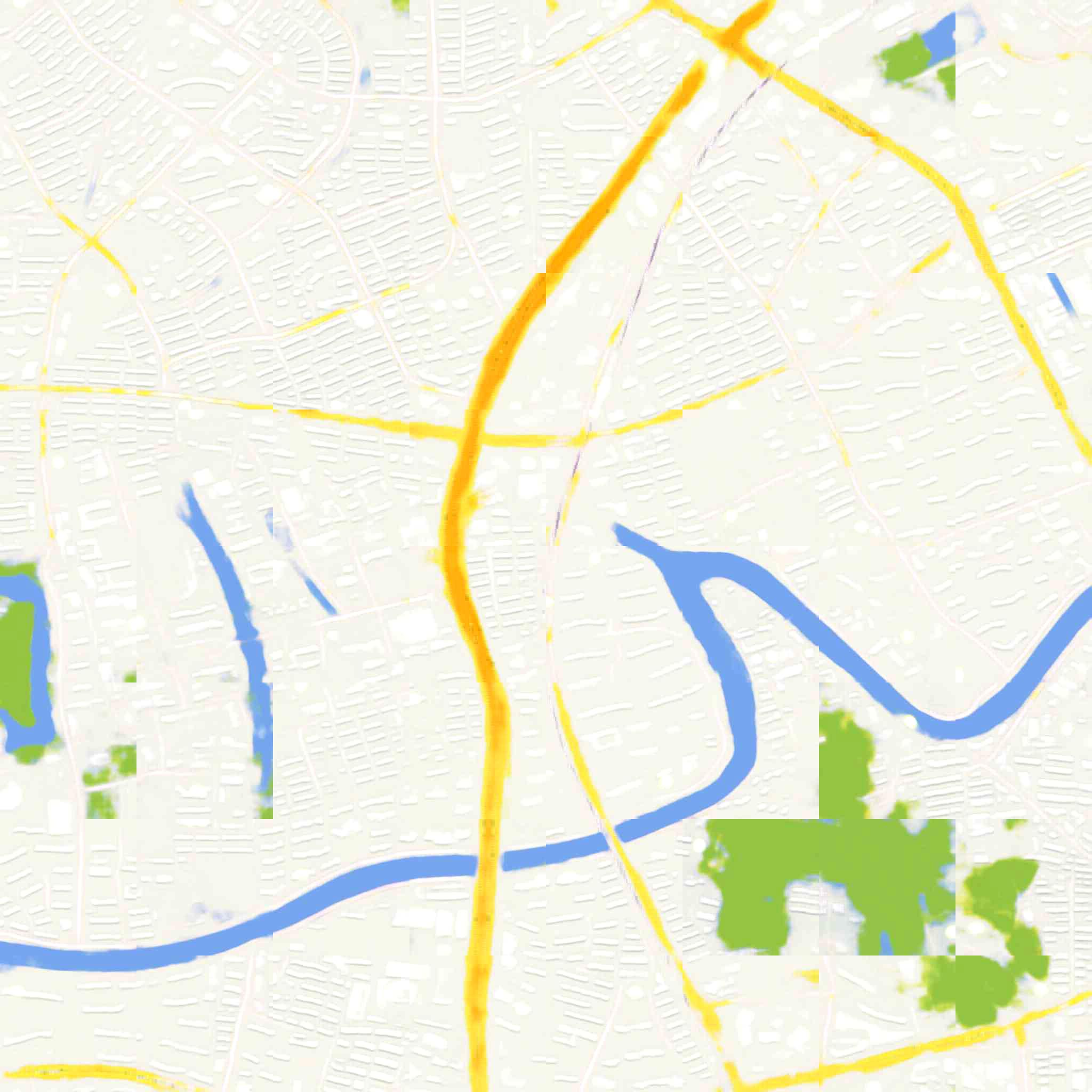}\hspace{-3mm}\quad
		\begin{sideways}\qquad \quad $2048\times 2048$\end{sideways}
		\\
(a) Aerial image & (b) Pix2Pix & (c) Pix2PixHD & (d) CycleGAN  & (e) GeoGAN & (f) SMAPGAN \\
		\begin{sideways}\qquad \quad 15-th\ level\end{sideways}
		\includegraphics[width=4.01cm, angle=0]{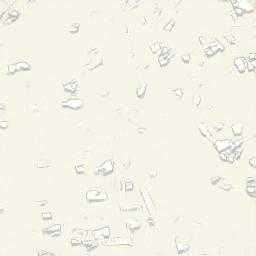}\hspace{-3mm}
		&\includegraphics[width=4.01cm, angle=0]{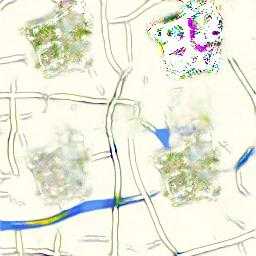}\hspace{-3mm}
		&\includegraphics[width=4.01cm, angle=0]{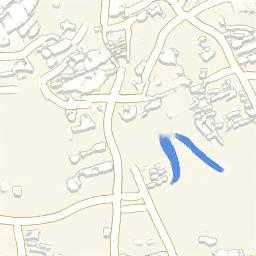}\hspace{-3mm}
		&\includegraphics[width=4.01cm, angle=0]{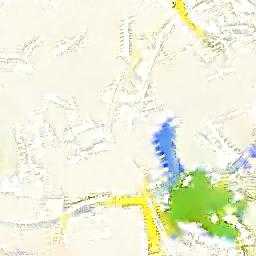}\hspace{-3mm}
		&\includegraphics[width=4.01cm, angle=0]{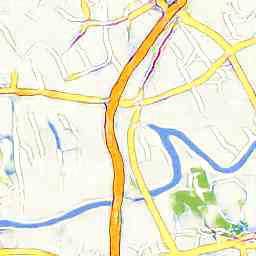}\hspace{-3mm}
		&\includegraphics[width=4.01cm, angle=0]{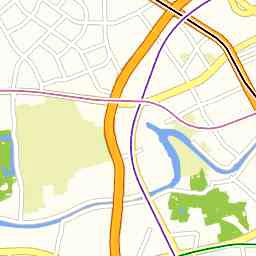}\hspace{-3mm}\quad
		\begin{sideways}\qquad \quad $256\times 256$\end{sideways}
		\\
		\begin{sideways}\qquad \quad 16-th\ level\end{sideways}
		\includegraphics[width=4.01cm, angle=0]{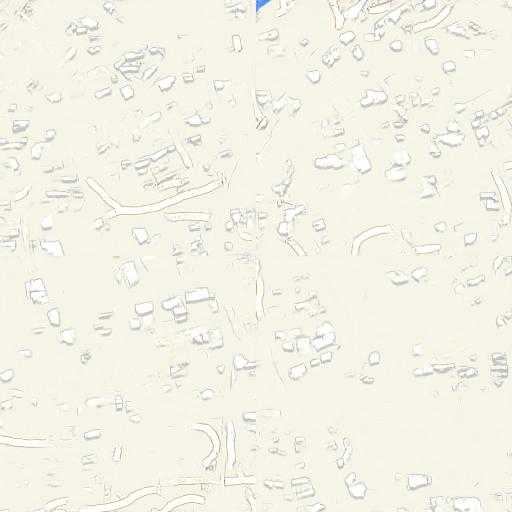}\hspace{-3mm}
		&\includegraphics[width=4.01cm, angle=0]{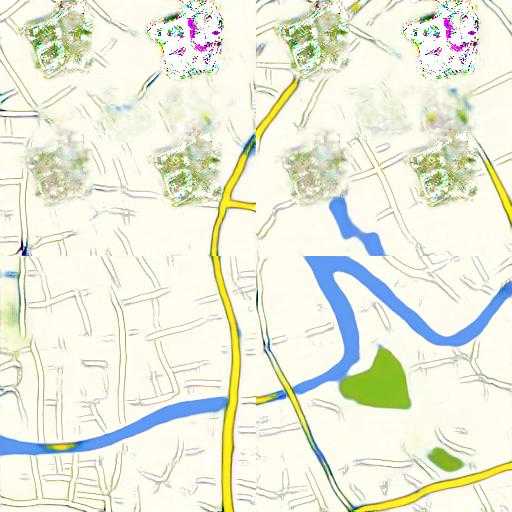}\hspace{-3mm}
		&\includegraphics[width=4.01cm, angle=0]{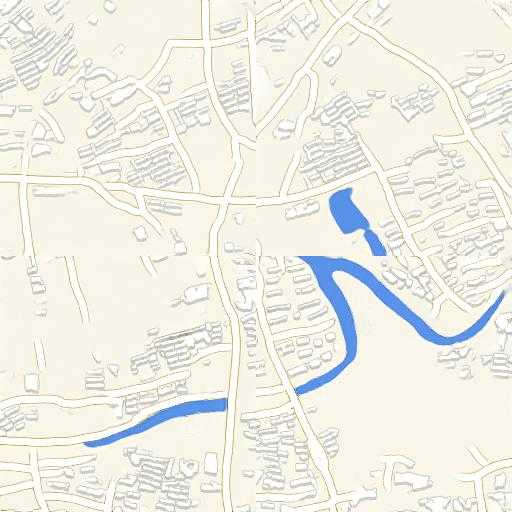}\hspace{-3mm}
		&\includegraphics[width=4.01cm, angle=0]{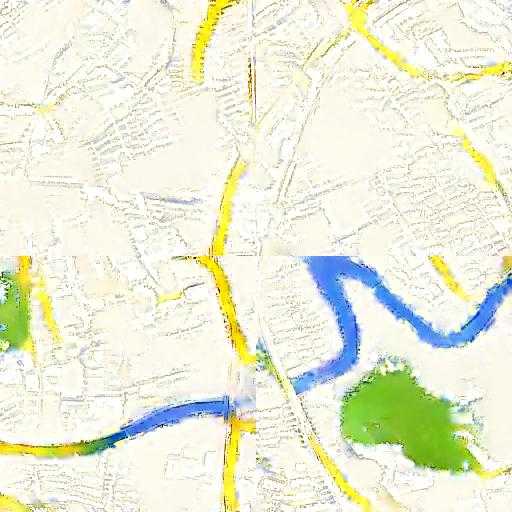}\hspace{-3mm}
		&\includegraphics[width=4.01cm, angle=0]{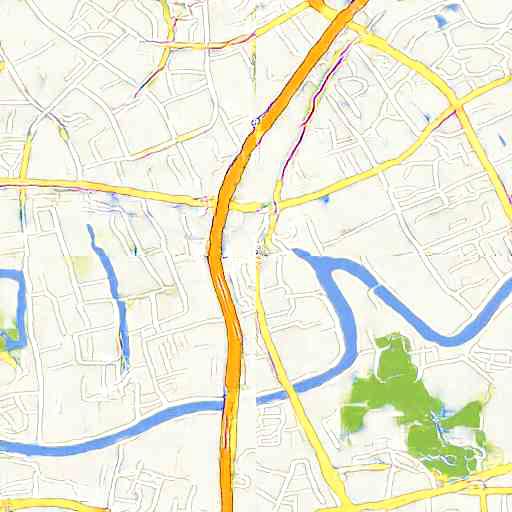}\hspace{-3mm}
		&\includegraphics[width=4.01cm, angle=0]{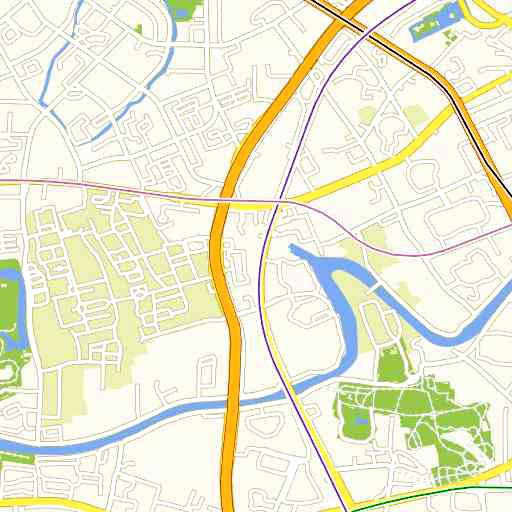}\hspace{-3mm}\quad
		\begin{sideways}\qquad \quad $512\times 512$\end{sideways}
		\\
		\begin{sideways}\qquad \quad 17-th\ level\end{sideways}
		\includegraphics[width=4.01cm, angle=0]{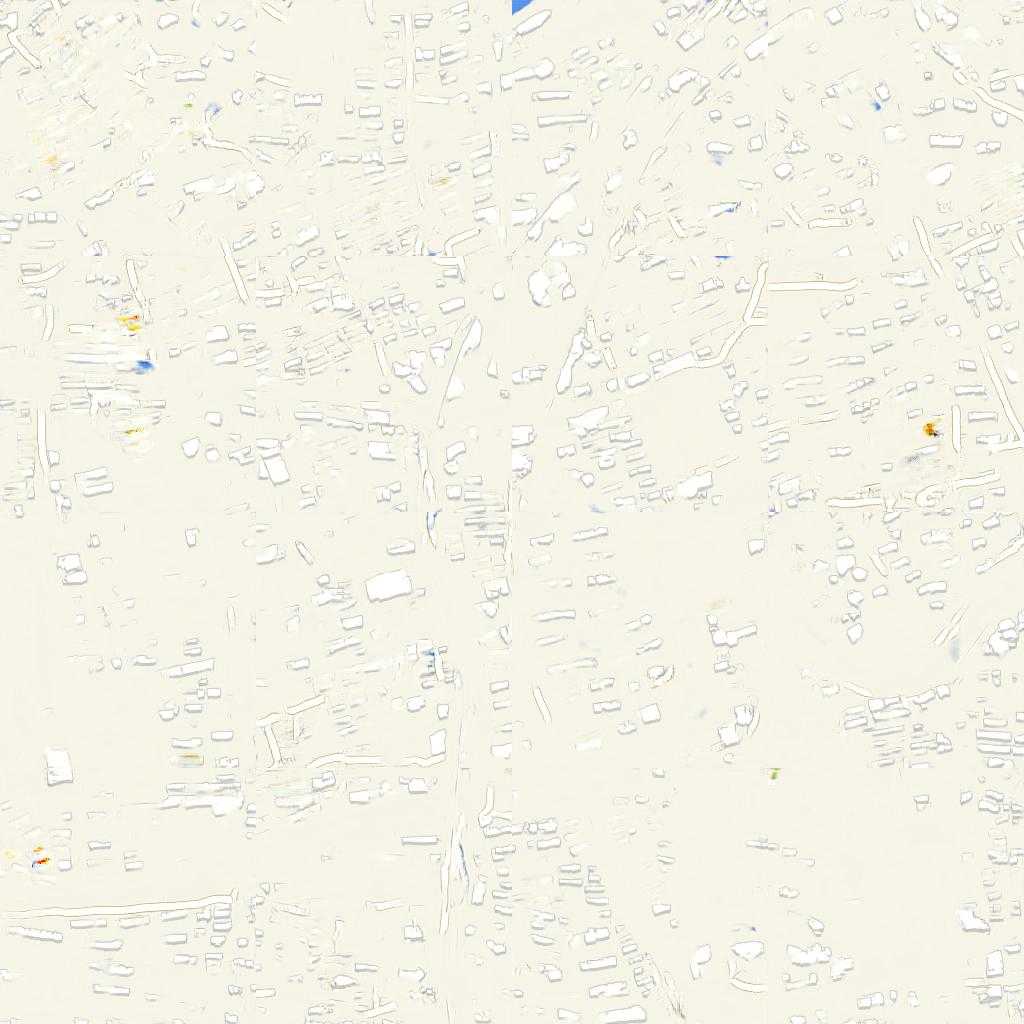}\hspace{-3mm}
		&\includegraphics[width=4.01cm, angle=0]{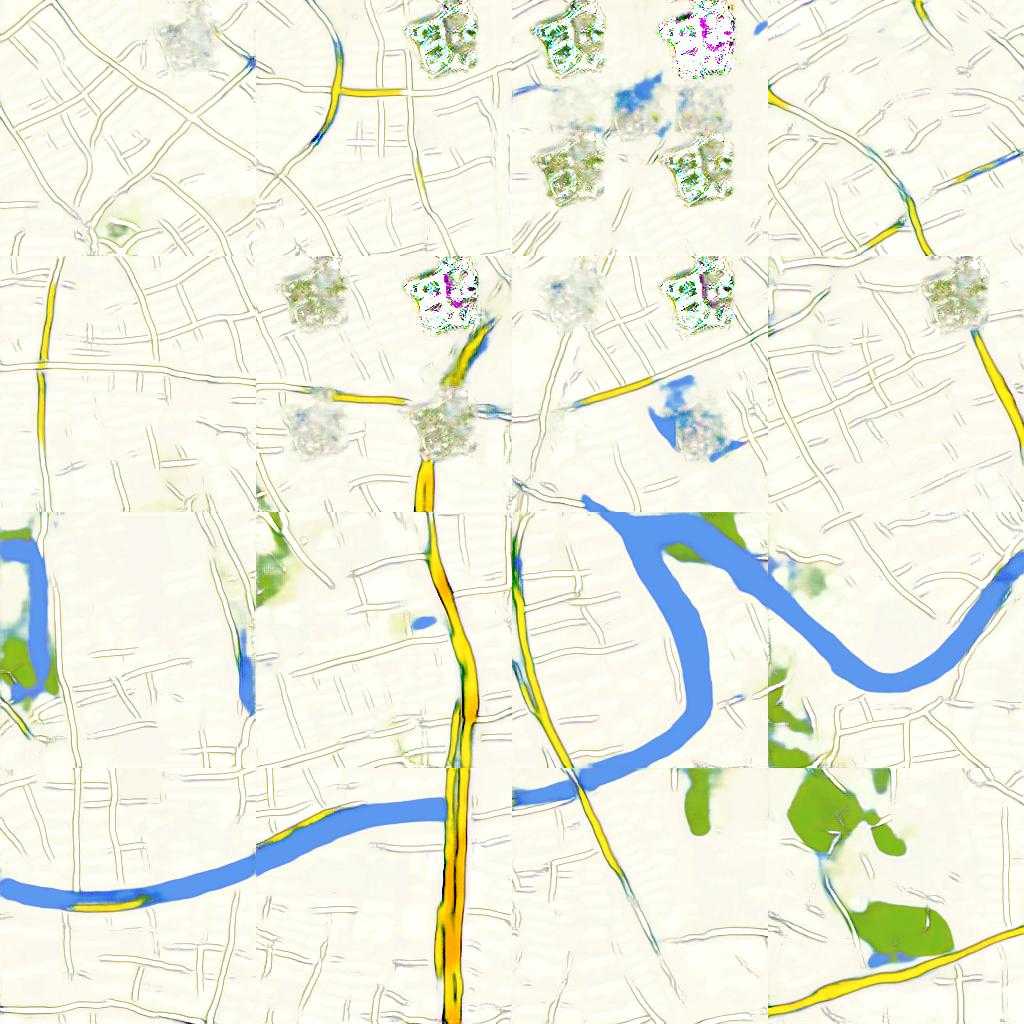}\hspace{-3mm}
		&\includegraphics[width=4.01cm, angle=0]{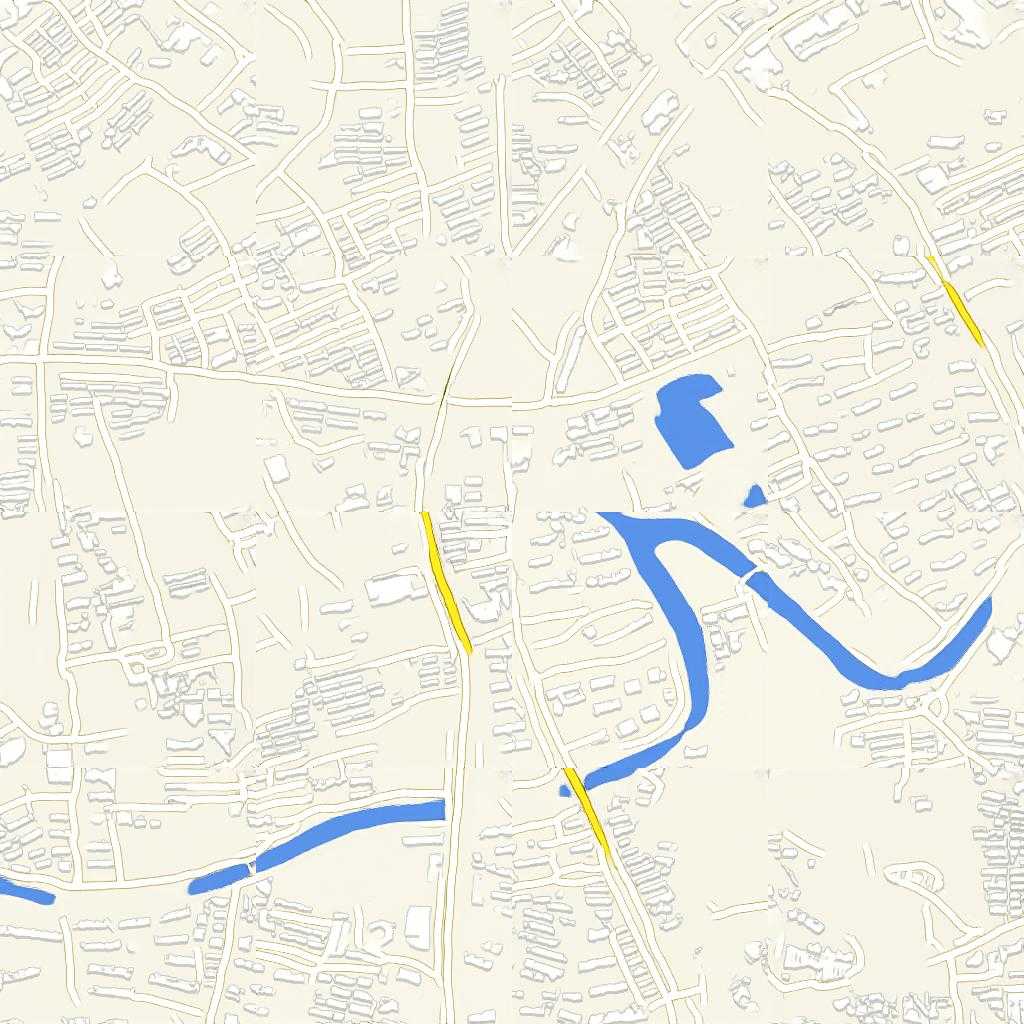}\hspace{-3mm}
		&\includegraphics[width=4.01cm, angle=0]{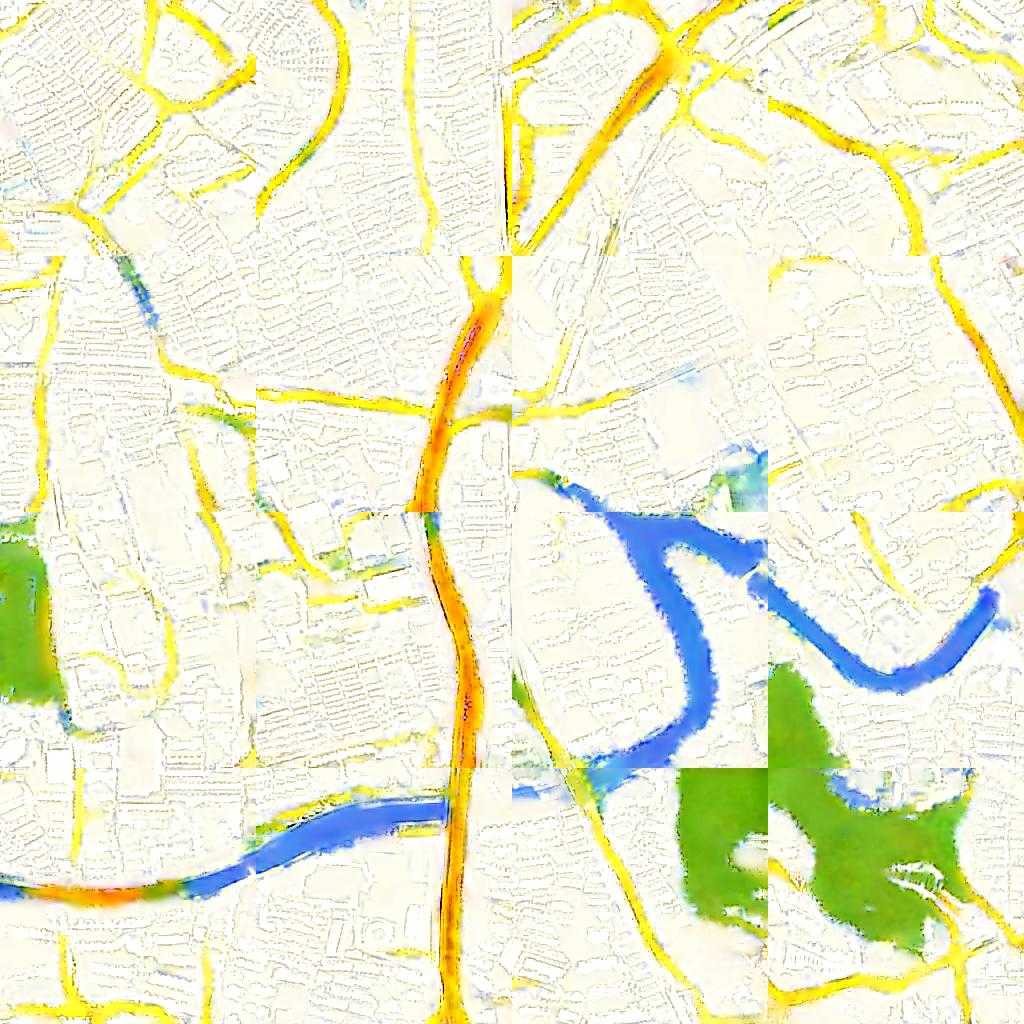}\hspace{-3mm}
		&\includegraphics[width=4.01cm, angle=0]{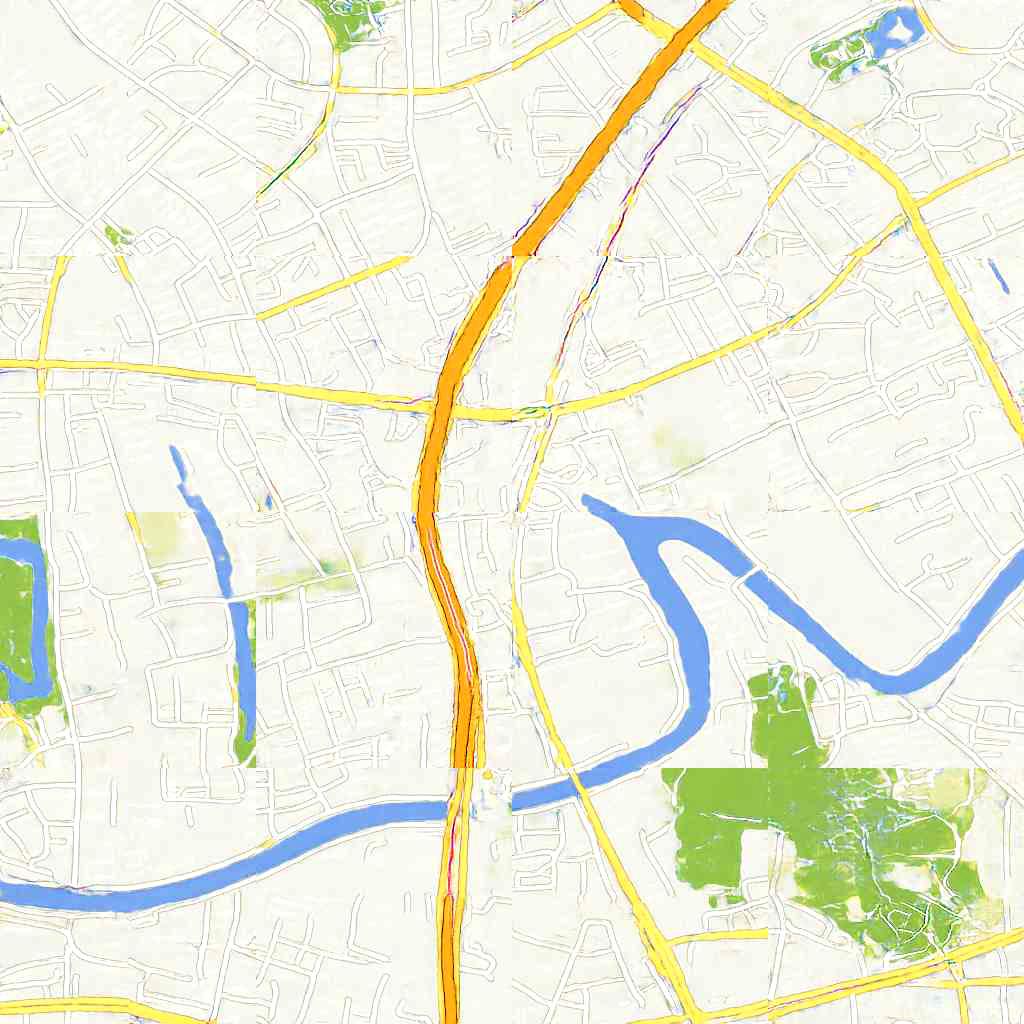}\hspace{-3mm}
		&\includegraphics[width=4.01cm, angle=0]{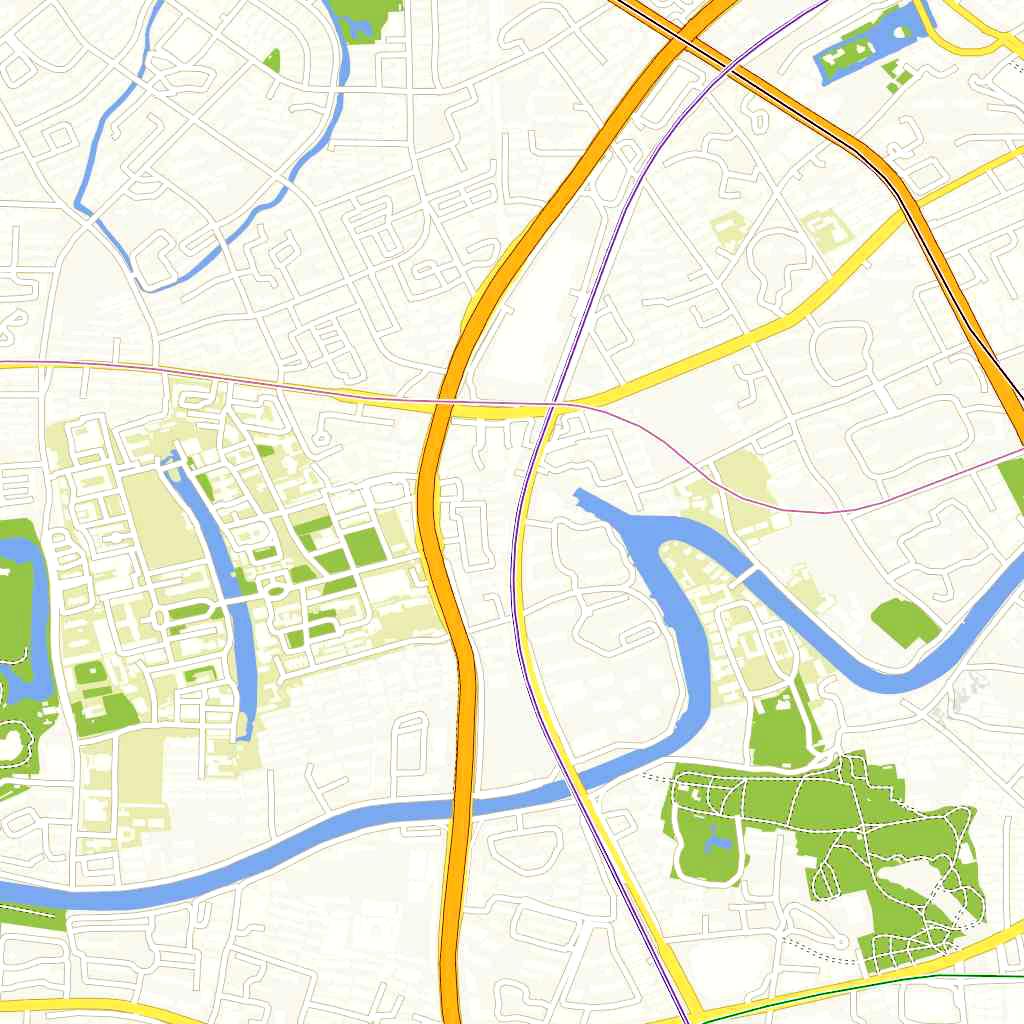}\hspace{-3mm}\quad
		\begin{sideways}\qquad \quad $1024\times 1024$\end{sideways}
		\\
		\begin{sideways}\qquad \quad 18-th\ level\end{sideways}
		\includegraphics[width=4.01cm, angle=0]{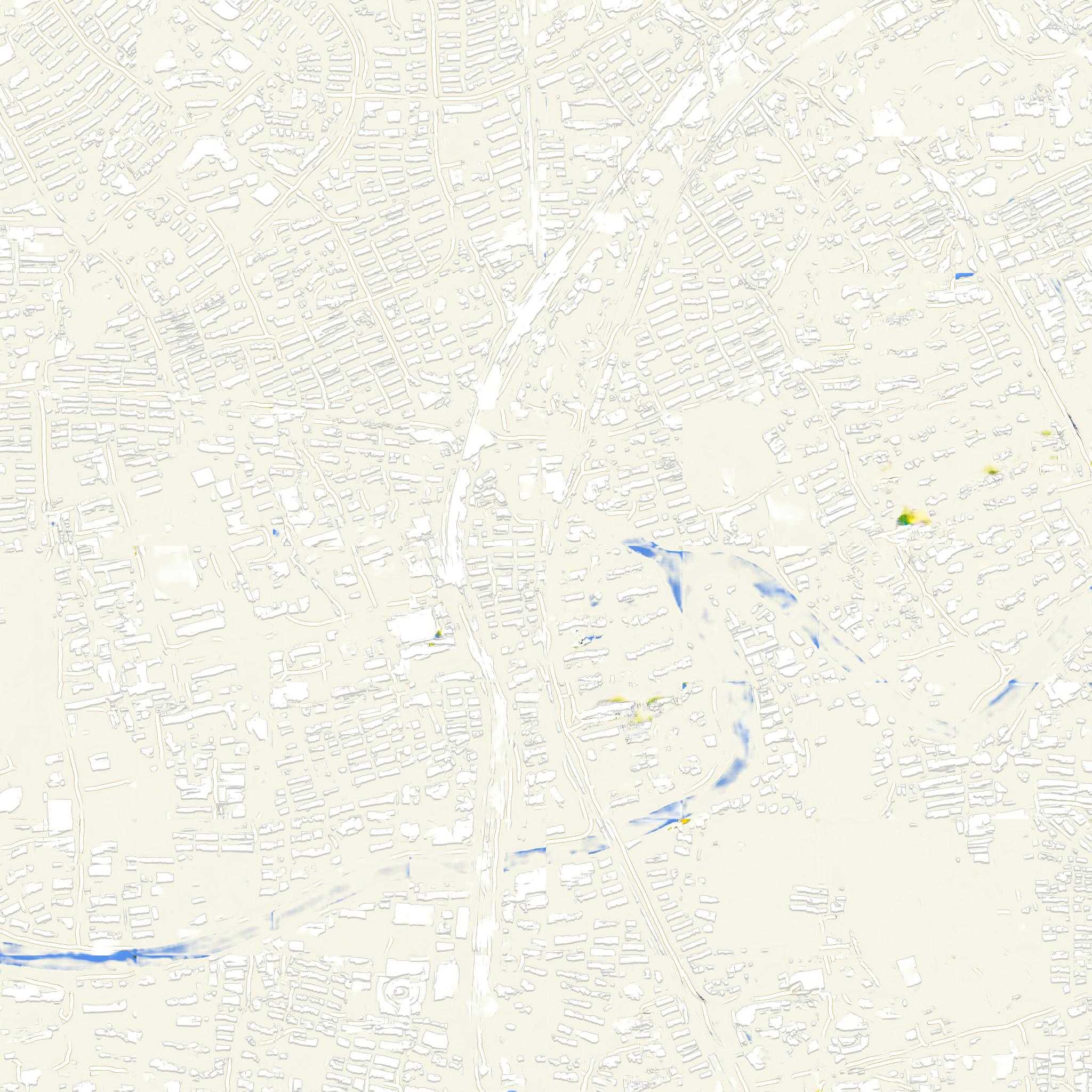}\hspace{-3mm}
		&\includegraphics[width=4.01cm, angle=0]{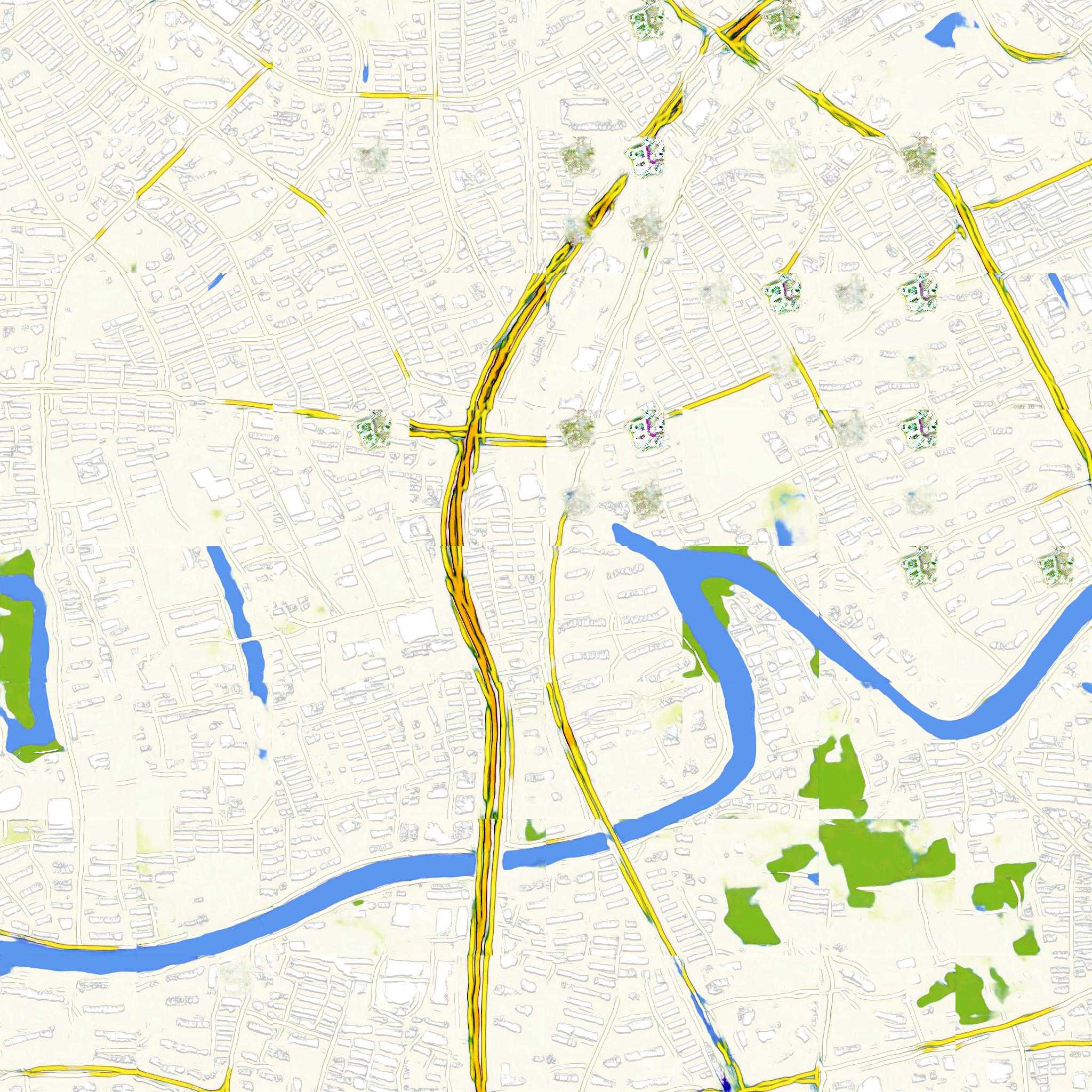}\hspace{-3mm}
		&\includegraphics[width=4.01cm, angle=0]{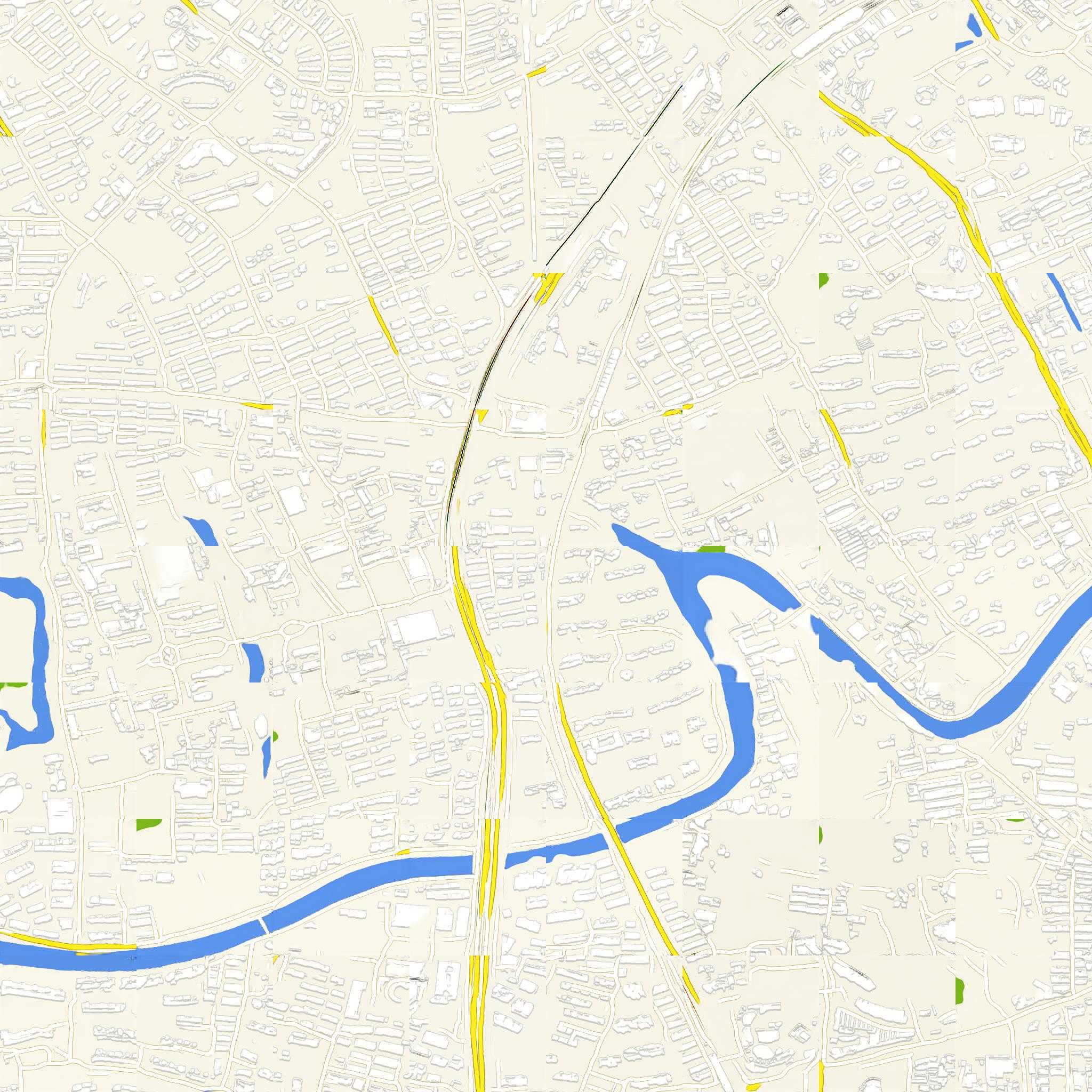}\hspace{-3mm}
		&\includegraphics[width=4.01cm, angle=0]{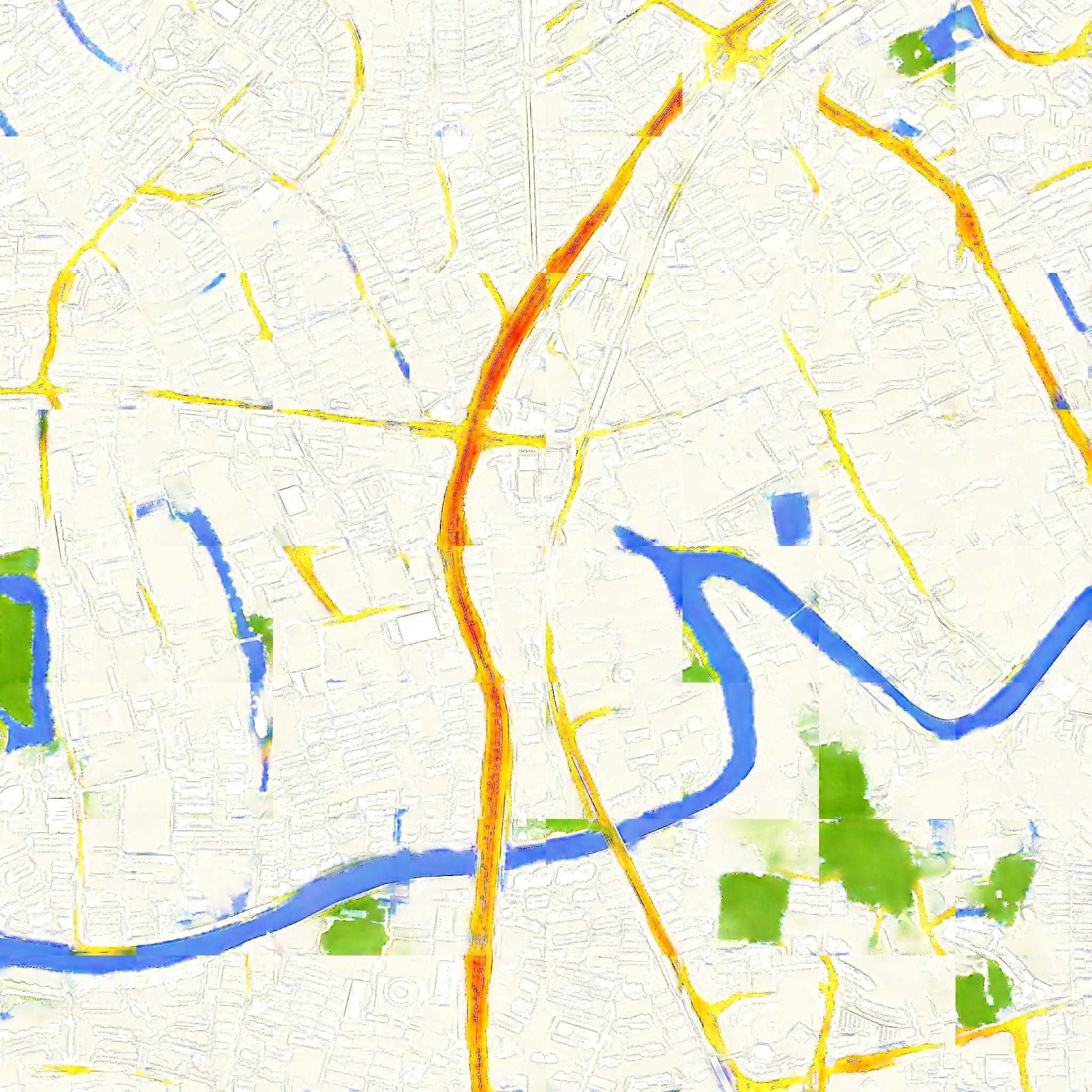}\hspace{-3mm}
		&\includegraphics[width=4.01cm, angle=0]{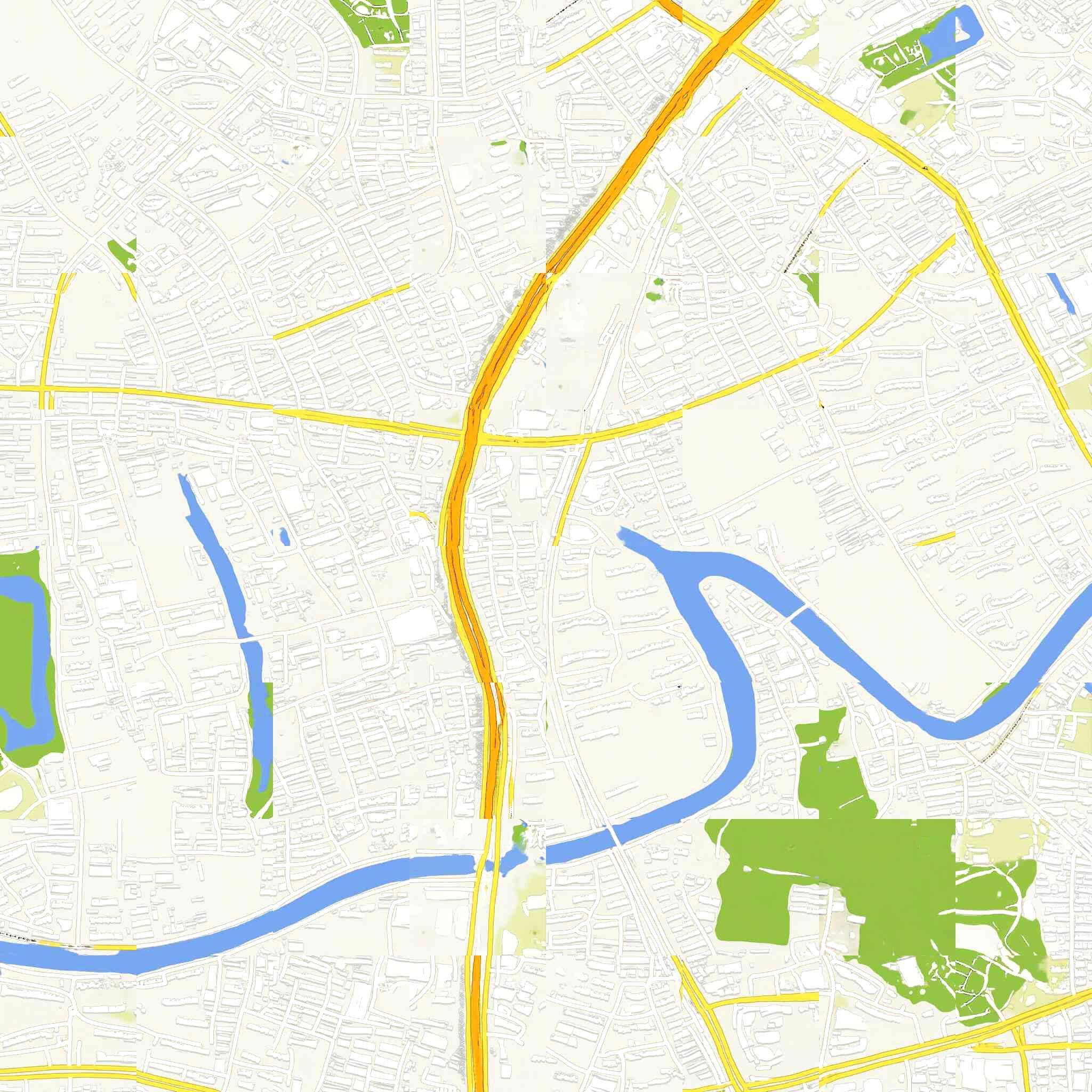}\hspace{-3mm}
		&\includegraphics[width=4.01cm, angle=0]{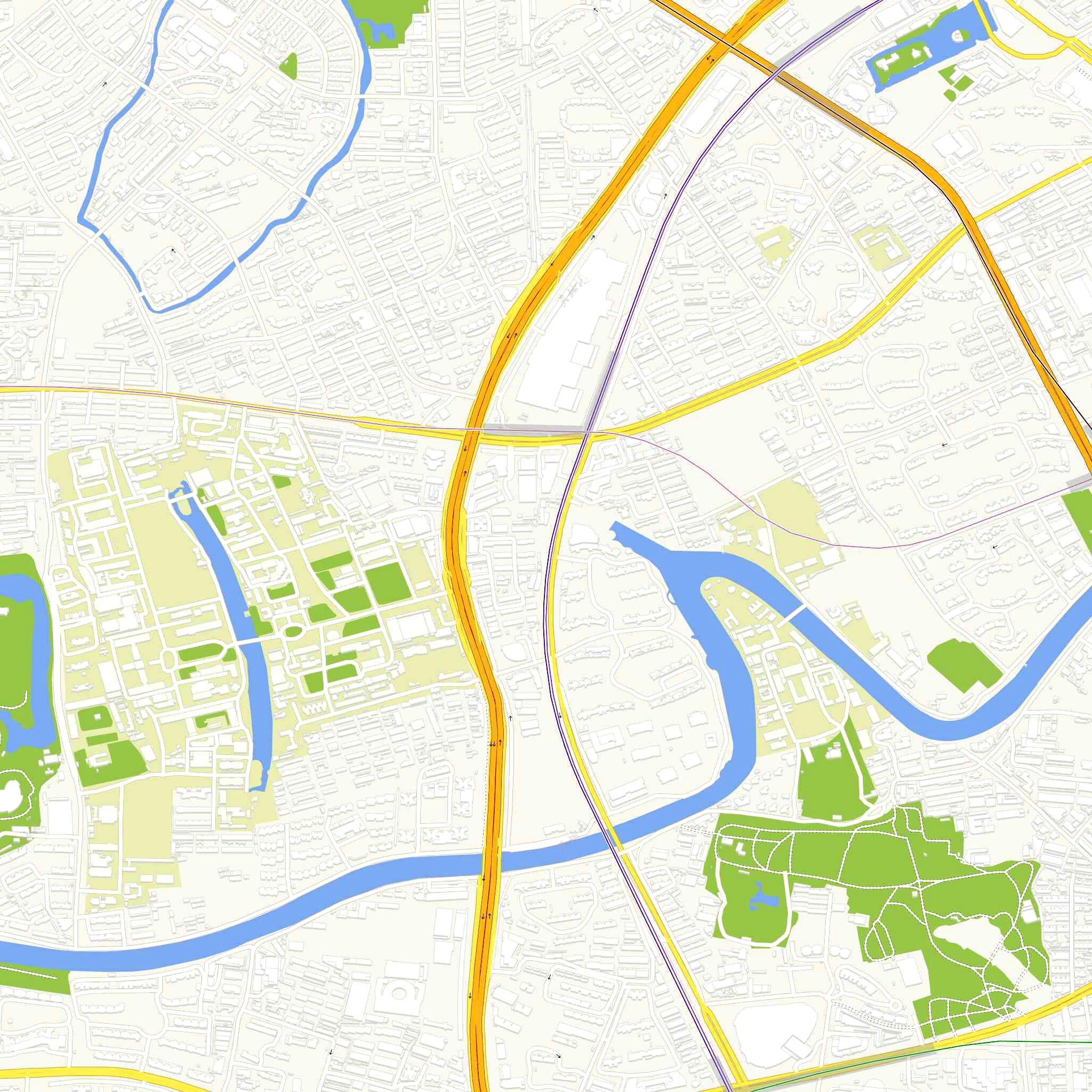}\hspace{-3mm}\quad
		\begin{sideways}\qquad \quad $2048\times 2048$\end{sideways}
		\\
(g) SPADE & (h) SelectionGAN  & (i) TSIT & (j) LPTN  & (k) Ours & (l) GT\\
	\end{tabular}}

\else
\includegraphics{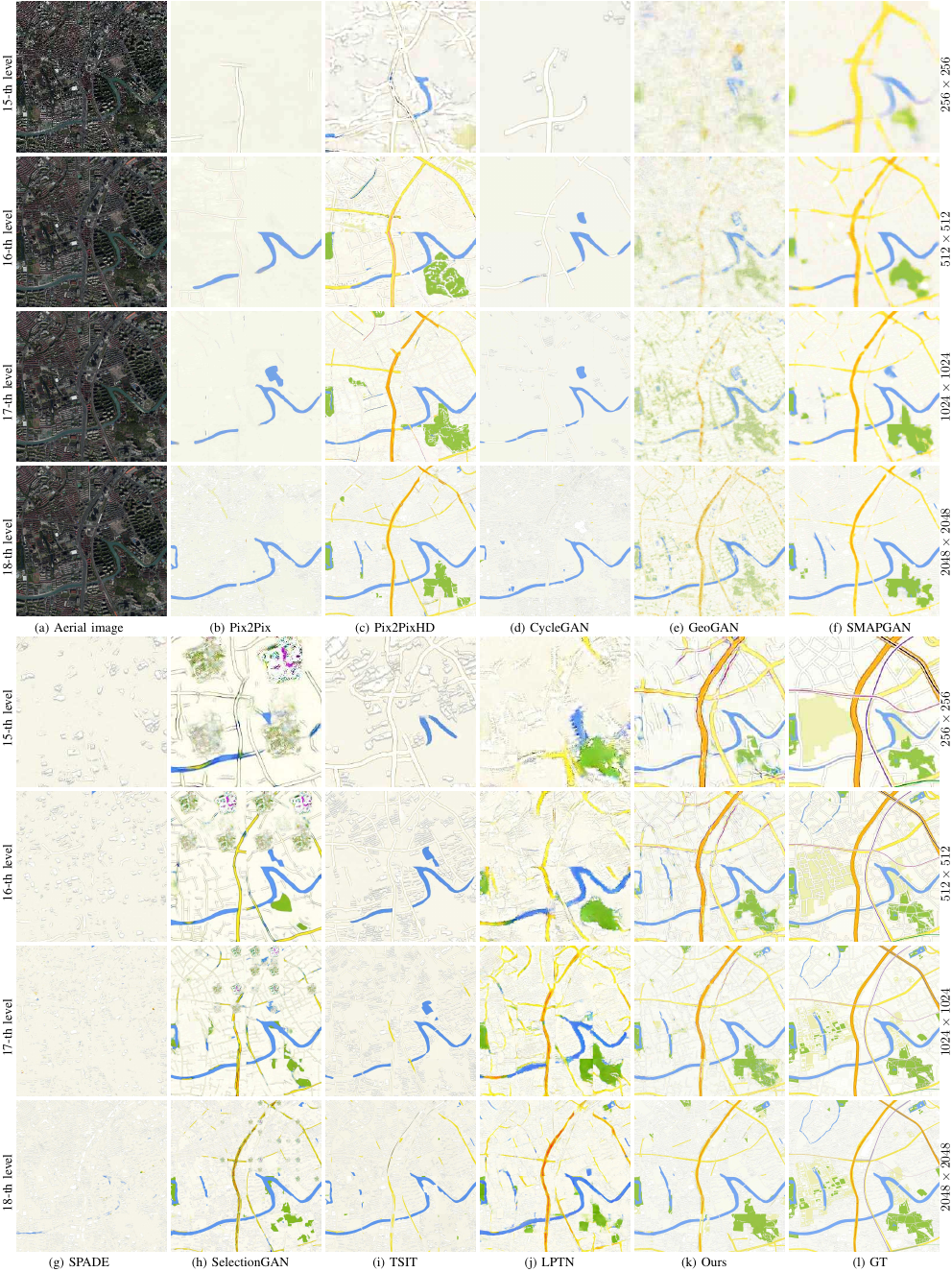}
\fi	
	\end{center}
	\vspace{-0mm}
	\caption{
	Visualized results of MLM-SH generated by our method and counterparts.
	Our method shows satisfactory accuracy and content consistency especially for rivers and roads in all zoom levels. 
	}
	\vspace{+2mm}
	\label{fig:visual_sh}
\end{figure*}

\begin{figure*}[tb!]
	\begin{center}
	
\iffalse
	\scalebox{0.66}{
	\begin{tabular}{cccccccc}
	\begin{sideways}\qquad \quad 15-th\ level\end{sideways}
				\includegraphics[width=4.01cm, angle=0]{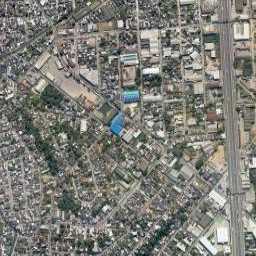}\hspace{-3mm}
				&\includegraphics[width=4.01cm, angle=0]{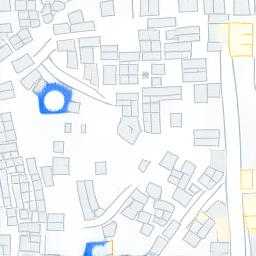}\hspace{-3mm}
				&\includegraphics[width=4.01cm, angle=0]{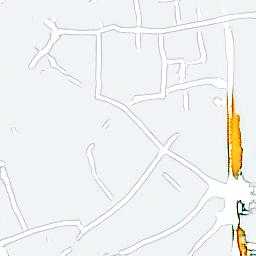}\hspace{-3mm}
				&\includegraphics[width=4.01cm, angle=0]{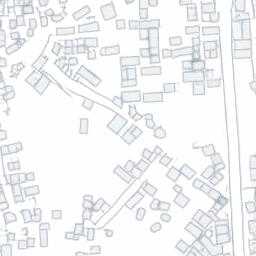}\hspace{-3mm}
				&\includegraphics[width=4.01cm, angle=0]{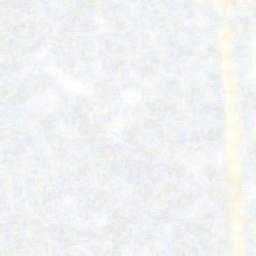}\hspace{-3mm}
				&\includegraphics[width=4.01cm, angle=0]{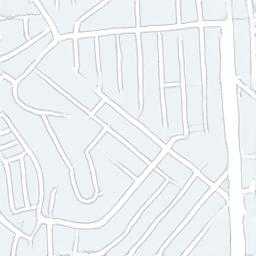}\hspace{-3mm}\quad
				\begin{sideways}\qquad \quad $256\times 256$\end{sideways}
				\\
				\begin{sideways}\qquad \quad 16-th\ level\end{sideways}
				\includegraphics[width=4.01cm, angle=0]{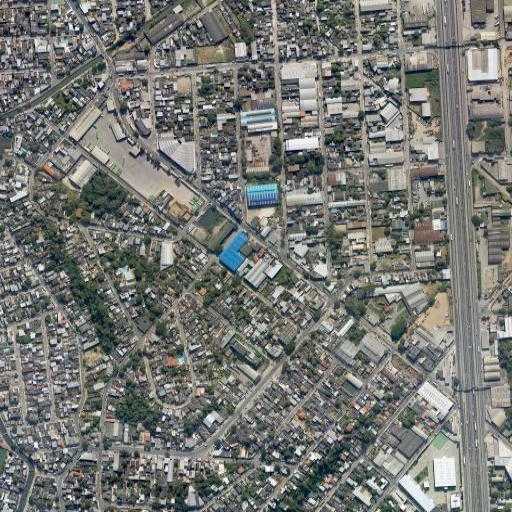}\hspace{-3mm}
				&\includegraphics[width=4.01cm, angle=0]{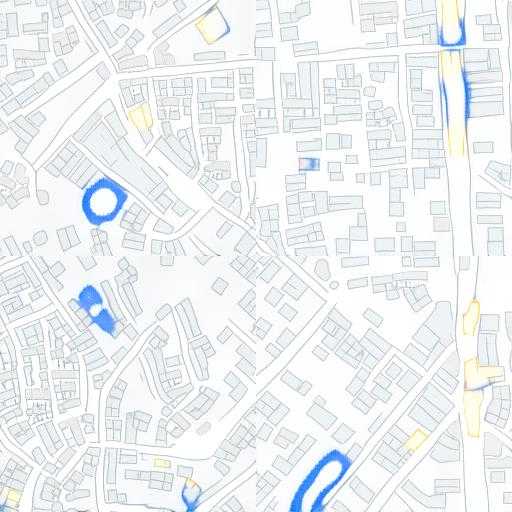}\hspace{-3mm}
				&\includegraphics[width=4.01cm, angle=0]{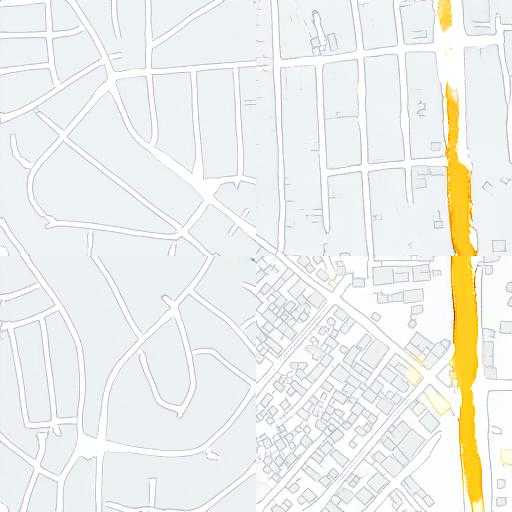}\hspace{-3mm}
				&\includegraphics[width=4.01cm, angle=0]{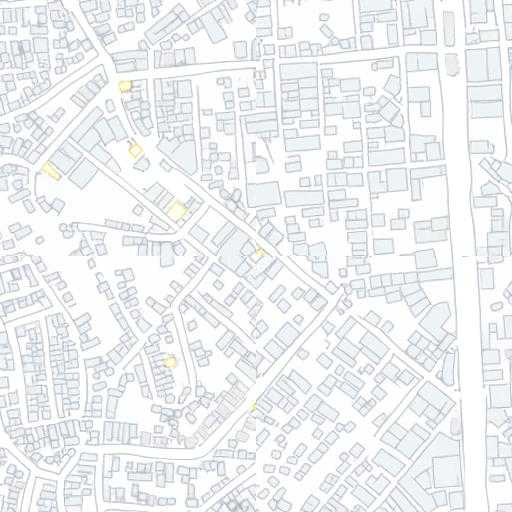}\hspace{-3mm}
				&\includegraphics[width=4.01cm, angle=0]{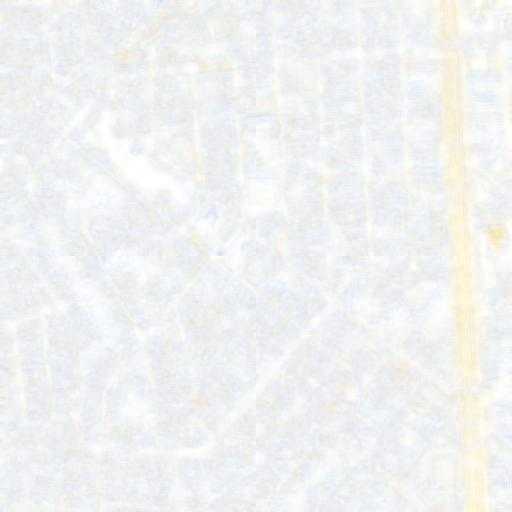}\hspace{-3mm}
				&\includegraphics[width=4.01cm, angle=0]{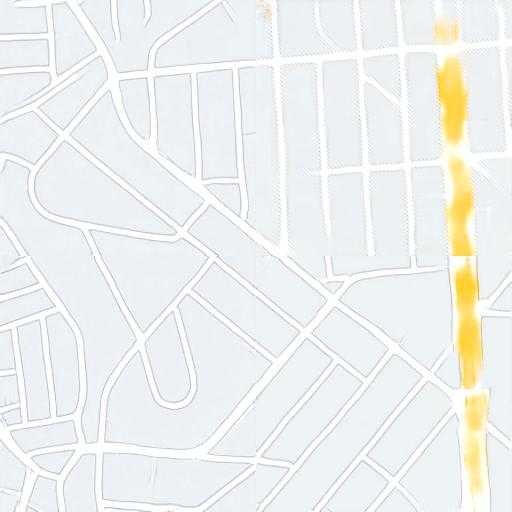}\hspace{-3mm}\quad
				\begin{sideways}\qquad \quad $512\times 512$\end{sideways}
				\\
				\begin{sideways}\qquad \quad 17-th\ level\end{sideways}
				\includegraphics[width=4.01cm, angle=0]{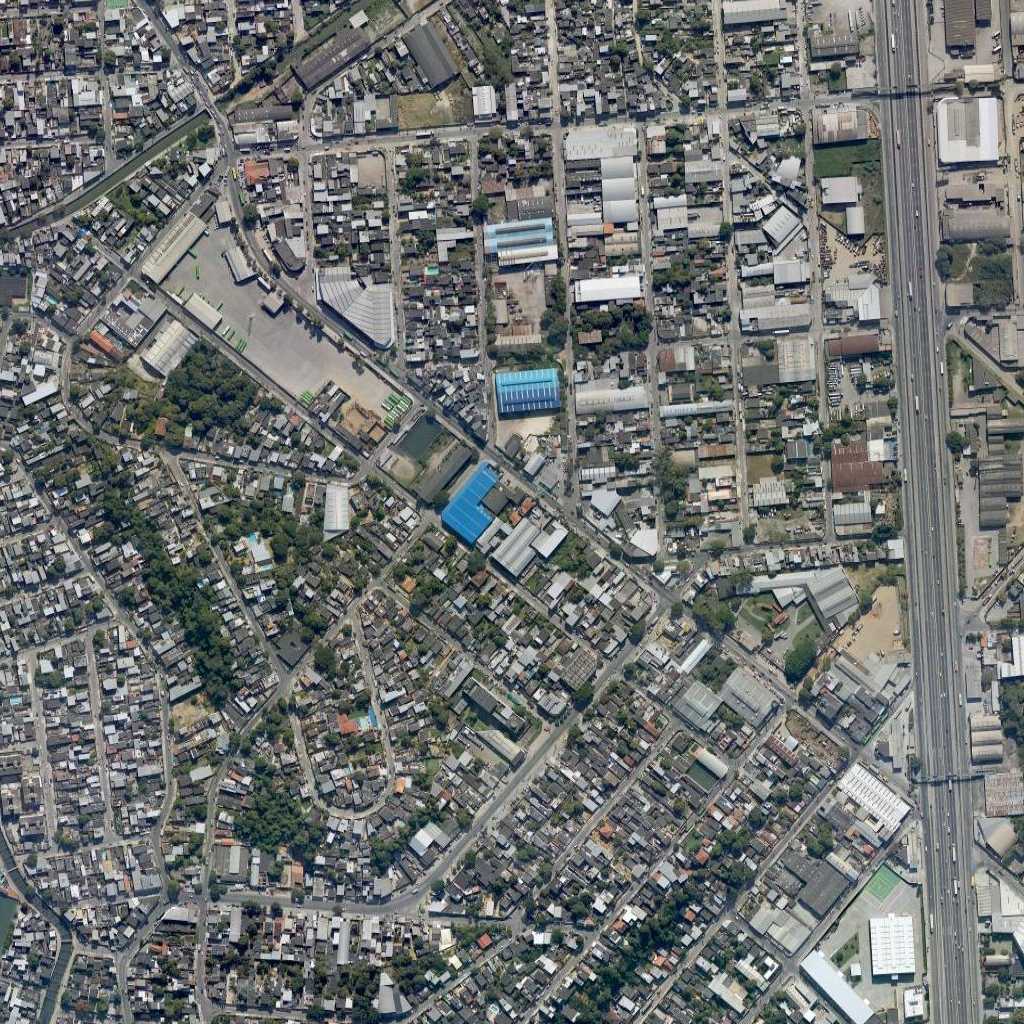}\hspace{-3mm}
				&\includegraphics[width=4.01cm, angle=0]{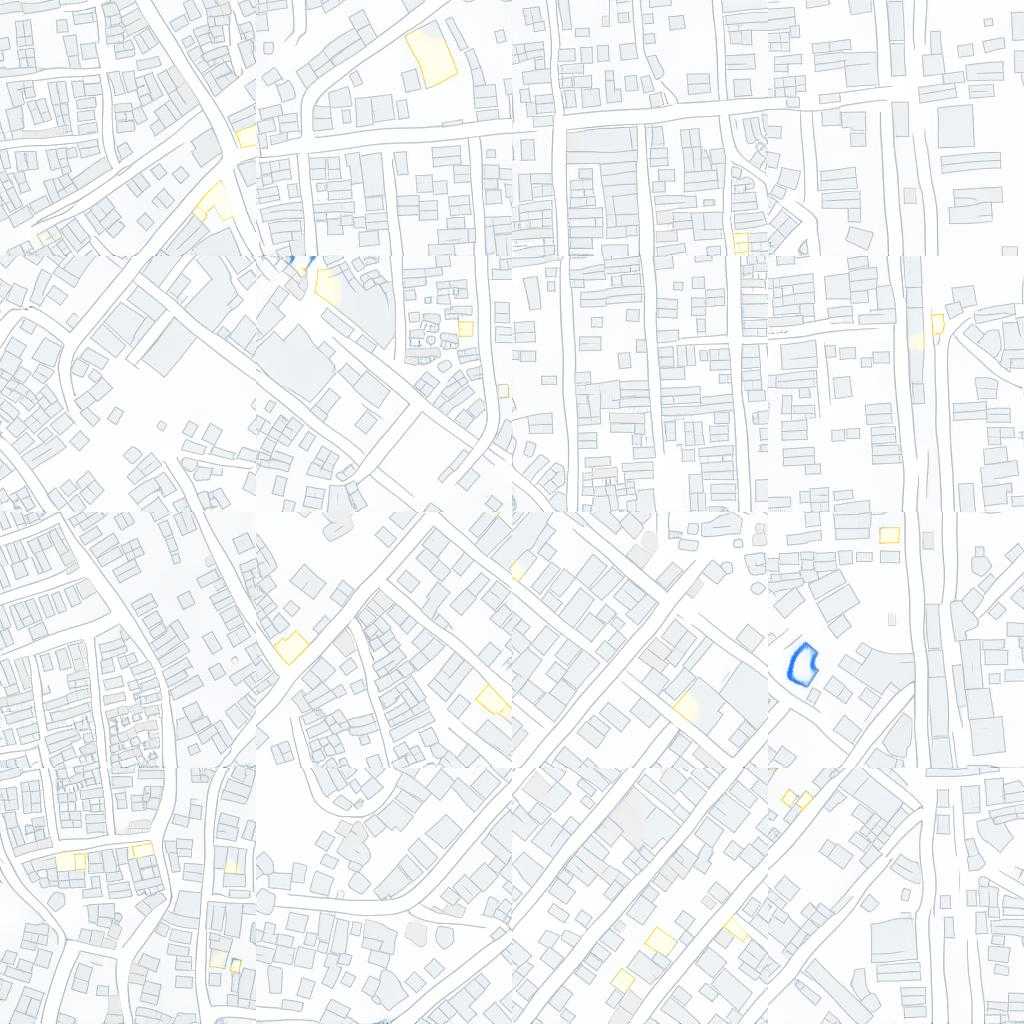}\hspace{-3mm}
				&\includegraphics[width=4.01cm, angle=0]{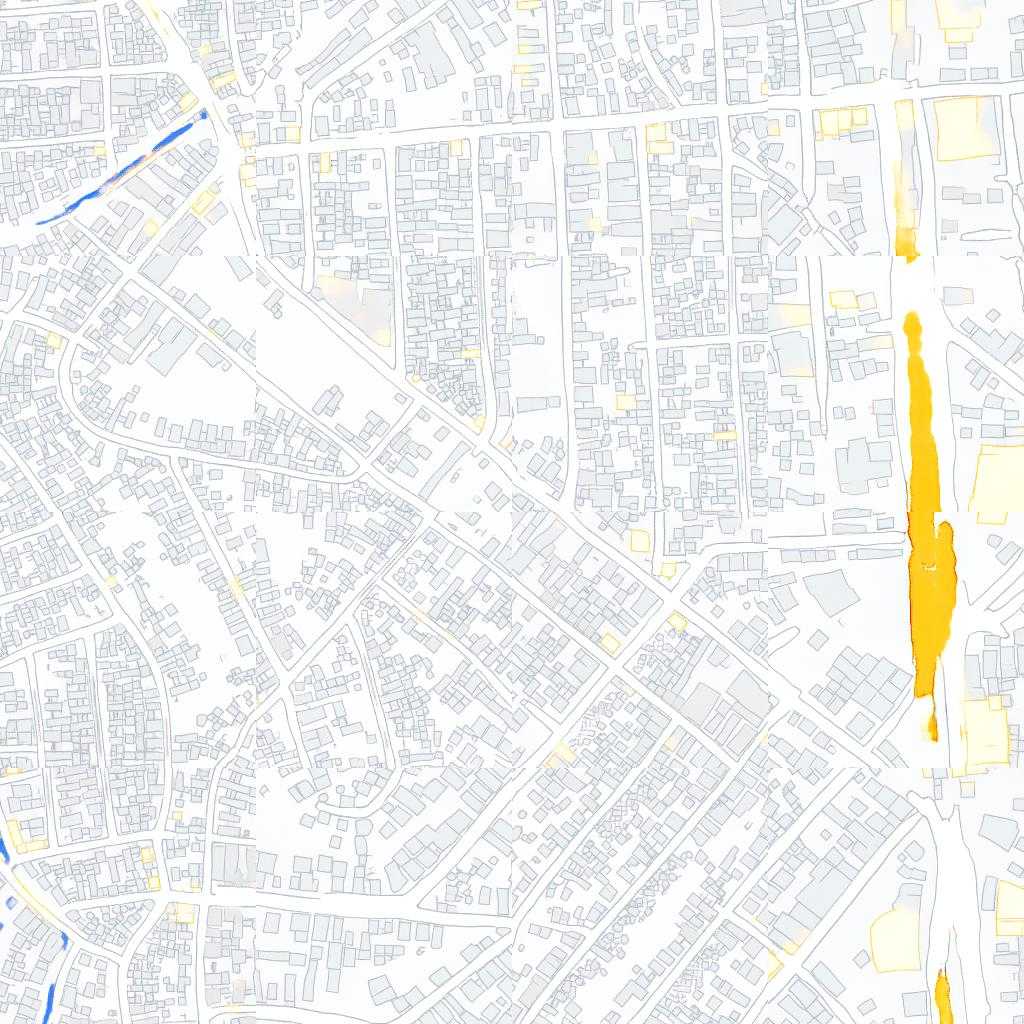}\hspace{-3mm}
				&\includegraphics[width=4.01cm, angle=0]{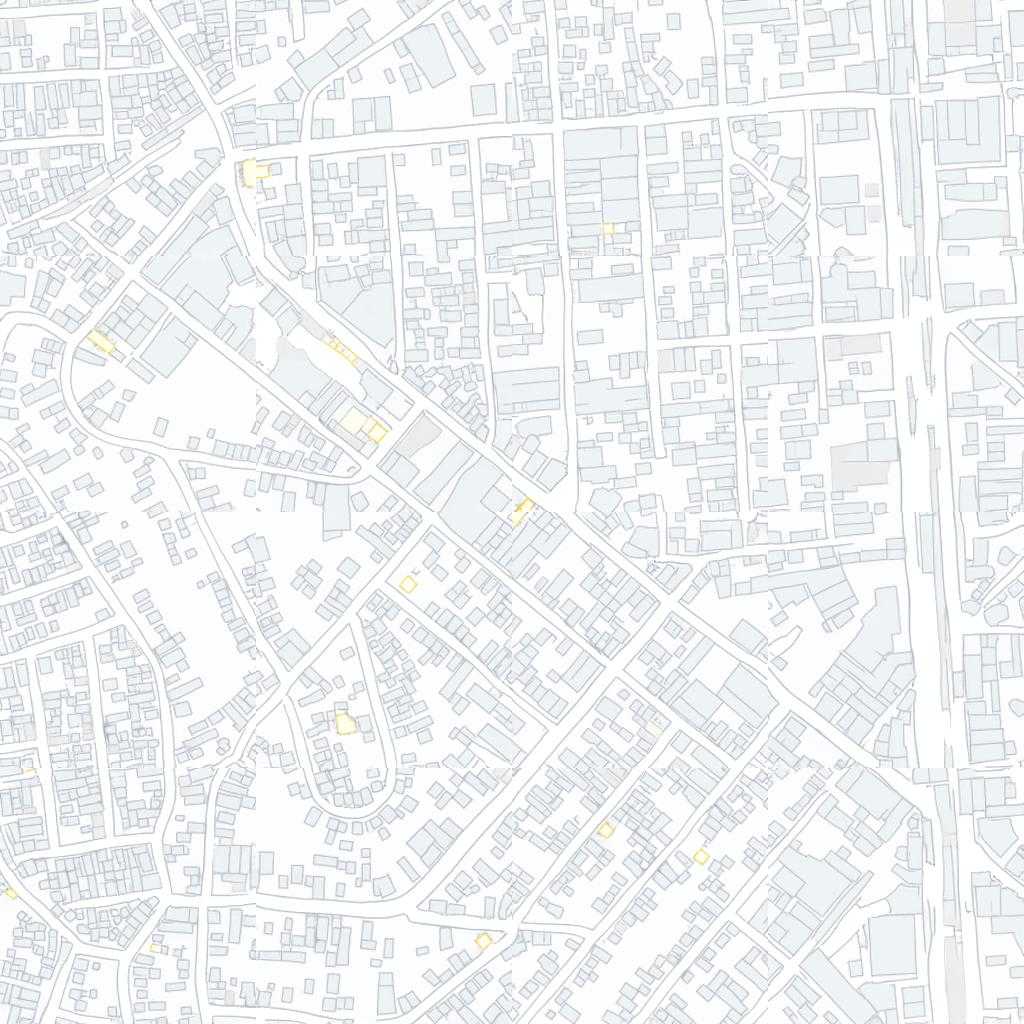}\hspace{-3mm}
				&\includegraphics[width=4.01cm, angle=0]{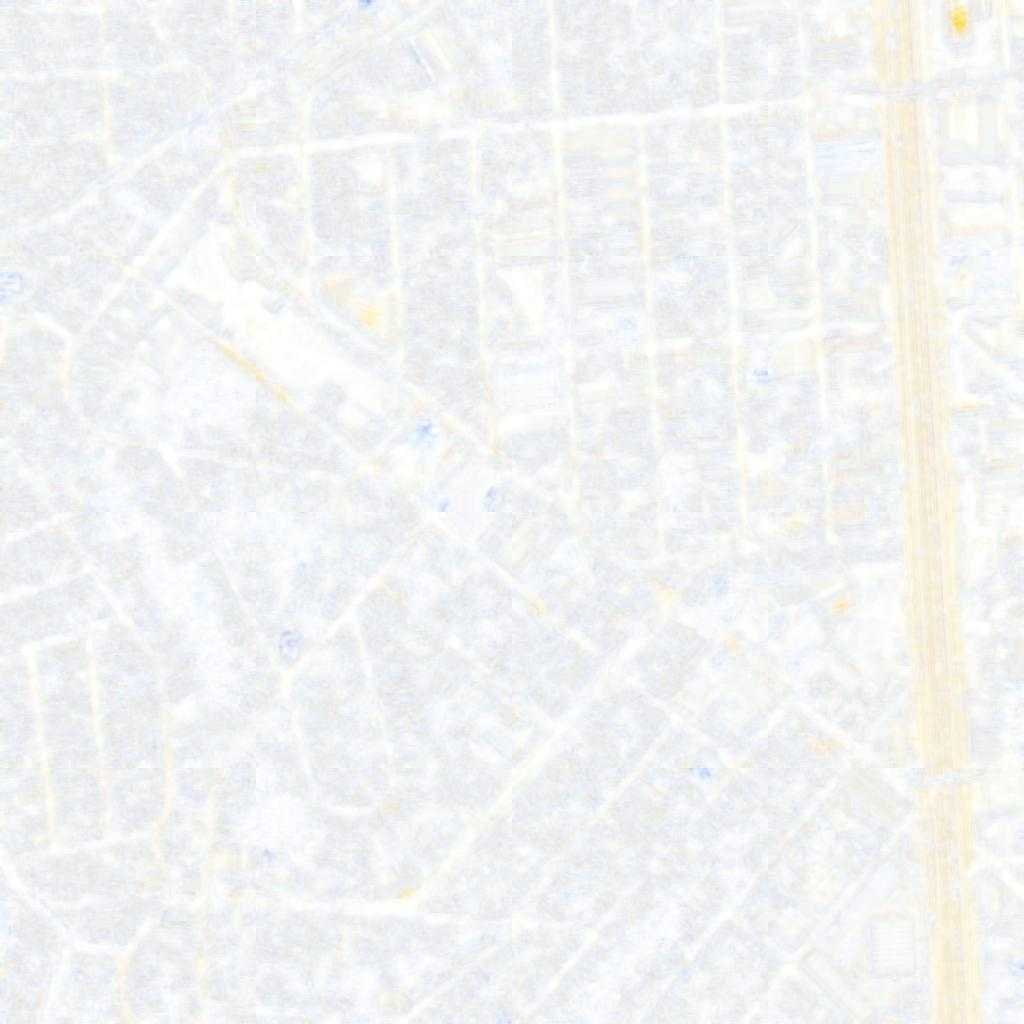}\hspace{-3mm}
				&\includegraphics[width=4.01cm, angle=0]{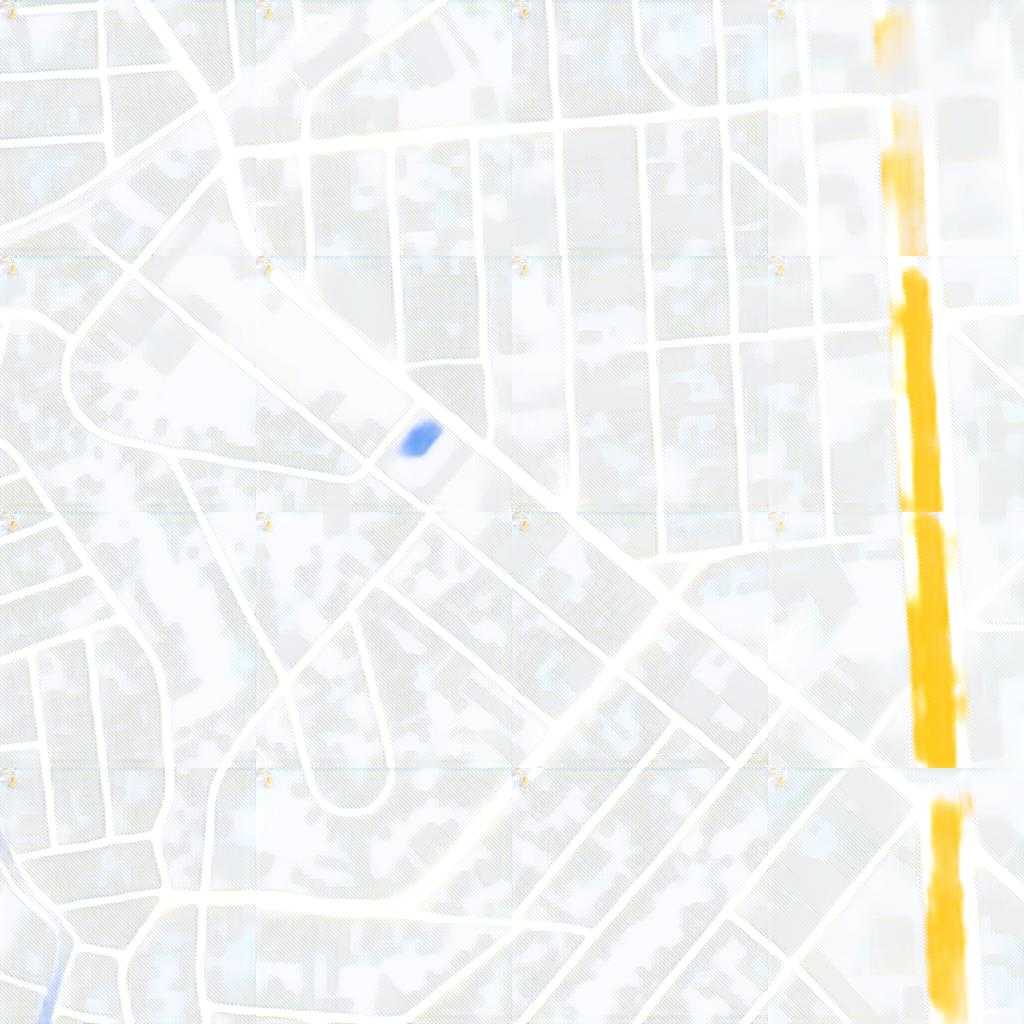}\hspace{-3mm}\quad
				\begin{sideways}\qquad \quad $1024\times 1024$\end{sideways}
				\\
				
				\begin{sideways}\qquad \quad 18-th\ level\end{sideways}
				\includegraphics[width=4.01cm, angle=0]{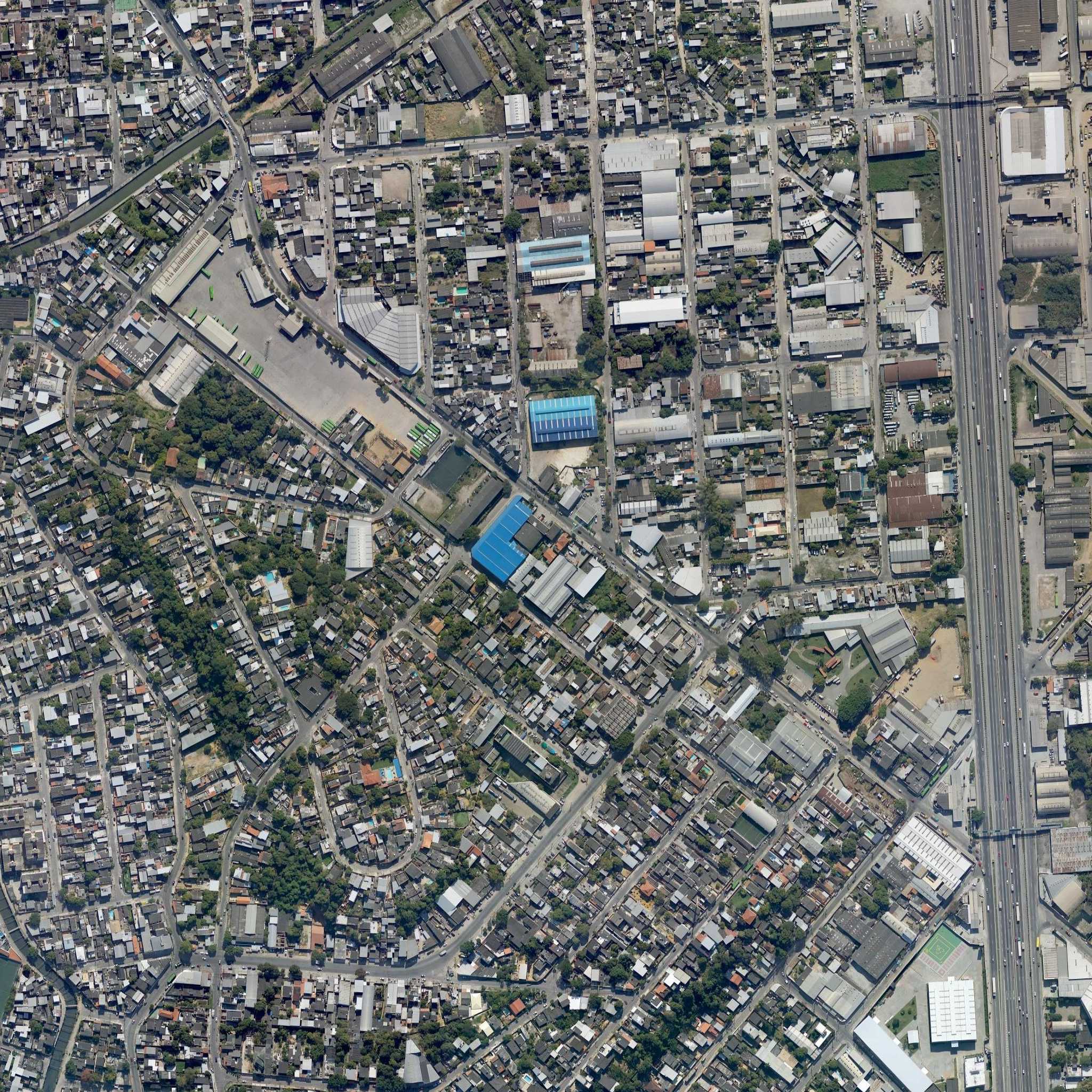}\hspace{-3mm}
				&\includegraphics[width=4.01cm, angle=0]{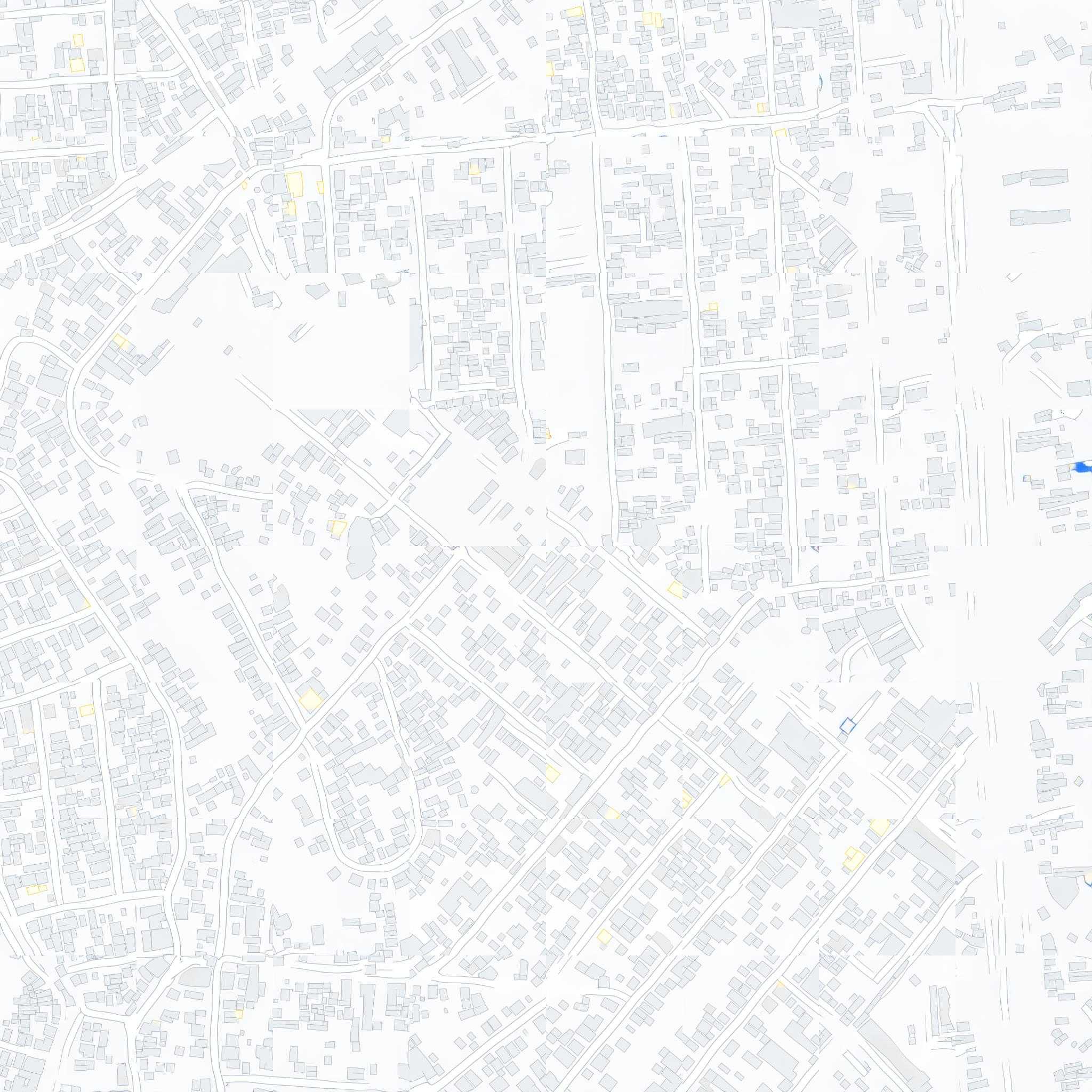}\hspace{-3mm}
				&\includegraphics[width=4.01cm, angle=0]{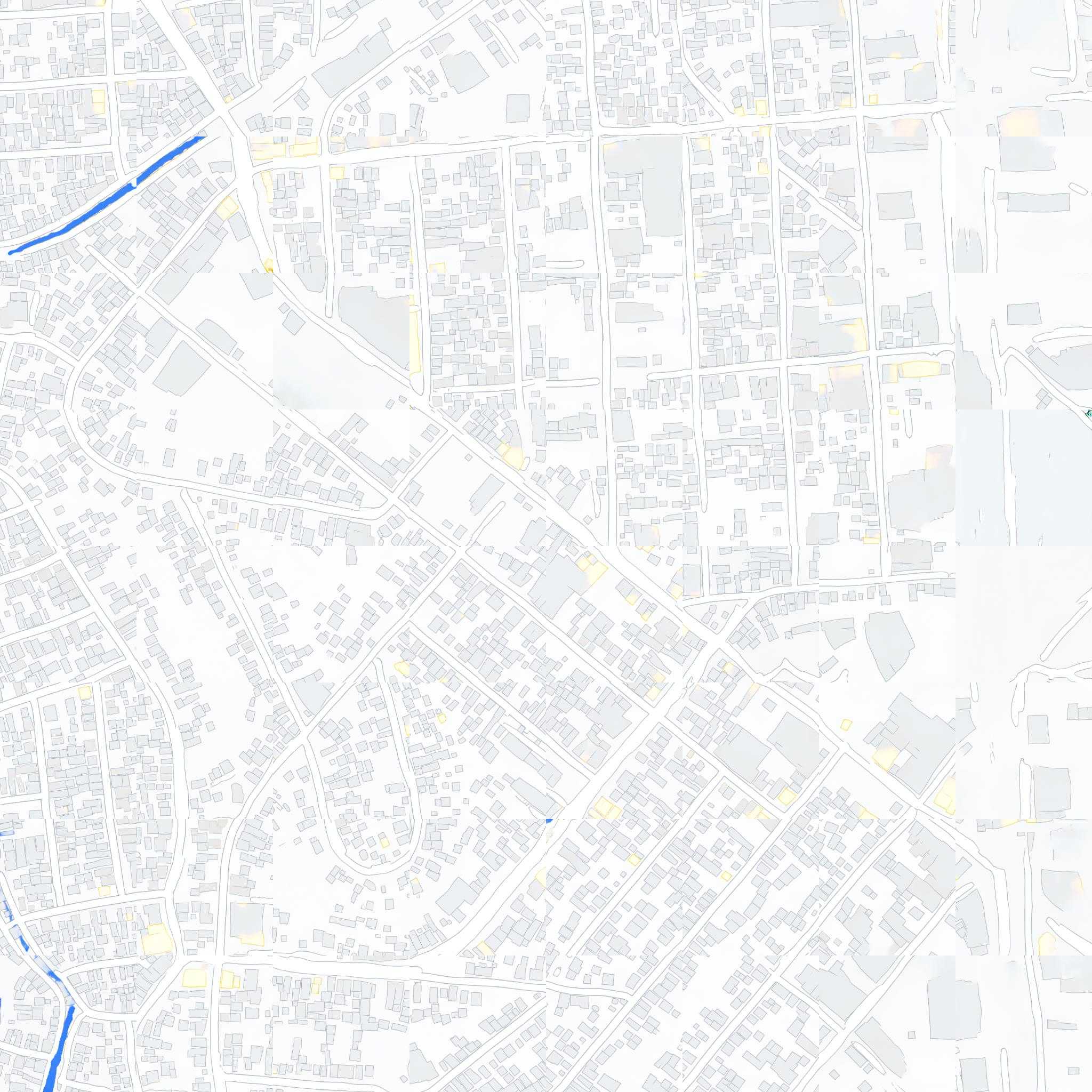}\hspace{-3mm}
				&\includegraphics[width=4.01cm, angle=0]{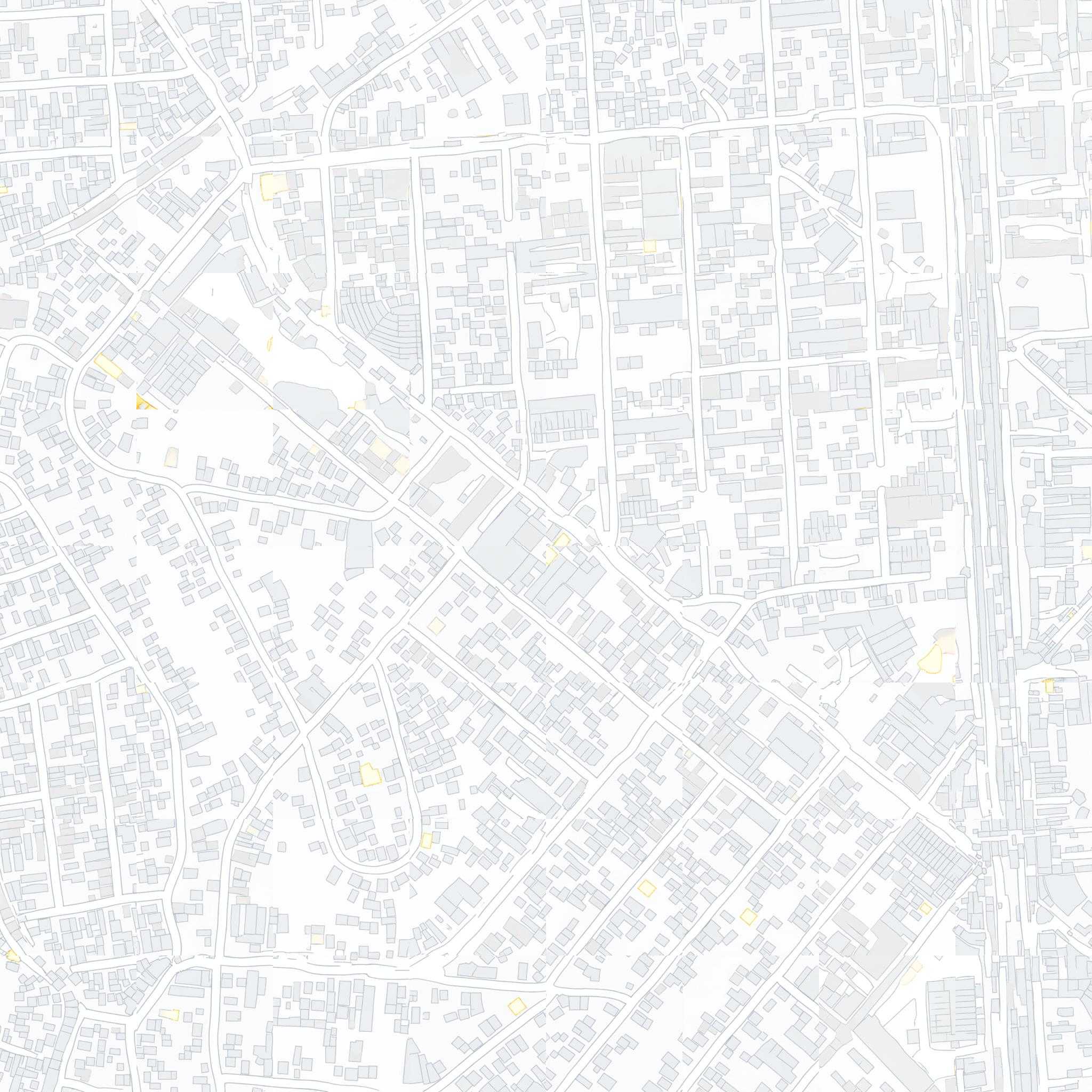}\hspace{-3mm}
				&\includegraphics[width=4.01cm, angle=0]{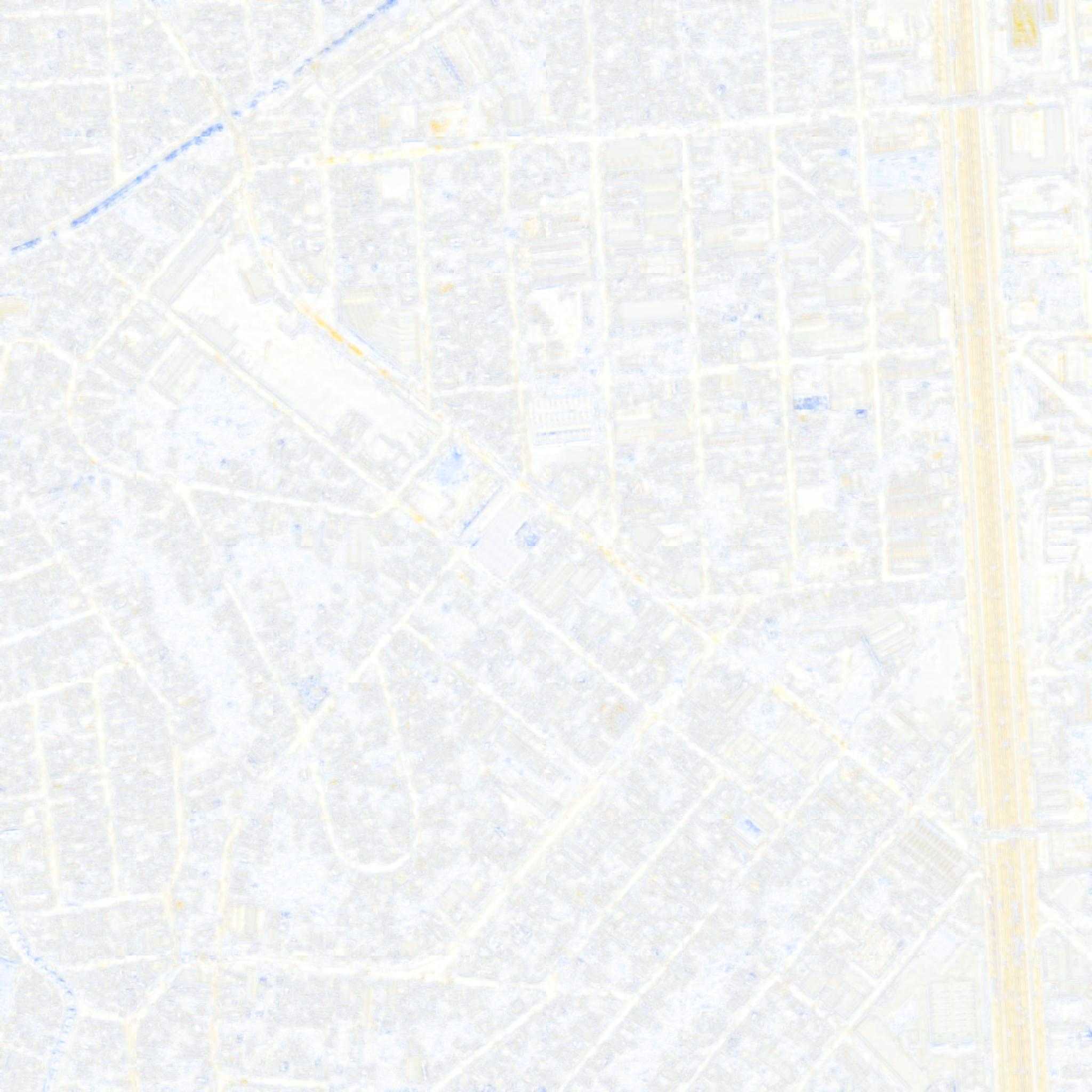}\hspace{-3mm}
				&\includegraphics[width=4.01cm, angle=0]{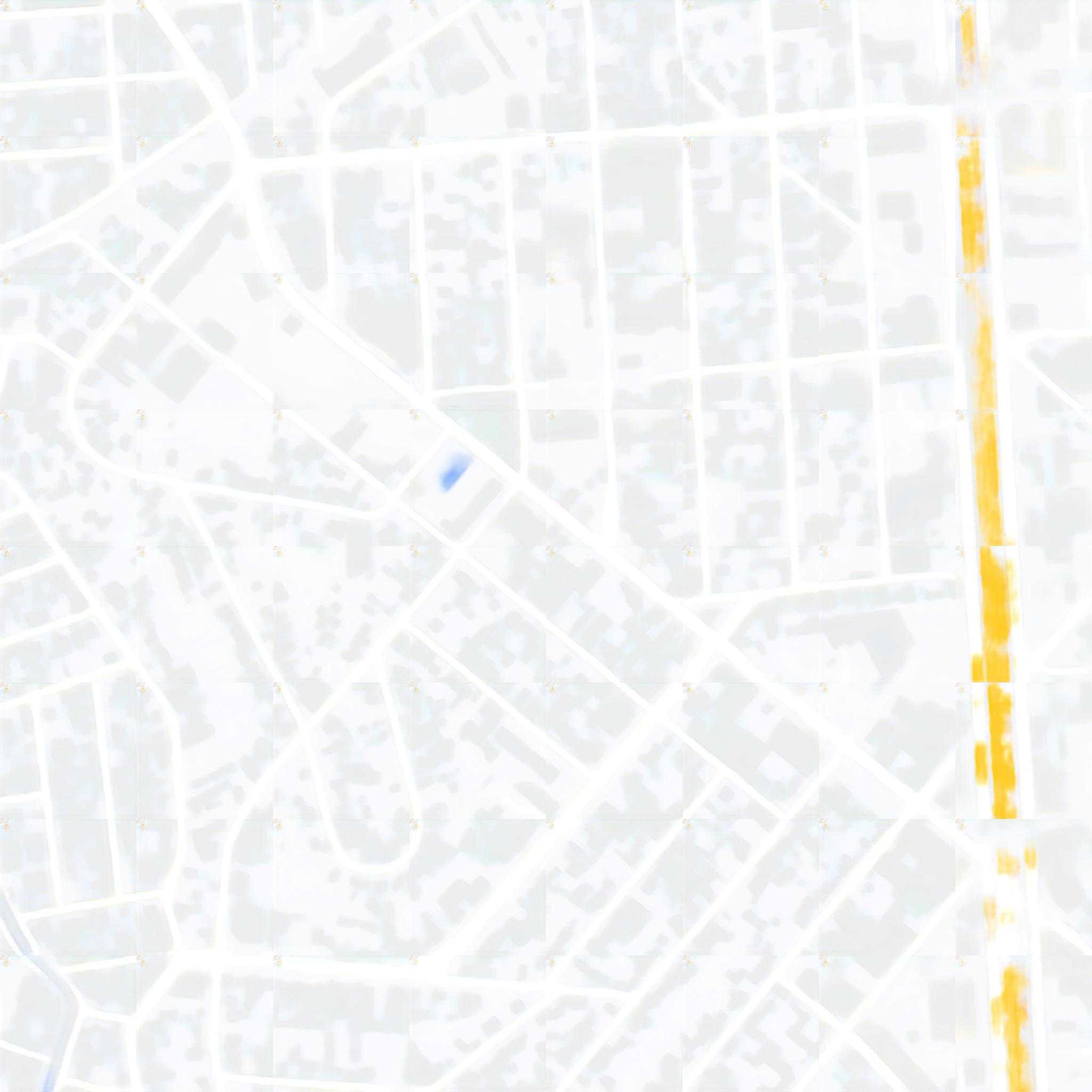}\hspace{-3mm}\quad
				\begin{sideways}\qquad \quad $2048\times 2048$\end{sideways}
				\\
(a) Aerial image & (b) Pix2Pix & (c) Pix2PixHD & (d) CycleGAN  & (e) GeoGAN & (f) SMAPGAN\\
				\begin{sideways}\qquad \quad 15-th\ level\end{sideways}
				\includegraphics[width=4.01cm, angle=0]{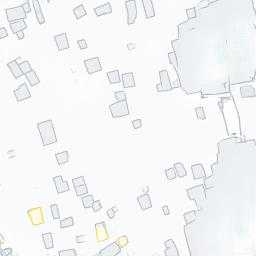}\hspace{-3mm}
				&\includegraphics[width=4.01cm, angle=0]{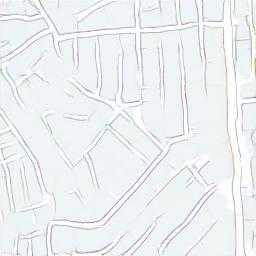}\hspace{-3mm}
				&\includegraphics[width=4.01cm, angle=0]{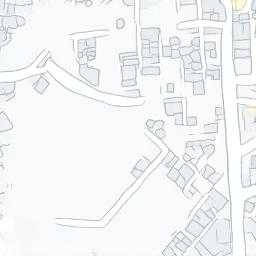}\hspace{-3mm}
				&\includegraphics[width=4.01cm, angle=0]{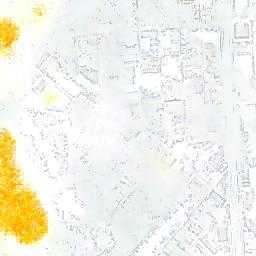}\hspace{-3mm}
				&\includegraphics[width=4.01cm, angle=0]{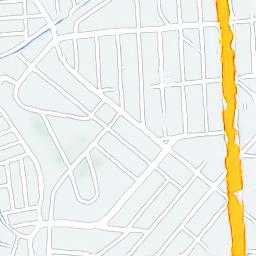}\hspace{-3mm}
				&\includegraphics[width=4.01cm, angle=0]{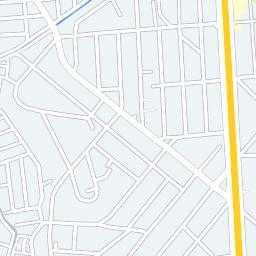}\hspace{-3mm}\quad
				\begin{sideways}\qquad \quad $256\times 256$\end{sideways}
				\\
				\begin{sideways}\qquad \quad 16-th\ level\end{sideways}
				\includegraphics[width=4.01cm, angle=0]{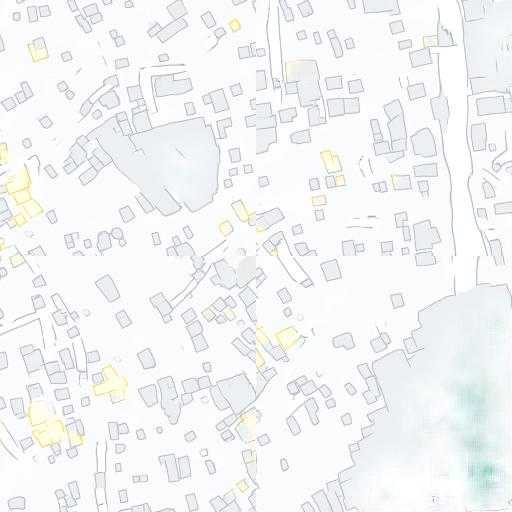}\hspace{-3mm}
				&\includegraphics[width=4.01cm, angle=0]{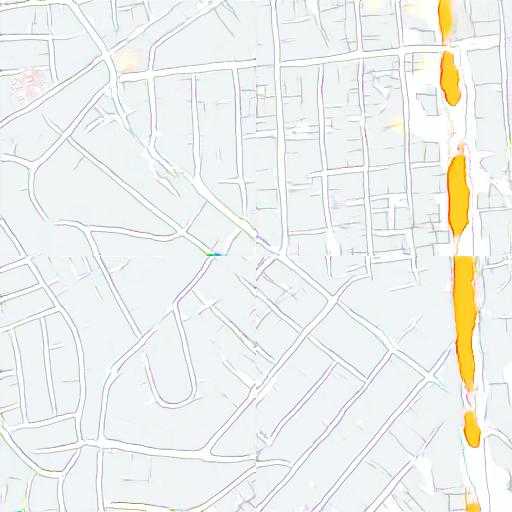}\hspace{-3mm}
				&\includegraphics[width=4.01cm, angle=0]{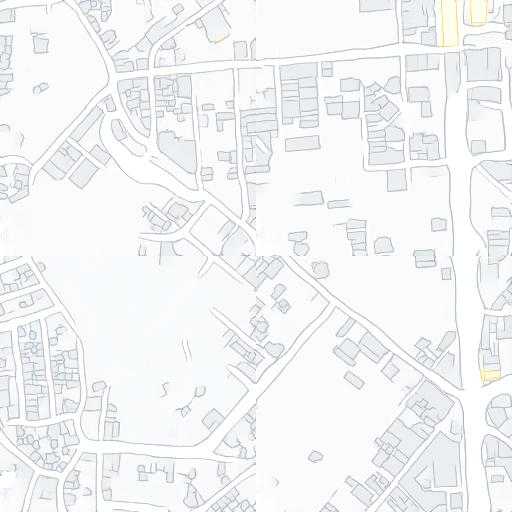}\hspace{-3mm}
				&\includegraphics[width=4.01cm, angle=0]{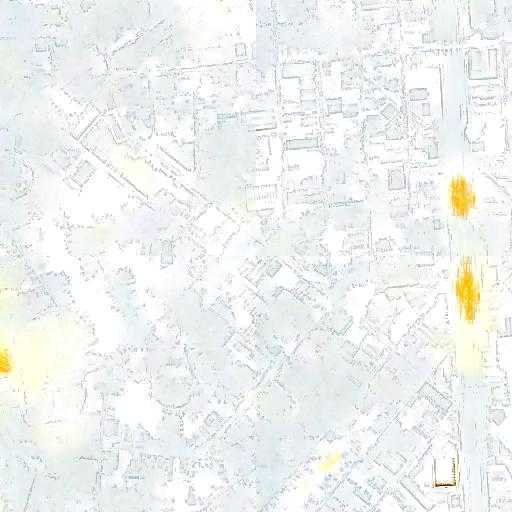}\hspace{-3mm}
				&\includegraphics[width=4.01cm, angle=0]{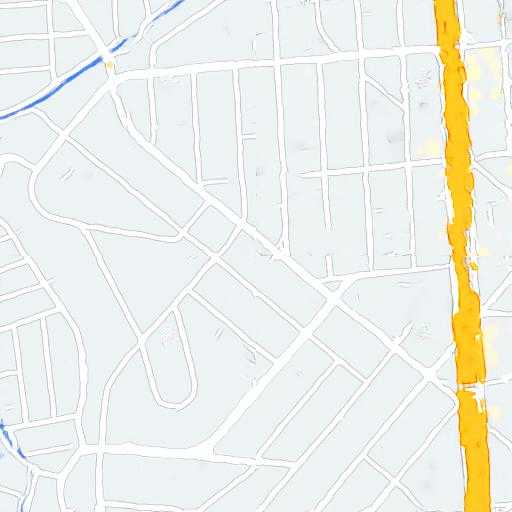}\hspace{-3mm}
				&\includegraphics[width=4.01cm, angle=0]{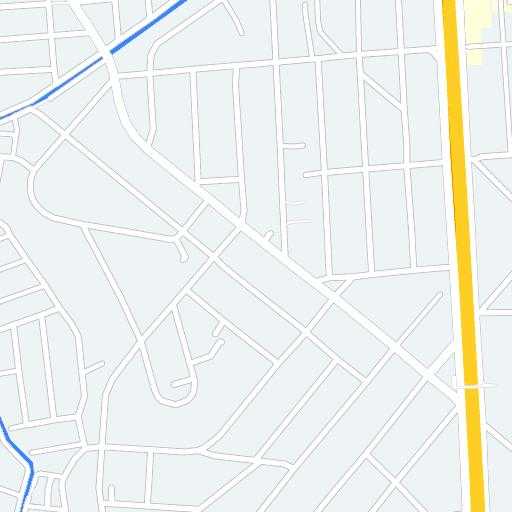}\hspace{-3mm}\quad
				\begin{sideways}\qquad \quad $512\times 512$\end{sideways}
				\\
				\begin{sideways}\qquad \quad 17-th\ level\end{sideways}
				\includegraphics[width=4.01cm, angle=0]{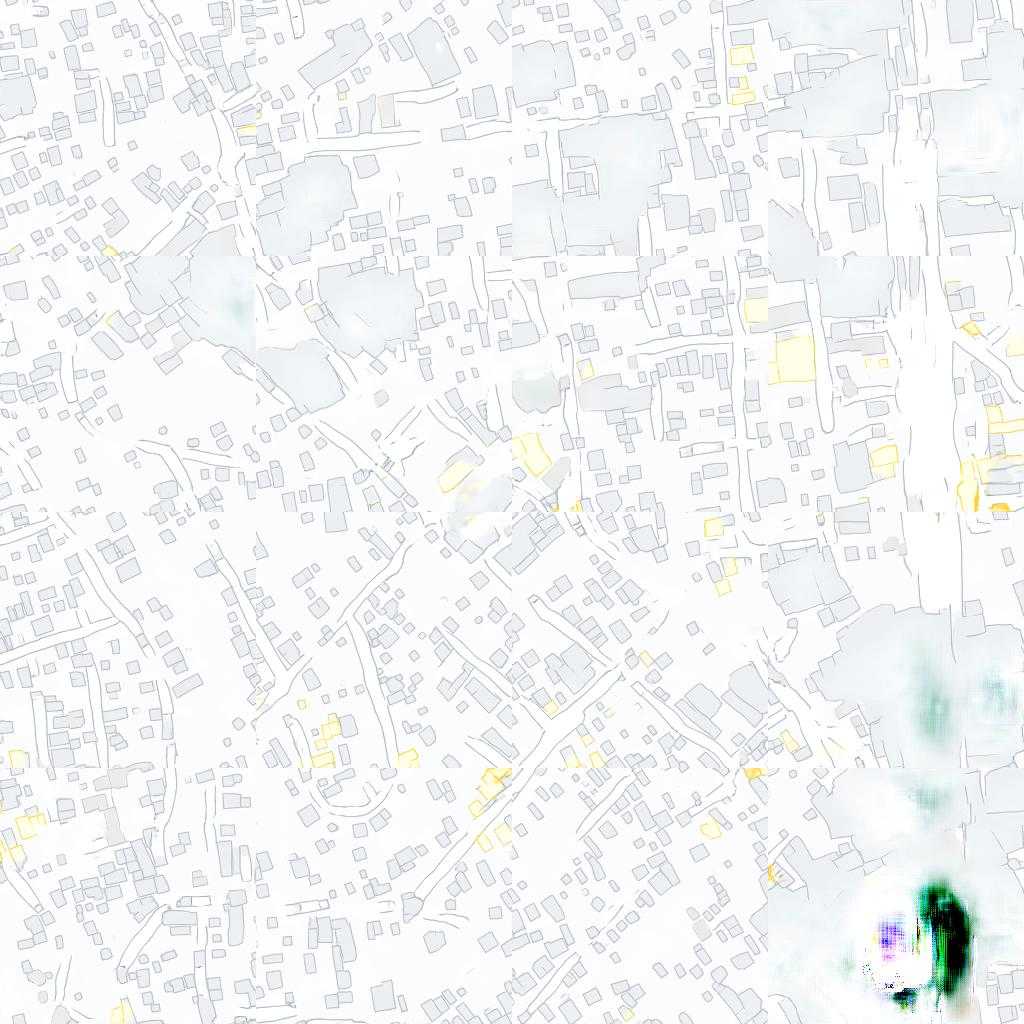}\hspace{-3mm}
				&\includegraphics[width=4.01cm, angle=0]{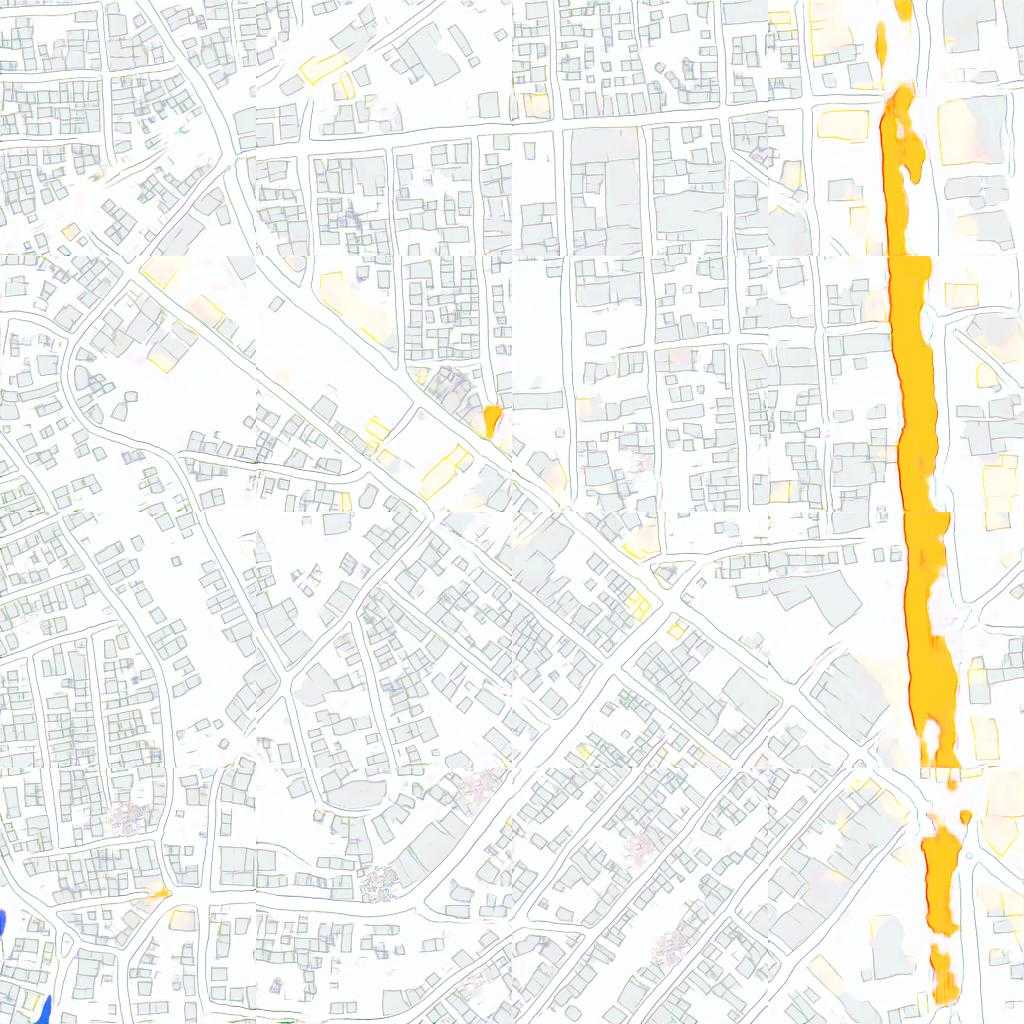}\hspace{-3mm}
				&\includegraphics[width=4.01cm, angle=0]{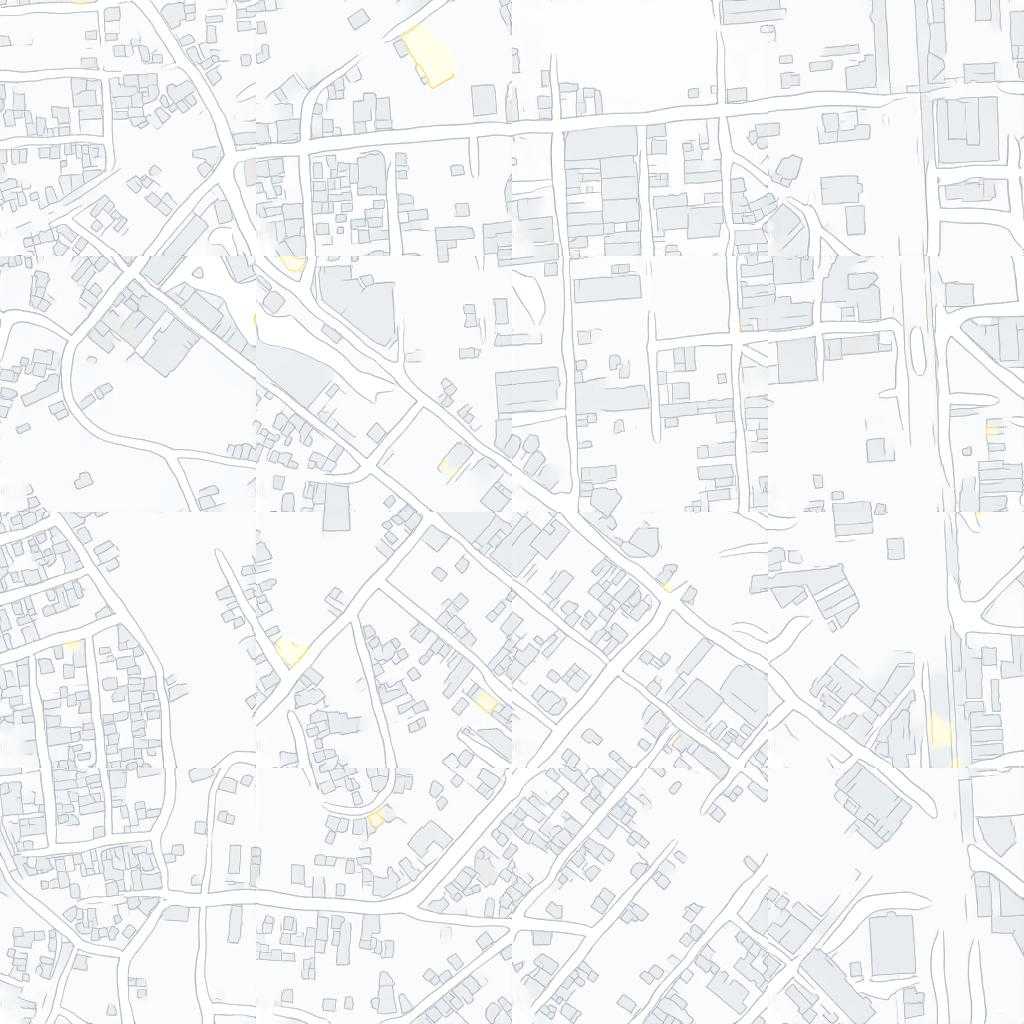}\hspace{-3mm}
				&\includegraphics[width=4.01cm, angle=0]{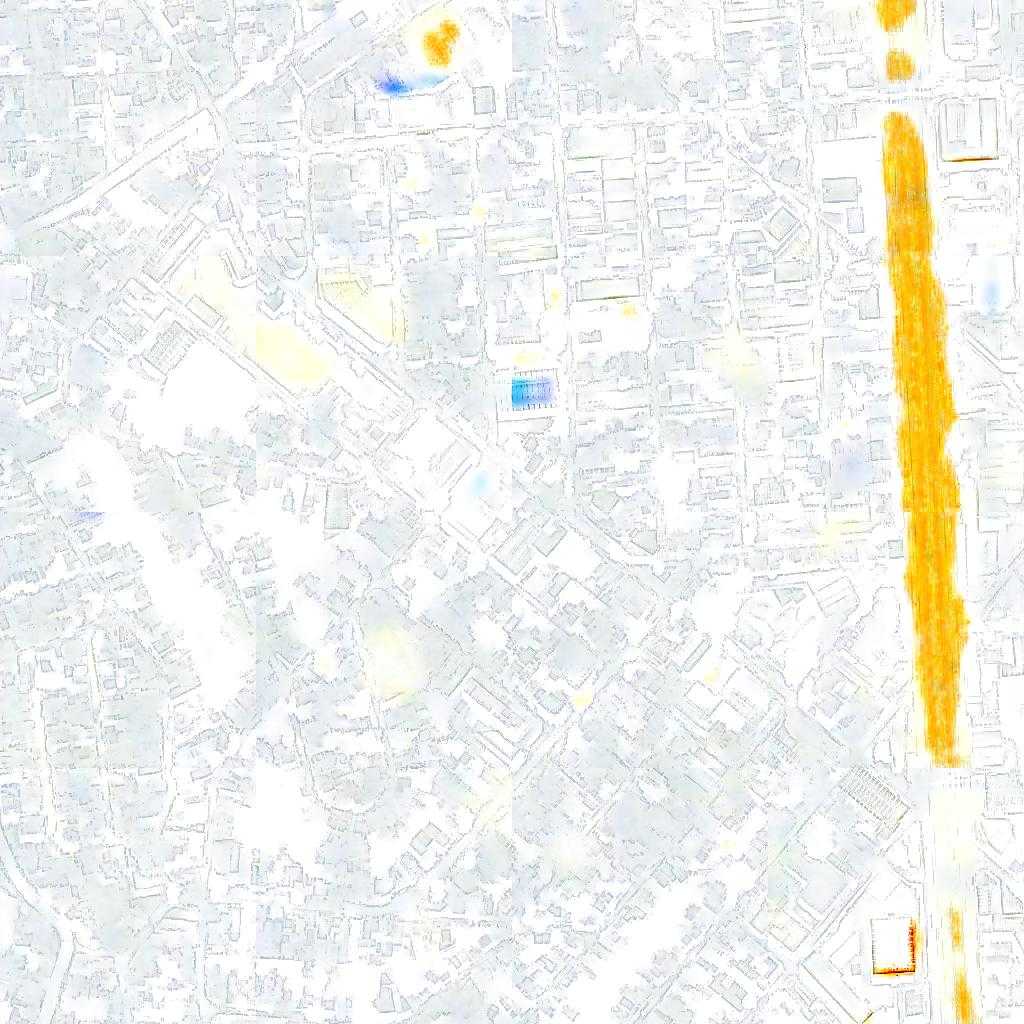}\hspace{-3mm}
				&\includegraphics[width=4.01cm, angle=0]{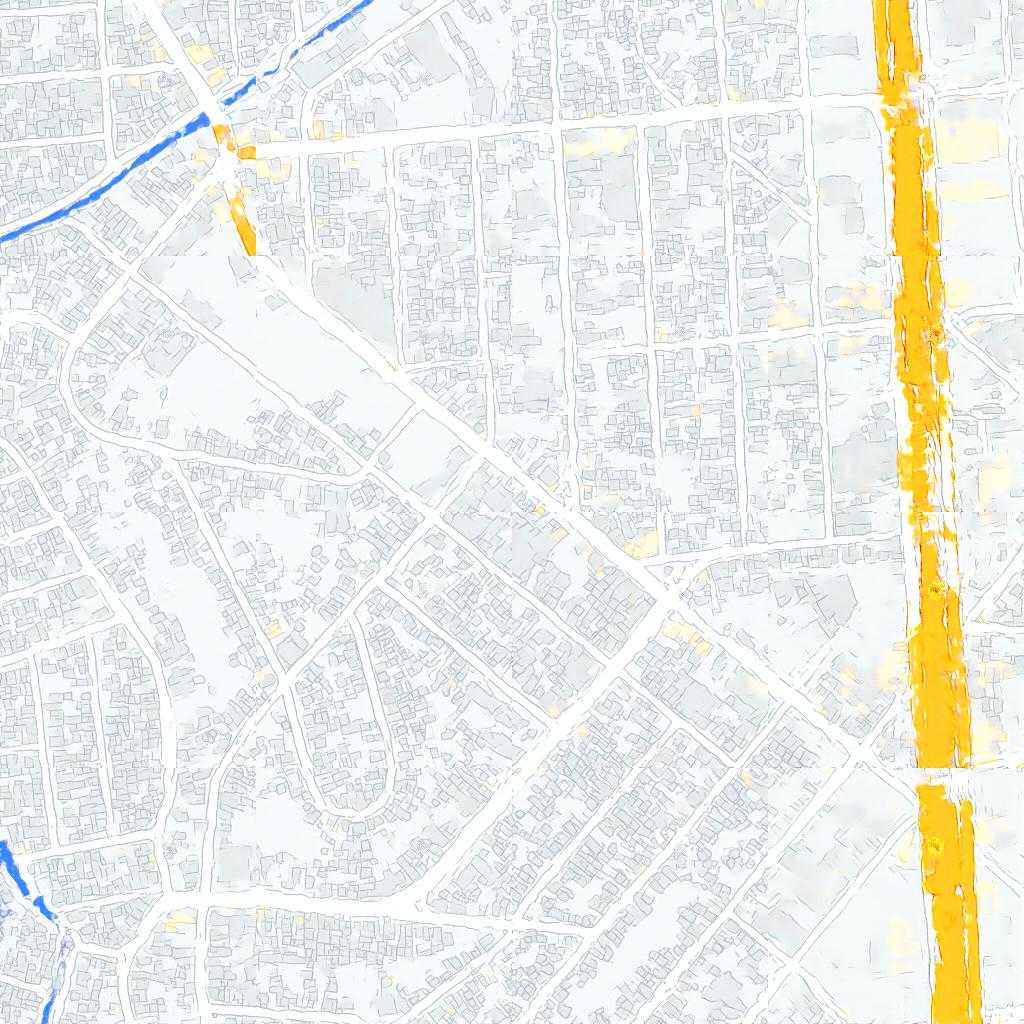}\hspace{-3mm}
				&\includegraphics[width=4.01cm, angle=0]{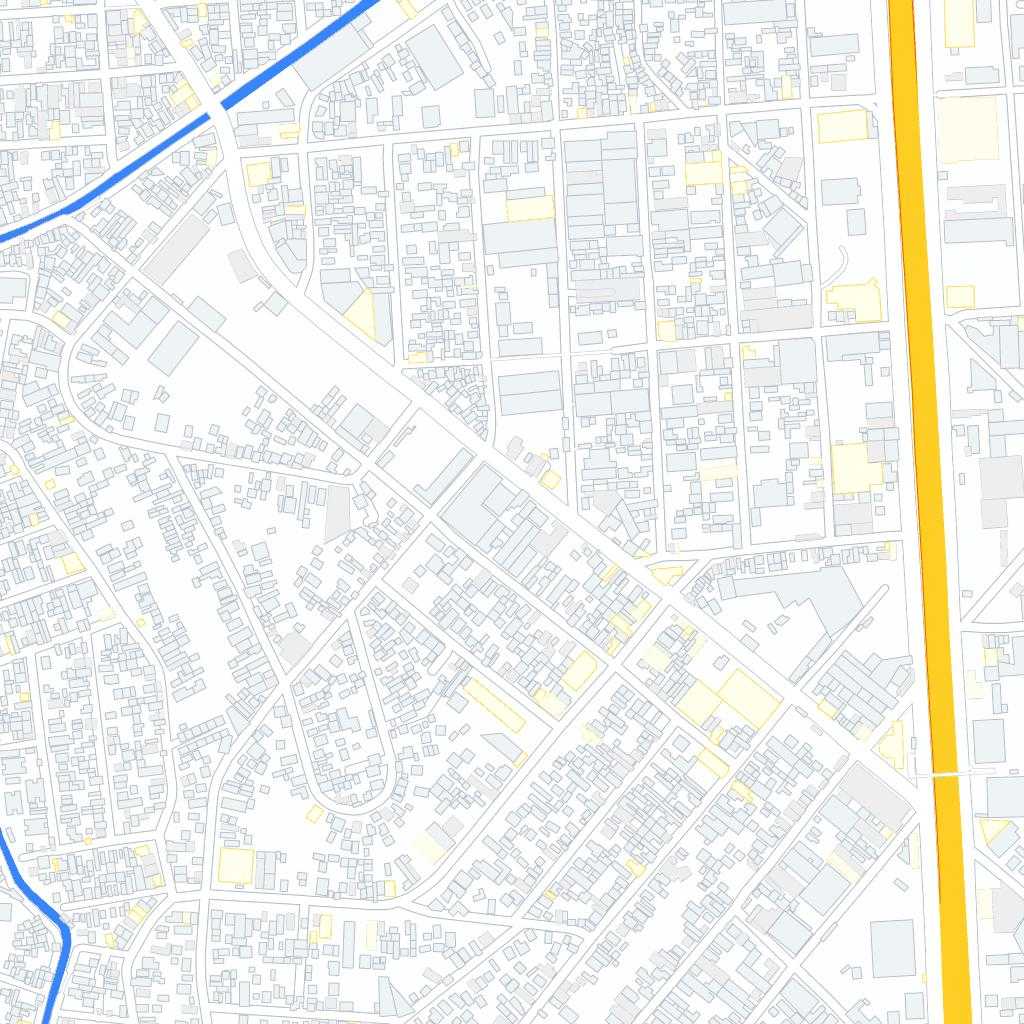}\hspace{-3mm}\quad
				\begin{sideways}\qquad \quad $1024\times 1024$\end{sideways}
				\\
				\begin{sideways}\qquad \quad 18-th\ level\end{sideways}
				\includegraphics[width=4.01cm, angle=0]{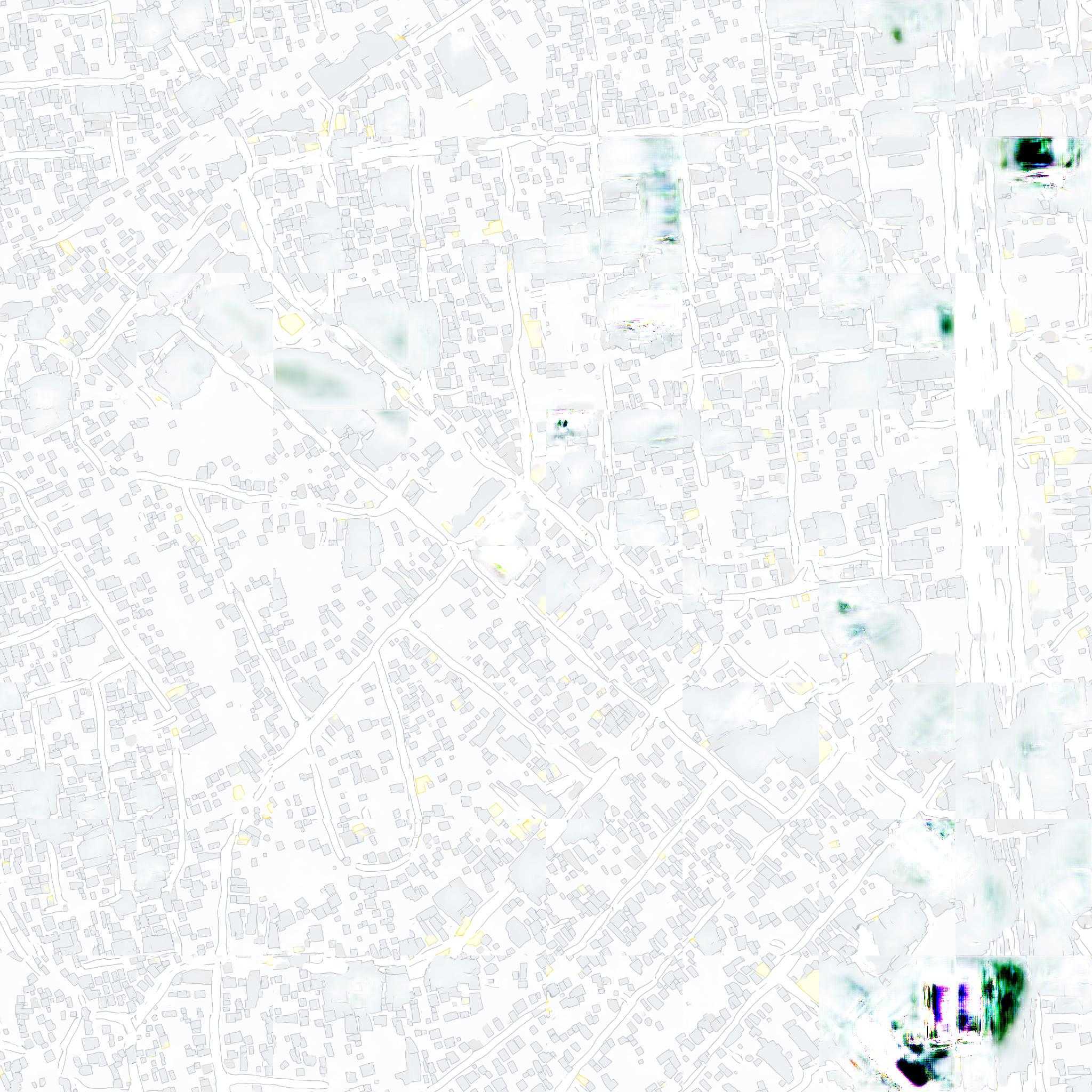}\hspace{-3mm}
				&\includegraphics[width=4.01cm, angle=0]{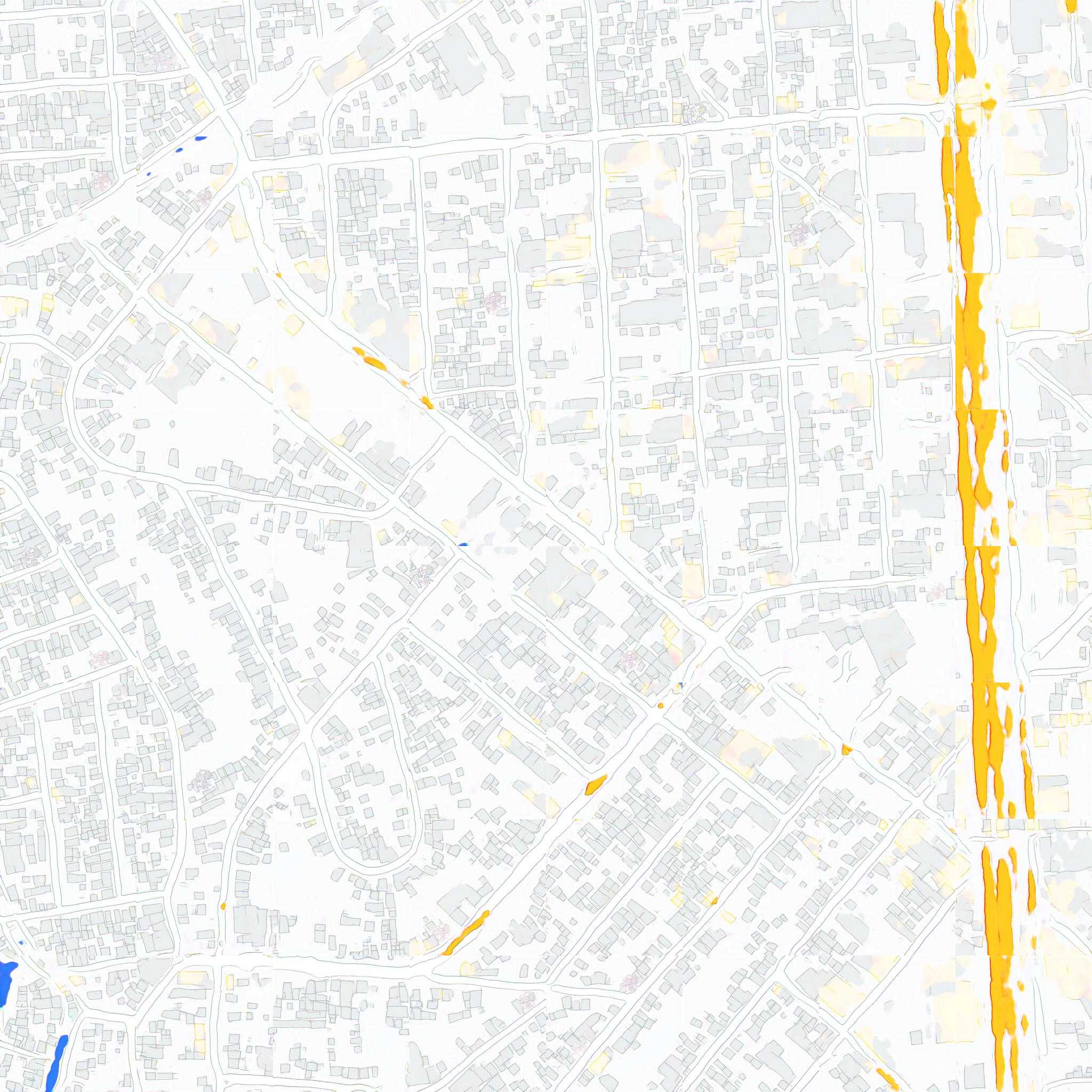}\hspace{-3mm}
				&\includegraphics[width=4.01cm, angle=0]{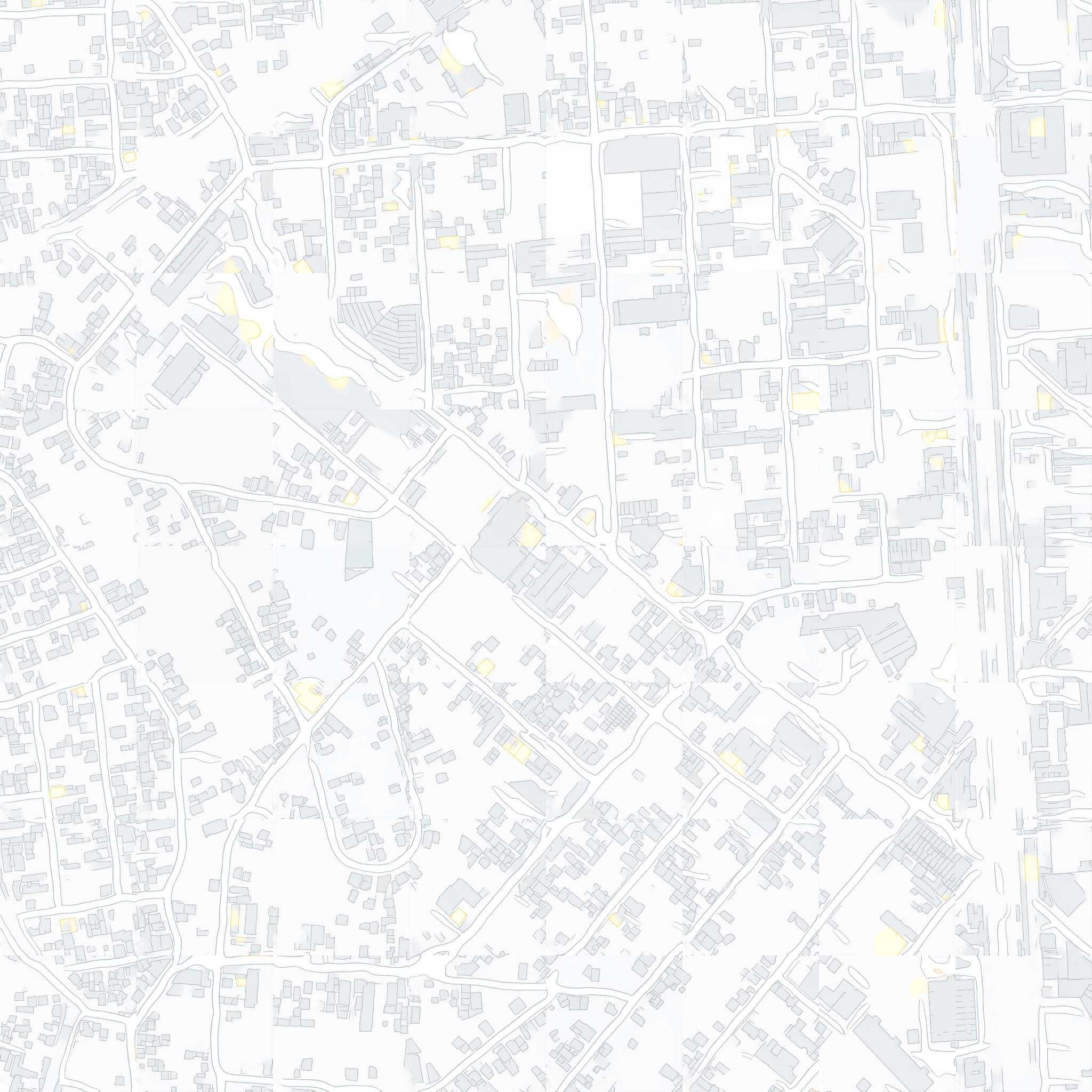}\hspace{-3mm}
				&\includegraphics[width=4.01cm, angle=0]{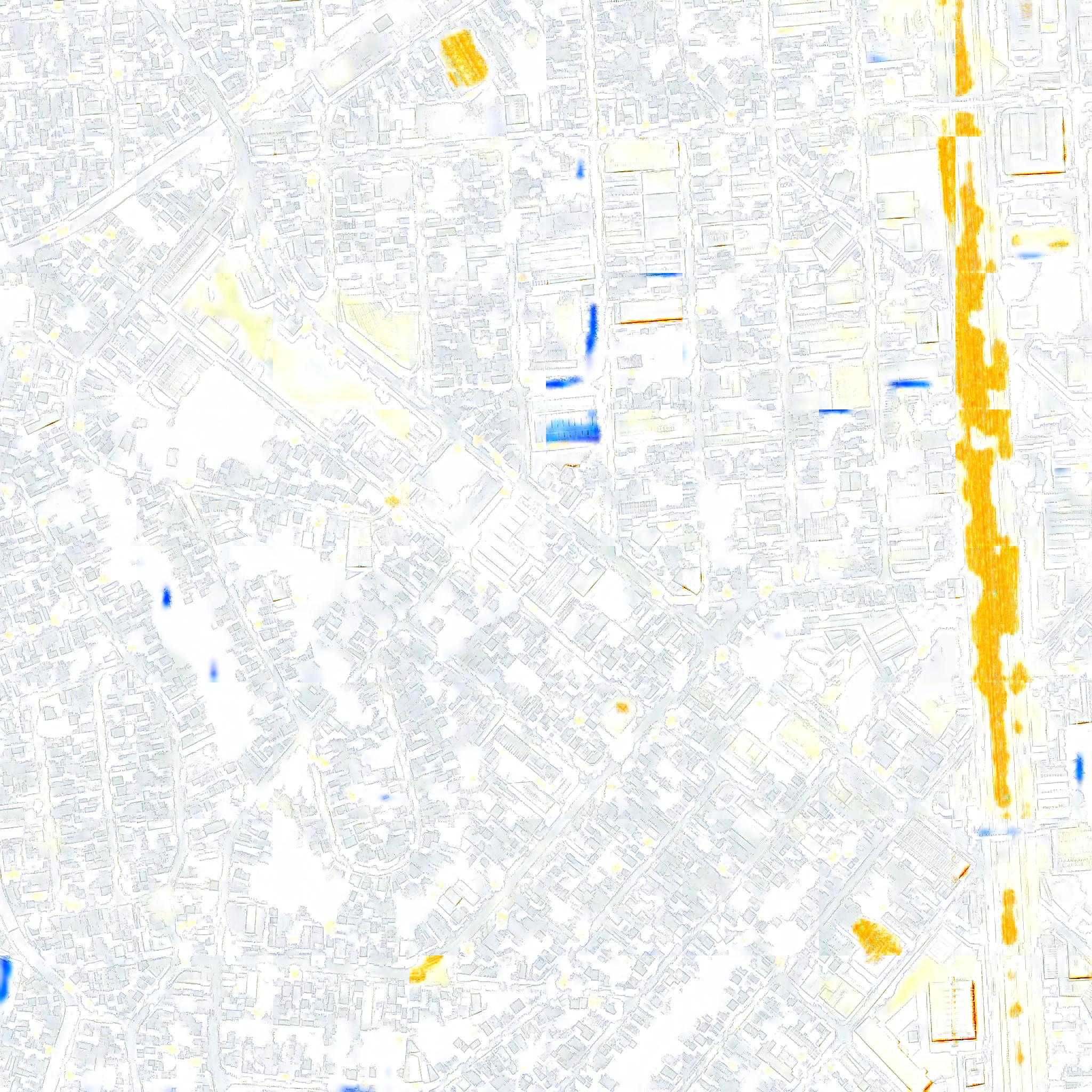}\hspace{-3mm}
				&\includegraphics[width=4.01cm, angle=0]{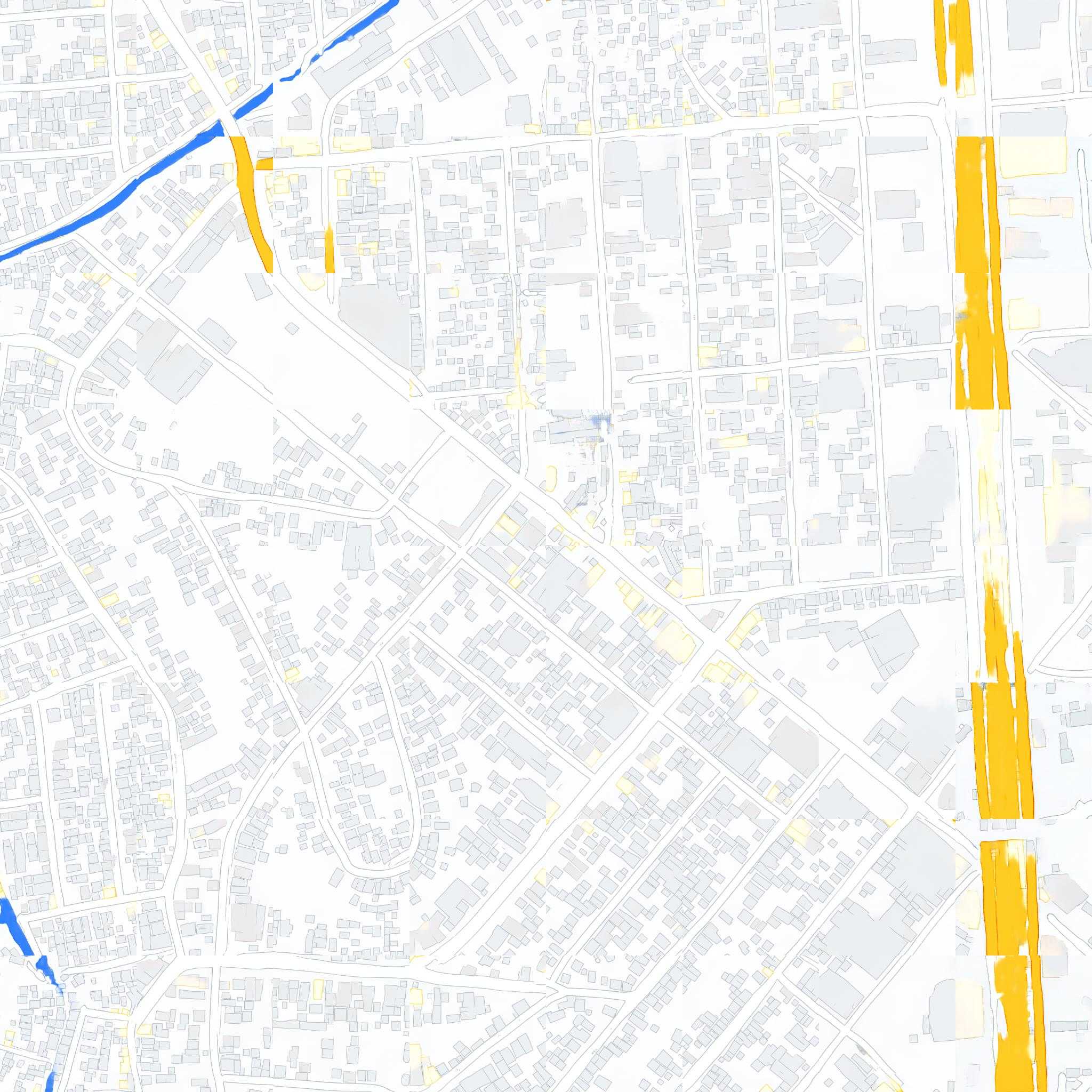}\hspace{-3mm}
				&\includegraphics[width=4.01cm, angle=0]{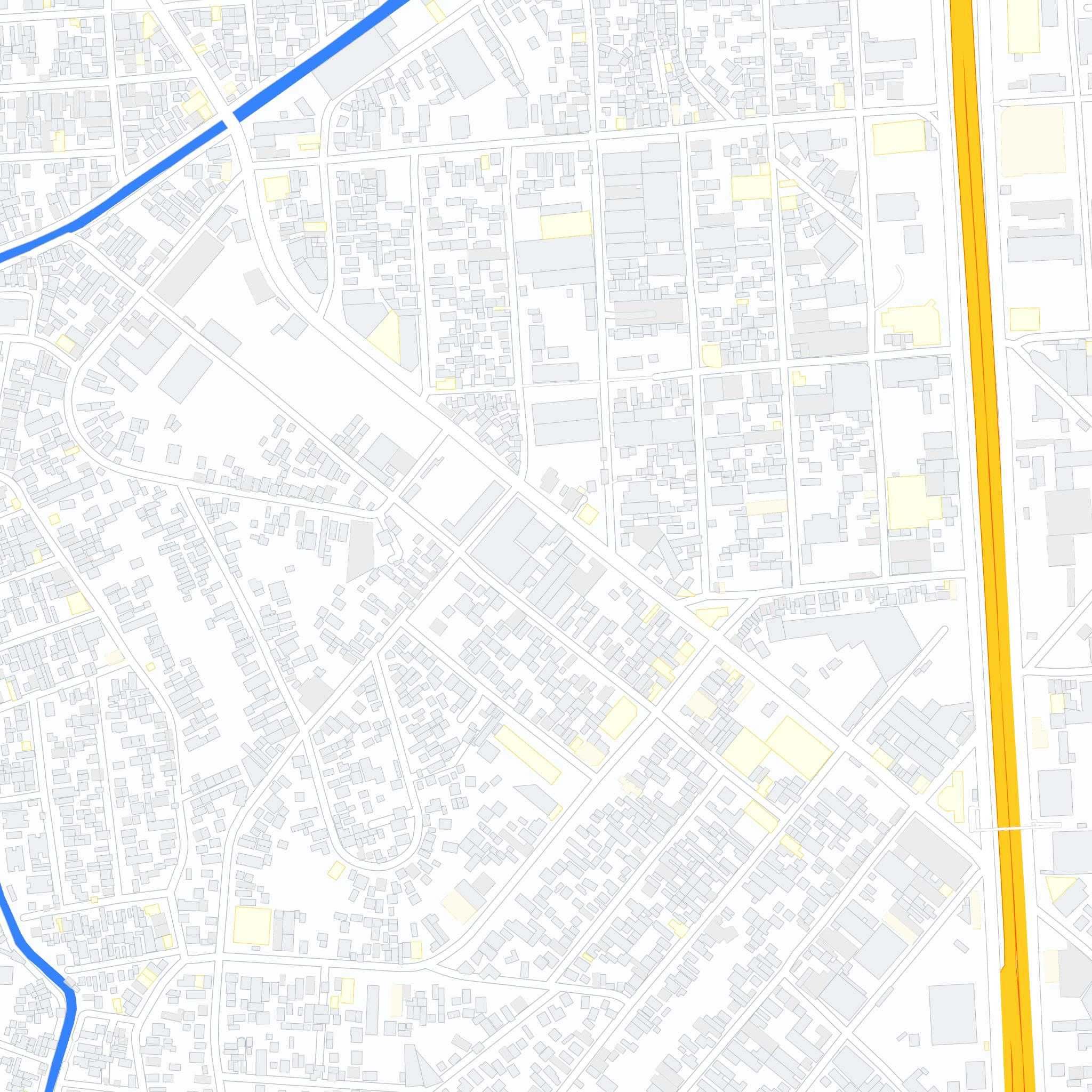}\hspace{-3mm}\quad
				\begin{sideways}\qquad \quad $2048\times 2048$\end{sideways}
				\\
(g) SPADE & (h) SelectionGAN  & (i) TSIT & (j) LPTN  & (k) Ours & (l) GT\\
	
	\end{tabular}}
	
\else
\includegraphics{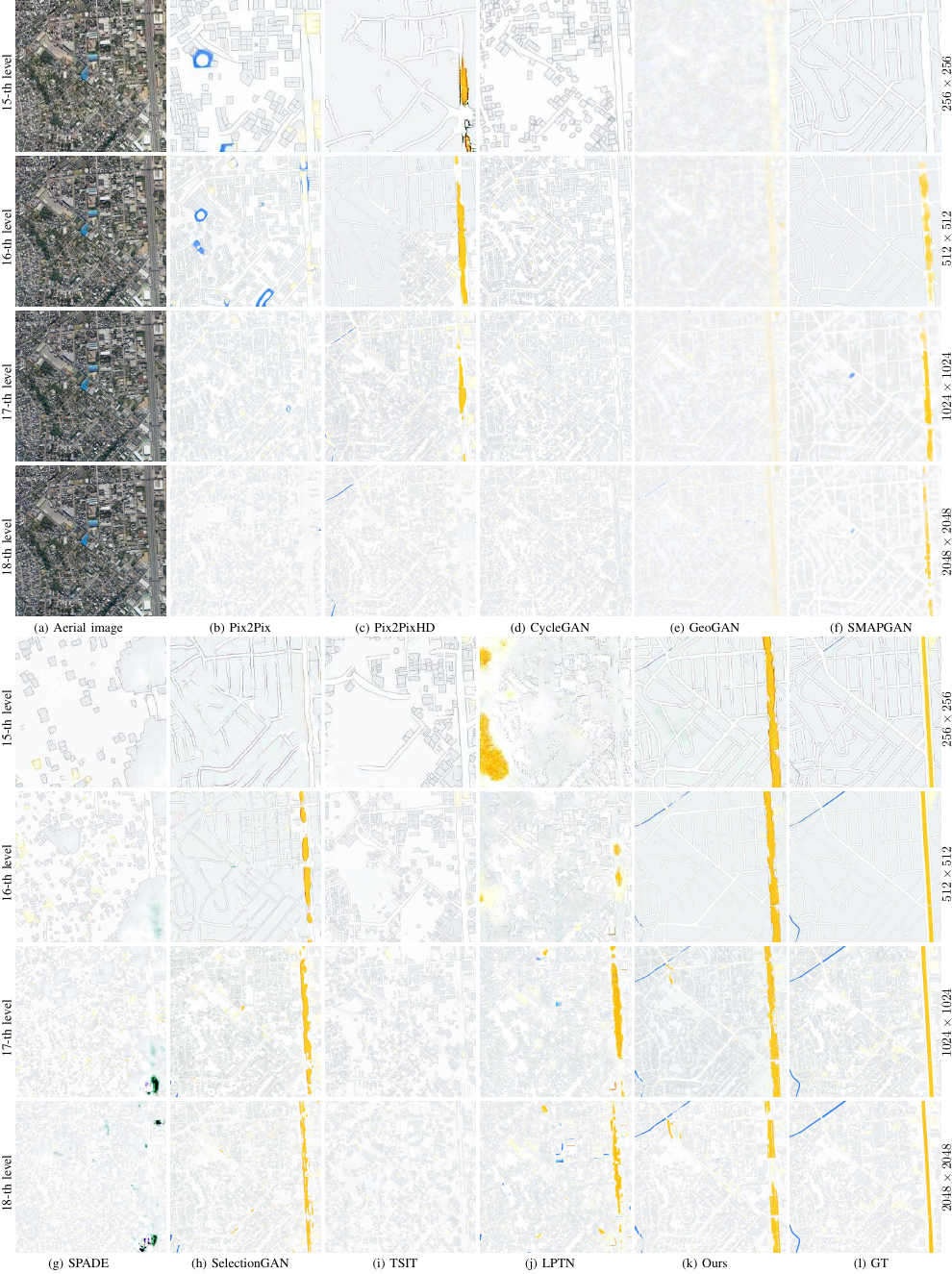}
\fi	
	\end{center}
	\vspace{-0mm}
	\caption{
	Visualized results of MLM-RIO generated by our method and counterparts.
	Our method shows satisfactory accuracy and content consistency, especially for rivers and roads in all zoom levels. 
	}
	\vspace{+2mm}
	\label{fig:visual_rio}
\end{figure*}

\begin{table*}[tb]
	\begin{center}
	\scalebox{1.1}{
		\begin{tabular}{c|c|c|c|c|ccccc}
			\toprule[1.0pt]
			\multirow{2}{*}{Level} & \multicolumn{3}{c|}{LAMG} & \multirow{2}{*}{CAMG}  
			& \multirow{2}{*}{FID$\downarrow$}& \multirow{2}{*}{KID$\downarrow$}& \multirow{2}{*}{KMMD$\downarrow$}& \multirow{2}{*}{IS$\uparrow$} &\multirow{2}{*}{PSNR$\uparrow$}  \\
			\cline{2-4}  
			& Drawer Module & Level ID & Semantic Module & & & & & \\		                            
			\midrule
			\multirow{4}{*}{Avg.}
			&$\surd$ & $\times$ & $\times$ & $\times$  & 265.2 & 0.1685 & 0.4578 & 2.952 & 20.65  \\
			&$\surd$ & $\surd$& $\times$ & $\times$  & 264.8 & 0.1654 & 0.4489 & 3.134 & 20.53  \\
			&$\surd$ & $\surd$ & $\surd$ & $\times$  & 174.0 & 0.0970 & 0.3042 & 3.244 & 20.26 \\
			&$\surd$ & $\surd$&  $\surd$ &  $\surd$  & \bf 137.7 &\bf 0.0805 & \bf 0.2957 & \bf 3.312 & \bf 22.01 \\
			\bottomrule[1.0pt]
		\end{tabular}}
	\end{center}
	\vspace{-0mm}
	\caption{Results of level-aware map generator (LAMG) and consistency-aware map generator (CAMG) on MLM-SH dataset for ablation study.}
	\vspace{-0mm}
	\label{tab:ablations1}
\end{table*}

\subsection{Evaluation Metrics}

\noindent{\bf Evaluation Metrics for Accuracy.}
To evaluate the accuracy and the similarity between generated maps and real maps for each zoom level, we use Fr{\'e}chet Inception Distance (FID)~\cite{FID}, Kernel Inception Distance (KID)~\cite{KID}, Kernel Maximum Mean Discrepancy (KMMD)~\cite{KMMD}, and Inception Score (IS)~\cite{IS} as metrics. 
FID and KID are shown to correlate well with the human judgment of visual quality and are often used to evaluate the quality of samples of GANs \cite{FID, KID}, KMMD is able to identify generative or noisy images from real images, and IS is a popular metric for image generation quality evaluation. 
In addition, we also adopt the classical Peak Signal-to-Noise Ratio (PSNR) as our metric. 
The lower FID, KID, KMMD and higher IS, PSNR indicate better results. 
The relevant details are shown in Table \ref{tab:metric}. 

\vspace{+1mm}
\noindent{\bf Evaluation Metrics for Consistency.}
To evaluate the content consistency in a simple way, we compare generated maps with real maps at each zoom level. 
Considering real maps keep their content consistent at different zoom levels, if generated maps at all zoom levels are accurate and close to real maps, they should also keep good content consistency, thus we evaluate the consistency by
\begin{equation}
	Metric^{c} = \frac{1}{K} \sum_{i=1}^{i=K}Metric(y_i, y^{map}_i), 
\end{equation}
where $y_i$,  $y_i^{map}$ is generated map and real map at the $i$-th zoom level, $Metric(.)$ indicates the metric we use for accuracy evaluation, \ie, FID, KID, KMMD, IS and PSNR, and $Metric^{c}$ is the result of consistency. 
In this way, the result of consistency can also show the overall accuracy. 
 


\subsection{Quantitative and Qualitative Evaluation. }
We compare our method with several previous state-of-the-art methods for single-level map generation including Pix2Pix~\cite{pix2pix}, Pix2PixHD~\cite{pix2pixhd}, CycleGAN~\cite{cyclegan}, GeoGAN~\cite{2019geogan}, SMAPGAN~\cite{2020smapgan}, SPADE~\cite{2019spade}, SelectionGAN~\cite{2019selectiongan}, TSIT~\cite{2020tsit} and LPTN~\cite{2021LPTN}.
{\color{black} As for the multi-level map generation method, there is still no relevant method for comparison. 
We note that all methods are trained with the same setting on the training set of the MLM dataset and evaluated on the testing set for fairness.}
{\color{black}
We only need to crop the aerial images into $256\times256$ patches as inputs, and do not need other preprocessing for the training or testing.}
Quantitative and qualitative results show our method outperforms prior works on the MLM dataset. 


As illustrated in Tables~\ref{tab:mainresults1} and \ref{tab:mainresults2}, our method outperforms all counterparts by a large margin on average results and most single-level results of FID, KID, KMMD and IS metrics. 
{\color{black}
Notice that IS uses an ImageNet pre-trained InceptionV3~\cite{InceptionV3} to calculate the realism of generated images but does not compare the generated map images with ground truth map images~\cite{IS}, so its result is not as reliable as FID, KID and KMMD.
Nevertheless, we provide IS result due to its popularity and for a more comprehensive comparison. 
}
As for PSNR, we also achieve better results than including Pix2Pix~\cite{pix2pix}, Pix2PixHD~\cite{pix2pixhd}, CycleGAN~\cite{cyclegan}, SPADE~\cite{2019spade}, SelectionGAN~\cite{2019selectiongan}, TSIT~\cite{2020tsit} and LPTN~\cite{2021LPTN}.
Though GeoGAN~\cite{2019geogan} and SMAPGAN~\cite{2020smapgan} achieve similar or better results than our method in PSNR, their visual quality is much worse than ours as shown in Figure~\ref{fig:visual_sh} and Figure~\ref{fig:visual_rio}. 
The reason may be that GeoGAN~\cite{2019geogan} and SMAPGAN~\cite{2020smapgan} tend to minimize the mean square error (MSE), but the ability of MSE (and PSNR) to capture perceptually relevant differences (\eg, high texture detail) is very limited as they are defined based on pixel-wise image differences \cite{SSIM, 2011modifiedpsnr, msssim}. 
And a series of researches have shown that a higher PSNR does not necessarily reflect perceptually better visual results \cite{srgan}. 

The Figure~\ref{fig:visual_sh} and Figure~\ref{fig:visual_rio} show that the visual quality of the proposed method outperforms all counterparts in two aspects. 
Firstly, the semantic module explicitly provides the pixel-wise understanding of input aerial images, so our method is able to generate clear and accurate maps that are similar to real maps, while visual results of counterparts tend to be blurred and inaccurate at each zoom level. 
{\color{black} Secondly, our method largely keeps the content consistency across all of the zoom levels with the help of the consistency-aware map generator, whereas the multi-level maps generated by counterparts suffer from severe content inconsistency.}
{\color{black}
Nonetheless, our method still has room for improvement to distinguish different types of roads (indicate as white, yellow and orange in map images).
These roads have the same semantic features and look very similar in aerial images, but they need to draw in different colors, which makes it very challenging. 
}

These results show the proposed method outperforms all counterparts on accuracy and consistency by a large margin quantitatively and qualitatively in multi-level map generation task, and demonstrate the effectiveness of the proposed method.



\subsection{Ablation Study}
We perform the ablation studies on MLM-SH for evaluating the effectiveness of each part in our method.  
Considering the drawer module is the basic module of the generator and the semantic module cannot generate maps independently, we take the performance of the drawer module as a baseline and evaluate the effectiveness of level identification, semantic module and consistency-aware map generator. 

{\color{black}

\vspace{+1mm}
\noindent{\bf Level Identification.}
Level identification provides information of zoom level, which is helpful for the generator to optimize the output map at each zoom level. 
As shown in Table~\ref{tab:ablations1}, it leads to improvements on FID, KID, KMMD and IS metrics. 
And the accuracy of generated maps with level identification are better than baseline in Figure~\ref{fig:Ablation Result}{\color{black}(c)} (\eg, the river in the 15-th zoom level is more clear). 
 
\vspace{+1mm}
\noindent{\bf Semantic Module.}
The semantic module is designed to provide pixel-wise understanding of aerial images and semantic information so as to keep the topological relationship among geographical elements. 
It can be seen in Table~\ref{tab:ablations1} that the semantic module largely improves FID, KID, KMMD and IS results. 
And Figure \ref{fig:Ablation Result}{\color{black}(d)} shows semantic module increases the accuracy of generated maps, especially for lower zoom levels. 
This shows the effectiveness of the semantic module. 


\vspace{+1mm}
\noindent{\bf Consistency-Aware Map Generator (CAMG).}
CAMG not only builds the connection between maps at different zoom levels, but also repairs and refines the initial generated map with the information provided by maps at higher zoom levels.
After we employ the consistency-aware map generator (CAMG), the overall accuracy and content consistency of generated maps are largely improved as shown in Table~\ref{tab:ablations1} and Figure~\ref{fig:Ablation Result}{\color{black}(e)}. 

\vspace{+1mm}
{\color{black}
\noindent{\bf Dataset for Multi-Level Map Generation.}
{\color{black} In order to demonstrate the effectiveness of our dataset for multi-level map generation, we perform a series of experiments with our MLM dataset and an aerial-to-map dataset provided by Pix2Pix~\cite{pix2pix}. }
The Pix2Pix dataset~\cite{pix2pix} is a well-known high-quality aerial-to-map dataset, which validates the feasibility of translating aerial images to single-level maps.

For efficiency, we choose TSIT~\cite{2020tsit}, SelectionGAN~\cite{2019selectiongan} and our level-aware map generator (LAMG) for experiments. 
The reason why LAMG was chosen instead of our entire model is that the Pix2pix dataset does not include multi-level data and cannot be used to train our entire model.
For fairness, we train these models on the Pix2pix dataset and the MLM-SH dataset respectively, and we evaluate them on the MLM-RJ dataset.
Notice that maps in MLM-SH and MLM-RJ have a very different style (\ie, domain gap) as shown in Figure~\ref{fig:SH} and Figure~\ref{fig:RIO}. 
Therefore, MLM-SH, MLM-RJ and Pix2Pix dataset~\cite{pix2pix} can be regarded as three independent datasets.

As shown in Table~\ref{tab:DatasetExperiment}, the models trained on the MLM-SH dataset achieve the best results on most metrics at each level. 
These results show that our dataset is effective and has better potential for multi-level map generation task, whereas the existing single-level map dataset cannot handle the needs of this task.
}

\section{Conclusion}

In this paper, we propose a novel method for multi-level map generation from aerial images.
By understanding aerial images pixel-wisely and building connections between maps at different zoom levels, it is able to generate accurate maps for each zoom level, and keep the content consistency between maps at different zoom levels. 
Quantitative and qualitative results both show our method achieves the best performance compared with previous state-of-the-art map generation methods. 
{\color{black}
In addition, we collect and produce a large-scale high-quality dataset called Multi-Level Map (MLM). 
The MLM dataset provides 18,700 samples of aerial images and maps from two representative cities in four zoom levels. 
It is able to serve as a benchmark and support the future research of multi-level map generation. 

In the future, we plan to collect and produce a multi-level map dataset with a larger scale, more cities and more zoom levels. 
Considering the development of multi-level map generation is still in its early stage, it is worth investigating the better model structure, loss function and evaluation metric specially designed for this task. }
We hope our dataset and preliminary work can serve as a foundation, and open new chances for future research.

{\small
\bibliographystyle{IEEEtran}
\bibliography{egbib.bib}
}


%

\ifCLASSOPTIONcaptionsoff
  \newpage
\fi



%


%

\iftrue
\begin{IEEEbiography}
[{\includegraphics[width=1in,height=1.25in,clip,keepaspectratio]{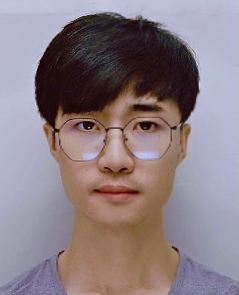}}]{Linwei Chen} received the B.S. degree in Mechanical Engineering and Automation from China University of Geosciences (Beijing), China, in 2019, and the M.S. degree in Software Engineering from Beijing Institute of Technology, China, in 2021. 
His research interests include image segmentation, object detection, and remote sensing.
\end{IEEEbiography}

\begin{IEEEbiography}
[{\includegraphics[width=1in,height=1.25in,clip,keepaspectratio]{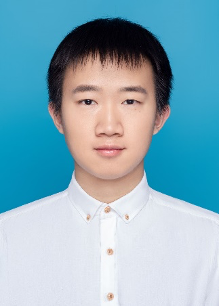}}]{Zheng Fang} \color{blue} received the B.S. degree in Computer Science and Technology from the Ocean University of China, China, in 2020. He is currently pursuing the M.S. degree in Computer Science and Technology at Beijing Institute of Technology, China.
His research focuses on the application of image translation and remote sensing.
\end{IEEEbiography}

\begin{IEEEbiography}
[{\includegraphics[width=1in,height=1.25in,clip,keepaspectratio]{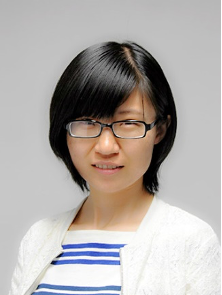}}]{Ying Fu} (Senior Member, IEEE) received the B.S. degree in Electronic Engineering from Xidian University, China, in 2009, the M.S. degree in Automation from Tsinghua University, China, in 2012, and the Ph.D. degree in Information Science and Technology from the University of Tokyo, Japan, in 2015. She is currently a Professor in the School of Computer Science and Technology, Beijing Institute of Technology. Her research interests include physics-based vision, image and video processing, and remote sensing.
\end{IEEEbiography}

\fi




\end{document}